\documentclass[letterpaper]{article}
\usepackage{aaai25} %
\usepackage{times} %
\usepackage{helvet} %
\usepackage{courier} %
\usepackage[hyphens]{url} %
\usepackage{graphicx} %
\usepackage{graphicx}  %
\usepackage{natbib}  %
\usepackage{caption}  %
\usepackage{algorithm}
\usepackage{algorithmicx}
\usepackage{algpseudocode}
\usepackage{placeins}

\usepackage{newfloat}
\usepackage{listings}
\DeclareCaptionStyle{ruled}{labelfont=normalfont,labelsep=colon,strut=off}

\floatstyle{ruled}
\newfloat{listing}{tb}{lst}{}
\floatname{listing}{Listing}

\nocopyright
\usepackage{booktabs}
\usepackage{xcolor}
\usepackage[inline]{enumitem}
\usepackage{amsmath}
\usepackage{amssymb}
\usepackage{cleveref}
\usepackage[export]{adjustbox}
\usepackage{tabularx}
\usepackage{makecell}
\usepackage{array}
\usepackage{fancybox}
\usepackage{multirow}
\usepackage{colortbl}
\usepackage{pdfpages}

\newlength{\ww}
\def\ShowNotes{}
\def\ShowProjectPage{}
\makeatletter
\renewcommand\@makefntext[1]{%
    \noindent\makebox[0pt][r]{\@thefnmark\,}#1}
\makeatother

\title{Click2Mask: Local Editing with Dynamic Mask Generation}

\author {
    Omer Regev,
    Omri Avrahami,
    Dani Lischinski
}
\affiliations {
    The Hebrew University of Jerusalem
}

\begin{document}
\maketitle

\newcommand{\red    }   [1]     {\textcolor{red}{#1}    }
\newcommand{\un     }   [1]     {_\textit{#1}           } 
\newcommand{\DALLE  }   [1]     {{DALL$\cdot$E$\!$~{#1}}}

\definecolor{darkred}{rgb}{0.7, 0.1,0.1}
\definecolor{darkgreen}{rgb}{0.1,0.6,0.1}
\definecolor{cyan}{rgb}{0.7,0.0,0.7}
\definecolor{otherblue}{rgb}{0.1,0.4,0.8}
\definecolor{maroon}{rgb}{0.76,.13,.28}
\definecolor{burntorange}{rgb}{0.81,.33,0}

\ifdefined\ShowNotes
  \newcommand{\colornote}[3]{{\color{#1}\textbf{#2} #3\normalfont}}
\else
  \newcommand{\colornote}[3]{}
\fi

\ifdefined\ShowProjectPage
    \newcommand{\projecttext}[1]{#1}
\else
    \newcommand{\projecttext}[1]{}
\fi

\ifdefined\ShowProjectPage
    \newcommand\blfootnote[1]{%
      \begingroup
      \renewcommand\thefootnote{}%
      \footnote{\hspace{0pt}#1}%
      \addtocounter{footnote}{-1}%
      \endgroup
    }
\else
    \newcommand{\blfootnote}[1]{}
\fi

\newcommand {\todo}[1]{\colornote{cyan}{todo:}{#1}}
\newcommand {\dlc}[1]{\colornote{darkred}{Dani:}{#1}}
\newcommand {\omer}[1]{{\color{otherblue}{#1}\normalfont}}
\newcommand {\omri}[1]{\colornote{darkgreen}{Omri:}{#1}}

\newcommand {\reqs}[1]{\colornote{red}{\tiny #1}}
\newcommand {\new}[1]{{\color{red}{#1}}}

\newcommand {\mask}         {M_t}
\newcommand {\potential}    {\Phi}
\newcommand {\thresh}       {\tau}
\newcommand {\prompt}       {p}
\newcommand {\coordinates}  {c}
\newcommand {\point}        {\coordinates}
\newcommand {\img}          {x}
\newcommand {\diffsteps}    {n}
\newcommand {\startstep}    {b}
\newcommand {\stopstep}     {k}
\newcommand {\laststep}     {l}
\newcommand {\est}[1]       {\tilde{#1}}
\newcommand {\eg}           {e.g.}
\newcommand {\ie}           {i.e.}
\newcommand {\etal}         {et al.}
\newcommand{\supl}          {appendix}

\newcommand {\emu}          {Emu Edit}
\newcommand {\mb}           {MagicBrush}
\newcommand {\ipp}          {InstructPix2Pix}
\newcommand {\ctm}          {Click2Mask}
\newcommand {\ctmb}         {\textbf{\ctm}}
\newcommand {\alphaclip}    {Alpha-CLIP}

\newcommand{\pw}                {40pt}
\newcommand{\prsize}            {scriptsize}
\newcommand{\methsize}          {scriptsize}
\newcommand{\defprsize}         {scriptsize}
\newcommand{\defmethsize}       {scriptsize}
\newcommand{\defbigprsize}      {footnotesize}
\newcommand{\defbigmethsize}    {footnotesize}
\newcommand{\ablsize}           {scriptsize}

\newcommand{\defrss}            {23pt}
\newcommand{\defcapspace}       {0px}
\newlength{\capspace}           {}
\newcommand{\figtitle}          {}
\newlength{\rss}                {}
\newlength{\rsm}                {}
\newlength{\rsb}                {}

\newcommand{\sizedtext}[2]{%
  \csname#1\endcsname{#2}%
}
    
\newcommand{\abltitlesplit}[2]{
  \rotatebox[origin=c]{90}{
    \begin{tabular}{>{\centering\arraybackslash}m{35pt}} %
    \scriptsize{{#1}}
    \\
    \scriptsize{{#2}}
    \end{tabular}
    }
}

\newcommand{\abltitle}[2]{
  \rotatebox[origin=c]{90}{\sizedtext{#1}{#2}}
}

\newcommand{\pr}[1]{
\begin{tabular}{>{\centering\arraybackslash}m{40pt}} %
  \textit{``#1''}
\end{tabular} &
}

\newcommand{\prtwolines}[3]{
\begin{tabular}[t]{@{}c@{}} 
    \sizedtext{#1}{``\textit{#2}}\\ \sizedtext{#1}{\textit{#3}"}
\end{tabular}%
}

\newcommand{\prs}[3][40pt]{
\begin{tabular}{>{\centering\arraybackslash}m{#1}} %
  \sizedtext{#2}{\textit{``#3''}}
\end{tabular} 
}

\newcommand{\pruntiny}[1]{
\begin{tabular}{>{\centering\arraybackslash}m{40pt}} %
  \tiny{\textit{``#1''}}
\end{tabular} 
}

\newcommand{\pruns}[3]{
\begin{tabular}{>{\centering\arraybackslash}m{#1}} %
  \sizedtext{#2}{\textit{``#3''}}
\end{tabular} 
}

\newcommand{\prpr}[2]{
\begin{tabular}{>{\centering\arraybackslash}m{50pt}} %
  \small
  \textit{``#1''} 
  \\ \\ \hline\\
  \small
  \textit{``#2''}
\end{tabular} &
}

\newcommand{\prprs}[4][50pt]{
\begin{tabular}{>{\centering\arraybackslash}m{#1}} %
    \sizedtext{#2}{\textit{``#3''}}
    \\ \\ \hline\\
    \sizedtext{#2}{\textit{``#4''}}
\end{tabular} &
}

\newcommand{\prprti}[2]{
\begin{tabular}{>{\centering\arraybackslash}m{50pt}} %
  \tiny
  \textit{``#1''} 
  \\ \\ \hline\\
  \tiny
  \textit{``#2''}
\end{tabular} &
}

\newcommand{\prunmergerule}[4]{
    \multicolumn{#2}{c}{\rule{0pt}{#3ex}\sizedtext{#1}{\textit{\textit{``#4''}}}}
}

\newcommand{\prunmerge}[3]{
    \multicolumn{#2}{c}{\sizedtext{#1}{\textit{``#3''}}}
}

\newcommand{\prprunmerge}[4]{
   \prunmerge{#1}{#2}{#3}  
   \\
   \prunmerge{#1}{#2}{#4}
}

\newcommand{\prprendline}[2]{
\begin{tabular}{>{\centering\arraybackslash}m{50pt}} %
  \small
  \textit{``#1''} 
  \\ \\ \hline\\
  \small
  \textit{``#2''}
\end{tabular} 
}

\begin{abstract}
  Recent advancements in generative models have revolutionized image generation and editing, making these tasks accessible to non-experts. This paper focuses on local image editing, particularly the task of adding new content to a loosely specified area. Existing methods often require a precise mask or a detailed description of the location, which can be cumbersome and prone to errors. We propose Click2Mask, a novel approach that simplifies the local editing process by requiring only a single point of reference (in addition to the content description). A mask is dynamically grown around this point during a Blended Latent Diffusion (BLD) process, guided by a masked CLIP-based semantic loss. Click2Mask surpasses the limitations of segmentation-based and fine-tuning dependent methods, offering a more user-friendly and contextually accurate solution. Our experiments demonstrate that Click2Mask not only minimizes user effort but also enables competitive or superior local image manipulations compared to SoTA methods, according to both human judgement and automatic metrics. Key contributions include the simplification of user input, the ability to freely add objects unconstrained by existing segments, and the integration potential of our dynamic mask approach within other editing methods.
\end{abstract}

\blfootnote{\textbf{Project page: } \textcolor{purple}{\url{https://omeregev.github.io/click2mask}}
\\[0.5em]
This is a preprint of the paper accepted at \textit{\textbf{AAAI 2025}}
\label{projectpage}}

\section{Introduction}
\label{sec:introduction}
Recent advances in generative models have revolutionized image generation and editing capabilities, enabling both streamlined workflows and accessibility for non-experts. The latest approaches utilize natural language to manipulate images either globally -- altering the content or style of the entire image -- or locally -- adding, removing, or modifying specific objects within a limited image region.

\begin{figure}[ht]
    \centering
    \setlength{\tabcolsep}{0.5pt}
    \renewcommand{\arraystretch}{0.5}
    \setlength{\ww}{0.195\columnwidth}
    \renewcommand{\prsize}{\defprsize}
    \renewcommand{\methsize}{\defmethsize}
    \setlength{\rss}{1pt}
    \setlength{\capspace}{\defcapspace}
    
    \begin{tabular}{c cccc}
        \sizedtext{\methsize}{Input} &
        \sizedtext{\methsize}{\emu} &
        \sizedtext{\methsize}{\mb} &
        \sizedtext{\methsize}{\DALLE 3} &
        \sizedtext{\methsize}{\ctmb} 
        \\
        [\rss]
        
        \raisebox{-.5\height}{\includegraphics[width=\ww]{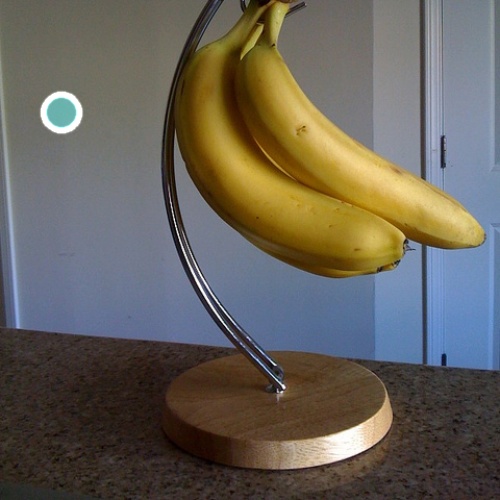}} &
        \raisebox{-.5\height}{\includegraphics[width=\ww]{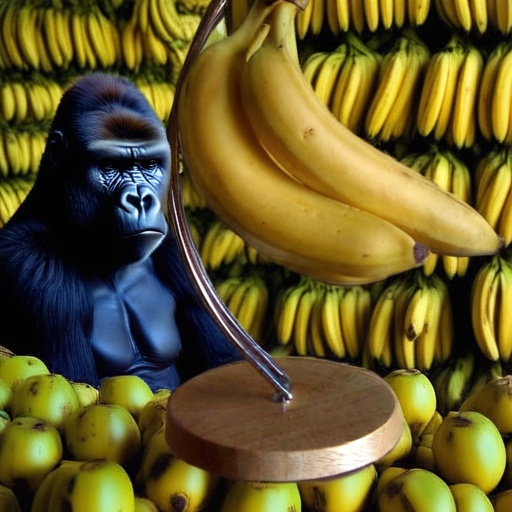}} &
        \raisebox{-.5\height}{\includegraphics[width=\ww]{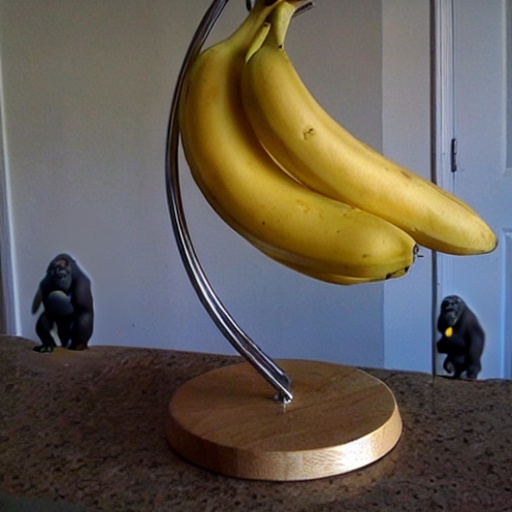}} &
        \raisebox{-.5\height}{\includegraphics[width=\ww]{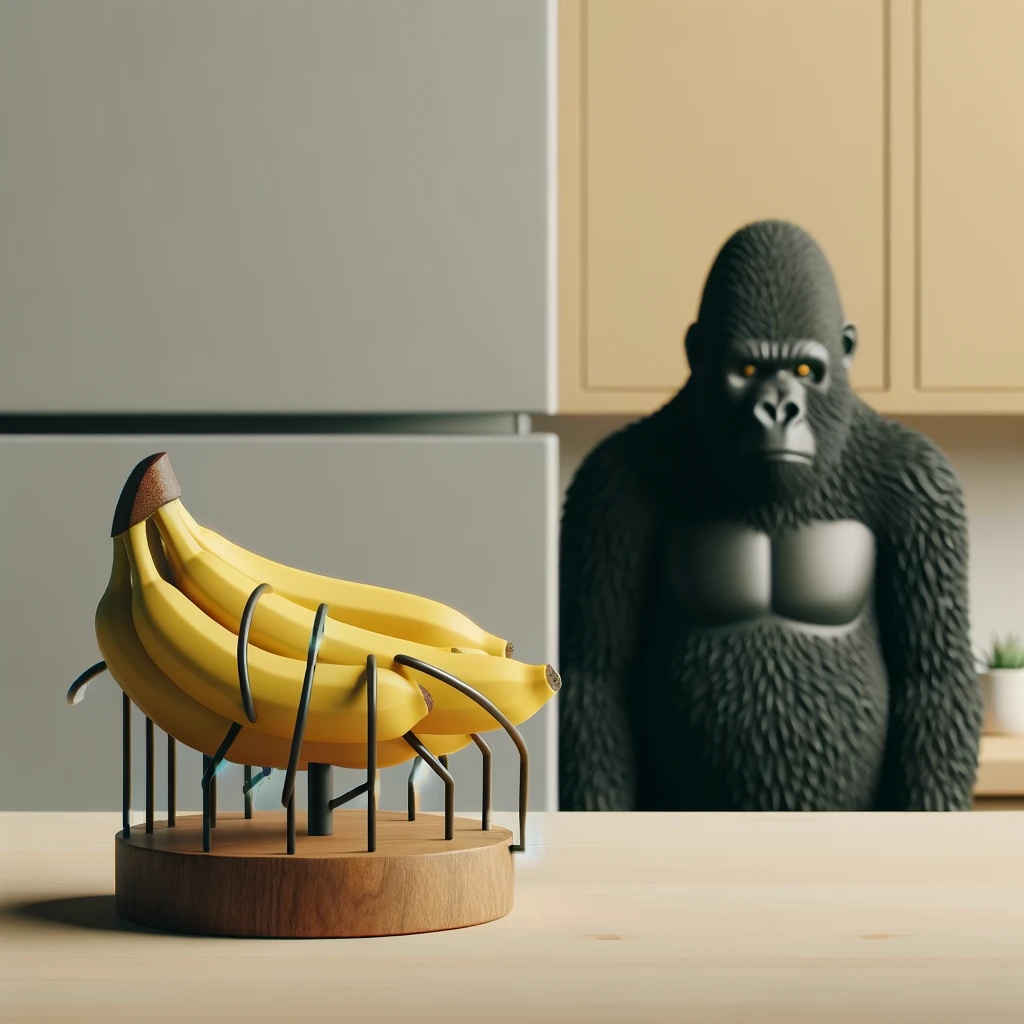}} &
        \raisebox{-.5\height}{\includegraphics[width=\ww]
        {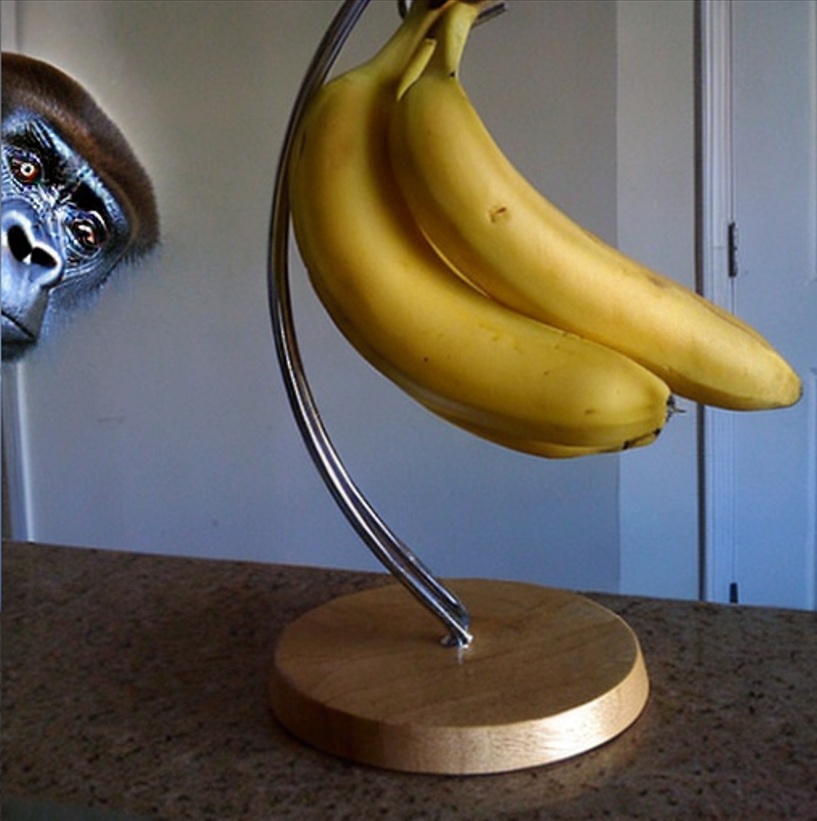}}  
        \\
        \prunmergerule{\prsize}{5}{1.5}{Add a gorilla in the background looking at the bananas}
        \\ 
        \prunmergerule{\prsize}{5}{1.5}{A gorilla looking at the bananas}
        \\
        \\
        \raisebox{-.5\height}{\includegraphics[width=\ww]{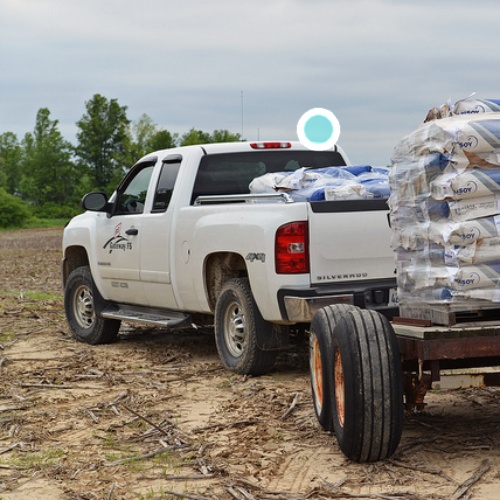}} &
        \raisebox{-.5\height}{\includegraphics[width=\ww]{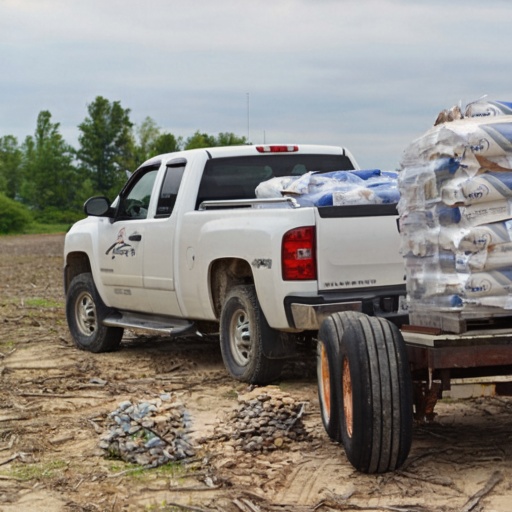}} &
        \raisebox{-.5\height}{\includegraphics[width=\ww]{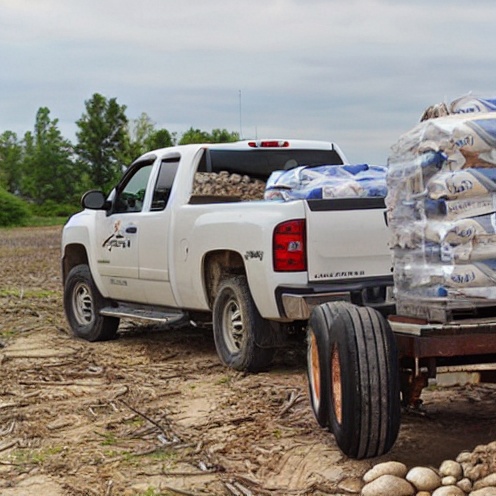}} &
        \raisebox{-.5\height}{\includegraphics[width=\ww]{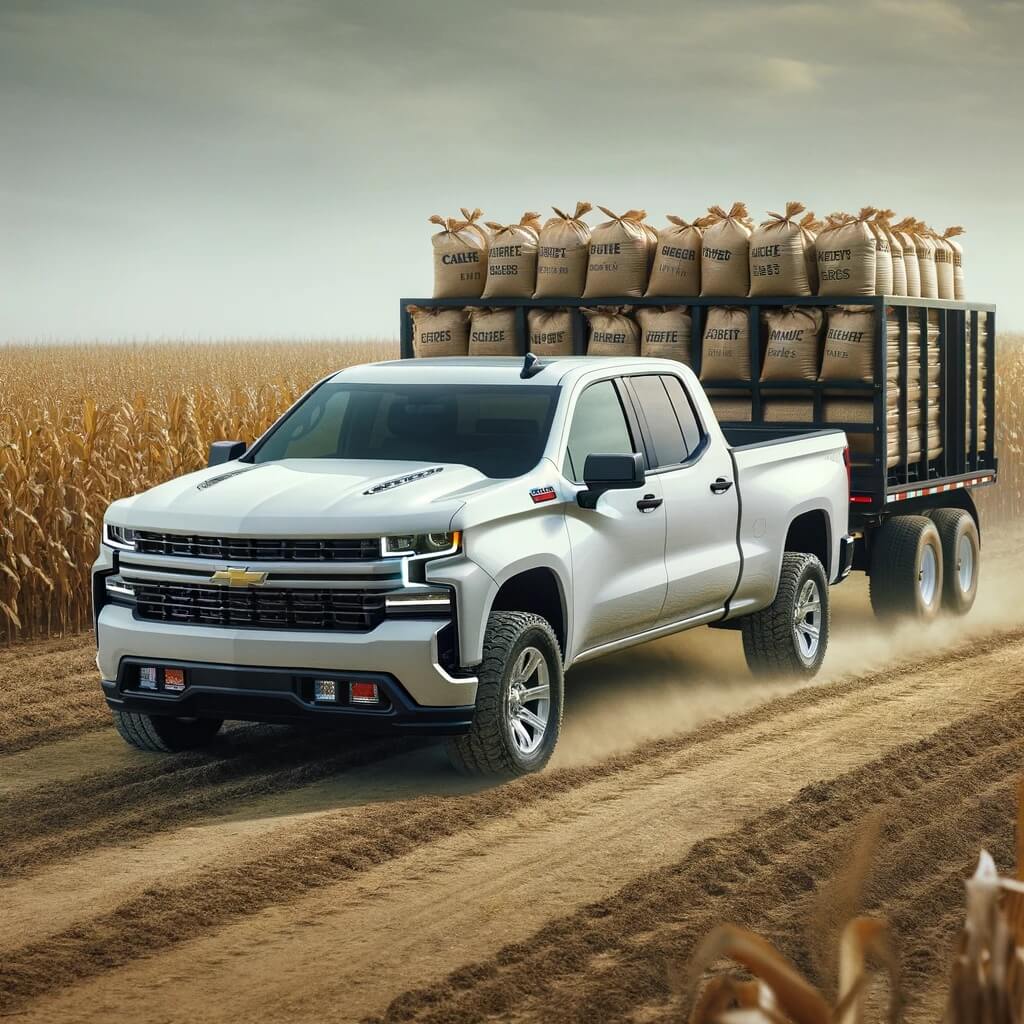}} &
        \raisebox{-.5\height}{\includegraphics[width=\ww]{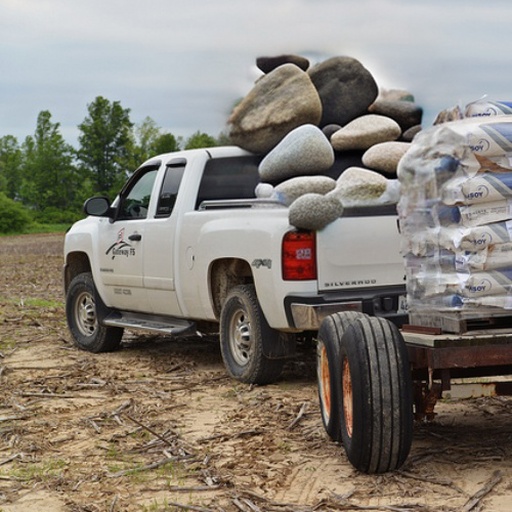}} 
        \\
        \prunmergerule{\prsize}{5}{1.5}{Add a pack of rocks to the back of the truck}
        \\
        \prunmergerule{\prsize}{5}{1.5}{A pack of rocks}

    \end{tabular}
  \captionof{figure}{\textbf{Comparisons to SoTA models.} A comparison of \emu\ \cite{sheynin2023emu}, \mb\ \cite{Zhang2023MagicBrush} and \DALLE{3} \cite{BetkerImprovingIG} with our model \ctmb. In each example, the top prompt was given to the other models, while \ctm\ received the simpler bottom prompt, in addition to the blue dot (mouse click) on the input. Other models completely change the image, or the background, fail to edit, or produce unrealistic results.}
  \label{fig:teaserfigure}
  \vspace{\capspace}
\end{figure}

\begin{figure*}[t]
    \centering
    \setlength{\tabcolsep}{0pt}
    \renewcommand{\arraystretch}{0.5}
    \setlength{\ww}{0.099\linewidth}
    \setlength{\rsb}{1cm}
    \setlength{\rsm}{1cm}
    \renewcommand{\prsize}{\defprsize}
    \renewcommand{\methsize}{\defmethsize}
    \setlength{\capspace}{\defcapspace} 
    
    \begin{tabular}{c c @{\hspace{0.01\columnwidth}}ccccccc @{\hspace{0.01\columnwidth}}c}
        \prs{\prsize}{A giraffe} &
        \raisebox{-.5\height}{\includegraphics[width=\ww]{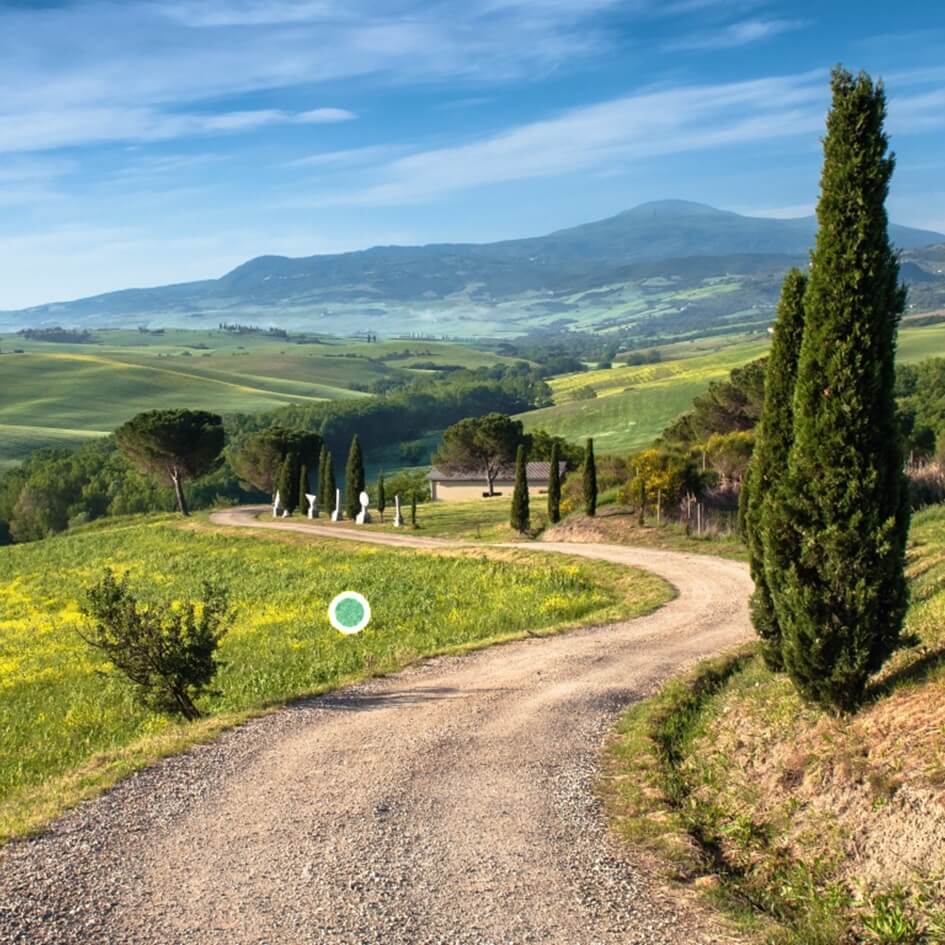}} &
        \raisebox{-.5\height}{\includegraphics[width=\ww]{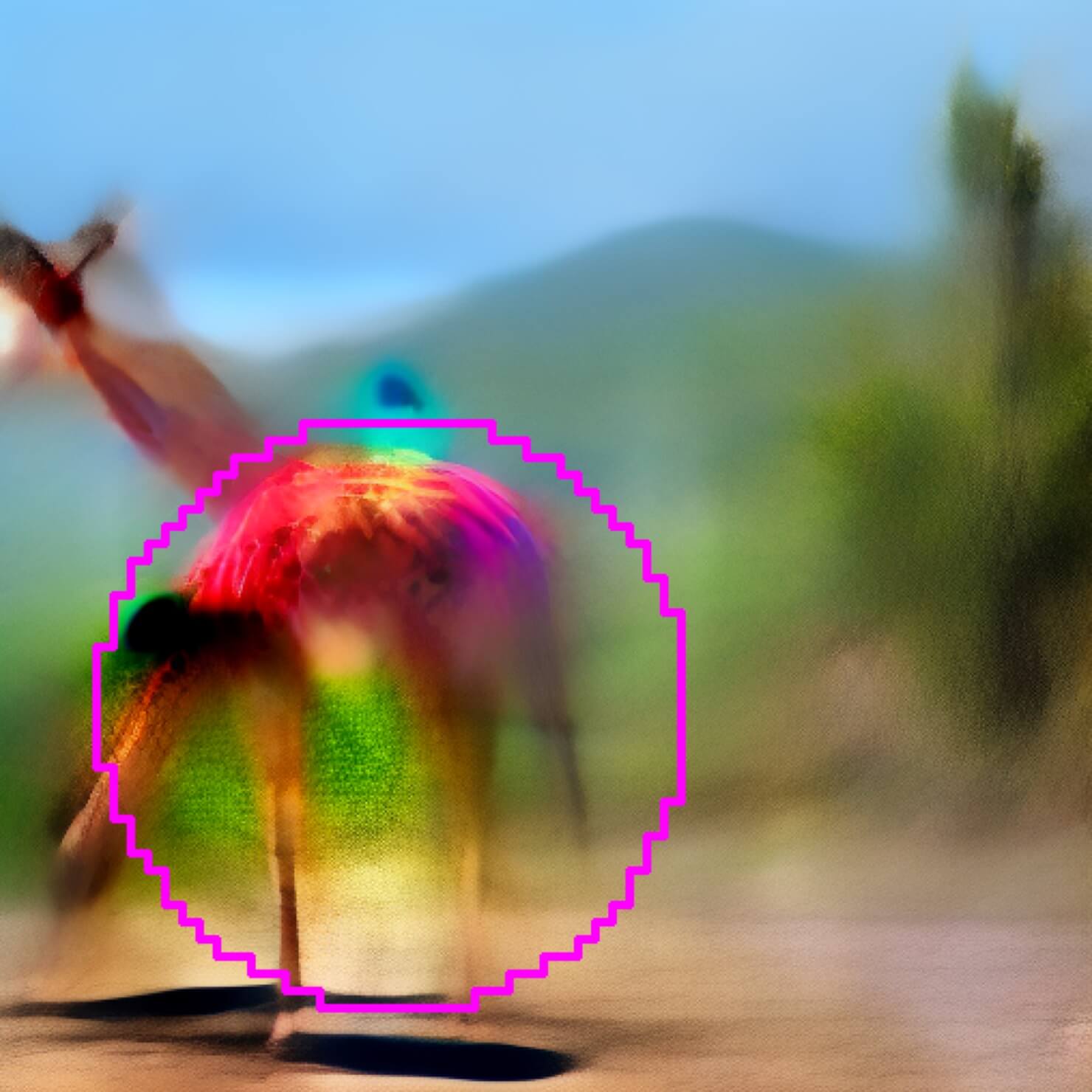}}  &
        \raisebox{-.5\height}{\includegraphics[width=\ww]{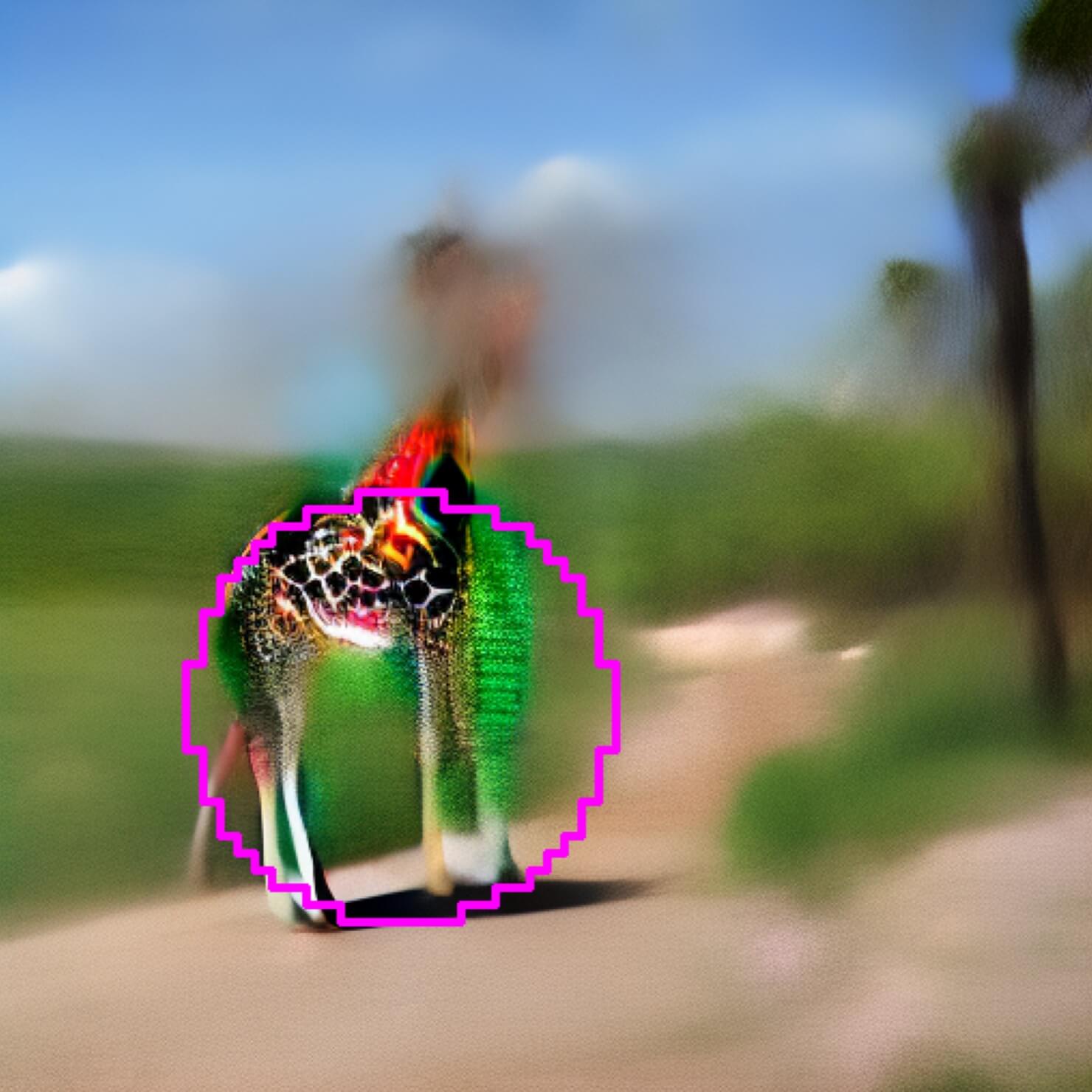}}  &
        \raisebox{-.5\height}{\includegraphics[width=\ww]{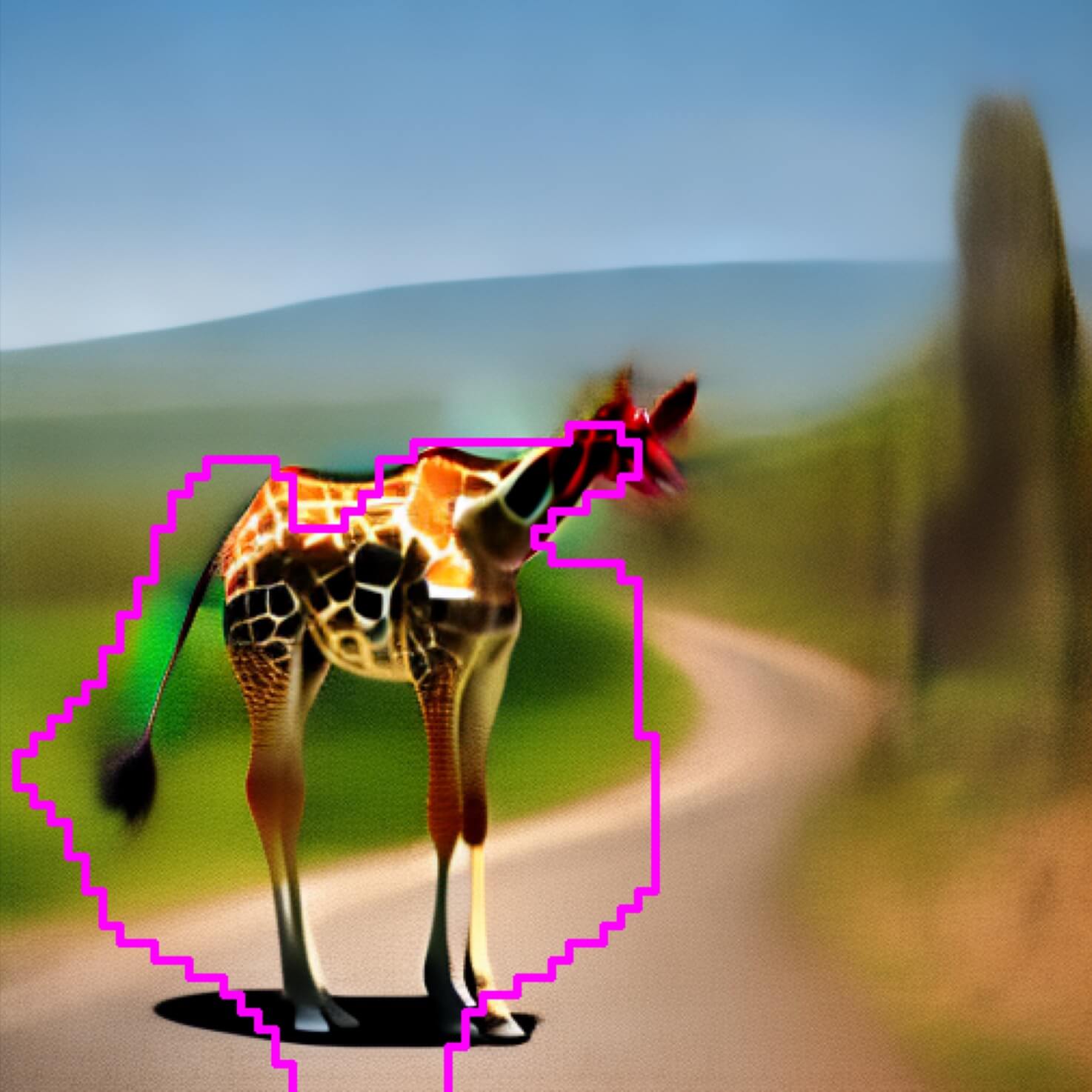}}  &
        \raisebox{-.5\height}{\includegraphics[width=\ww]{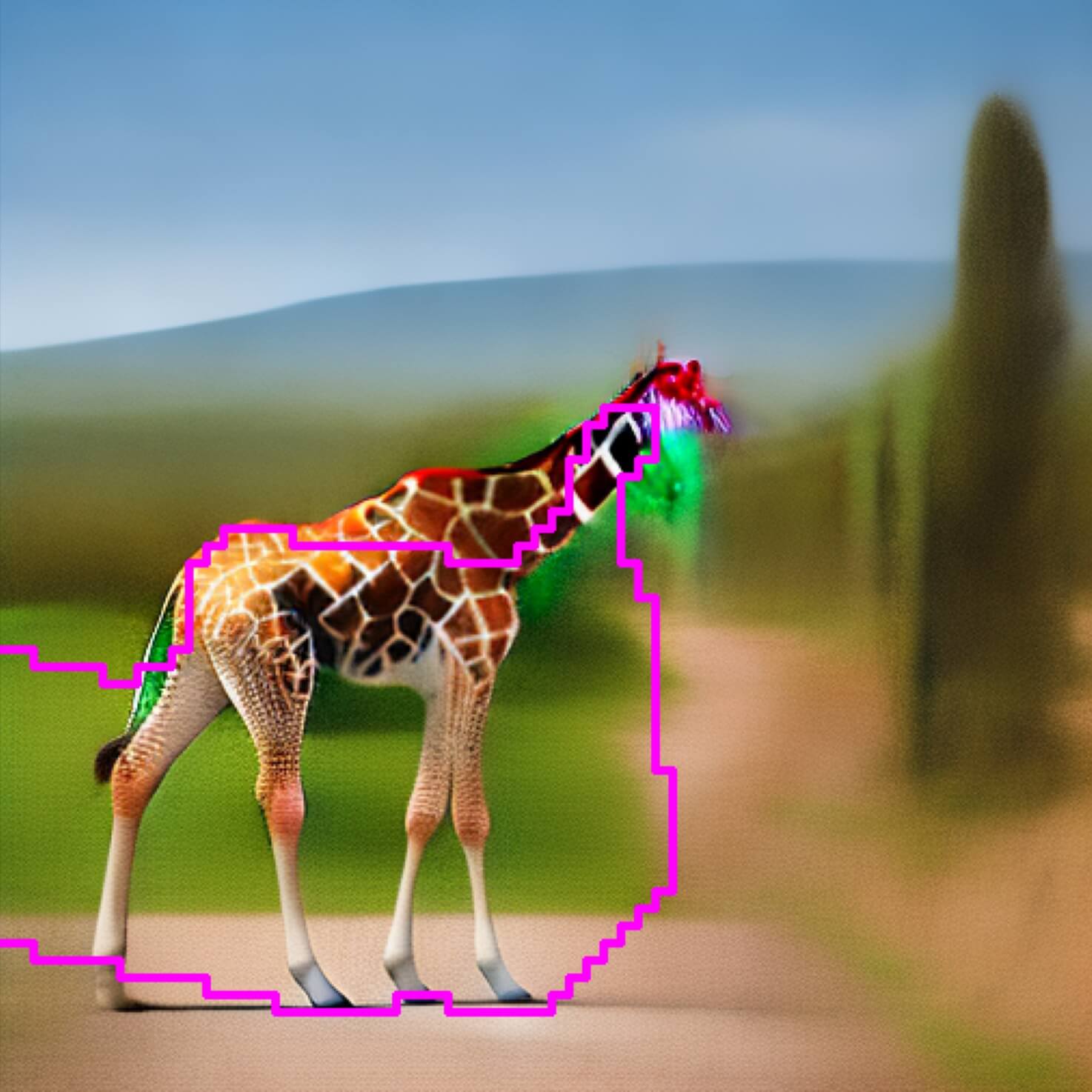}} &
        \raisebox{-.5\height}{\includegraphics[width=\ww]{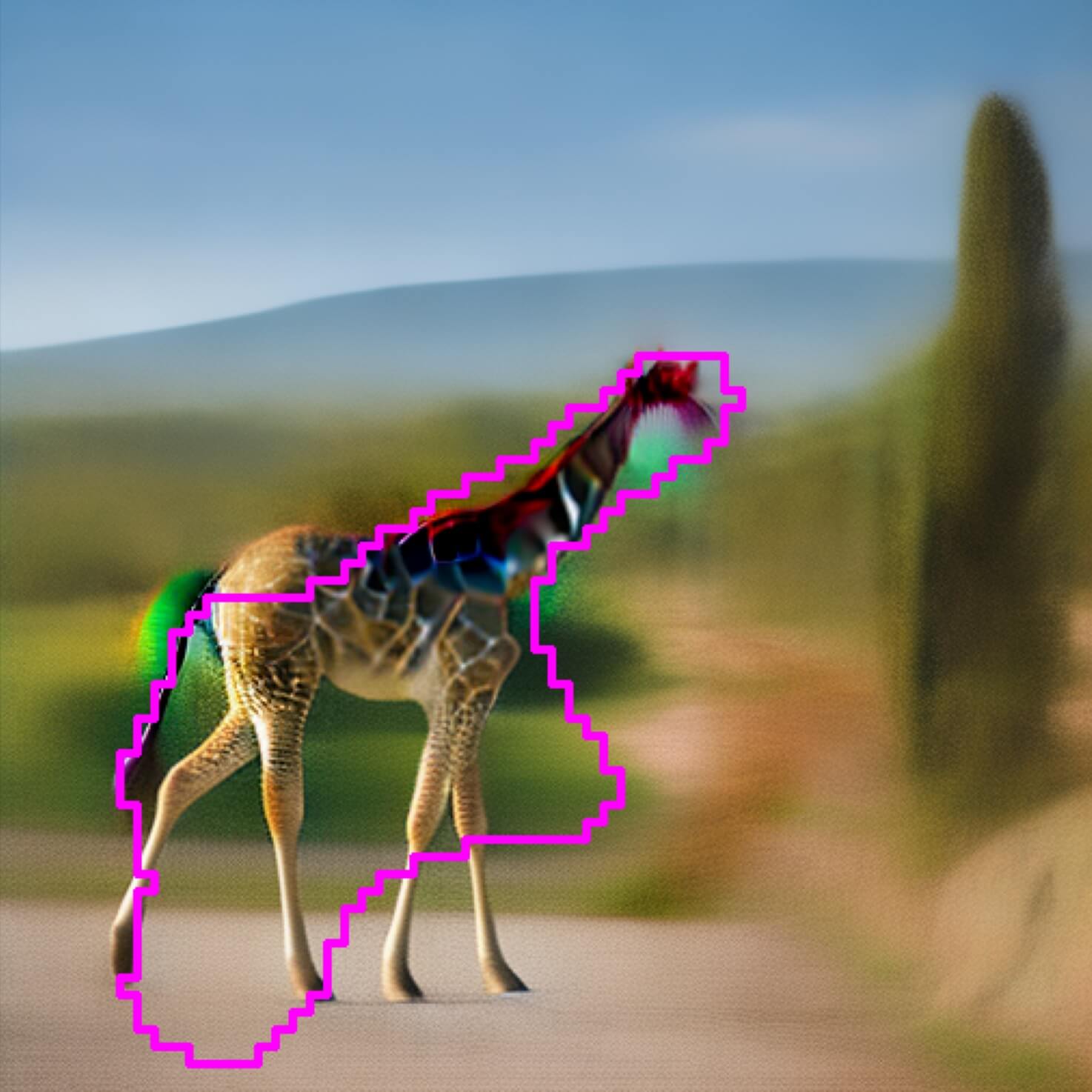}} &
        \raisebox{-.5\height}{\includegraphics[width=\ww]{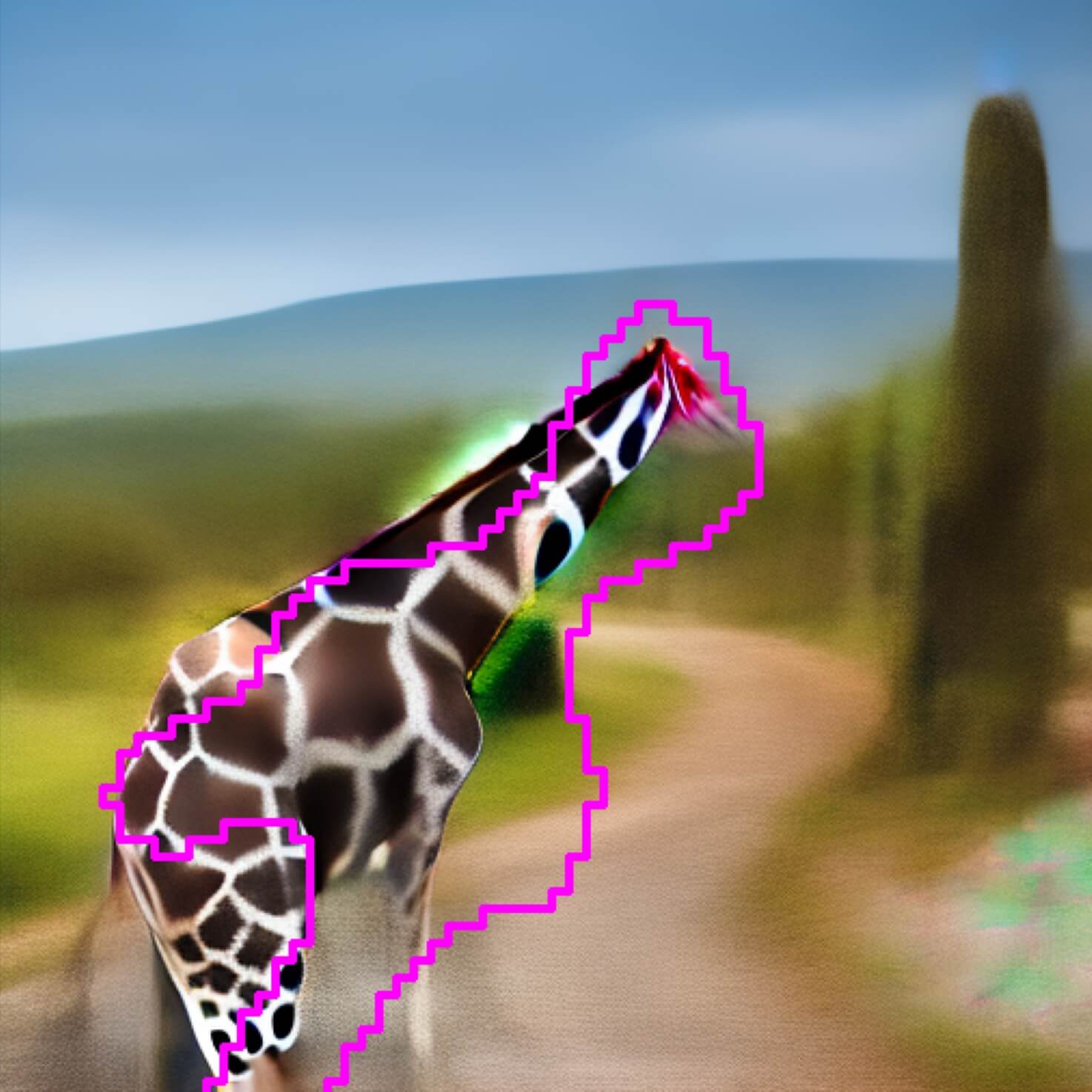}} &
        \raisebox{-.5\height}{\includegraphics[width=\ww]{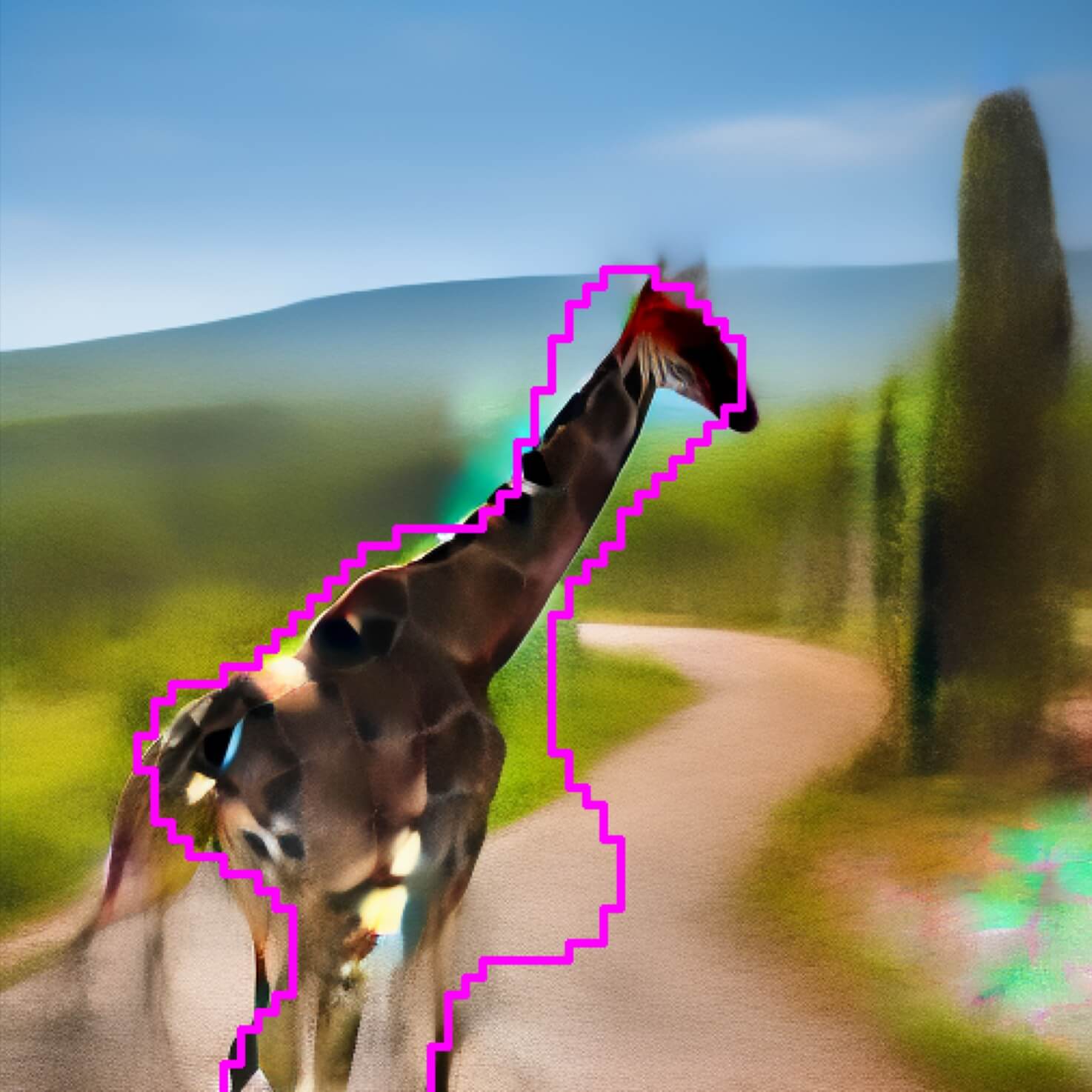}} &
        \raisebox{-.5\height}{\includegraphics[width=\ww]{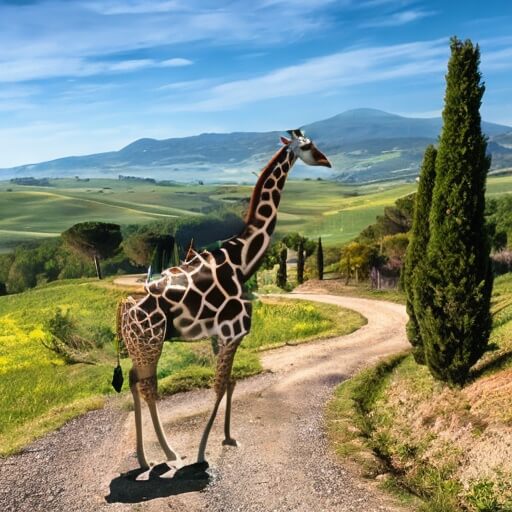}} 
        \\
        [\rsm]

        \prs{\prsize}{Snowy mountains} &
        \raisebox{-.5\height}{\includegraphics[width=\ww]{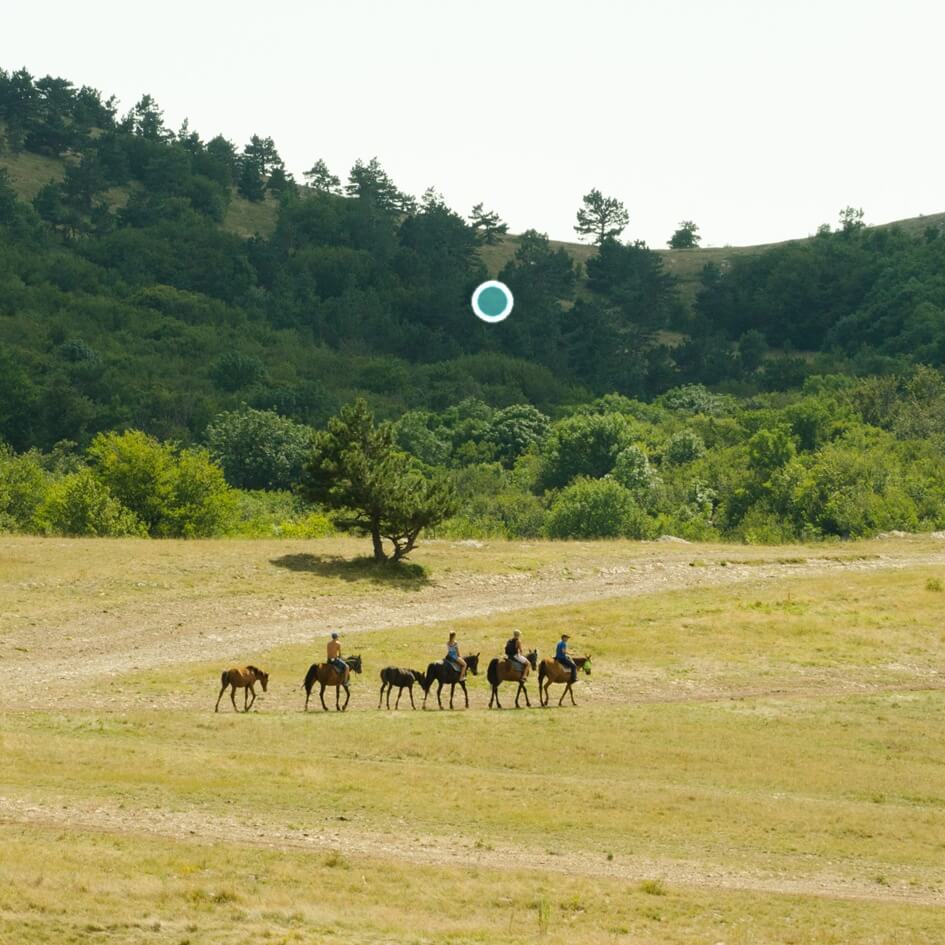}} &
        \raisebox{-.5\height}{\includegraphics[width=\ww]{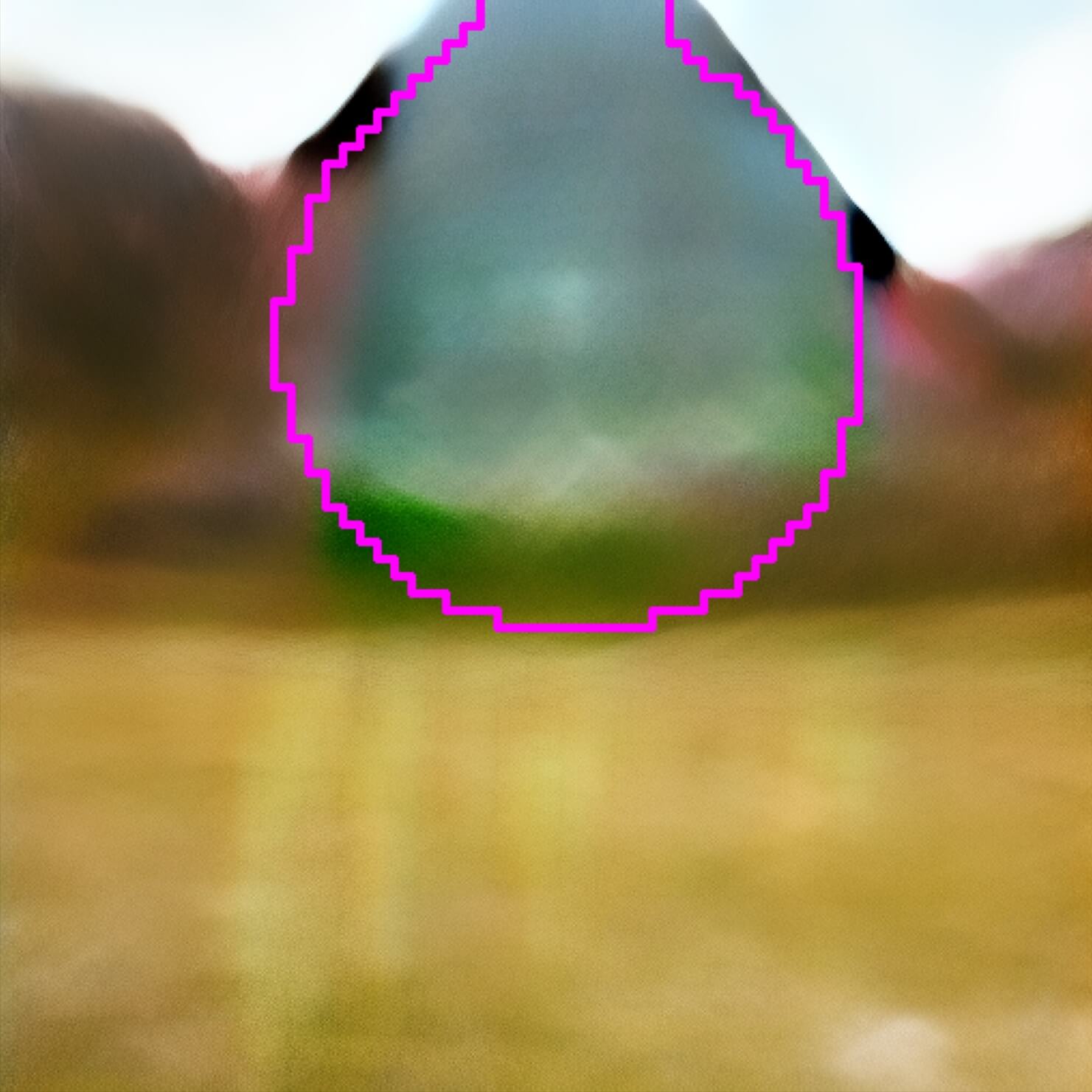}}  &
        \raisebox{-.5\height}{\includegraphics[width=\ww]{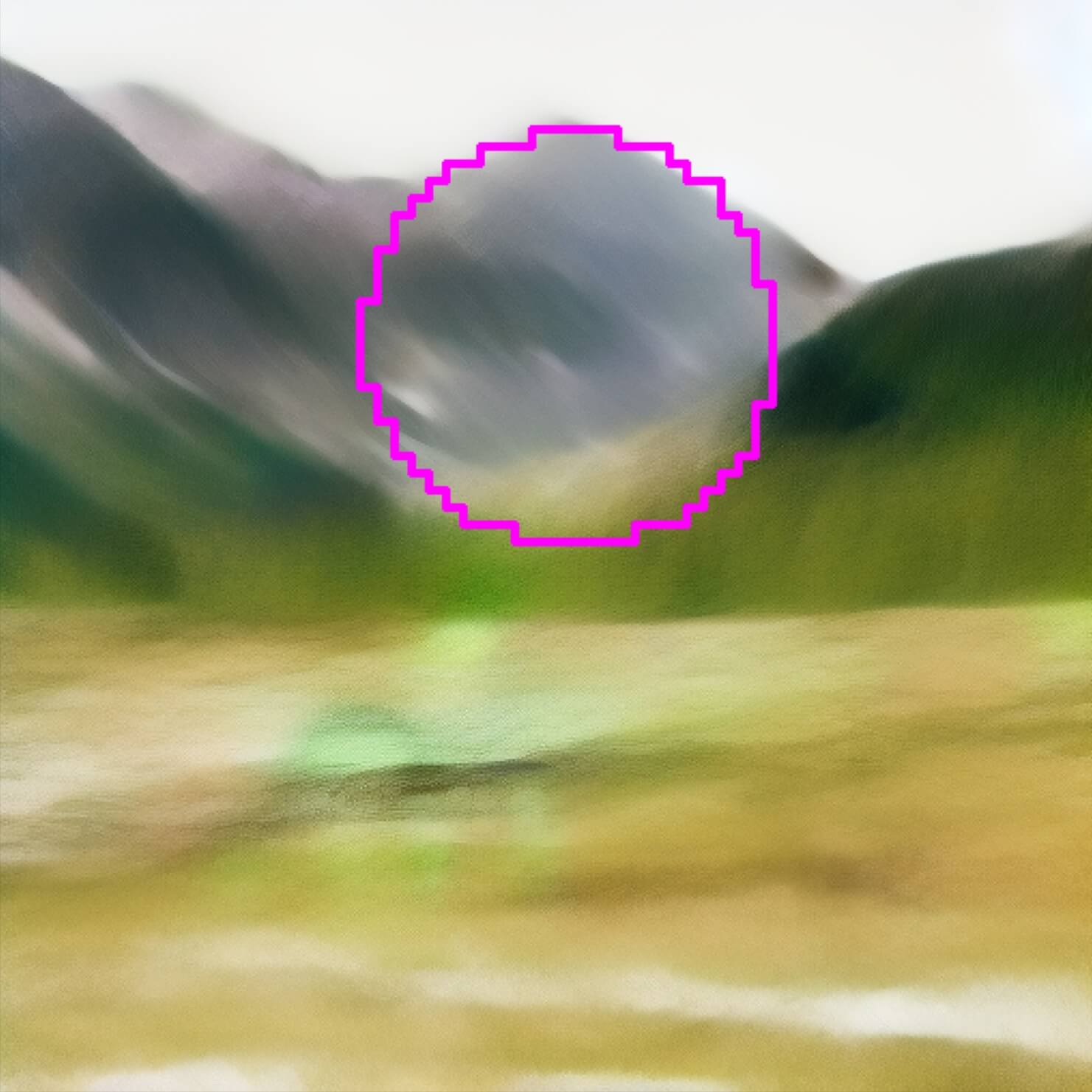}}  &
        \raisebox{-.5\height}{\includegraphics[width=\ww]{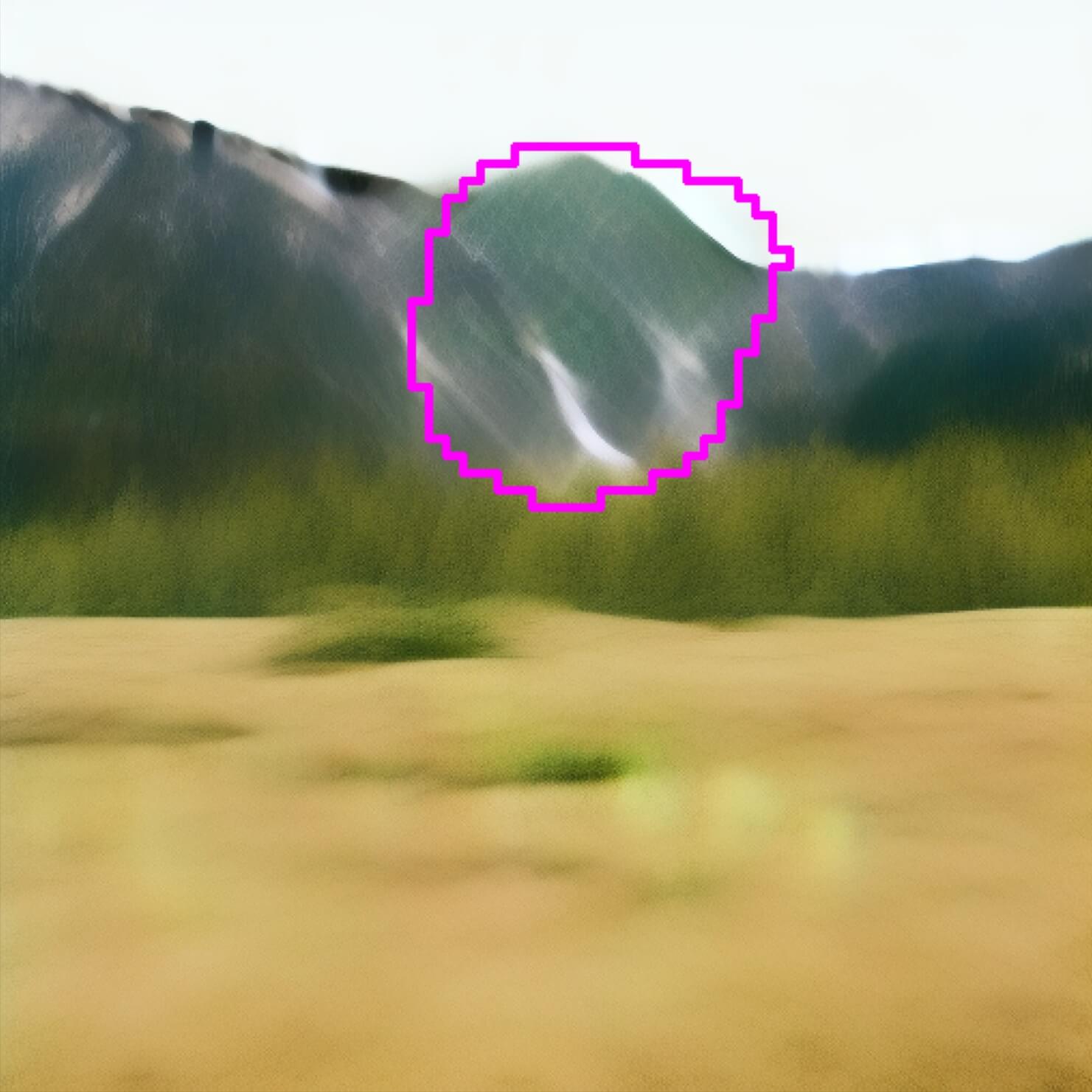}}  &
        \raisebox{-.5\height}{\includegraphics[width=\ww]{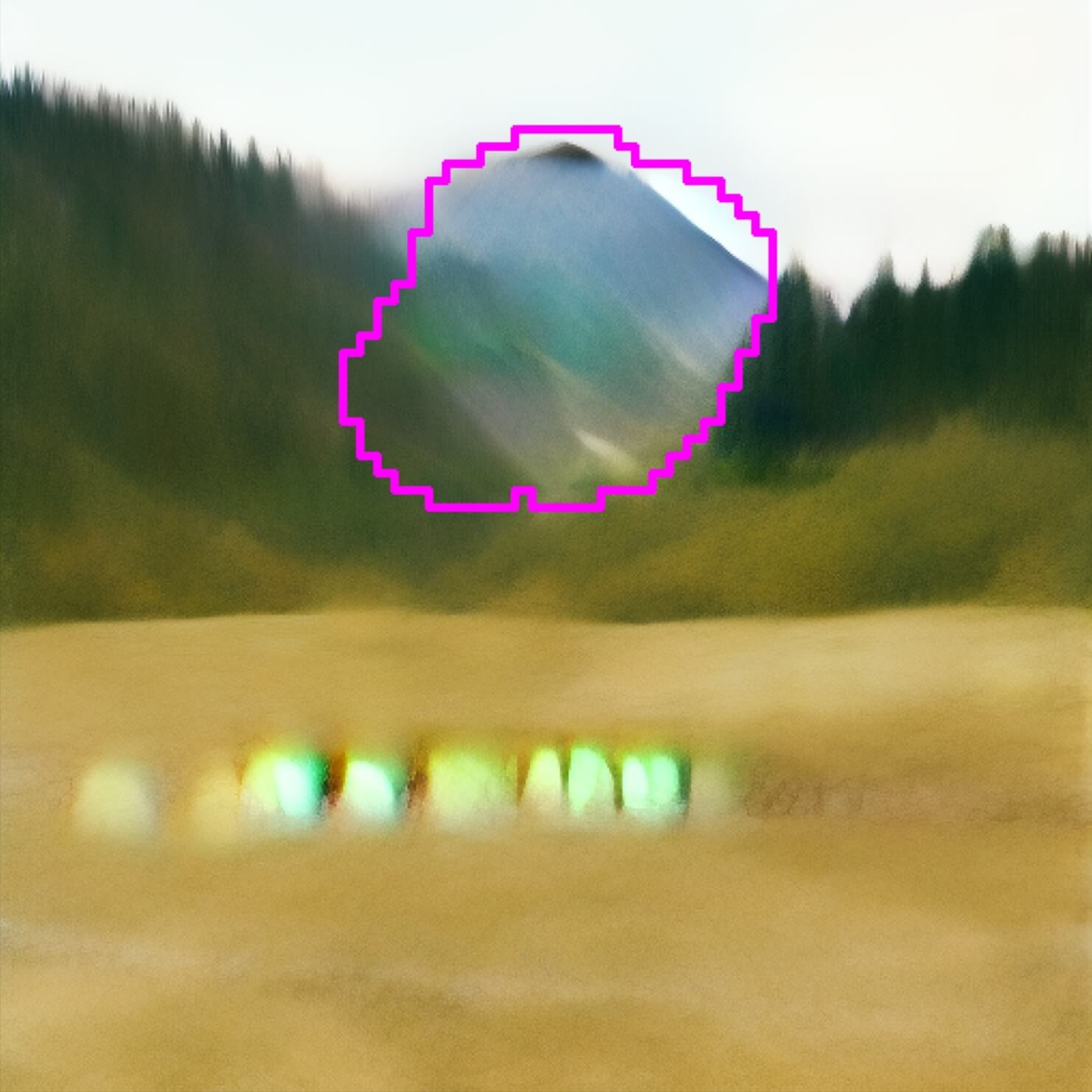}} &
        \raisebox{-.5\height}{\includegraphics[width=\ww]{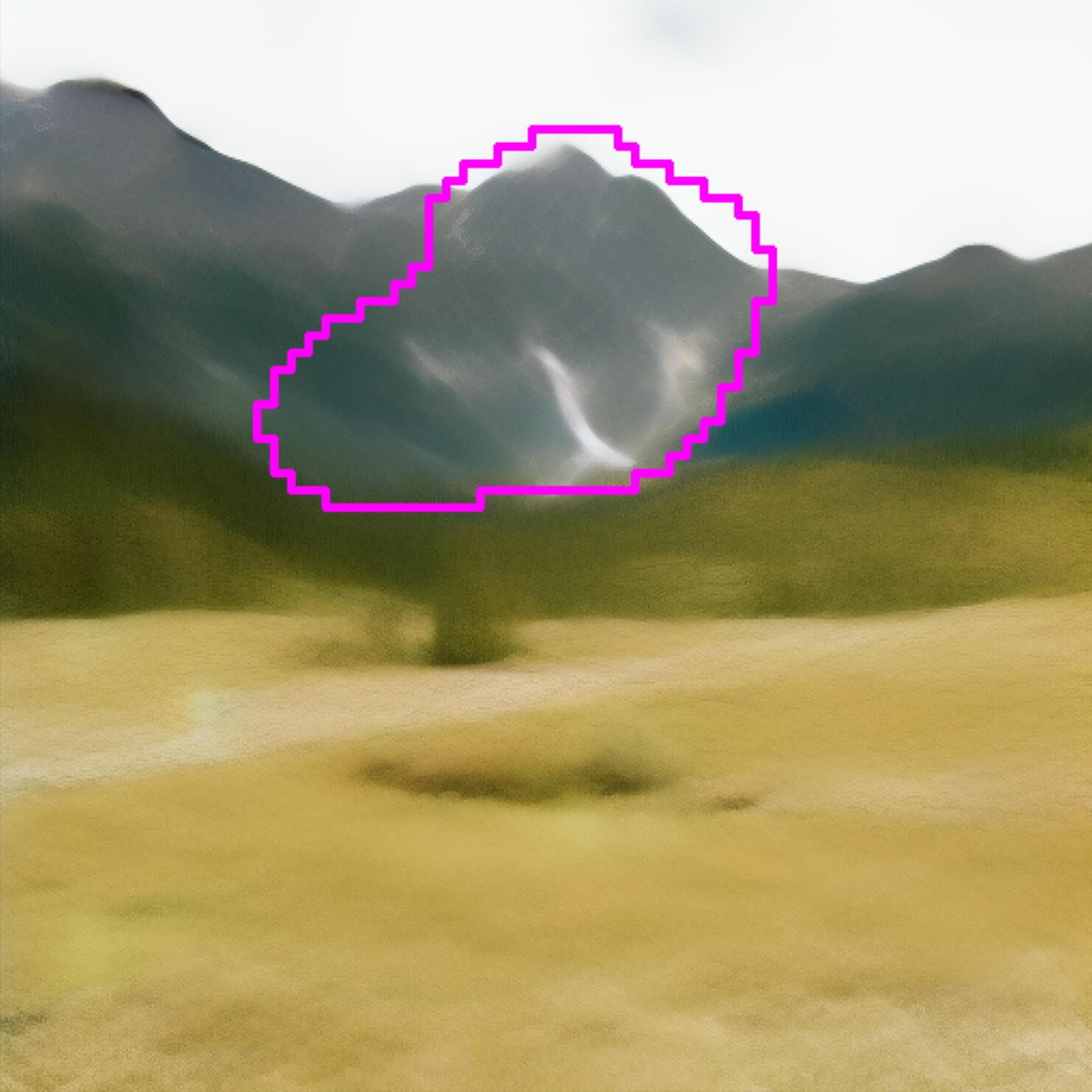}} &
        \raisebox{-.5\height}{\includegraphics[width=\ww]{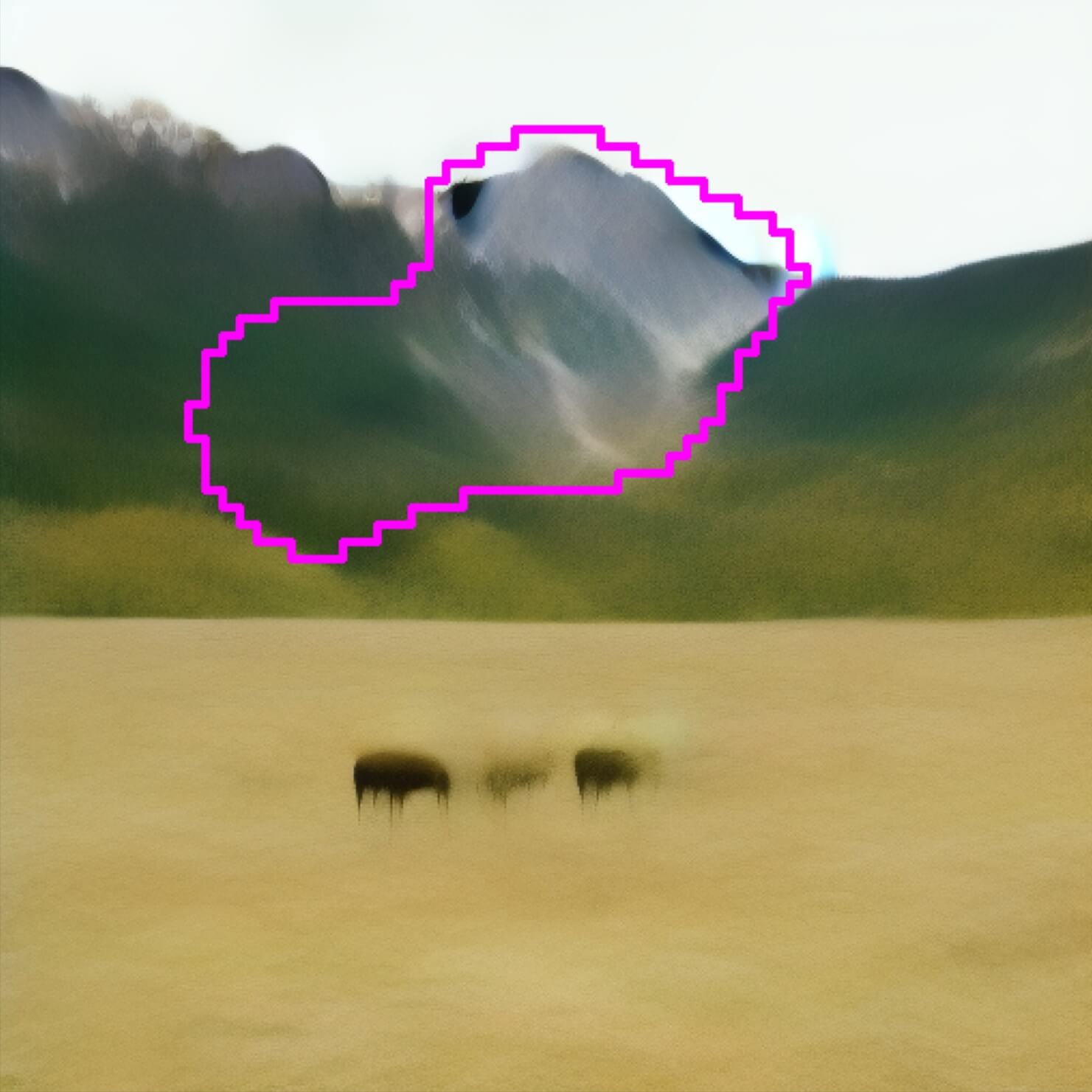}} &
        \raisebox{-.5\height}{\includegraphics[width=\ww]{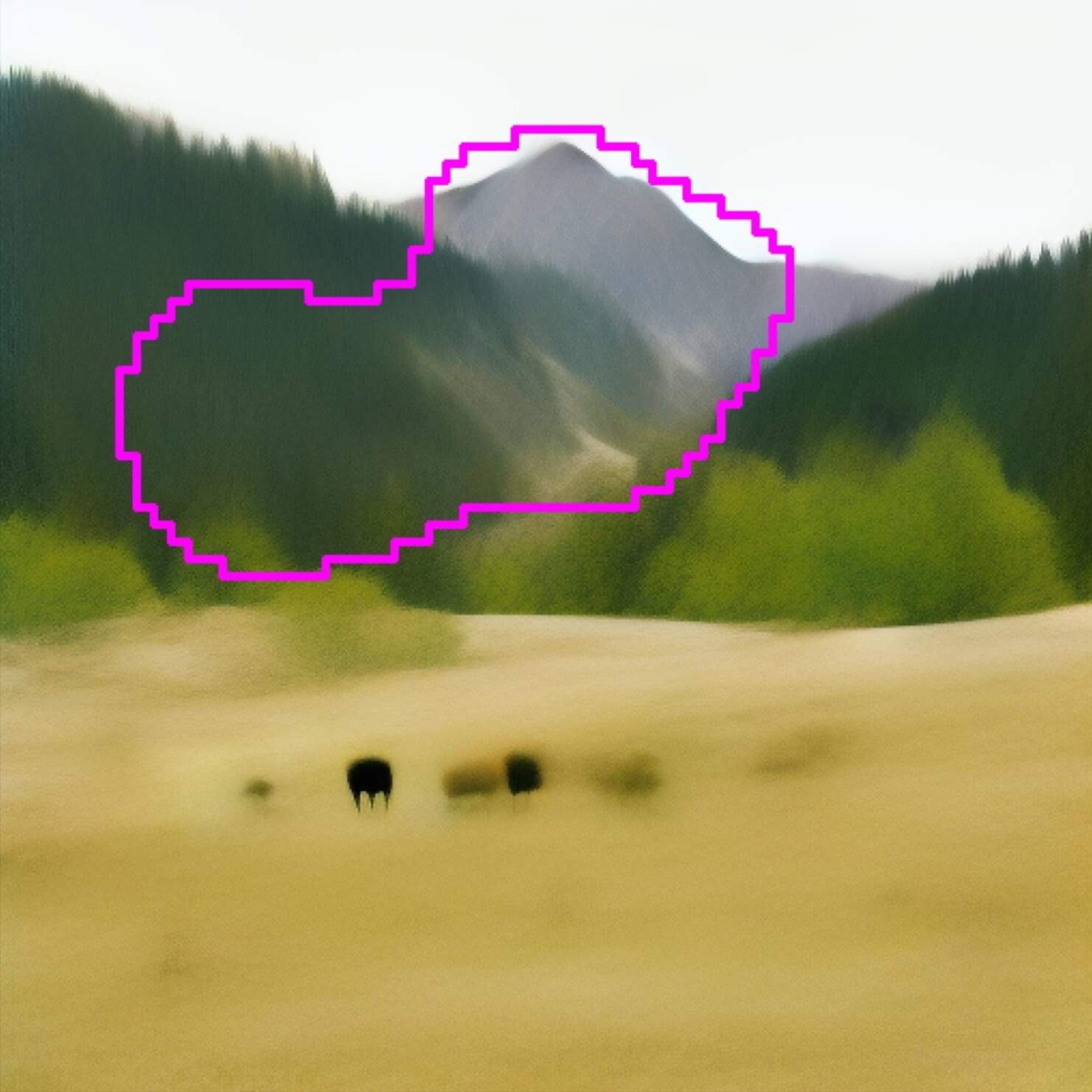}} &
        \raisebox{-.5\height}{\includegraphics[width=\ww]{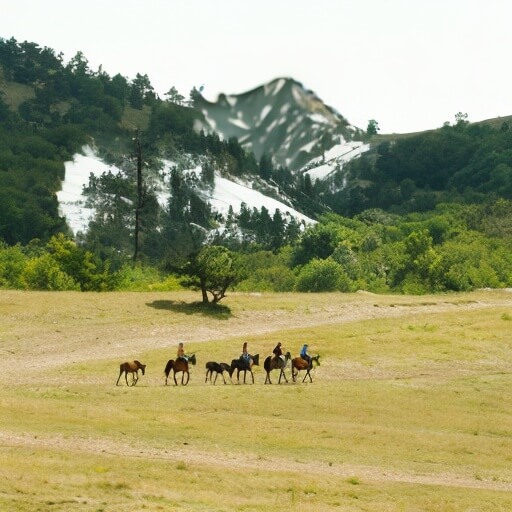}} 
        \\
        [\rsb]
        \sizedtext{\methsize}{Prompt} &  \sizedtext{\methsize}{Input} &  \sizedtext{\methsize}{30\%} &  \sizedtext{\methsize}{37\%} &  \sizedtext{\methsize}{44\%} &  \sizedtext{\methsize}{45\%} &  \sizedtext{\methsize}{46\%} &  \sizedtext{\methsize}{47\%} & 
        \sizedtext{\methsize}{48\%} \scriptsize{(final $\mask$)} &  \sizedtext{\methsize}{Output}
        
    \end{tabular}
    
    \caption{\textbf{Mask evolution}. A visualization of the mask evolution throughout the diffusion process. Leftmost image is input with clicked point, rightmost image is the final \ctm\ output. Intermediate images are decoded latents $\est{z}\un{fg}$ at several diffusion steps, where the purple outline depicts the contour of current (upscaled) mask $\mask$. Percentages indicate the step out of 100 diffusion steps, with the last being the final evolved mask.}
    \label{fig:mask_evolution}
    \vspace{\capspace}
\end{figure*}

In this work, we focus on local editing, specifically on the task of adding new content in a local area. Similar to DragDiffusion \cite{shi2023dragdiffusion} for movement and MagicEraser \cite{li2024magicerasererasingobjectssemanticsaware} for removal, this focused scope leverages specialization to tackle the unique challenges of local editing. To accomplish such edits, some existing methods require users to provide explicit precise masks \cite{Avrahami_2022_CVPR, ramesh2022hierarchical, avrahami2023blendedlatent, wang2023imagen, xie2022smartbrush}, which is tedious and may yield unexpected results due to lack of mask precision.
Other methods describe the desired manipulations in natural language, 
as an edit instruction \cite{brooks2022instructpix2pix, sheynin2023emu}, or by providing a caption and the desired change \cite{bar2022text2live, kawar2023imagic, hertz2022prompt, tumanyan2022plugandplay}. These methods also require user expertise, and their results may suffer from ambiguous or imprecise prompts. Moreover, they fail to ensure that the changes to the image are confined to a local area, or that they occur at all, as demonstrated in \Cref{fig:teaserfigure}.

\begin{figure}[hbt!]
    \centering
    \setlength{\tabcolsep}{0.5pt}
    \renewcommand{\arraystretch}{0.6}
    \setlength{\ww}{0.24\columnwidth}
    \renewcommand{\methsize}{\defmethsize}
    \renewcommand{\prsize}{\defprsize}
    \renewcommand{\pw}{0.95\ww}
    \setlength{\rss}{1cm}
    \setlength{\rsm}{9pt}
    \setlength{\rsb}{16pt}
    \setlength{\capspace}{\defcapspace}

    \begin{tabular}{c c c c}
        \raisebox{-.5\height}{\includegraphics[width=\ww]{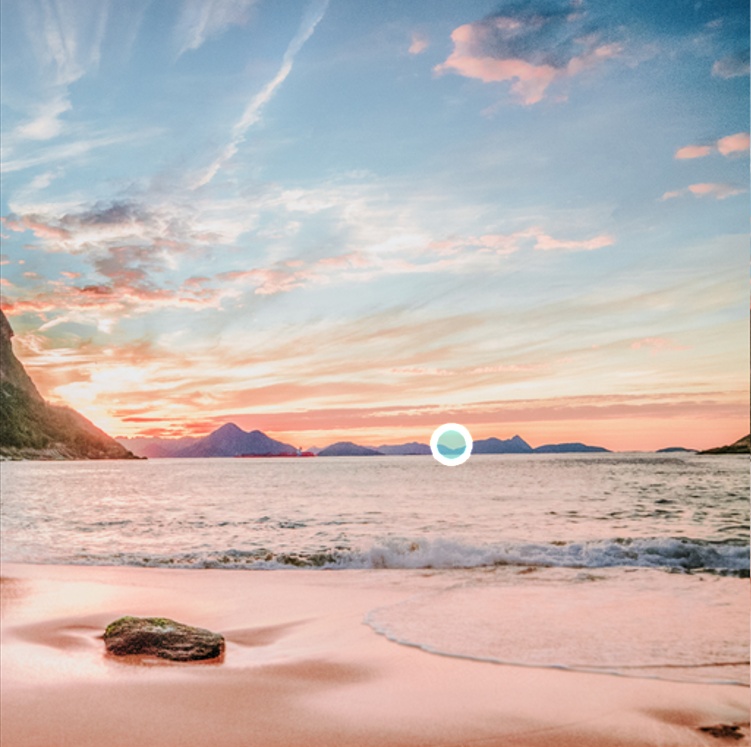}}   &
        \raisebox{-.5\height}{\includegraphics[width=\ww]{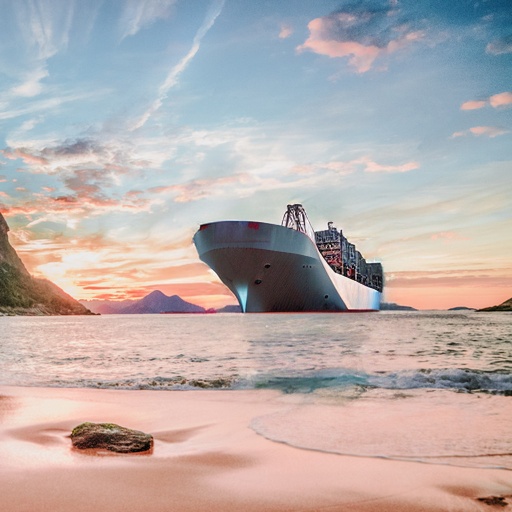}}            &
        \raisebox{-.5\height}{\includegraphics[width=\ww]{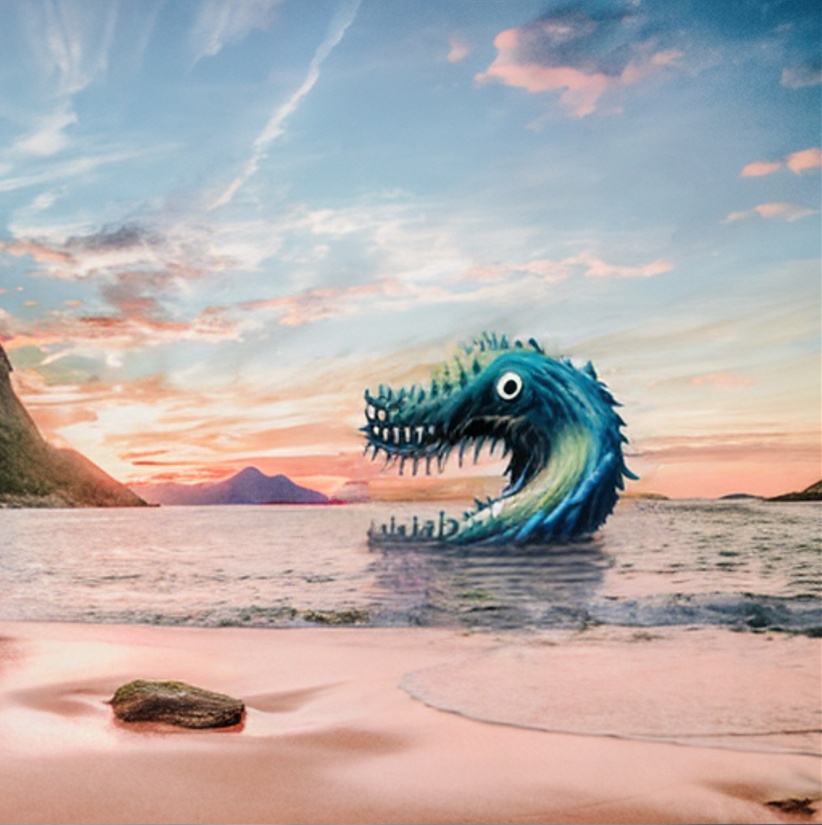}}             &
        \raisebox{-.5\height}{\includegraphics[width=\ww]{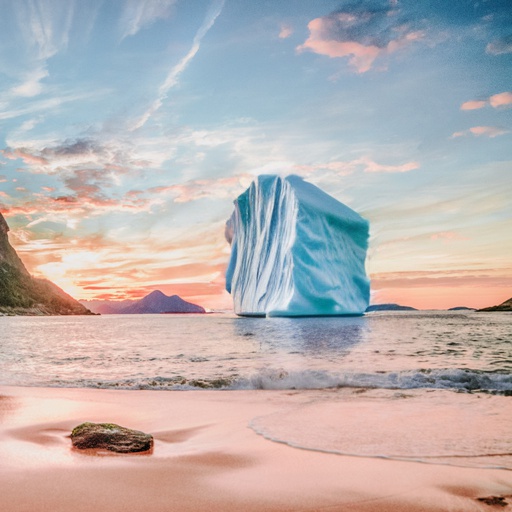}}
        \\
        [\rss]
       
         \sizedtext{\methsize}{Input} & \pruns{\pw}{\prsize}{A big ship} & \pruns{\pw}{\prsize}{A sea monster} & \pruns{\pw}{\prsize}{Iceberg}
        \\
        [\rsm]
        
        \raisebox{-.5\height}{\includegraphics[width=\ww]{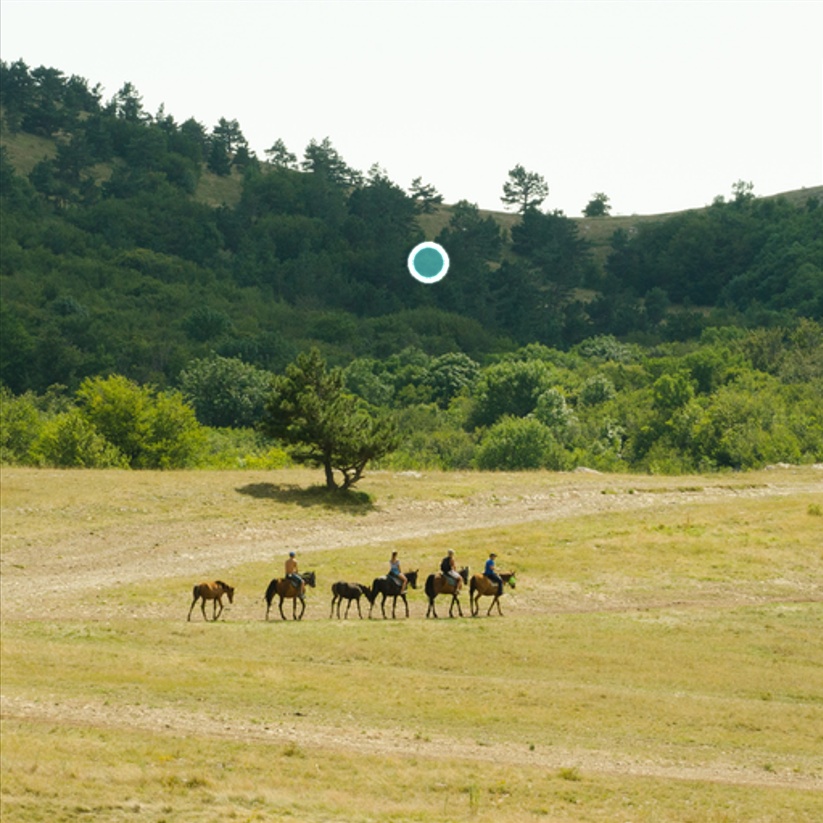}}       &
        \raisebox{-.5\height}{\includegraphics[width=\ww]{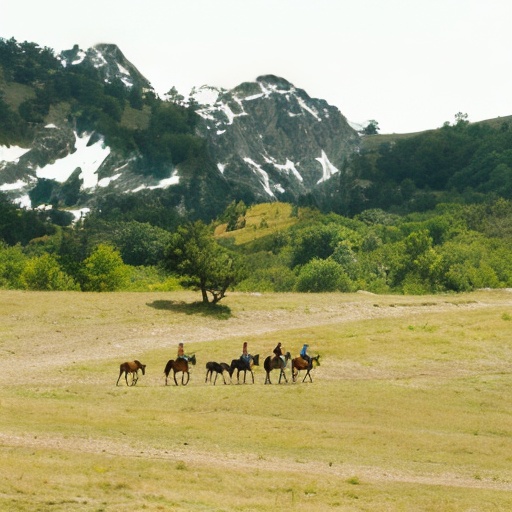}}                &
        \raisebox{-.5\height}{\includegraphics[width=\ww]{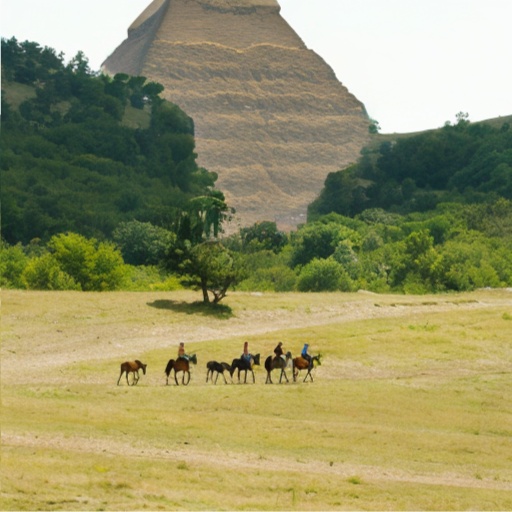}}                &
        \raisebox{-.5\height}{\includegraphics[width=\ww]{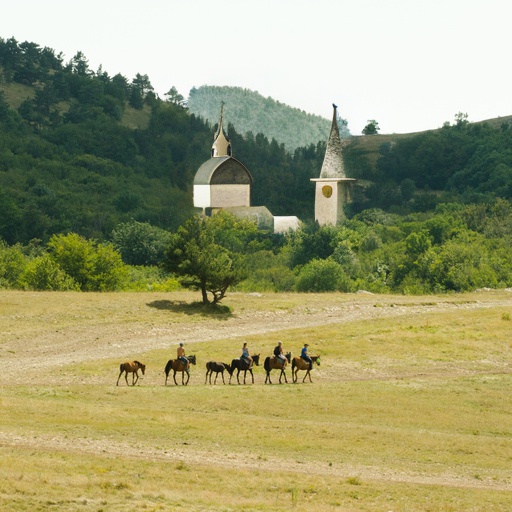}}
        \\
        [\rss]
        
         \sizedtext{\methsize}{Input} & \prtwolines{\prsize}{Snowy}{mountains} & \prtwolines{\prsize}{Great Pyramid}{of Giza} & \pruns{\pw}{\prsize}{A town}
        \\
        [\rsb]
        
        \raisebox{-.5\height}{\includegraphics[width=\ww]{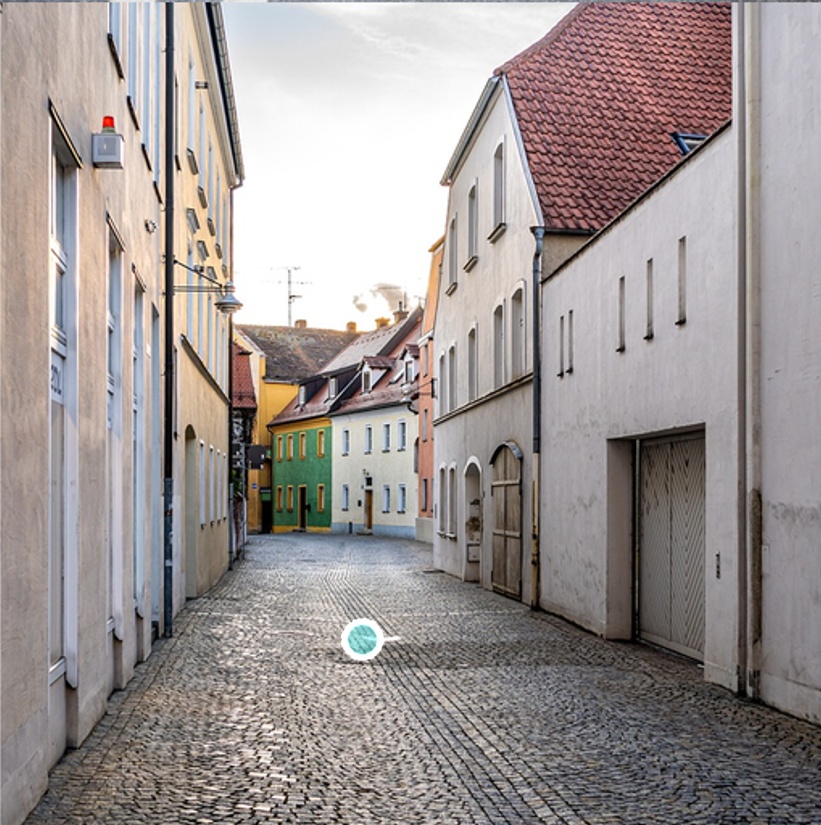}}     &
        \raisebox{-.5\height}{\includegraphics[width=\ww]{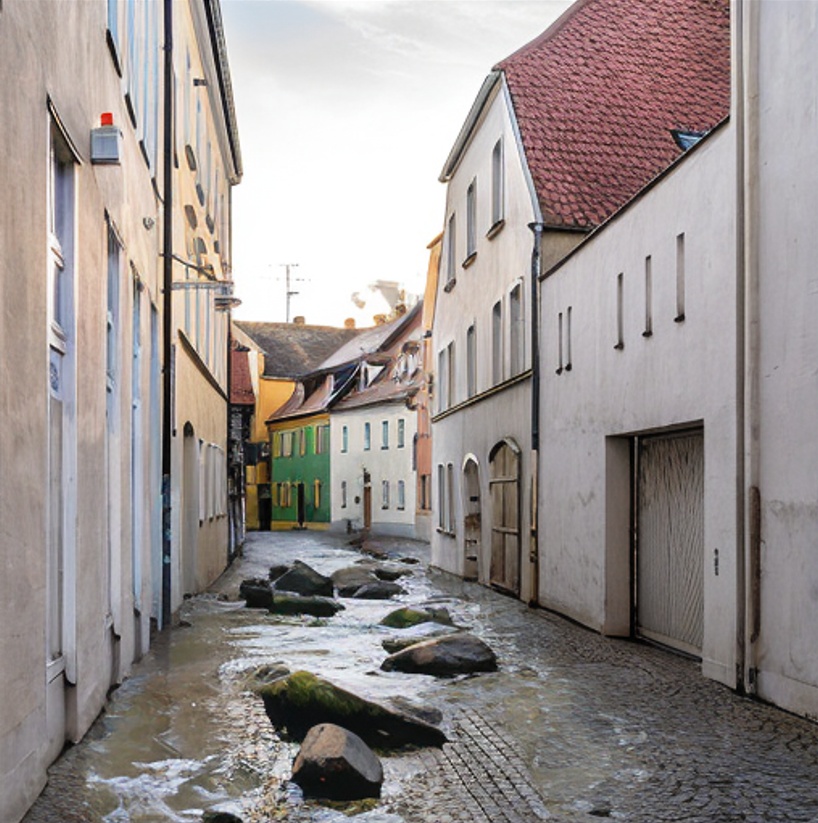}}              &
        \raisebox{-.5\height}{\includegraphics[width=\ww]{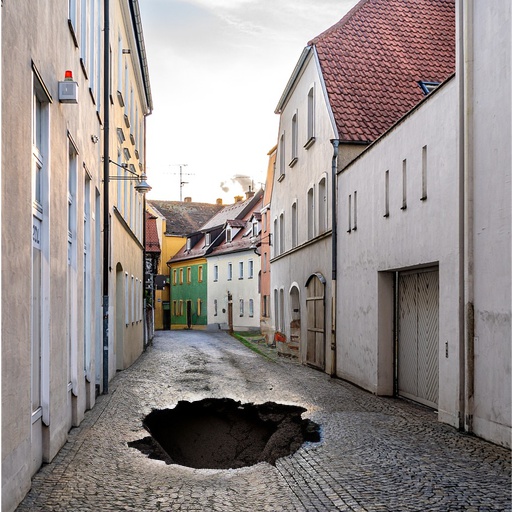}}               &
        \raisebox{-.5\height}{\includegraphics[width=\ww]{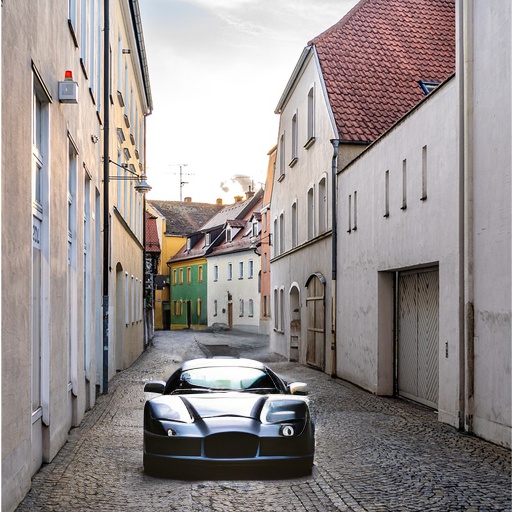}}            
        \\
        [\rss]
        
         \sizedtext{\methsize}{Input} & \pruns{\pw}{\prsize}{A river} & \prtwolines{\prsize}{A hole in}{the ground} & \pruns{\pw}{\prsize}{A sports car}
        \\
        [\rsb]
            
        \raisebox{-.5\height}{\includegraphics[width=\ww]{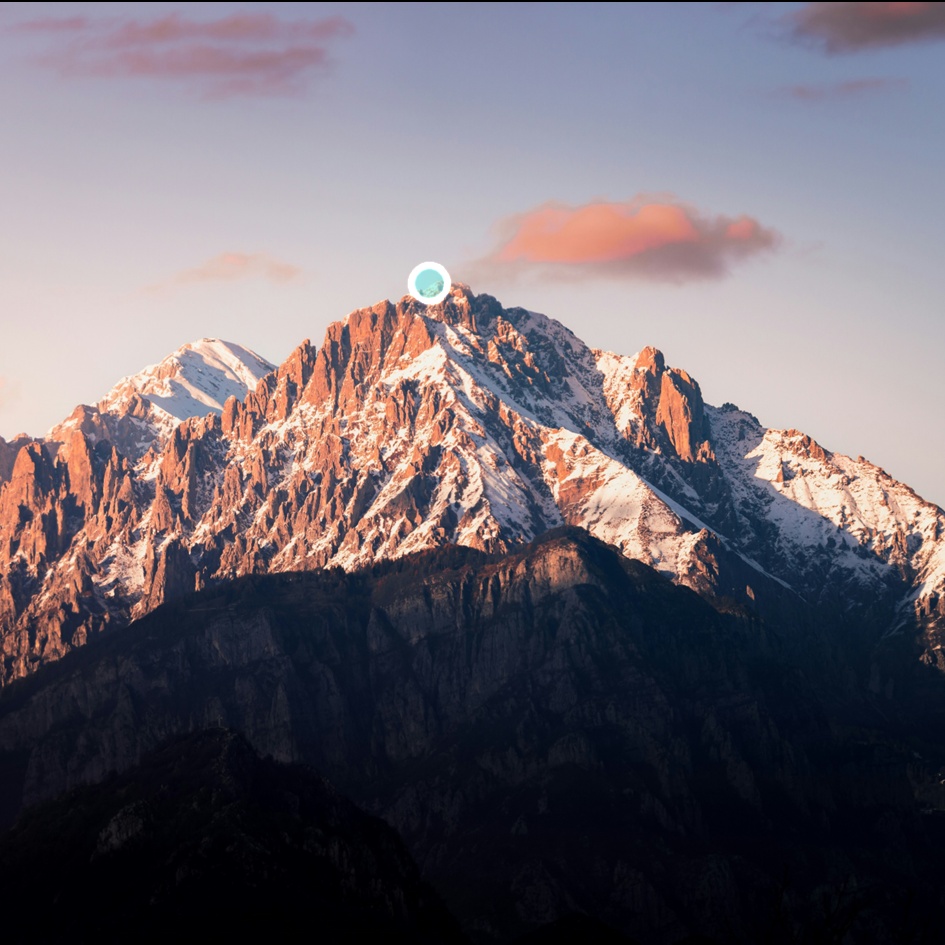}}   &
        \raisebox{-.5\height}{\includegraphics[width=\ww]{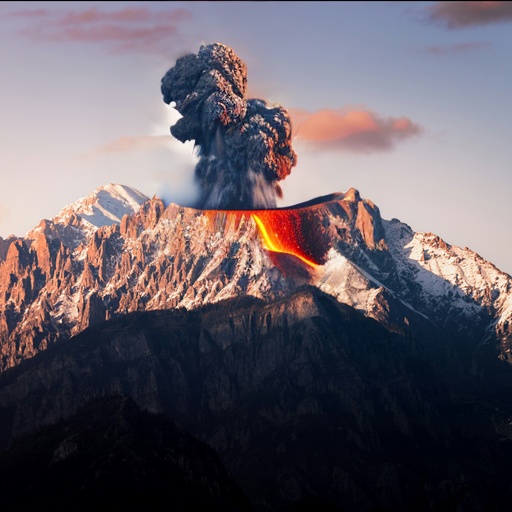}}            &
        \raisebox{-.5\height}{\includegraphics[width=\ww]{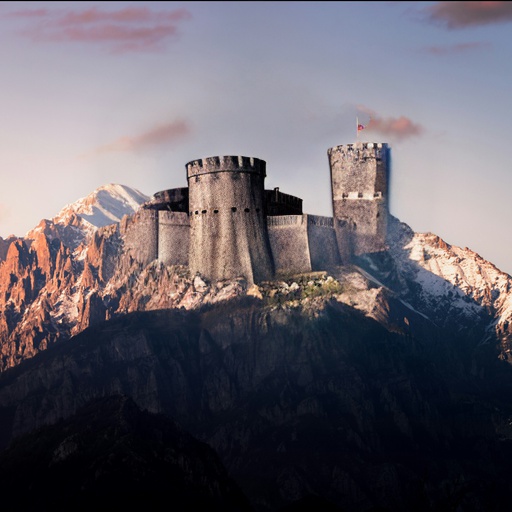}}           &
        \raisebox{-.5\height}{\includegraphics[width=\ww]{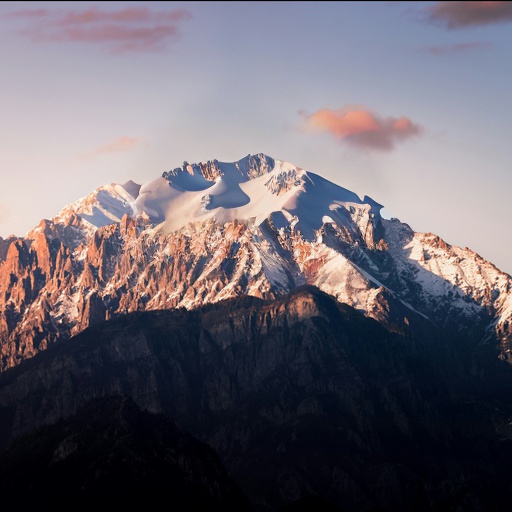}} 
        \\
        [\rss]
        
         \sizedtext{\methsize}{Input} & \prtwolines{\prsize}{A volcano}{erupting} & \pruns{\pw}{\prsize}{A fortress} & \pruns{\pw}{\prsize}{Skiing slopes}
        \\ 
        [\rsb]

        \raisebox{-.5\height}{\includegraphics[width=\ww]{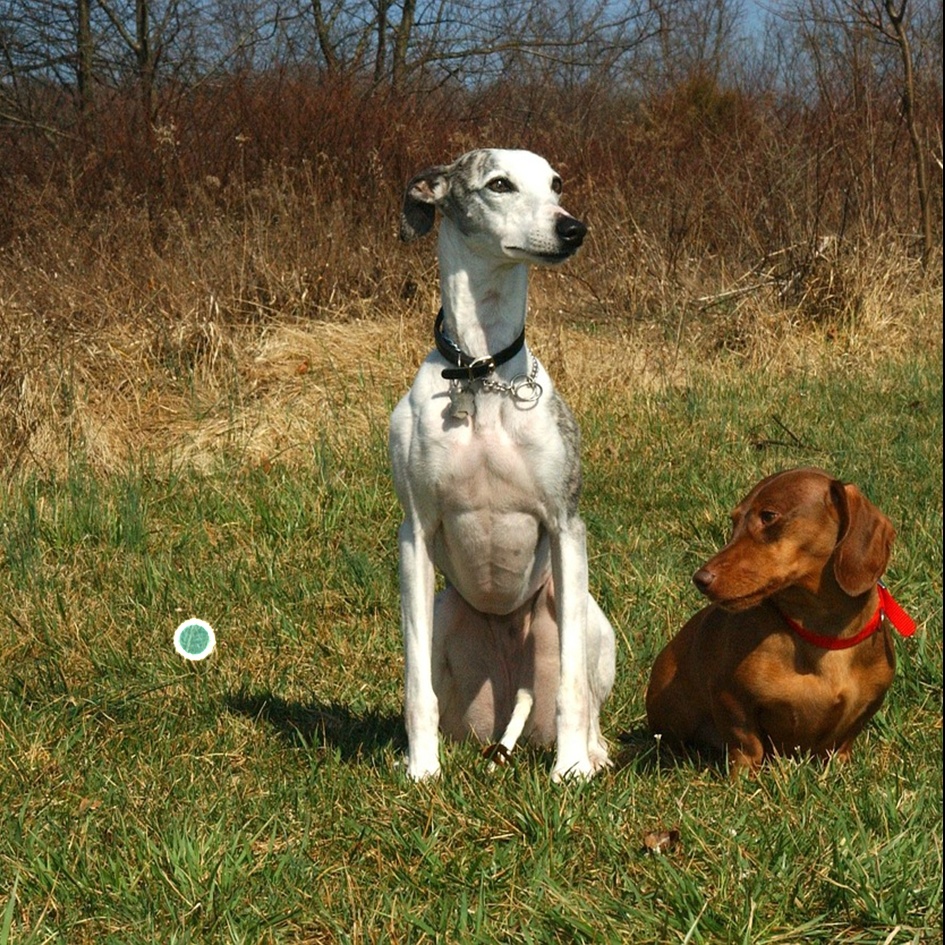}}   &
        \raisebox{-.5\height}{\includegraphics[width=\ww]{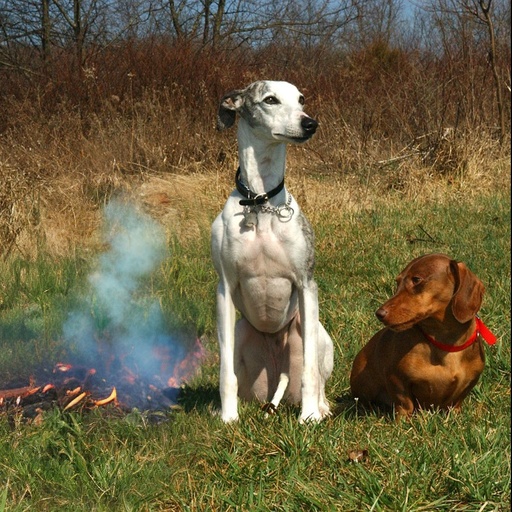}}            &
        \raisebox{-.5\height}{\includegraphics[width=\ww]{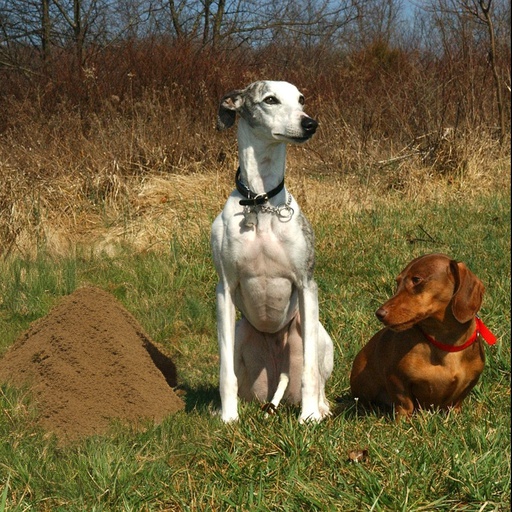}}               &
        \raisebox{-.5\height}{\includegraphics[width=\ww]{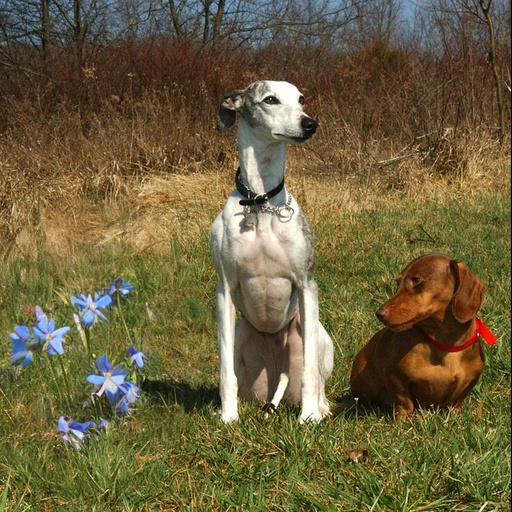}} 
        \\   
        [\rss]
        
         \sizedtext{\methsize}{Input} & \pruns{\pw}{\prsize}{A bonfire} & \pruns{\pw}{\prsize}{A pile of sand} & \pruns{\pw}{\prsize}{flowers}
        \\
        [\rsm]
                
        \raisebox{-.5\height}{\includegraphics[width=\ww]{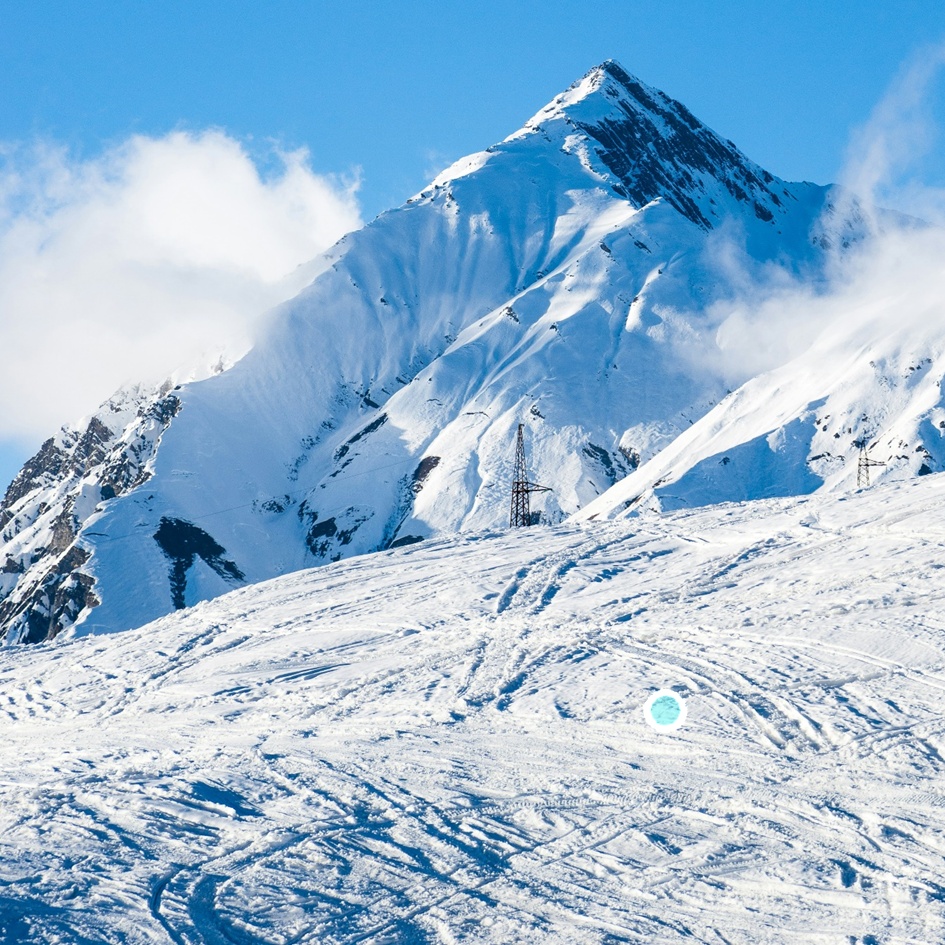}}       &
        \raisebox{-.5\height}{\includegraphics[width=\ww]{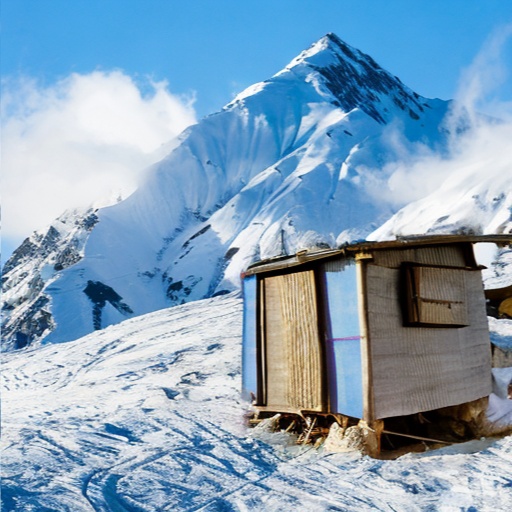}}                &
        \raisebox{-.5\height}{\includegraphics[width=\ww]{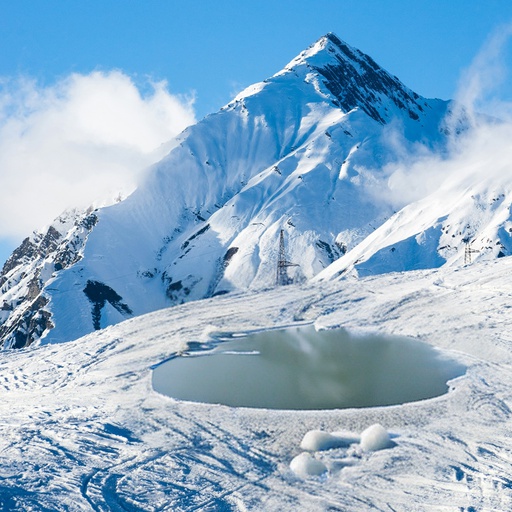}}                & 
        \raisebox{-.5\height}{\includegraphics[width=\ww]{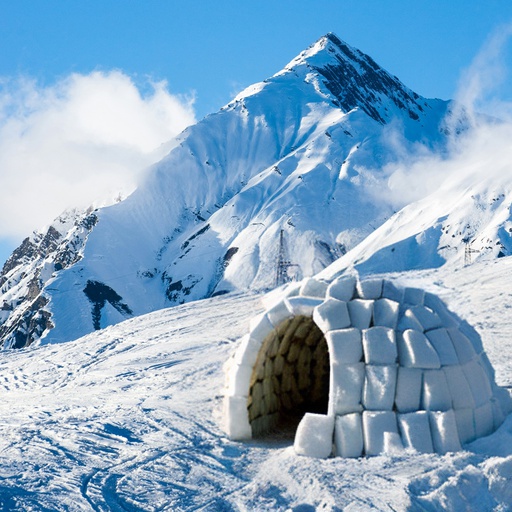}}  
        \\
        [\rss]
        
         \sizedtext{\methsize}{Input} & \pruns{\pw}{\prsize}{A small hut} & \pruns{\pw}{\prsize}{Iced lake} & \pruns{\pw}{\prsize}{An igloo}
    \end{tabular}
    
    \caption{\textbf{Examples of \ctm\ outputs.} The leftmost column is the input image with clicked point. The other columns are \ctmb\ outputs given the prompts below.}
    \label{fig:our_results}
    \vspace{\capspace}
\end{figure}

To overcome the aforementioned shortcomings, we introduce Click2Mask, a novel approach that simplifies user interaction by requiring only a single point of reference rather than a detailed mask or a description of the target area. The provided point gives rise to a mask that dynamically evolves through a Blended Latent Diffusion (BLD) process \cite{Avrahami_2022_CVPR, avrahami2023blendedlatent}, where the evolution is guided by a semantic loss based on Alpha-CLIP \cite{sun2023alphaclip}. This process (\Cref{fig:mask_evolution}) enables local edits that are both precise and contextually relevant (\Cref{fig:teaserfigure,fig:our_results}).

Unlike segmentation-based methods that depend on pre-existing objects \cite{couairon2022diffedit, xie2023edit, wang2023instructedit, zou2024efficient}, Click2Mask does not confine the edit area to the boundaries of an existing segment. Furthermore, in contrast to editing approaches that require fine-tuning the diffusion model \cite{wang2023imagen, xie2022smartbrush, kawar2023imagic, avrahami2023bas}, we employ pre-trained models, and only perform context dependent optimization on the mask.

Our experiments demonstrate that Click2Mask not only reduces the effort required by users but also achieves competitive or superior results compared to state-of-the-art methods in local image manipulation. 

In summary, our contributions are: 
\begin {enumerate*} [label=(\roman*)]
\item Reduction of user effort by eliminating the need for precise mask outlines, or overly descriptive prompts.
\item Ability to add objects in a free-form manner, unconstrained by boundaries of existing objects or segments.
\item Our dynamically evolving mask approach is not a stand-alone method, but rather it can be embedded as a mask generation of the fine-tuning step within other methods that internally employ a mask, such as \emu\ \cite{sheynin2023emu} which currently generates multiple masks (a precise mask using DINO \cite{Caron_2021_ICCV} and SAM \cite{kirillov2023segany}, an expanded version of it, and a bounding box), and filters the best result from multiple images produced using these masks.
\end {enumerate*}

\section{Related Work}
\label{sec:related_work}
In recent years, much work has been done on image generation, with diffusion-based models (DMs) \cite{ho2020denoising,song2020denoising,dhariwal2021diffusion,rombach2022highresolution,ramesh2022hierarchical,saharia2022photorealistic} facilitating a host of SoTA text-guided image editing methods and capabilities. 

\textbf{Mask-based approaches.} Text-guided image manipulation may naturally be limited to a specific region using a mask.
In the context of DMs this was first explored in Blended Diffusion \cite{Avrahami_2022_CVPR}, where a user-provided mask is used to blend images throughout a denoising process with a text-guided noisy image. 
This approach was later incorporated into Latent Diffusion \cite{rombach2022highresolution} by performing the blending in latent space. The resulting Blended Latent Diffusion (BLD) method \cite{avrahami2023blendedlatent} serves as the basis for our work and described in more detail in \Cref{sec:background}.
GLIDE \cite{nichol2022glide}, Imagen Editor~\cite{wang2023imagen} and SmartBrush~\cite{xie2022smartbrush} fine-tuned the DM for image inpainting, by obscured training images or by conditioning on a mask. 
However, user-provided masks have a major disadvantage: the success of the edit depends on the exact shape of the mask, which can be tedious and time-consuming for a user to create. 

\textbf{Mask-free approaches.} Both Text2Live \cite{bar2022text2live}, which generates a composite layer, and Imagic \cite{kawar2023imagic}, which interpolates target text and optimized source embeddings, fine-tune the generative model for each image, which is quite costly, contrary to our work.
Several works use attention injection, such as Plug-and-Play \cite{tumanyan2022plugandplay} and Prompt-to-Prompt \cite{hertz2022prompt}, where the latter requires a time-consuming caption of the input image, unlike our method.
Most of these methods focus on altering a certain object (by replacement, removal or style change), or applying global changes (style or content), in contrast to our focus on adding objects freely.  

\textbf{Instruction-based approaches.} 
Other methods can add objects in a free manner. 
\ipp\ \cite{brooks2022instructpix2pix} (subsequently fine-tuned by \mb\ \cite{Zhang2023MagicBrush}) produces (instruction, image) pairs, used to train an instruction-conditioned DM. 
\emu\ \cite{sheynin2023emu} is a more recent model trained on a wide range of learned task embeddings to enable instruction-based image editing, however, it is not publicly available.
\DALLE{3}\ \cite{BetkerImprovingIG} is also proprietary, 
and modifies the entire image as demonstrated in \Cref{fig:teaserfigure}. \DALLE{3} and \DALLE{2}\ \cite{ramesh2022hierarchical} apparently support masked inpainting, but we are unaware of a publicly available way to apply it to general real images.
MGIE \cite{fu2024guiding} train a DM, utilizing a MLLM to derive expressive instructions. 
These methods require the user to specify the desired localization in words, which has a few shortcomings. On the user's side, this requires effort, and it can be difficult or impossible to describe the precise location. From the model's side, failure to visually ground the text-specified location may fail to perform the desired edit, and/or make unintended changes in other locations instead.

\textbf{Segmentation-based approaches.} 
Segmentation methods have been utilized to overcome the need for a precise user-provided mask.
DiffEdit \cite{couairon2022diffedit} and Edit Everything \cite{xie2023edit} generate segmentation-based masks by utilizing conditionings on diffusion steps, or SAM \cite{kirillov2023segany}, but require an input image caption, which is painstaking.
InstructEdit \cite{wang2023instructedit}, which uses Grounding DINO \cite{liu2023grounding} and SAM to generate a mask, does not require one, but requires a description of the object to alter. This can cause errors due to failure of the model to localize. 
InstDiffEdit \cite{zou2024efficient} generates masks based on attention maps during denoising.  

The segmentation-based methods, however, suffer from a few limitations: 
\begin {enumerate*} [label=(\roman*)]
\item Such models need to ``lock'' on an existing object or segment; consequently, in most cases they alter objects, but do not add new free-form ones, which is our focus.
\item These methods typically require the user to provide an input caption or a description of the altered object.
\end {enumerate*}

In contrast to all the above, our work enables \emph{adding} objects to real images (as opposed to merely altering existing ones), without having to provide a precise mask, to describe the input image, or target image, and without being constrained to boundaries of existing objects or segments. 
We aim to enable edits where the manipulated area is not well-defined in advance, and a free-form alteration is required.

\begin{figure*}[ht]
    \centering
    \setlength{\tabcolsep}{0.5pt}
    \renewcommand{\arraystretch}{0.6}
    \setlength{\ww}{0.09\linewidth}
    \renewcommand{\prsize}{\defprsize}
    \renewcommand{\methsize}{\defmethsize}
    \setlength{\rsm}{9pt}
    \setlength{\capspace}{\defcapspace}
    
    \begin{tabular}{c cccc @{\hskip 0.5\ww} c cccc} 
        \sizedtext{\methsize}{Input} &
        \sizedtext{\methsize}{\emu} &
        \sizedtext{\methsize}{\mb} &
        \sizedtext{\methsize}{\ipp} &
        \sizedtext{\methsize}{\ctmb} & 
        \sizedtext{\methsize}{Input} &
        \sizedtext{\methsize}{\emu} &
        \sizedtext{\methsize}{\mb} &
        \sizedtext{\methsize}{\ipp} &
        \sizedtext{\methsize}{\ctmb} 
        \\
    
        \raisebox{-.5\height}{\includegraphics[width=\ww]{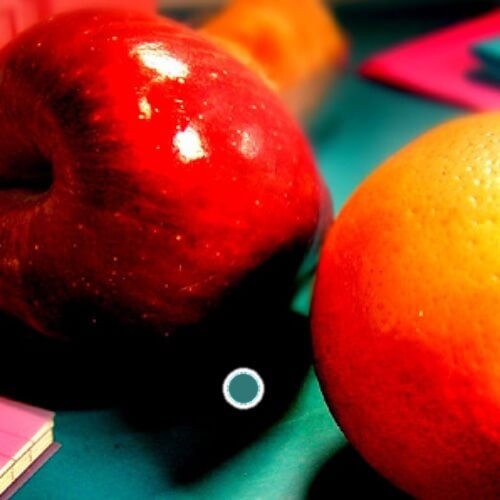}} &
        \raisebox{-.5\height}{\includegraphics[width=\ww]{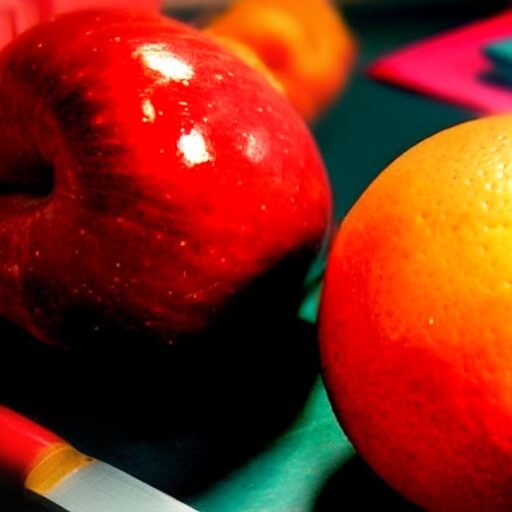}} &
        \raisebox{-.5\height}{\includegraphics[width=\ww]{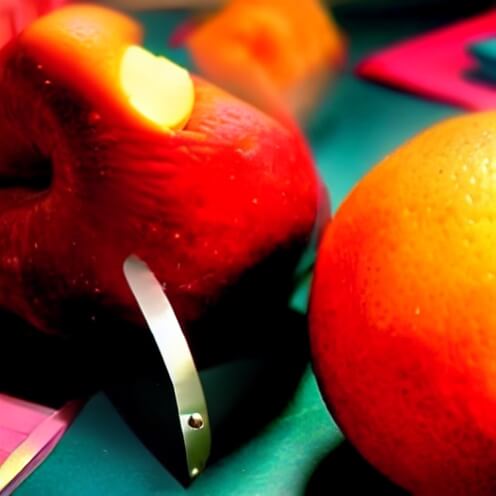}} &
        \raisebox{-.5\height}{\includegraphics[width=\ww]{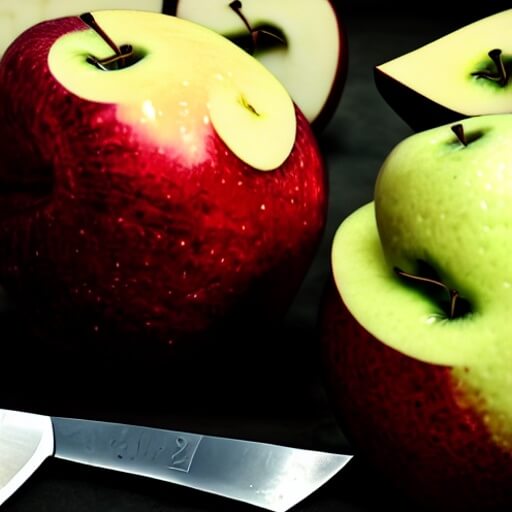}} &
        \raisebox{-.5\height}{\includegraphics[width=\ww]{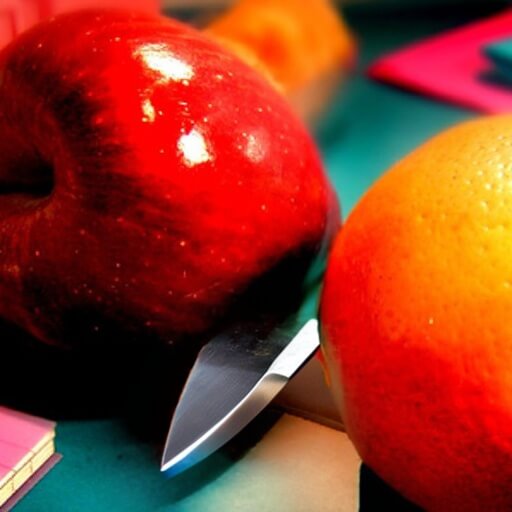}} &

        \raisebox{-.5\height}{\includegraphics[width=\ww]{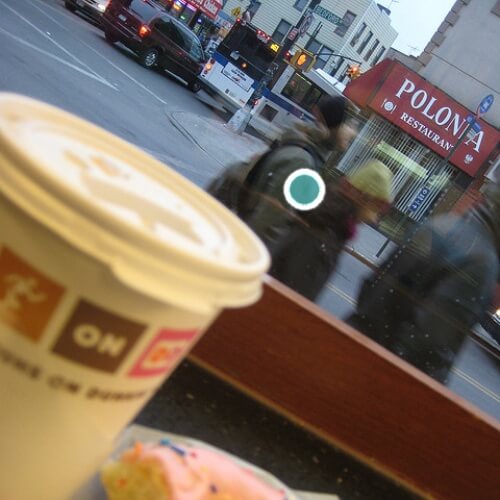}} &
        \raisebox{-.5\height}{\includegraphics[width=\ww]{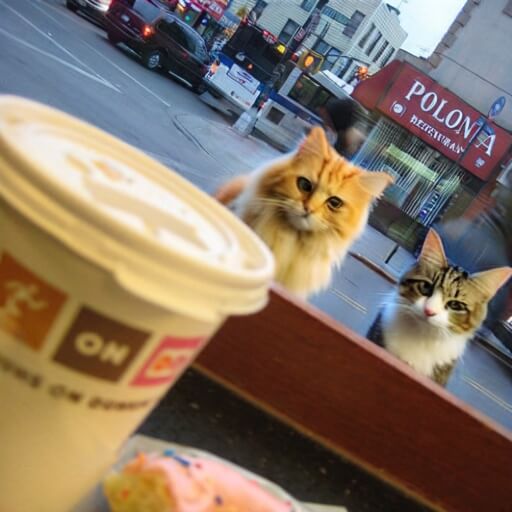}} &
        \raisebox{-.5\height}{\includegraphics[width=\ww]{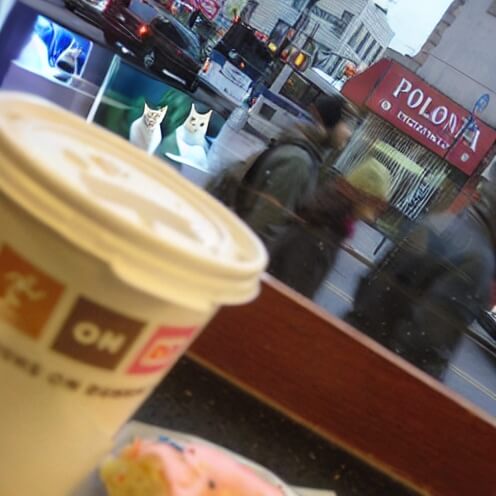}} &
        \raisebox{-.5\height}{\includegraphics[width=\ww]{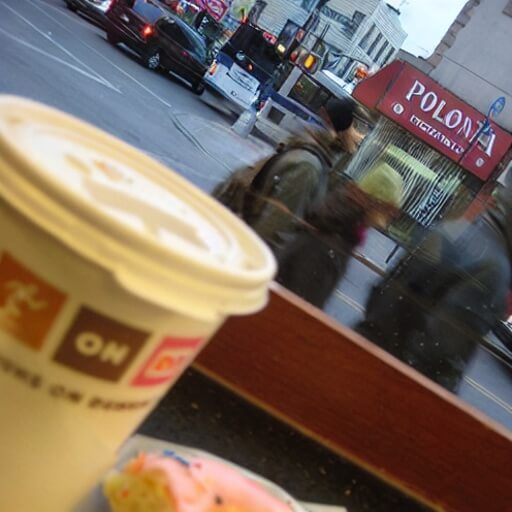}} &
        \raisebox{-.5\height}{\includegraphics[width=\ww]{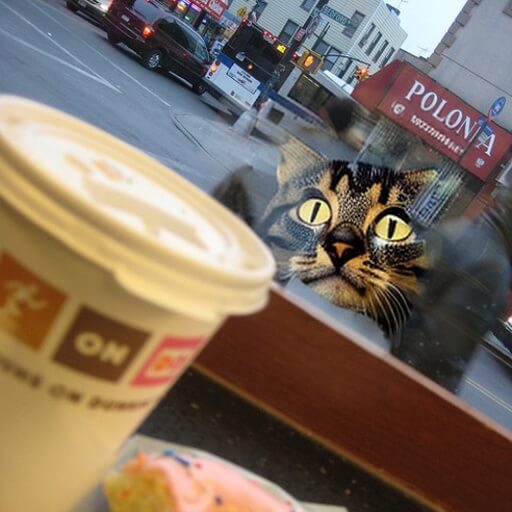}}
        \\
       
        \multicolumn{5}{c}{\rule{0pt}{1.5ex}\sizedtext{\prsize}{\textit{``Have a knife laying between the orange and apple"}}}&
        \prunmerge{\prsize}{5}{Add a cat behind the glass window looking at the food}
        \\
        \prunmergerule{\prsize}{5}{1.5}{A knife} &
        \prunmerge{\prsize}{5}{A cat behind the glass window looking at the food}
        \\
        [\rsm]

        \raisebox{-.5\height}{\includegraphics[width=\ww]{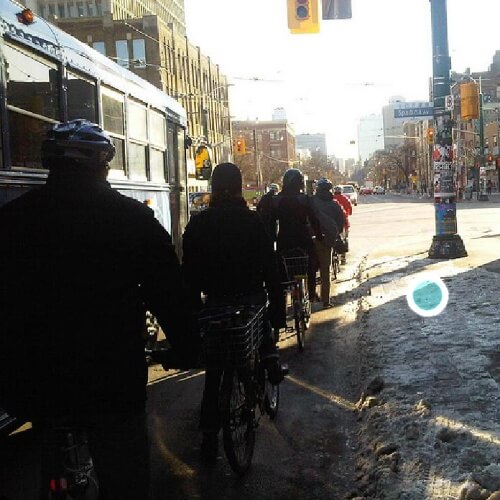}} &
        \raisebox{-.5\height}{\includegraphics[width=\ww]{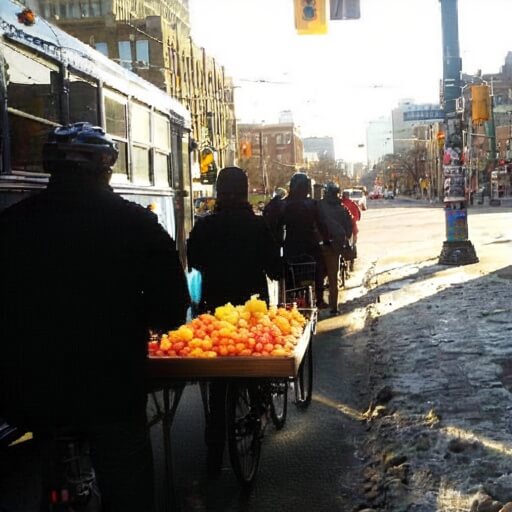}} &
        \raisebox{-.5\height}{\includegraphics[width=\ww]{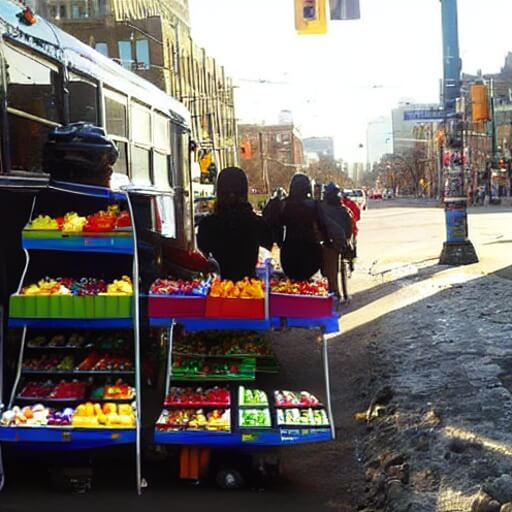}} &
        \raisebox{-.5\height}{\includegraphics[width=\ww]{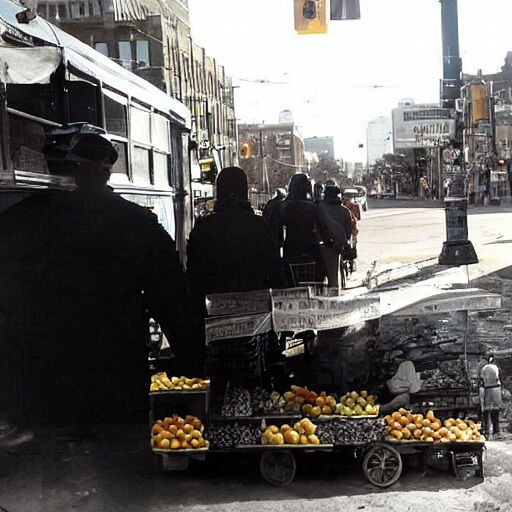}} &
        \raisebox{-.5\height}{\includegraphics[width=\ww]{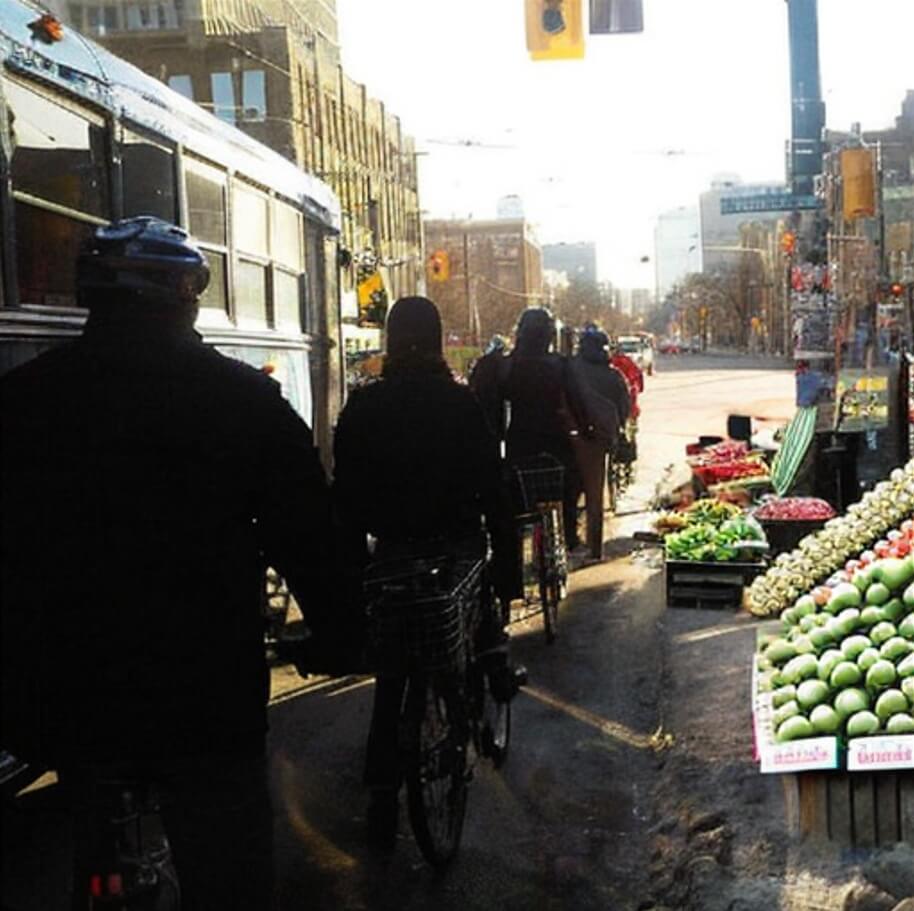}} &

        \raisebox{-.5\height}{\includegraphics[width=\ww]{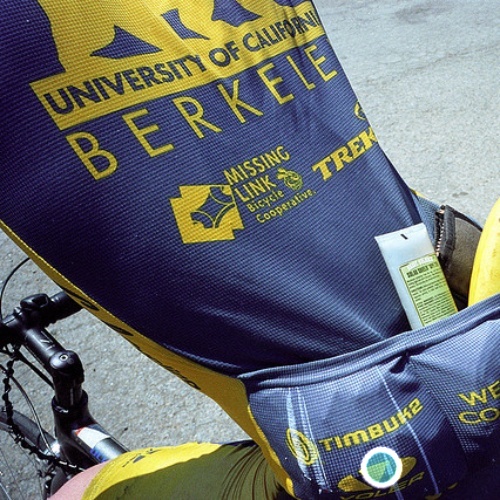}} &
        \raisebox{-.5\height}{\includegraphics[width=\ww]{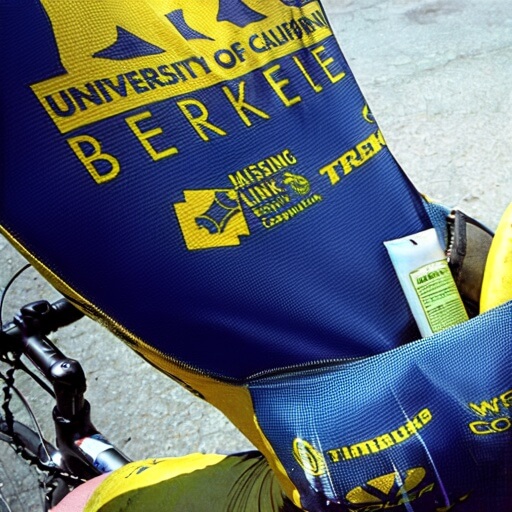}} &
        \raisebox{-.5\height}{\includegraphics[width=\ww]{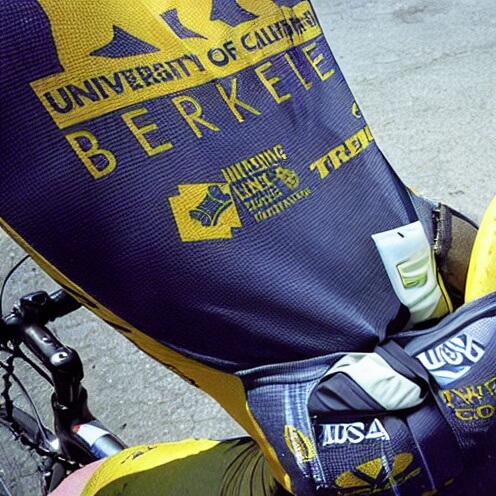}} &
        \raisebox{-.5\height}{\includegraphics[width=\ww]{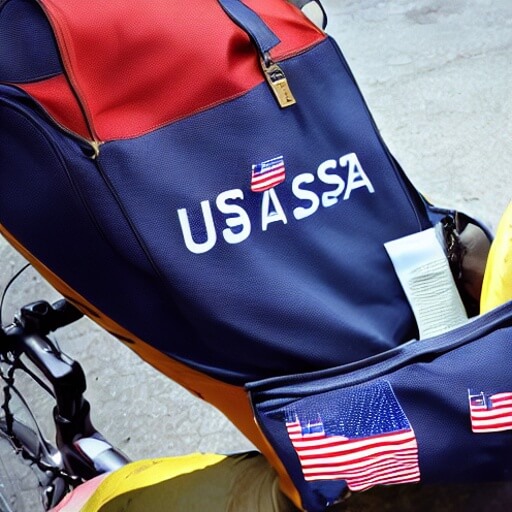}} &
        \raisebox{-.5\height}{\includegraphics[width=\ww]{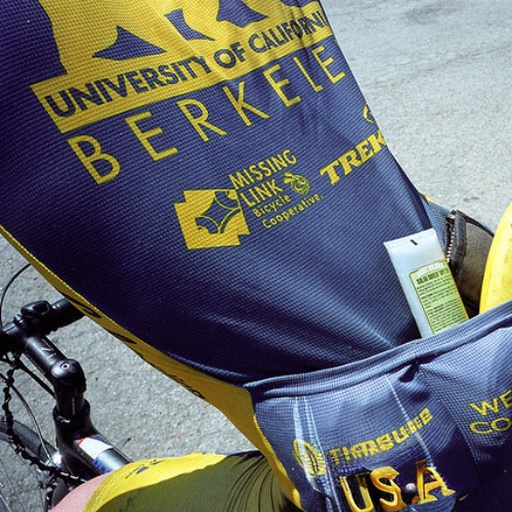}} 
        \\
        \multicolumn{5}{c}{\rule{0pt}{1.5ex}\sizedtext{\prsize}{\textit{``Add a fruit stand to the right of the image"}}}&
        \prunmerge{\prsize}{5}{Add USA for the bag}
        \\
        \prunmergerule{\prsize}{5}{1.5}{A fruit stand} &
        \prunmerge{\prsize}{5}{USA}
        \\
        [\rsm]
       
        \raisebox{-.5\height}{\includegraphics[width=\ww]{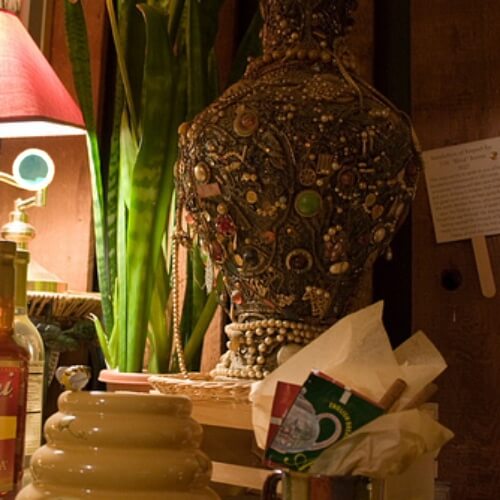}} &
        \raisebox{-.5\height}{\includegraphics[width=\ww]{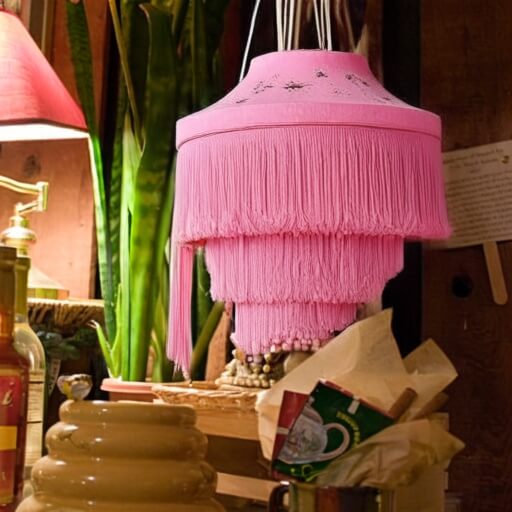}} &
        \raisebox{-.5\height}{\includegraphics[width=\ww]{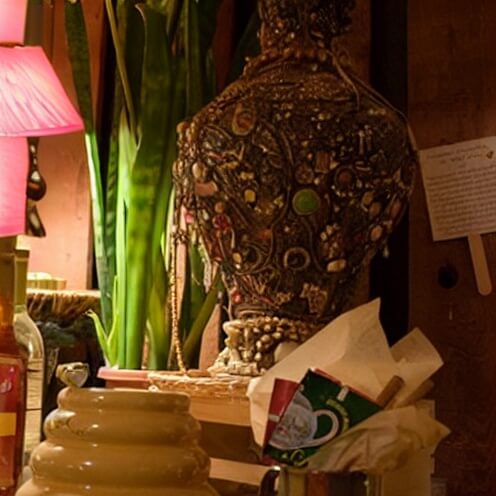}} &
        \raisebox{-.5\height}{\includegraphics[width=\ww]{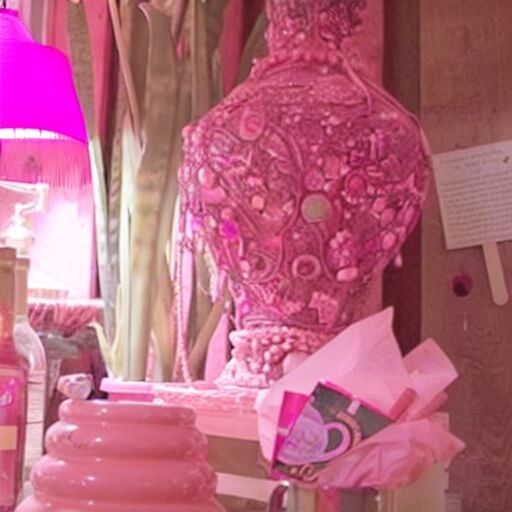}} &
        \raisebox{-.5\height}{\includegraphics[width=\ww]{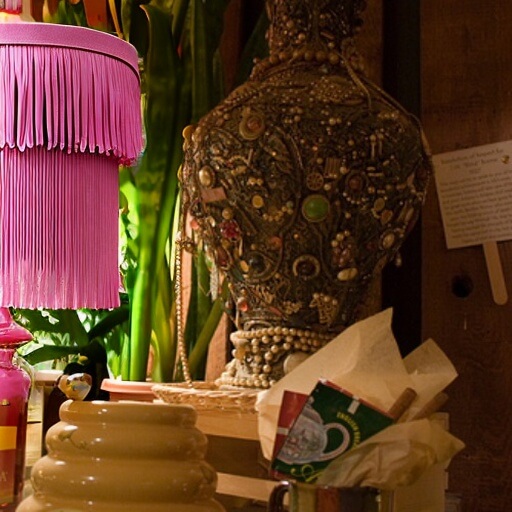}}  &

        \raisebox{-.5\height}{\includegraphics[width=\ww]{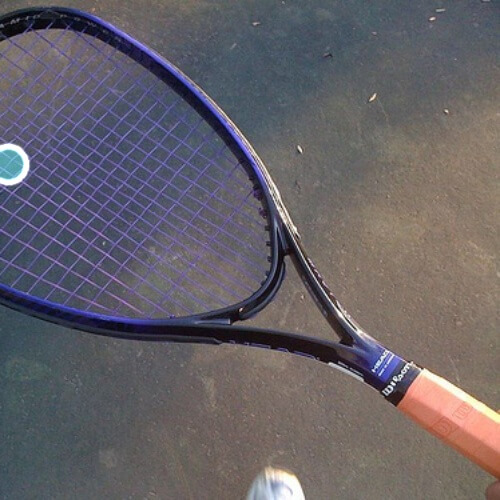}} &
        \raisebox{-.5\height}{\includegraphics[width=\ww]{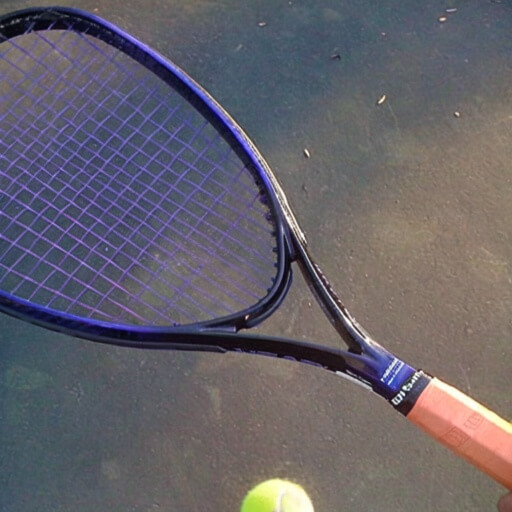}} &
        \raisebox{-.5\height}{\includegraphics[width=\ww]{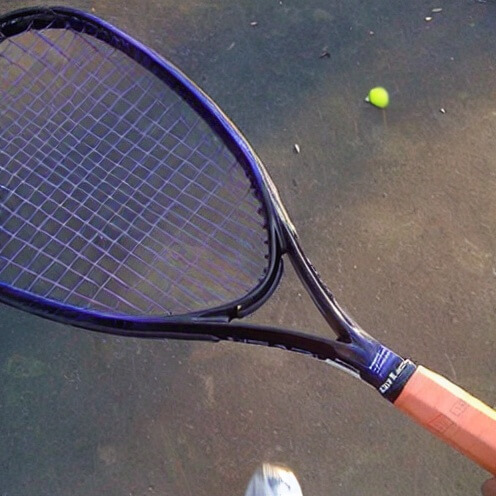}} &
        \raisebox{-.5\height}{\includegraphics[width=\ww]{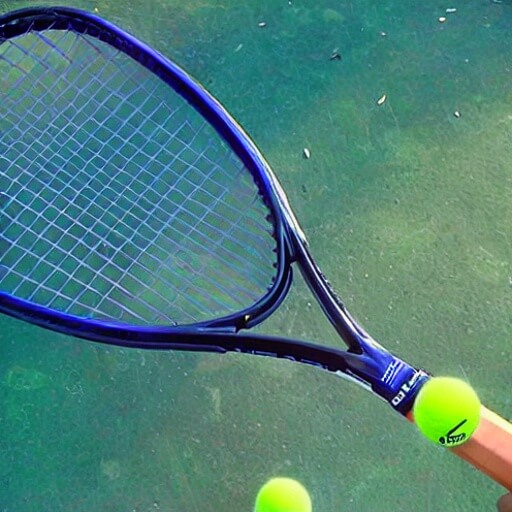}} &
        \raisebox{-.5\height}{\includegraphics[width=\ww]{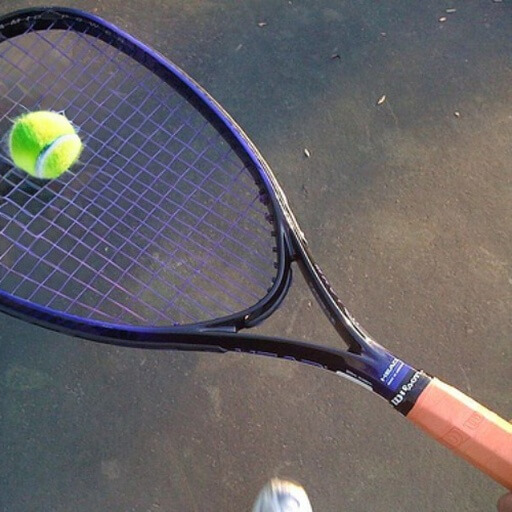}} 
        \\
        \multicolumn{5}{c}{\rule{0pt}{1.5ex}\sizedtext{\prsize}{\textit{``Add fringe to the pink lampshade"}}}&
        \prunmerge{\prsize}{5}{Add a tennis ball on top of the racket}
        \\
        \prunmergerule{\prsize}{5}{1.5}{Fringe} &
        \prunmerge{\prsize}{5}{A tennis ball}

    \end{tabular}
    
    \caption{\textbf{Comparisons with SoTA methods.} Comparisons of \emu\ \cite{sheynin2023emu}, \mb\ \cite{Zhang2023MagicBrush} and \ipp\ \cite{brooks2022instructpix2pix} with our model \ctmb. Upper prompts were given to baselines, and lower ones to \ctm. The inputs contain the clicked point given to \ctm. As \Cref{fig:compare_issues} shows, baselines often modify unrelated objects, make global changes, misplace elements, or replace rather than add objects. See \supl\ for more comparisons.}
    \vspace{\capspace}
    \label{fig:comparison_1}
\end{figure*}

\section{Blended Latent Diffusion} 
\label{sec:background}
Blended Latent Diffusion (BLD) \cite{avrahami2023blendedlatent} is a method for local text-guided image manipulation, based on Latent Diffusion Models (LDMs) \cite{rombach2022highresolution} and Blended Diffusion \cite{Avrahami_2022_CVPR}.
Given a source image $x$, a guiding text prompt $\prompt$, and a binary mask $m$, the model blends the source latents (obtained by DDIM inversion \cite{song2020denoising}) with the prompt-guided latents throughout the LDM process, to derive a blended final output. 

Initially, inputs are converted to a latent space. A variational auto-encoder \cite{Kingma2013vae} with encoder $E(\cdot)$ and decoder $D(\cdot)$, encodes $x$ to latent space, s.t. $z_\textit{init} = E(x)$. In addition, $m$ is downsampled to $m_\textit{latent}$ in order to meet latent spatial dimensions.

In each BLD step $t$, the following occurs:
\begin{enumerate}
    \item The latent resulting from the previous step, $z_{t+1}$, undergoes denoising conditioned by the prompt $p$, to yield $z_\textit{fg}$ (we refer to the generated content as \emph{foreground}, or \emph{fg}). 
    \item The original image latent $z_\textit{init}$ is noised to step $t$, yielding $z_\textit{bg}$ (we refer to the original content as \emph{background}, \emph{bg}).
    \item The next step $z_{t}$ is obtained by blending $z_\textit{fg}$ and $z_\textit{bg}$ using $m_\textit{latent}$: 
\begin{equation}
    z_{\textit{t}} = z_{\textit{fg}} \odot m_{\textit{latent}} + z_{\textit{bg}} \odot (1 - m_{\textit{latent}})
\label{bld_blend}
\end{equation}
where $\odot$ denotes element-wise multiplication. 
\end{enumerate}
After the final step, the output $z_{0}$ is decoded to obtain the final edited image $\hat{x} = D(z_{0})$.

However, because information is lost during the VAE encoding, the decoded final output $\hat{x}$, might exhibit some artifacts when the unmasked region has important fine-detailed content (such as faces, text, etc.). Avrahami \etal~\shortcite{avrahami2023blendedlatent} solve this issue by optionally fine-tuning the decoder weights for each image after the denoising steps, and using these weights to infer the final result. In our experiments, we found that this optional background preservation process is no longer necessary (possibly due to improvements in the Stable Diffusion VAE),
and a final blending with Gaussian feathering suffices (refer to \Cref{fig:abl_background_preservation} in \supl).

\section{Method}
\label{sec:method}

Given an image, a text prompt, and a user-indicated location (e.g., via a mouse click), our goal is to modify the image according to the prompt in an unspecified area roughly surrounding the provided point.
We utilize Blended Latent Diffusion (BLD) \cite{avrahami2023blendedlatent} as our image editing backbone, but rather than providing it with a fixed mask at the outset, we evolve a mask dynamically throughout the diffusion process. 
We initialize the process with a large mask around the indicated point, and gradually \emph{contract} the mask towards the center, while guiding the rate of contraction along the mask boundary using a semantic alignment loss based on \alphaclip\ \cite{sun2023alphaclip}.

\begin{figure}[t!]
    \centering
    \setlength{\tabcolsep}{0pt}
    \renewcommand{\arraystretch}{0.5}
    \setlength{\ww}{0.15\columnwidth}
    \renewcommand{\prsize}{\defprsize}
    \renewcommand{\methsize}{\defmethsize}
    \setlength{\capspace}{\defcapspace}

    \renewcommand{\figtitle}{
        \sizedtext{\methsize}{Input}&
        \begin{tabular}{>{\centering\arraybackslash}m{40pt}} %
			\sizedtext{\methsize}{Generated Mask}
		\end{tabular} &
        \sizedtext{\methsize}{Output} 
    }

    \begin{tabular}{ccc @{\hskip 0.4\ww} ccc} 
        \figtitle & \figtitle 
        \\

        \raisebox{-.5\height}{\includegraphics[width=\ww]{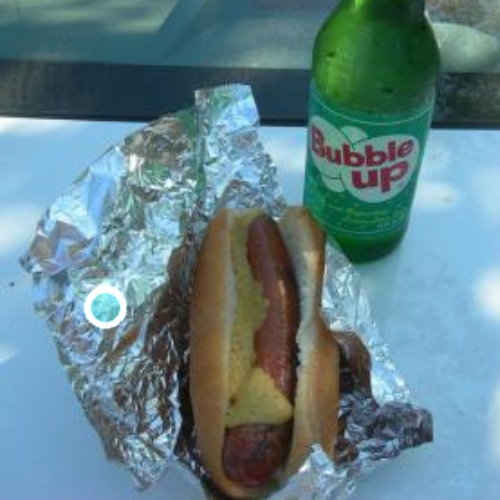}} &
        \raisebox{-.5\height}{\includegraphics[width=\ww]{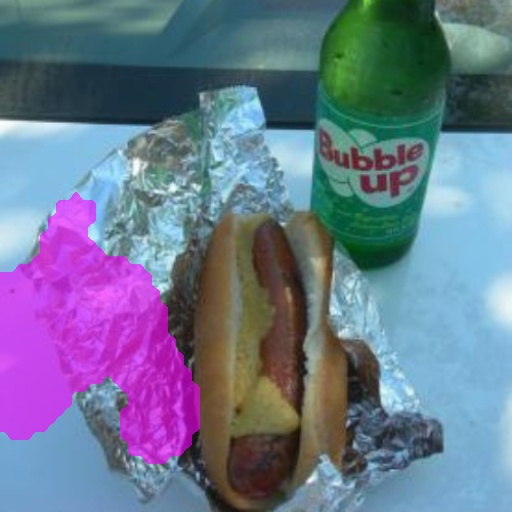}} &
        \raisebox{-.5\height}{\includegraphics[width=\ww]{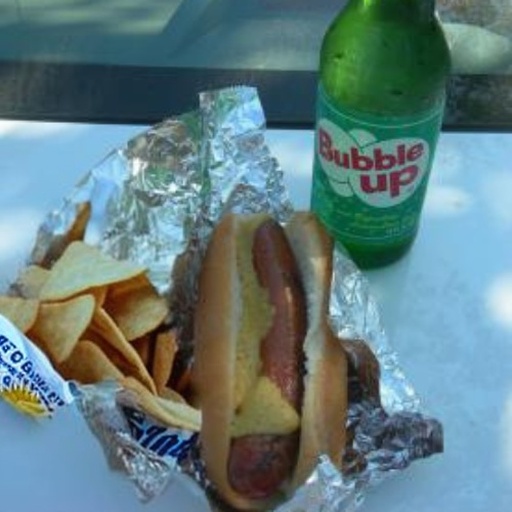}} &

        \raisebox{-.5\height}{\includegraphics[width=\ww]{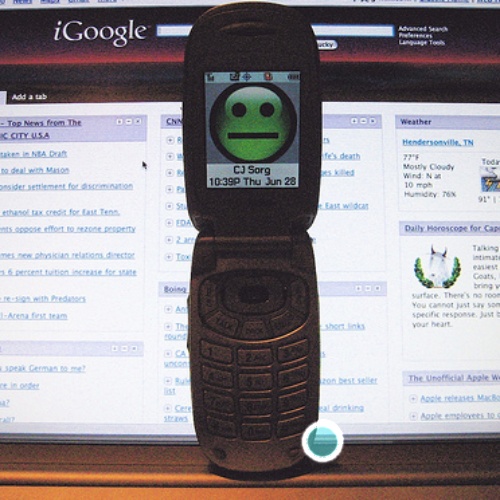}} &
        \raisebox{-.5\height}{\includegraphics[width=\ww]{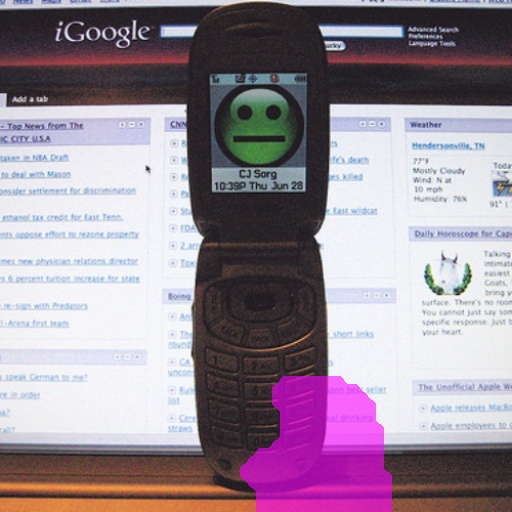}} &
        \raisebox{-.5\height}{\includegraphics[width=\ww]{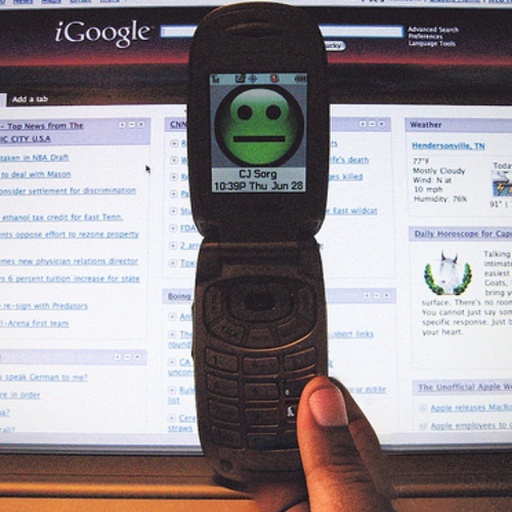}} 
        \\
        \multicolumn{3}{c}{\rule{0pt}{1.5ex}\sizedtext{\prsize}{\textit{``A bag of chips"}}} &
        \prunmerge{\prsize}{3}{A person's hand holding the phone} 
        \\
        \\

        \raisebox{-.5\height}{\includegraphics[width=\ww]{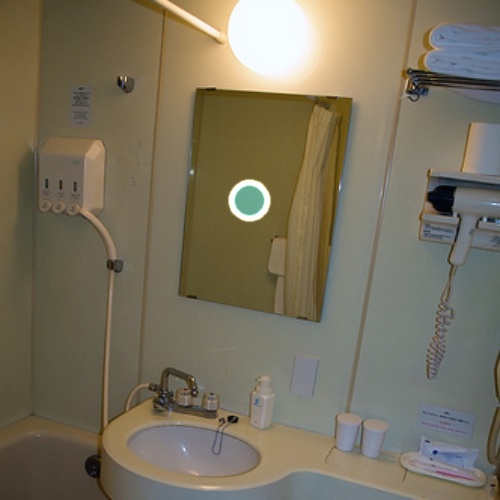}} &
        \raisebox{-.5\height}{\includegraphics[width=\ww]{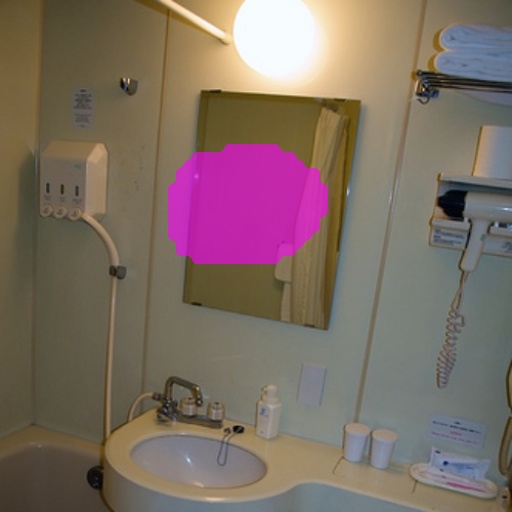}} &
        \raisebox{-.5\height}{\includegraphics[width=\ww]{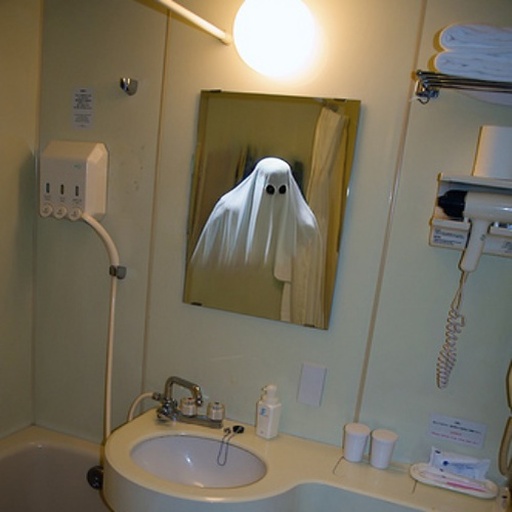}}    &  

        \raisebox{-.5\height}{\includegraphics[width=\ww]{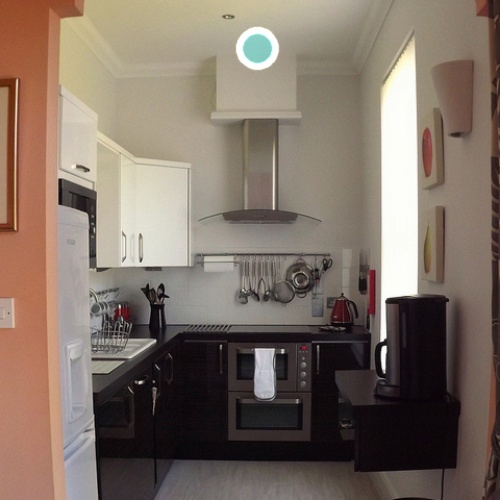}}   &
        \raisebox{-.5\height}{\includegraphics[width=\ww]{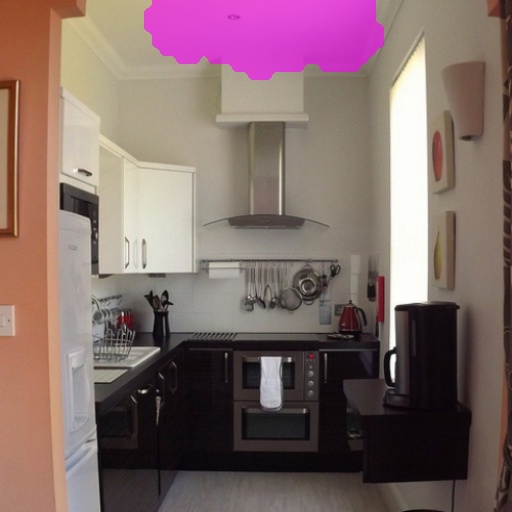}} &
        \raisebox{-.5\height}{\includegraphics[width=\ww]{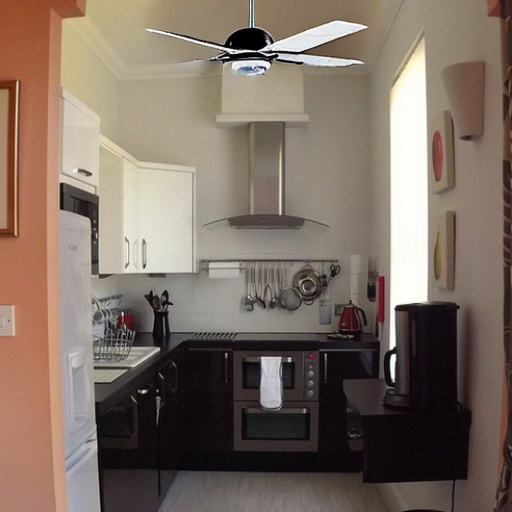}}   
        \\
        \multicolumn{3}{c}{\rule{0pt}{1.5ex}\sizedtext{\prsize}{\textit{``A reflection of a ghost"}}} &
        \prunmerge{\prsize}{3}{A ceiling fan}
        \\
        \\

        \raisebox{-.5\height}{\includegraphics[width=\ww]{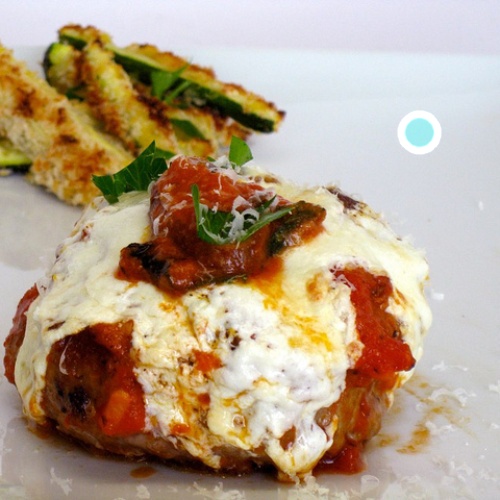}}   &
        \raisebox{-.5\height}{\includegraphics[width=\ww]{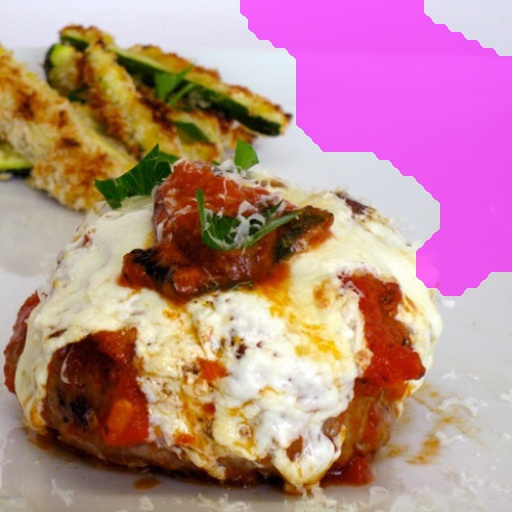}} &
        \raisebox{-.5\height}{\includegraphics[width=\ww]{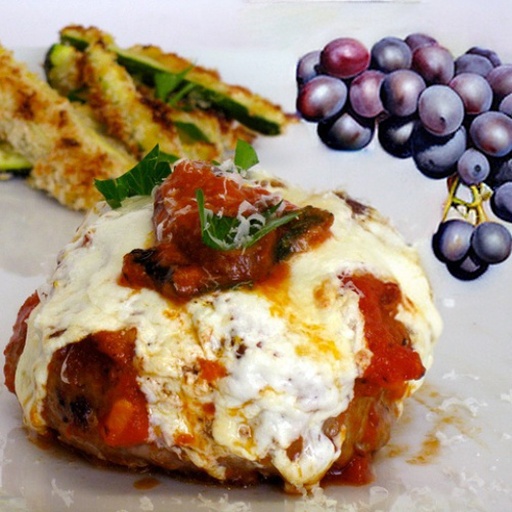}} &

        \raisebox{-.5\height}{\includegraphics[width=\ww]{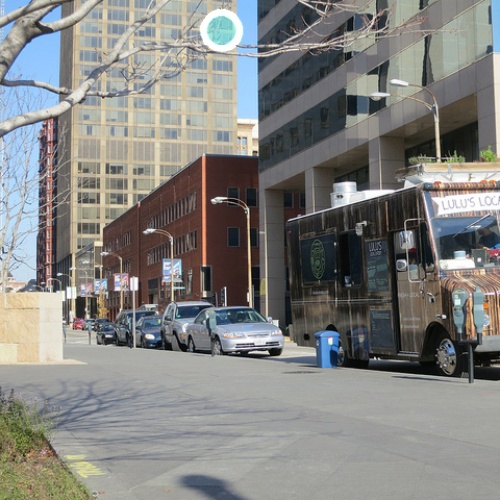}} &
        \raisebox{-.5\height}{\includegraphics[width=\ww]{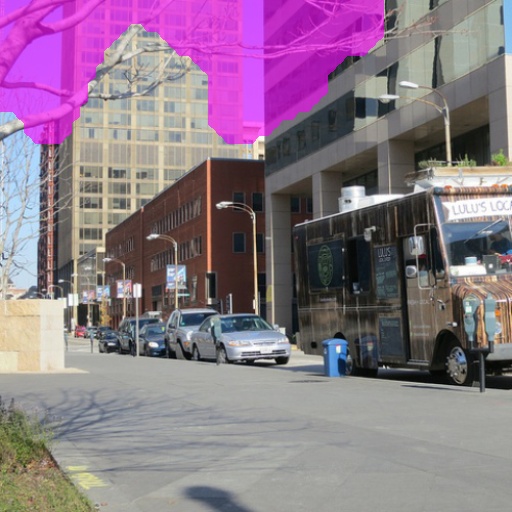}} &
        \raisebox{-.5\height}{\includegraphics[width=\ww]{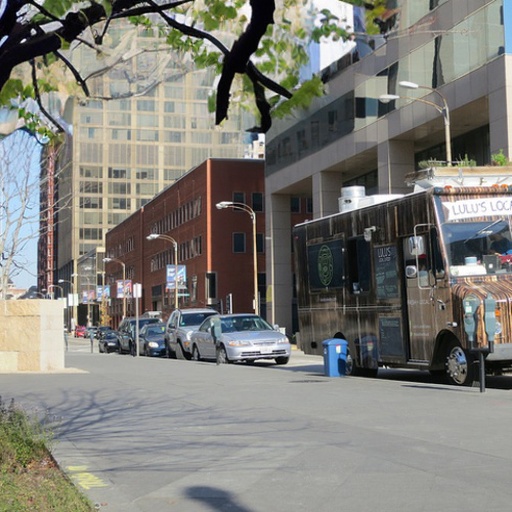}} 
        \\
         \multicolumn{3}{c}{\rule{0pt}{1.5ex}\sizedtext{\prsize}{\textit{``A bunch of grapes"}}} &
        \prunmerge{\prsize}{3}{Leaves on the trees}
        \\

    \end{tabular}
    
    \caption{\textbf{Examples of generated masks.} For each triplet, given an input image with clicked point (left) and a prompt (below), a purple overlay shows the generated mask (middle). The rightmost image is Click2Mask output.}
    \label{fig:generated_masks}
    \vspace{\capspace}
\end{figure}

This iterative process results in a mask whose shape and size are determined by both the text prompt, the content, and the structure of the original input image. 
Furthermore, the shape of the mask adjusts itself to the emerging object, as the mask's evolution is determined by the gradients obtained by the semantic alignment loss (see \Cref{sec:dyn_mask_evolution}), which in turn depend on the shape of the object being generated (see \Cref{fig:mask_evolution} for mask evolution illustration, and \Cref{fig:generated_masks} for examples of generated masks).
Once the mask has settled into its final form, we run BLD once more, using the final mask to generate the final result. Our method is outlined in \Cref{alg:click2mask} and illustrated in \Cref{fig:method_illusration}.

\subsection{Dynamic Mask Evolution}
\label{sec:dyn_mask_evolution}

Given an image $\img$, a text prompt $\prompt$, and a user-provided location $\coordinates$, we aim to modify $\img$, so as to align with $\prompt$, in proximity to $\coordinates$. 
We start by encoding the input image $z\un{init} = E(\img)$. We also create a 2D potential height-field $\potential$ in latent space, which is initialized to a Gaussian around $\coordinates$.

We now perform the BLD process, where at each step $t$ we obtain a binary mask $\mask$ by thresholding the potential $\potential$ using a threshold $\thresh$. The mask evolves dynamically through the BLD process, since the threshold $\thresh$ and the potential $\potential$ are both updated at each step: the threshold $\thresh$ increases, while the potential $\potential$ is elevated — starting from a specific step, as explained later — in important areas to ensure they remain above the threshold. This prevents the mask from shrinking in spatial areas that emerge as important for alignment of the generated new content with the guiding prompt $\prompt$. As a consequence, the mask evolves into a shape determined by the newly generated object. 

\begin{figure}[ht]
    \centering
    \setlength{\tabcolsep}{0pt}
    \renewcommand{\arraystretch}{0.5}
    \setlength{\ww}{0.132\columnwidth}
    \renewcommand{\ablsize}{scriptsize}
    \renewcommand{\rss}{17pt}
    \setlength{\capspace}{\defcapspace}
    
    \begin{tabular}{c@{\hspace{0.004\columnwidth}} c cccc @{\hspace{0.004\columnwidth}}c}  
        \abltitle{\ablsize}{No enlarge}
        \raisebox{-.5\height}{\includegraphics[width=\ww]{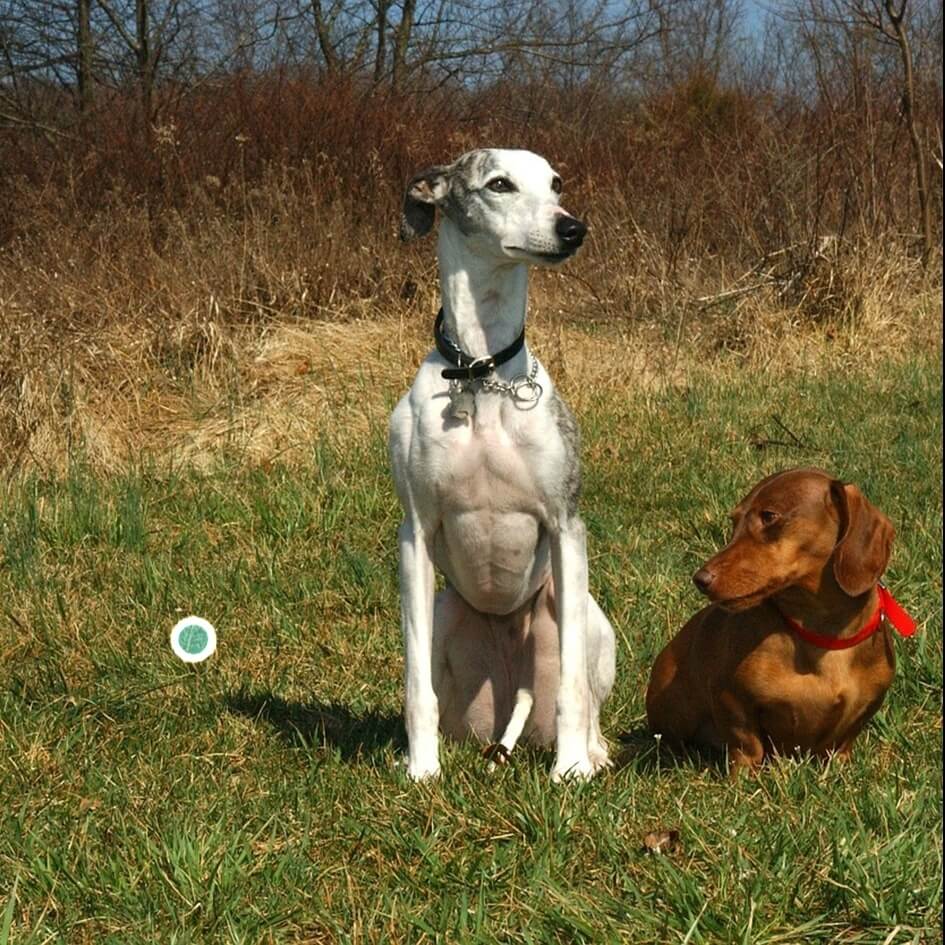}} &
        \raisebox{-.5\height}{\includegraphics[width=\ww]{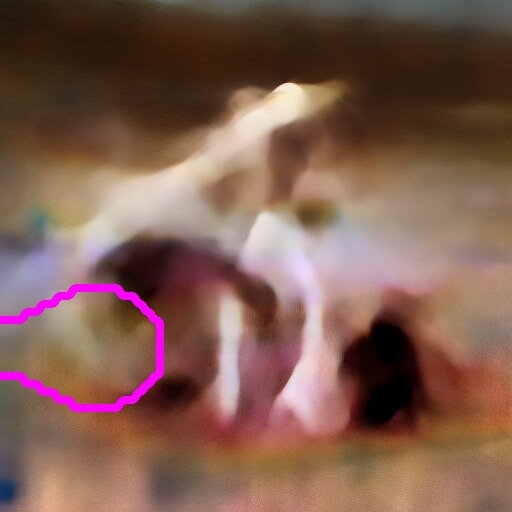}}  &
        \raisebox{-.5\height}{\includegraphics[width=\ww]{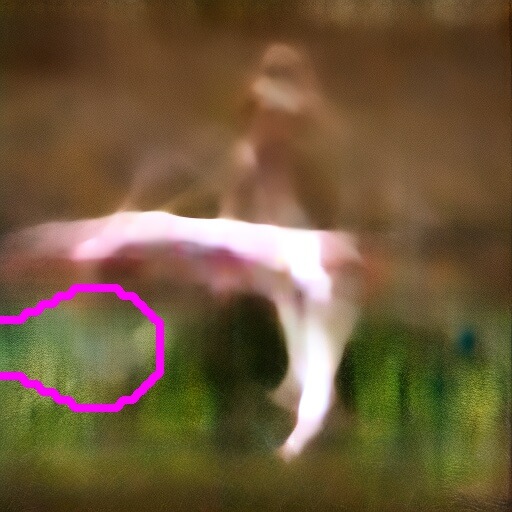}}  &
        \raisebox{-.5\height}{\includegraphics[width=\ww]{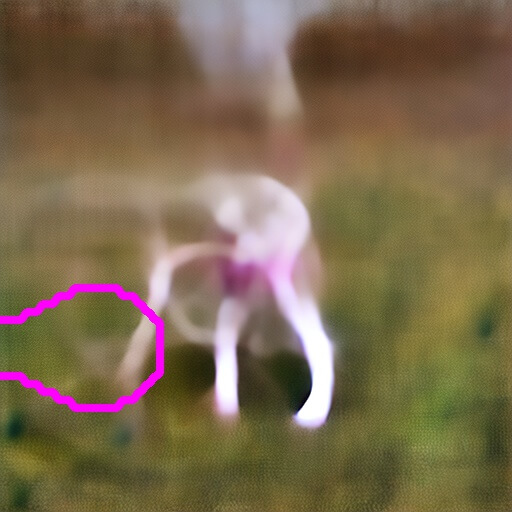}} &
        \raisebox{-.5\height}{\includegraphics[width=\ww]{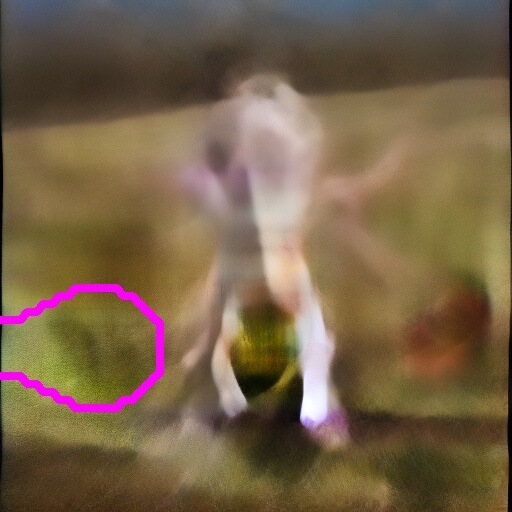}} &
        \raisebox{-.5\height}{\includegraphics[width=\ww]{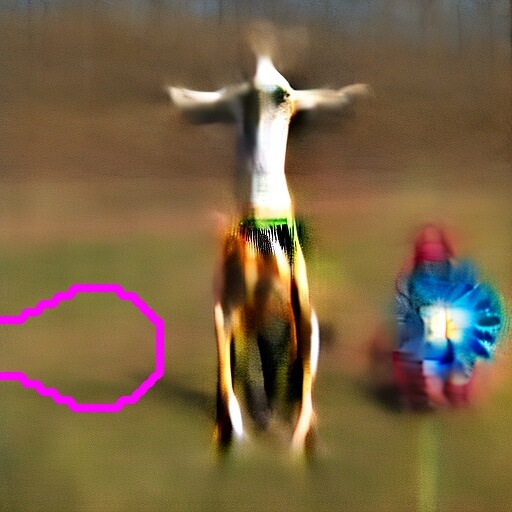}} &
        \raisebox{-.5\height}{\includegraphics[width=\ww]{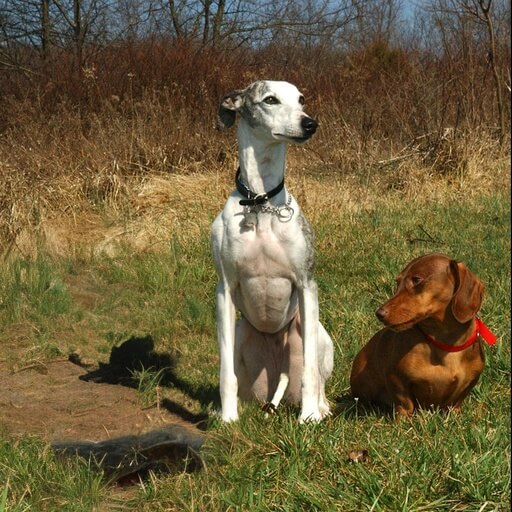}} 
        \\

        \abltitle{\ablsize}{\text{Ours}}
        \rule{0pt}{4ex}
        \raisebox{-.5\height}{\includegraphics[width=\ww]{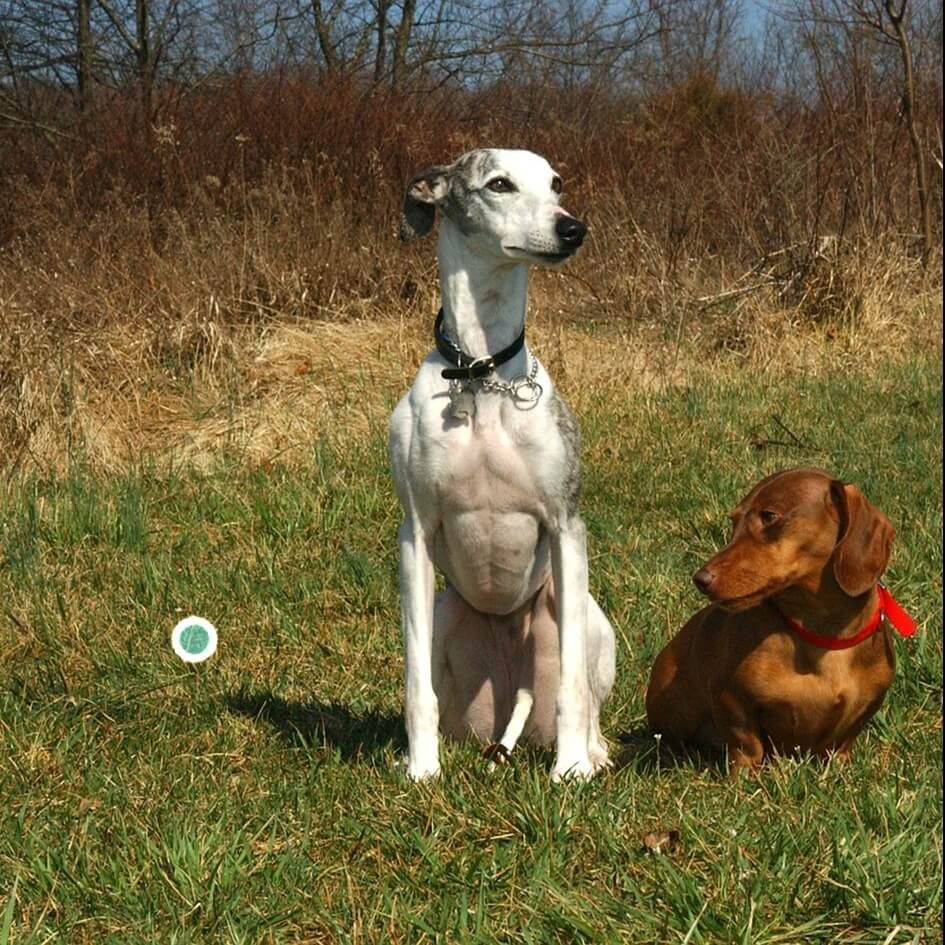}} &
        \raisebox{-.5\height}{\includegraphics[width=\ww]{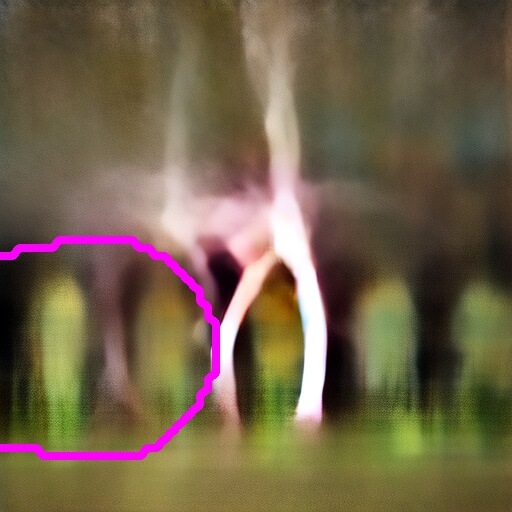}}  &
        \raisebox{-.5\height}{\includegraphics[width=\ww]{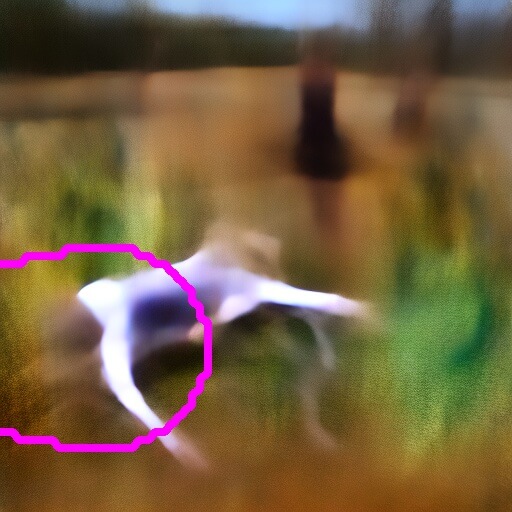}}  &
        \raisebox{-.5\height}{\includegraphics[width=\ww]{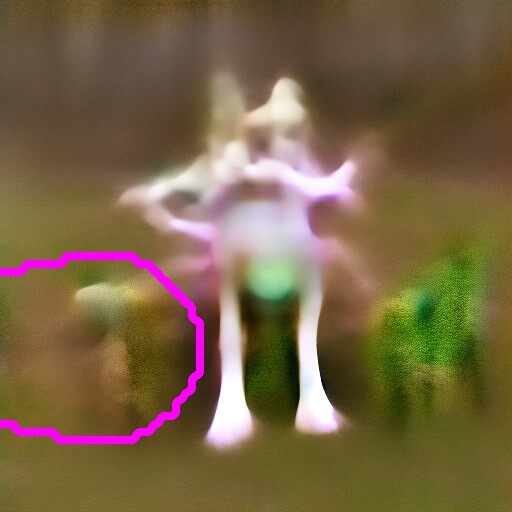}} &
        \raisebox{-.5\height}{\includegraphics[width=\ww]{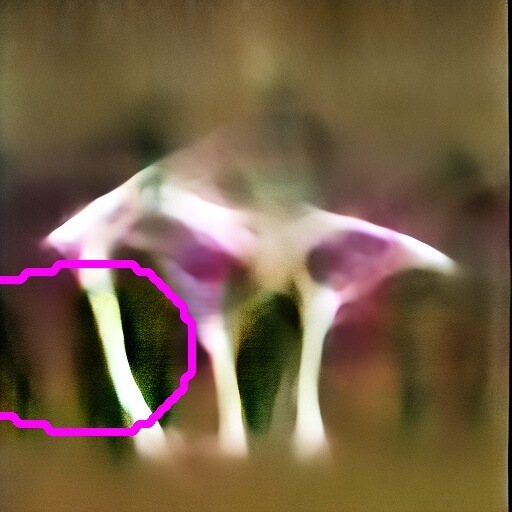}} &
        \raisebox{-.5\height}{\includegraphics[width=\ww]{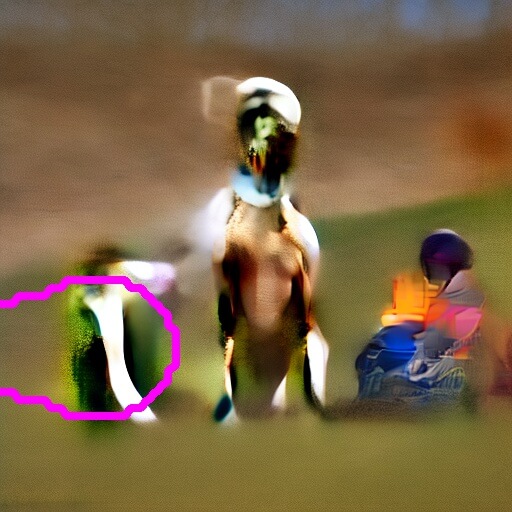}} &
        \raisebox{-.5\height}{\includegraphics[width=\ww]{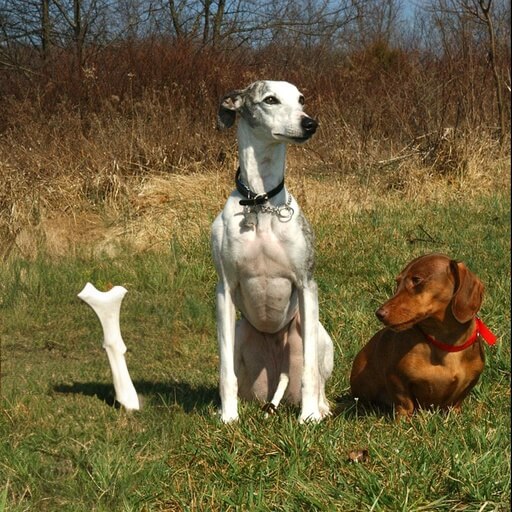}} 
        \\
        [\rss]

        \scriptsize{Input}& \scriptsize{27\%}& \scriptsize{29\%}&  \scriptsize{31\%}& \scriptsize{34\%}& \scriptsize{40\%}& \scriptsize{Output}

    \end{tabular}

    \caption{\textbf{Ablation study: No early mask enlargement.} 
    As explained in \Cref{sec:method}, we start with a large mask ($\sim$16\% of the image), to capture the desired edit in $\mask$. Top: $\mask$ (purple contours on decoded $\est{z}\un{fg}$s, throughout diffusion steps indicated by \%s) evolves without an initial enlargement, and the diffusion guides the white dog to the prompt \textit{``Huge bone"}, while the small $\mask$ fails to capture the bone. Bottom: \ctm's enlarged $\mask$ captures the guided content although the dog is also initially identified as the bone.
    }
    \label{fig:ablation_no_dilation}
    \vspace{\capspace}
\end{figure}

Commencing the blending at 25\% of the diffusion steps, the initial threshold value $\thresh\un{init}$ is relatively low, such that $\mask$ is sufficiently large at the beginning ($\sim$16\% of the image). This enables BLD to capture the desired edit, as demonstrated in \Cref{fig:ablation_no_dilation} (this idea was originally introduced in BLD to cope with the case of small or thin input masks). On the other hand, to prevent overly large masks that could result in large-scale changes failing to blend seamlessly with the original content, we increase $\tau$ rapidly at the beginning, and delay first potential elevation step (denoted $\startstep$) to 40\% of the total diffusion steps. This ensures potential elevation starts late enough to control mask size but still early enough, when the blended image is noisy and can be modified. We stop mask evolution when the spatial structure is nearly determined (at 50\% of the total diffusion steps, denoted $\laststep$).

The potential elevation is obtained by generating the estimated final image $\est{x}_{0}$ at each step, and calculating the cosine distance between the CLIP \cite{radford2021learning} embeddings of $\est{x}_{0}$ and the guidance prompt $\prompt$. $\est{x}_{0}$ is obtained by blending a predicted final foreground latent $\est{z}\un{fg}$, with the original latent background $z\un{init}$:
\begin{equation}
  \est{z}_{0} = \est{z}\un{fg} \odot \mask + z\un{init} \odot (1 - \mask)
\end{equation}
The decoded $\est{x}_{0} = D(\est{z}_{0})$ is passed alongside the current mask $\mask$ and the prompt $\prompt$ to \alphaclip\, to focus on the area of $\mask$. The gradient of the cosine distance with respect to the latent mask pixels is then calculated by backpropagating through the CLIP embeddings and the decoder. The larger the absolute gradient of the cosine distance (\ie\, CLIP loss) with respect to a pixel in $\mask$, the more significant this location is for the alignment of the generated content to the prompt $\prompt$. 
Adding the absolute gradient values $G$ to $\potential$, elevates important areas in the $\potential$ height-field (around $\mask$'s contour for stable evolution -- \Cref{fig:ablation_elevate_all,fig:ablation_no_outer_elevation} in \supl). 

Halfway through the mask evolution steps (denoted $\stopstep$), we initiate an optional stoppage of $\mask$'s evolution if the \alphaclip\ loss does not decrease in subsequent steps.

Starting from the first $\potential$ elevation step $\startstep$, after each update of $\mask$, we restart the BLD process, letting it proceed from the beginning to the current step $t$, using the mask $\mask$ as a fixed mask. This is to allow pixels that were added (or removed) in $\mask$ to affect the generated image from the beginning (see \Cref{fig:ablation_rerun} in \supl).

We then apply \Cref{bld_blend} and blend $z\un{fg}$ with $z\un{bg}$ using the mask $\mask$, which provides $z_{t-1}$, the input to next step. 

After all mask evolution steps have been completed, we perform a final BLD run using the final $\mask$ with several seeds to obtain several candidate results, where the best one is filtered by \alphaclip. 
As noted earlier, rather than fine-tuning the VAE decoder weights to preserve the original background details outside the mask, we employ instead a simple Gaussian mask feathering when blending the BLD output and the original input image (in pixel space).

\newcommand{\algofont}[1]{\fontsize{9pt}{10pt}\selectfont #1\normalsize}

\begin{figure}[t]
\setlength{\capspace}{-20px}
\begin{algorithm}[H]
    \footnotesize
    \caption{
    \ctmb} 
    \label{alg:click2mask}
    
    \begin{algorithmic}
        \algofont{
        \State
        \textbf{Given:} \textbf{models} $\textit{LDM} = \{\textit{noise}(z, t), \textit{denoise}(z, \prompt, t) \rightarrow (z_{t}, z_{0}) \}$, $\textit{VAE} = \{E(\img), D(z)\}$, 
        $BLD = \{(\img, \prompt, m, t) \rightarrow {z_{t}}\}$, $\textit{Alpha-CLIP} = \{\alpha\un{CLIP}(\img, m, \prompt) \rightarrow Sim\un{CLIP}\}$, and \textbf{hyper parameters} $\{\thresh_{\diffsteps \ldots \laststep}, lr\}$ with schedulers $\{\diffsteps, \startstep, \stopstep, \laststep\}$
        \State
        
        \State \textbf{Input:} input image $\img$, text prompt $\prompt$, target coordinates $\coordinates$
        \State \textbf{Output:} edited image $\widehat{\img}$ that 
        matches the prompt $\prompt$ in proximity of $\coordinates$, and complies to $\img$ outside edited region 
        
        \State
        \State $\potential = \textit{Gaussian(\coordinates)}$
        \State $z\un{init} = E(\img)$
        \State $z_{\diffsteps} \sim \textit{noise}(z\un{init}, \diffsteps)$
        \For{all $t$ from $\diffsteps$ to $\laststep$}
            \State $z_{\textit{bg}} \sim \textit{noise}(z\un{init}, t)$
            \State $z\un{fg}, \est{z}\un{fg} \sim \textit{denoise}(z\un{t}, \prompt, t)$
            
            \State $G = 0$
            \If {$t < \startstep$} 
                \State $\est{z}_{0} = \est{z}\un{fg} \odot \mask + z\un{init} \odot (1 - \mask)$
                \State $S_{t} \sim \alpha\un{CLIP}(D(\est{z}_{0}), \textit{upscale}(\mask), \prompt)$
                \State $G \sim |\textit{gradients}(S_{t}, \mask)|$
                \State $z\un{fg} \sim \textit{BLD}(\img, \prompt, \mask, t)$

            \EndIf  
            \State \textbf{if } $t < \stopstep$ and $S_{t} > S_{t+1}$ \textbf{then exit loop}
            \State $\mask = (\potential + G * lr) > \thresh_{t}$
            \State $z_{t} = z\un{fg} \odot \mask + z\un{bg} \odot (1 - \mask)$

        \EndFor

        \State $\widehat{z} \sim \textit{BLD}(\img, \prompt, \mask, 0)$
        \State \Return $D(\widehat{z})$
        }

    \end{algorithmic}
\end{algorithm}
\vspace{\capspace}
\end{figure}

\begin{figure*}
    \centering
    \setlength{\ww}{2\columnwidth}
    \setlength{\capspace}{\defcapspace}
    
    \includegraphics[width=\ww]{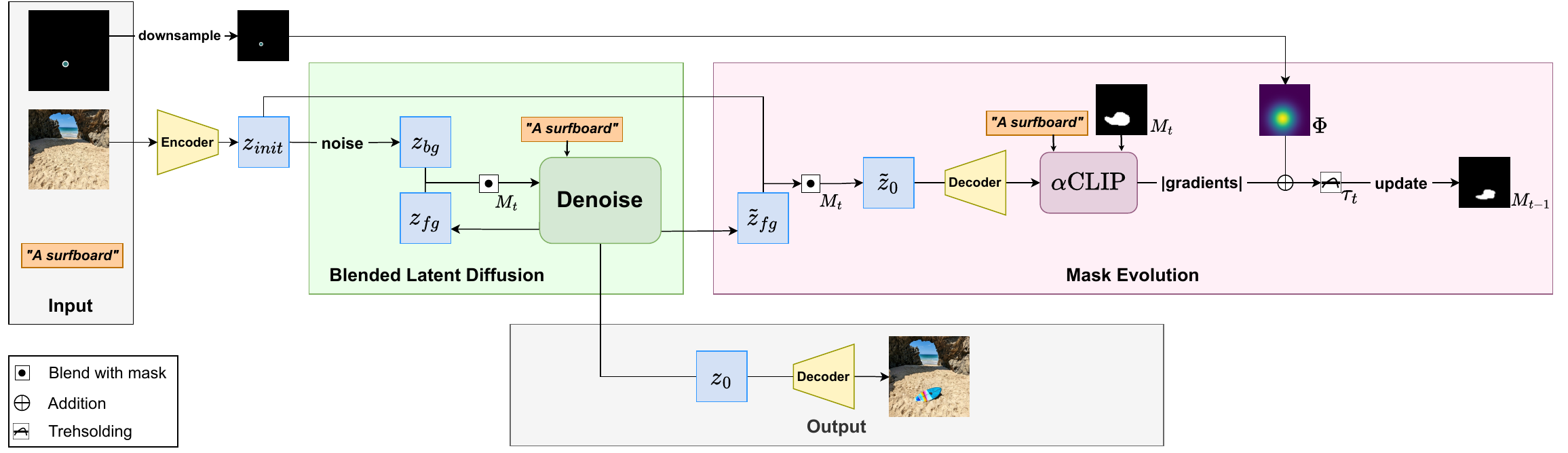}
   
    \caption{\textbf{\ctm:} An illustration of our method as described in 
    \Cref{alg:click2mask}. The \textbf{green block} is BLD process, performing diffusion steps while blending noised input latents with text guided latents. 
    The \textbf{pink block} is the mask evolution process, where we utilize \alphaclip\ to evaluate the gradients with respect to the mask $\mask$ pixels, using them to update $\mask$, obtaining $M_{t-1}$.}
   
    \label{fig:method_illusration}
    \vspace{\capspace}
\end{figure*}

\section{Results}
\label{sec:results}

\begin{figure}[ht]
    \centering
    \setlength                  {\tabcolsep}{0.5pt}
    \renewcommand               {\arraystretch}{0.5}
    \setlength{\ww}             {0.196\columnwidth}
    \newcommand{\rowspace}      {\rule{0pt}{2ex}}
    \renewcommand{\prsize}      {\defprsize}
    \renewcommand{\methsize}    {\defmethsize}
    \setlength{\capspace}{\defcapspace}
    
    \definecolor{nochange}{RGB} {255, 14, 14}
    \definecolor{replace}{RGB}  {105, 53, 255}
    \definecolor{place}{RGB}    {54, 178, 23} 
    \definecolor{whole}{RGB}    {98, 98, 98}
    \definecolor{other}{RGB}    {200, 16, 200}

    \newcommand{\signsize}  {\large}
    \newcommand{\nochange}  {\textcolor{nochange}{\signsize{$\varnothing$}}}
    \newcommand{\replace}   {\textcolor{replace}{\signsize{$\Leftrightarrow$}}}
    \newcommand{\place}     {\textcolor{place}{\signsize{$\mathbf{\looparrowleft}$}}}
    \newcommand{\whole}     {\textcolor{whole}{\signsize{$\blacksquare$}}}
    \newcommand{\other}     {\textcolor{other}{\signsize{$\barwedge$}}}

    \begin{tabular}{c cc cc}   
        \sizedtext{\methsize}{Input} &
        \sizedtext{\methsize}{\emu} &
        \sizedtext{\methsize}{\mb} &
        \sizedtext{\methsize}{\ipp} &
        \sizedtext{\methsize}{\ctmb}
        \\

        \raisebox{-.5\height}{\includegraphics[width=\ww]{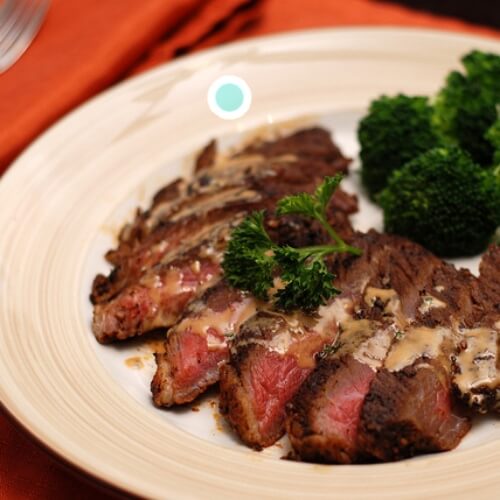}} &
        \raisebox{-.5\height}{\includegraphics[width=\ww]{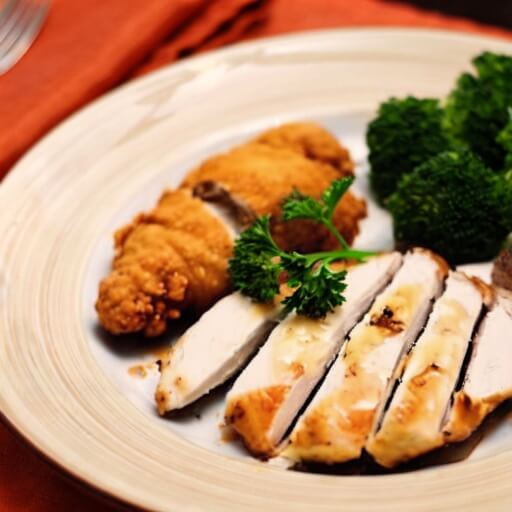}}  &
        \raisebox{-.5\height}{\includegraphics[width=\ww]{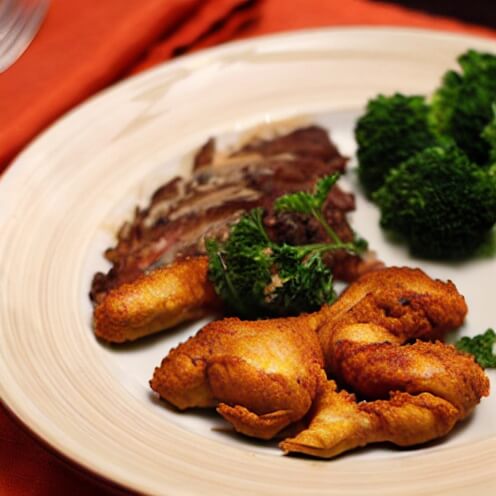}} &
        \raisebox{-.5\height}{\includegraphics[width=\ww]{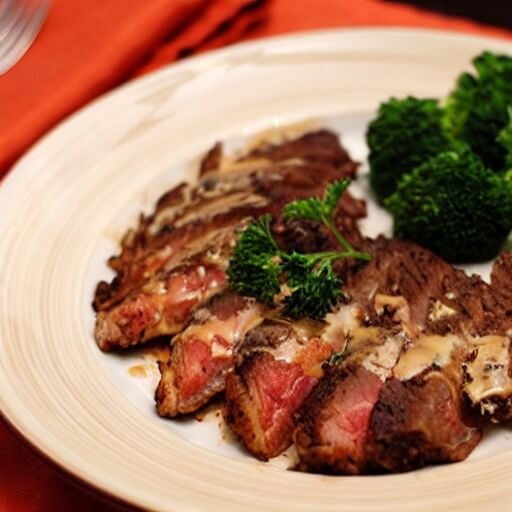}} &
        \raisebox{-.5\height}{\includegraphics[width=\ww]{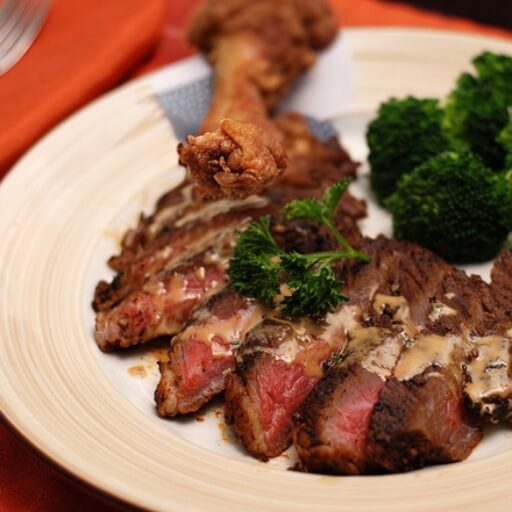}}
        \\
        & \replace & \replace & \nochange &  
        \\
        \prprunmerge{\prsize}{5}{Add a piece of fried chicken to the plate}{Piece of fried chicken}
        \\
        \\
        \\
        
        \raisebox{-.5\height}{\includegraphics[width=\ww]{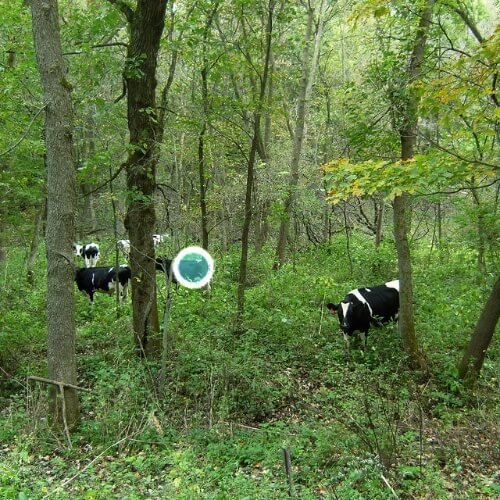}} &
        \raisebox{-.5\height}{\includegraphics[width=\ww]{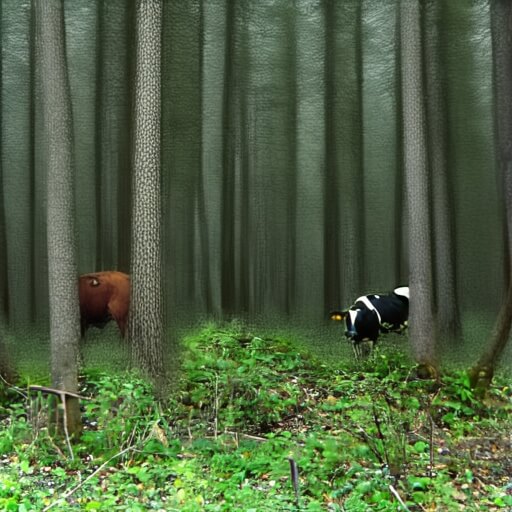}} &
        \raisebox{-.5\height}{\includegraphics[width=\ww]{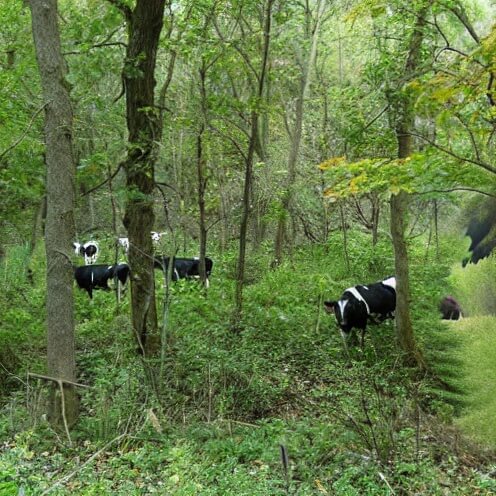}}  &
        \raisebox{-.5\height}{\includegraphics[width=\ww]{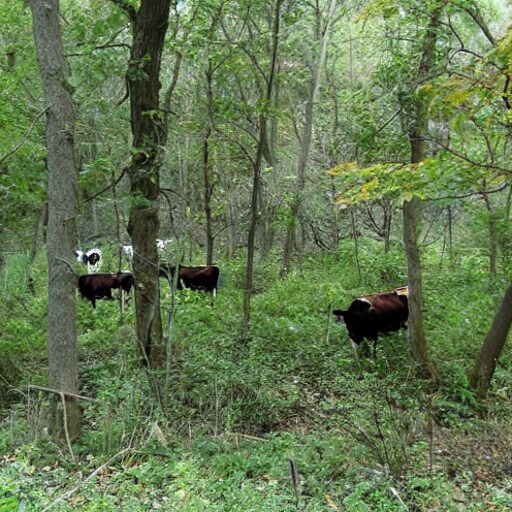}} &
        \raisebox{-.5\height}{\includegraphics[width=\ww]{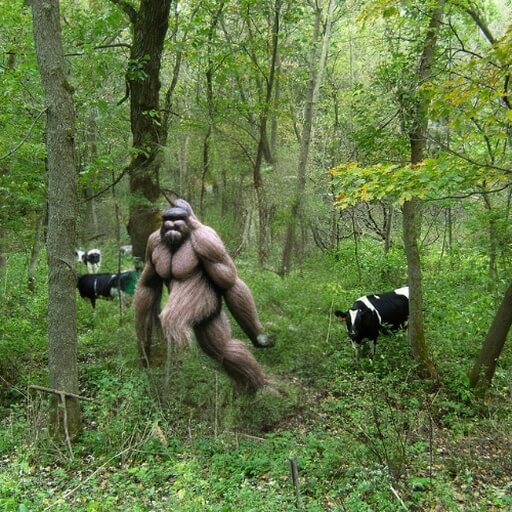}}
        \\
        \rowspace
        & \other , \whole & \other & \other & 
        \\
        \prprunmerge{\prsize}{5}{Add Bigfoot in the background along-side one of the cows}{Bigfoot}
        \\
        \\
        \\

        \raisebox{-.5\height}{\includegraphics[width=\ww]{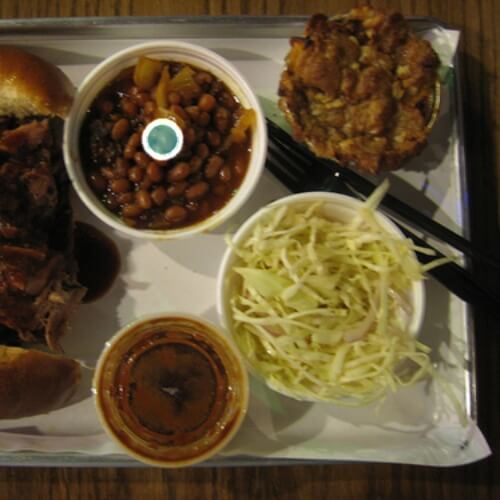}} &
        \raisebox{-.5\height}{\includegraphics[width=\ww]{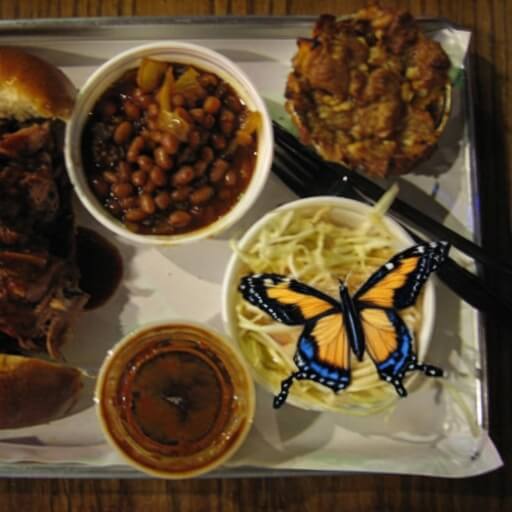}}  &
        \raisebox{-.5\height}{\includegraphics[width=\ww]{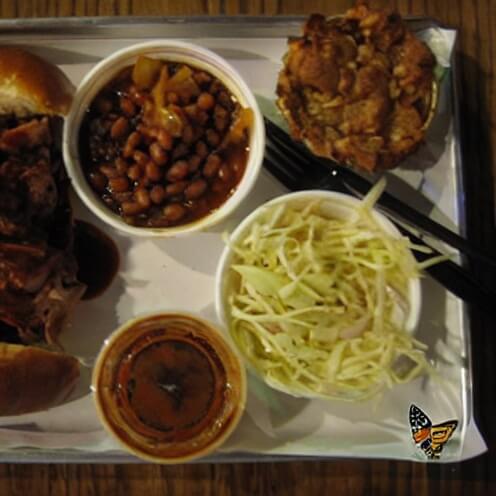}}  &
        \raisebox{-.5\height}{\includegraphics[width=\ww]{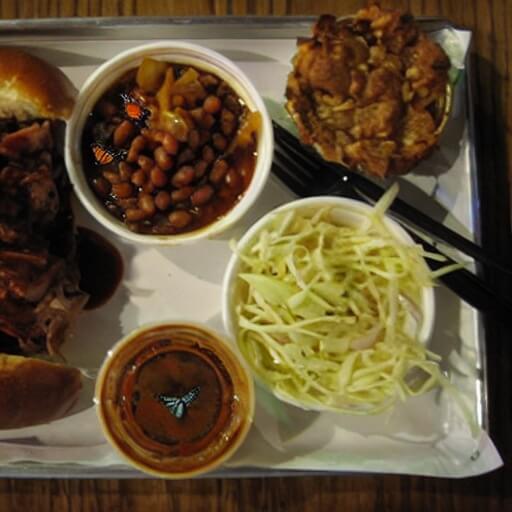}}  &
        \raisebox{-.5\height}{\includegraphics[width=\ww]{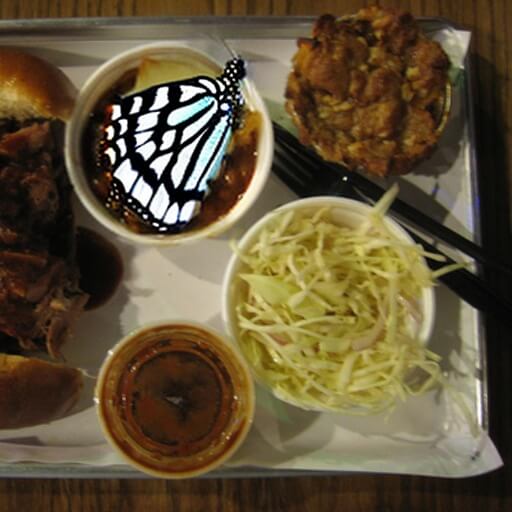}}
        \\
        \rowspace
        & \place & \place & \place &  
        \\
        \prprunmerge{\prsize}{5}{Add a butterfly on top of the beans}{A butterfly}
        \\
        \\
        \\

        \raisebox{-.5\height}{\includegraphics[width=\ww]{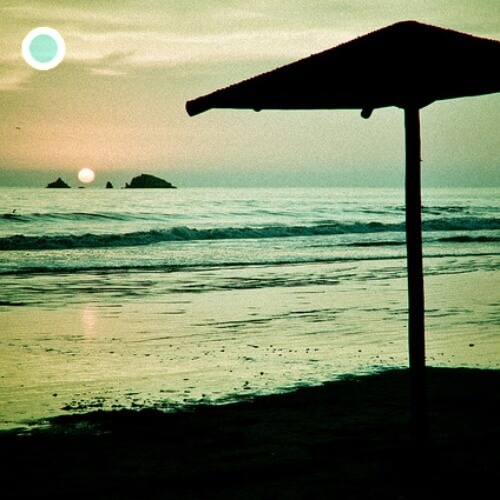}} &
        \raisebox{-.5\height}{\includegraphics[width=\ww]{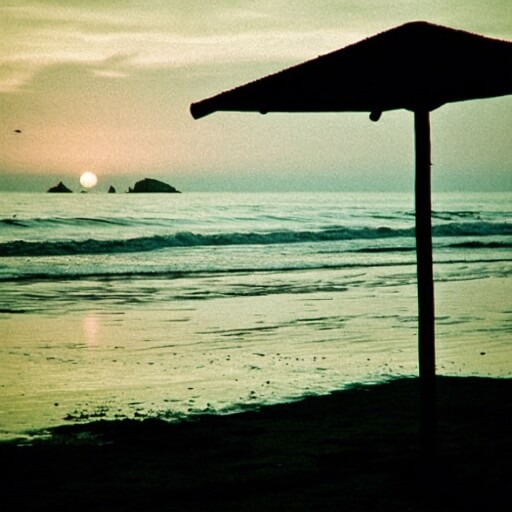}} &
        \raisebox{-.5\height}{\includegraphics[width=\ww]{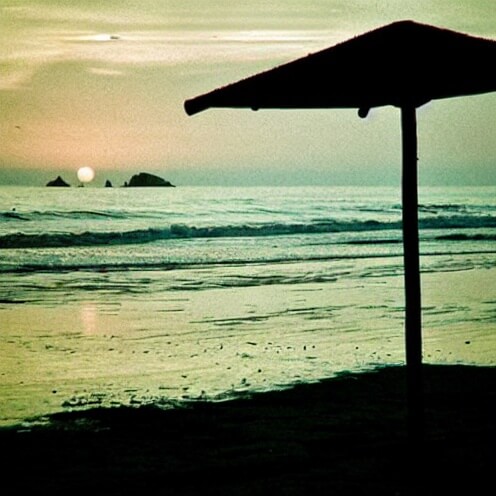}}  &
        \raisebox{-.5\height}{\includegraphics[width=\ww]{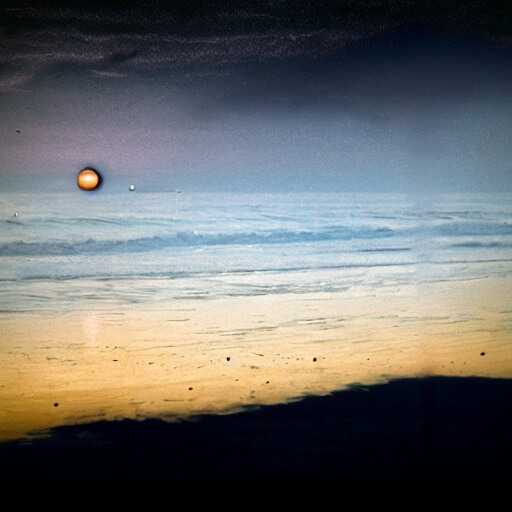}}  &
        \raisebox{-.5\height}{\includegraphics[width=\ww]{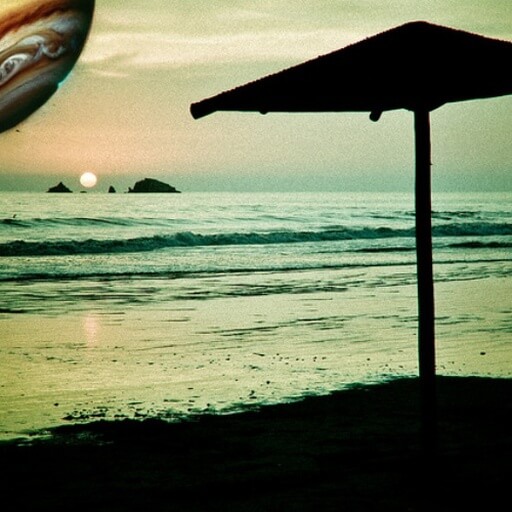}}
        \\
        & \nochange & \nochange & \whole & 
        \\
        \prprunmerge{\prsize}{5}{Add Jupiter to the sky}{Jupiter}
        \\

    \end{tabular}
    
    \caption{\textbf{Failure cases of baselines}. Baselines suffer occasionally from replacing an existing object instead of adding one (\replace), misplacing the object (\place), modifying other objects (\other), altering the image globally (\whole), or failing to produce an edit (\nochange). For additional comparisons to baselines, see \Cref{fig:comparison_1} and \supl.}
    \label{fig:compare_issues}
    \vspace{\capspace}
\end{figure}

Given that our method is mask-free, we compare ourselves to mask-free image editing methods, with the slight difference being that a \textit{clicked point} replaces location-describing text in the prompt. As our paradigm is novel and lacks a directly aligned method for comparison, using a single click instead of detailed text is a reasonable trade-off. To begin with, we compare to \mb, which is the SoTA method among the open-source models. In addition, we compare to \emu, which is the SoTA among closed-source models. Since we are unable to run \emu\ ourselves, we must rely on the \emu\ Benchmark \cite{sheynin2023emu}, which includes images generated by \emu. This benchmark contains images with several categories of editing instructions, such as adding objects, removing objects, altering style, etc. As our focus is adding objects to images, we filtered the dataset by the ``addition'' instruction. This resulted in 533 items, from which we randomly sampled an evaluation subset of 100 samples. 

We perform the following fixed routine for each sample: 
\begin {enumerate*} [label=(\roman*)] \item Removed the word that instructs addition (\eg, ``Add'', ``Insert''), \item removed the part that describes the edit location, and instead \item clicked on the image to direct the editing location. For instance, the instruction ``Add a black baseball cap to the man on the left'' becomes ``A black baseball cap'' (non-localized instruction).
\end{enumerate*} 

Following \emu\ \cite{sheynin2023emu} and BLD \cite{avrahami2023blendedlatent}, each sample run produces multiple results internally (comprising three mask evolutions, each followed by a batch of 8 outputs), and outputs the best result, as determined automatically using \alphaclip\ scoring.

To evaluate our results, we compared these 100 outputs generated by \ctm, with the outputs generated by \emu\ and by \mb\ (which ran with the original edit instructions).
We conducted the evaluation through a user study (\Cref{human_study}), as well as through automatic metrics (\Cref{metrics_study}). In both cases, our method outperformed the SoTA methods.

\subsection{Human Evaluation}
\label{human_study}
We conducted a user study, where participants were given a random batch of survey items out of 200 total items (100 items comparing to each model). Each item included an input image, the original edit instruction, and a pair of edited images: one generated by our model, and the other generated by either \emu\ or \mb. Participants were asked to rank which of the edited images performed better according to three criteria: executing the instruction, not adding any other edits or artifacts, and generating a realistic image. 
The survey was completed by 149 participants. Each of the 200 items was rated by at least 5 users, where the average rate was 15.67 users in \emu, and 8.06 users in \mb. 

In order to compare \ctm\ vs.~\emu, as well as \ctm\ vs.~\mb, while taking into account ``ties'' (ratings stating equal performance on an item, or items with equal ratings to both methods), we analyzed the results using the following metrics: \begin {enumerate*} [label=(\Alph*)] \item The percentage of items in which each method was preferred by the majority, disregarding ties. \item For each item we counted if the majority voted for a tie, and if so marked it as a ``tied item''. For the other ``non-tied items'', we conducted the same majority vote analysis described in A. \item The number of total ratings for each method. \end{enumerate*} In each parameter our method surpassed the closed-source SoTA method \emu, and the open-sourced SoTA \mb, as shown in \Cref{tab:user_study_comparison}. See \Cref{fig:teaserfigure,fig:comparison_1} (and \Cref{fig:comparison_2,fig:comparison_3,fig:comparison_4,fig:comparison_5} in \supl) for qualitative comparisons to baselines alongside \ipp, and \Cref{fig:compare_issues} for a detailed comparison. Statistical significance analysis is provided in the \supl.

\subsection{Automatic Metrics}
\label{metrics_study}
Utilizing the input captions and output captions (describing the desired output) provided in \emu\ benchmark, a variety of metrics were used to assess each method's outputs on the sampled items:
\begin {enumerate*} [label=(\roman*)]
    \item Directional CLIP~\cite{Gal2022} similarity ($\text{CLIP}_\textit{direct}$) To measure the alignment between changes in input and output images and their corresponding captions.
    \item CLIP similarity between the output image and output caption to evaluate alignment with desired outputs $\text{CLIP}_\textit{out}$).
    \item Mean L1 pixel distance between input and output images, to measure the amount of change in the entire image (L1).
    \item In addition, we present a new metric, Edited Alpha-CLIP ($\alpha\text{CLIP}_\textit{edit}$).  
\end{enumerate*}

\subsubsection{Edited Alpha-CLIP.}
Besides evaluating the images \emph{globally}, it is beneficial to evaluate the \emph{edited region}. We offer an Edited Alpha-CLIP procedure to overcome the lack of input or output masks in \emu\ and \mb: we extract a mask specifying the edited area in the generated image, and calculate the \alphaclip\ similarity between the masked generated image and the instruction (removing words describing addition and edit locations as mentioned in \Cref{sec:results}). See \Cref{edited_alphaclip_mask_extraction} and \Cref{fig:edited_alpha_clip} in \supl\ for details and extracted masks demonstrations.

\Cref{tab:automatic_metrics_comparison} shows that our method surpassed both \emu\ and \mb\ in all metrics: higher scores in all CLIP-based metrics, indicating stronger similarities, and lower L1 distance indicating better compliance with input image.

\begin{table}
    \renewcommand{\arraystretch}{1.2}
    \centering
    \setlength{\tabcolsep}{2.75mm} 
    \setlength{\capspace}{-5px}
    
    \fontsize{7}{9}\selectfont
        \begin{tabular}{>{\columncolor[gray]{0.95}}l | c | c c| c}
            \multicolumn{1}{c}{}& \multicolumn{1}{c}{(A)} & \multicolumn{2}{c}{(B)} & \multicolumn{1}{c}{(C)}
            \\
            \toprule
            Method & 
            \% Majority &
            \makecell{\% Tied\\items} &
            \makecell{\% Majority\\from non-tied} &
            \makecell{\# Total \\ votes}
            \\
            
            \midrule

            \emu& 
            42.86\% &
            \multirow{2}{*}{35\%} &
            47.69\% &
            416
            \\
            
            \ctmb & 
            \textbf{57.14\%} &
            &
            \textbf{52.31\%} &
            \textbf{465}
            \\

           \midrule
           
            \mb& 
            16.30\% &
            \multirow{2}{*}{27\%} &
            15.07\% &
            148
            \\
            
            \ctmb & 
            \textbf{83.70\%} &
            &
            \textbf{84.93\%} &
            \textbf{362}
            \\
            
            \bottomrule
        \end{tabular}
    \caption{\textbf{Human evaluation results.} Comparisons of (A): \% of items each method received majority votes, disregarding ties. (B): \% of items the majority voted as tie (left), and \% of items --~out of the other non-tied items~-- each method received majority votes (right). (C): Total votes. Refer to \Cref{sec:results} for details.
    }
    \label{tab:user_study_comparison}
    \vspace{\capspace}
\end{table}

\begin{table}
    \centering

    \renewcommand{\arraystretch}{1.2}
    \setlength{\tabcolsep}{1.9mm}  
    \setlength{\capspace}{-5px}
    
    \fontsize{8}{9}\selectfont
    
        \begin{tabular}{>{\columncolor[gray]{0.95}}l | c c c c}
            \toprule
            
            Method & 
            $\text{CLIP}_\textit{direct}\uparrow$ &
            $\text{CLIP}_\textit{out}\uparrow$ &
            $\alpha\text{CLIP}_\textit{edit}\uparrow$ &
            L1 $\downarrow$
            \\
            
            \midrule

            \emu& 
            0.150 &
            0.331 &
            0.186 &
            0.046
            \\

            \mb& 
            0.095 &
            0.324 &
            0.166 &
            0.049
            \\

            \ctmb & 
            \textbf{0.204} &
            \textbf{0.334} &
            \textbf{0.195} &
            \textbf{0.027}
            \\
            
            \bottomrule
        \end{tabular}
    \caption{\textbf{Automatic metrics results.} Evaluation using automatic metrics. $\text{CLIP}_\textit{direct}$ measures consistency between changes (from input to output) in images and captions, $\text{CLIP}_\textit{out}$ measures similarity between output image and output caption, $\alpha\text{CLIP}_\textit{edit}$ measures similarity to the non-localized instruction in the edited area, and L1 measures the alignment with the input image. See \Cref{sec:results} for details.}
    \label{tab:automatic_metrics_comparison}
    \vspace{\capspace}
\end{table}

\subsection{Ablation Study}
\label{sec:ablation_study}

We conducted several ablation studies to analyze the impact of various components on the overall performance of our model. \Cref{fig:ablation_no_dilation} demonstrates the need for a sufficiently large mask on early diffusion steps. See additional ablation studies in \Cref{additional_ablation} accompanied by \Cref{fig:abl_background_preservation,fig:ablation_rerun,fig:ablation_elevate_all,fig:ablation_no_outer_elevation,fig:ablation_continuous,fig:abl_naive}.

\section{Model Limitations}
\label{sec:model_limitations}

\begin{figure}[ht]
    \centering
    \setlength{\tabcolsep}{0pt}
    \renewcommand{\arraystretch}{0.5}
    \setlength{\ww}{0.195\columnwidth}
    \renewcommand{\methsize}{\defmethsize}
    \renewcommand{\prsize}{\defprsize}
    \renewcommand{\pw}{0.95\ww}
    \renewcommand{\rss}{\defrss}
    \setlength{\capspace}{\defcapspace}

    \begin{tabular}{c @{\hspace{0.01\columnwidth}}c c c c}
        \\
        \raisebox{-.5\height}{\includegraphics[width=\ww]{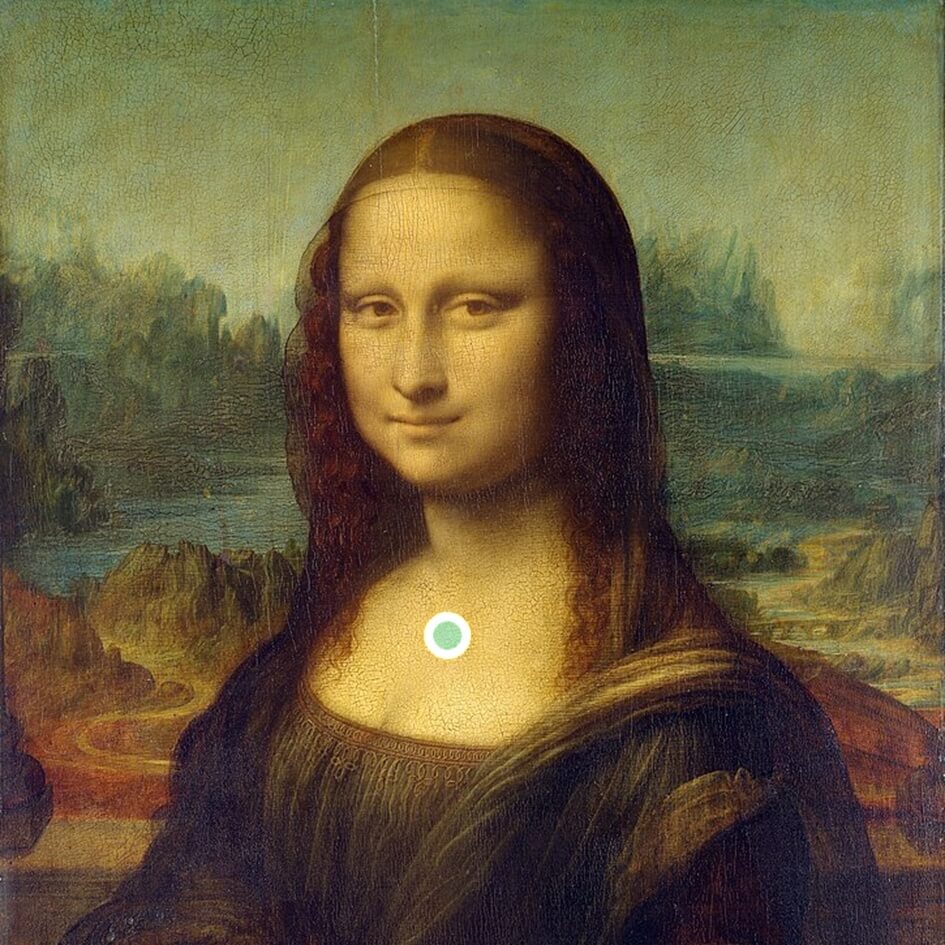}}    &
        \raisebox{-.5\height}{\includegraphics[width=\ww]{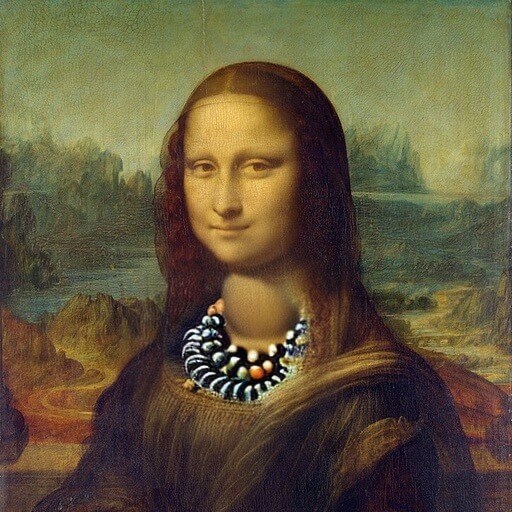}}                 &
        \raisebox{-.5\height}{\includegraphics[width=\ww]{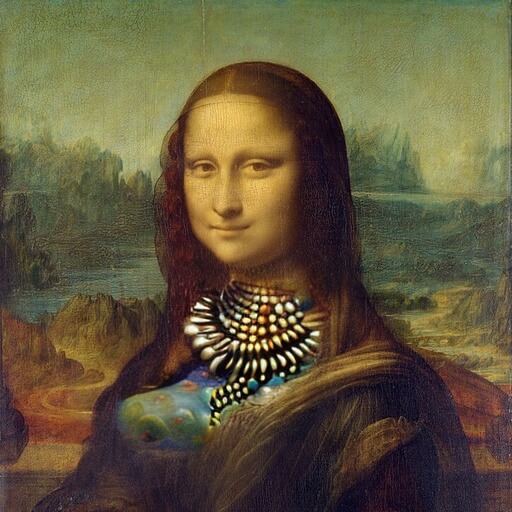}}                 &
        \raisebox{-.5\height}{\includegraphics[width=\ww]{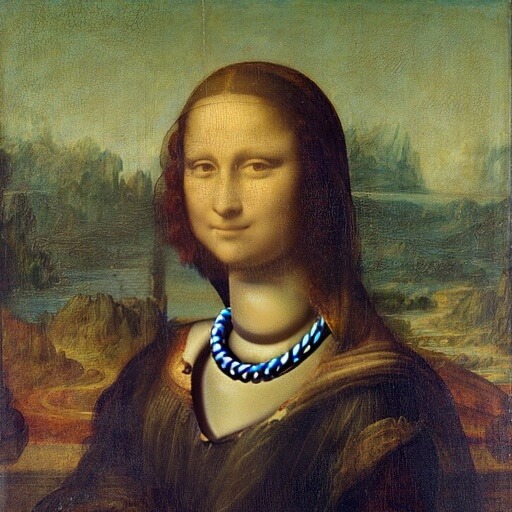}}                 &
        \raisebox{-.5\height}{\includegraphics[width=\ww]{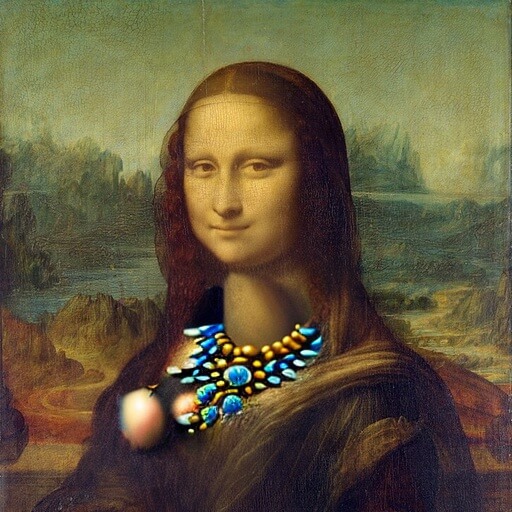}}
        \\
        [\rss]
        \sizedtext{\methsize}{Input}
        &
        \prunmerge{\prsize}{4}{A golden necklace}
        \\
        \\
        \raisebox{-.5\height}{\includegraphics[width=\ww]{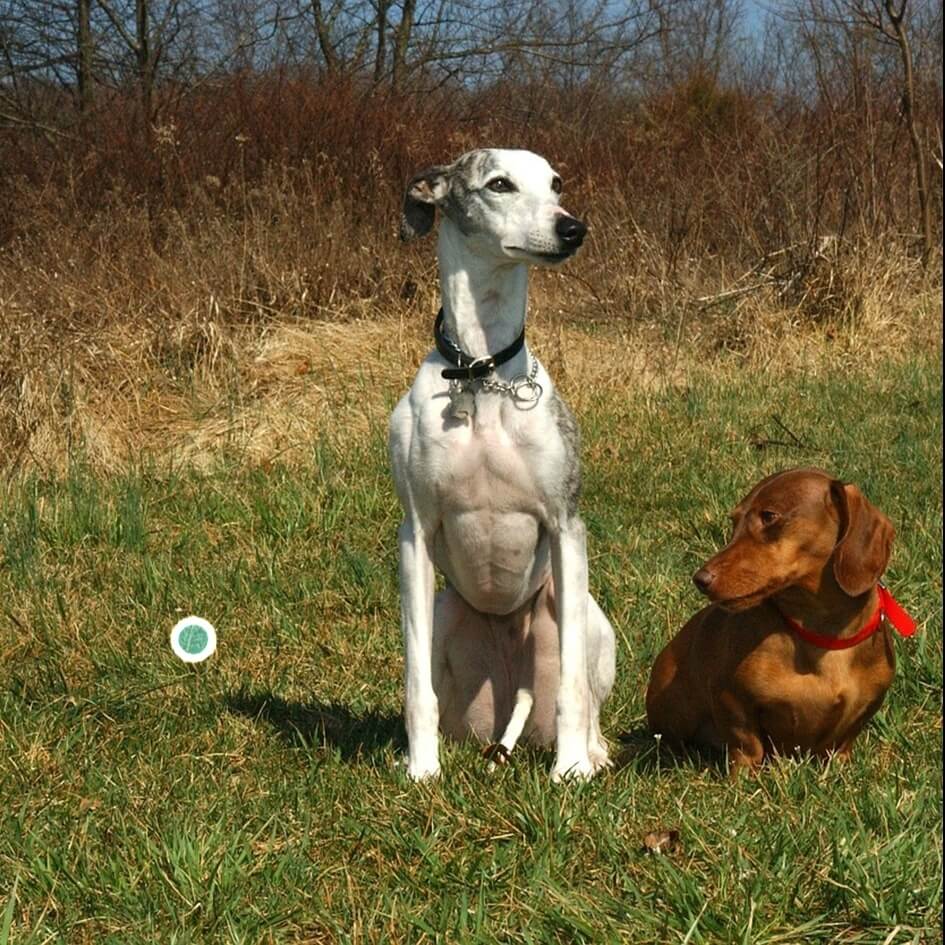}}          &
        \raisebox{-.5\height}{\includegraphics[width=\ww]{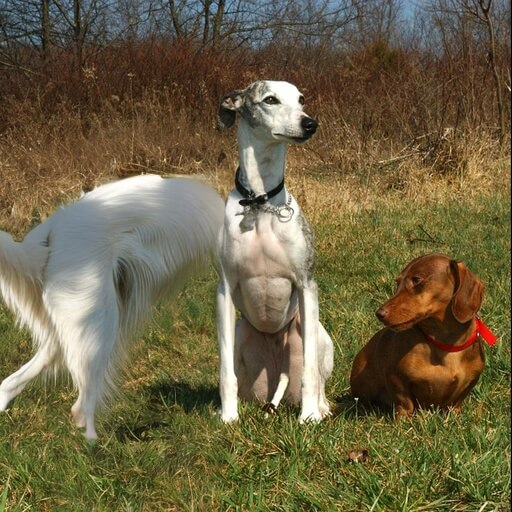}}                  &
        \raisebox{-.5\height}{\includegraphics[width=\ww]{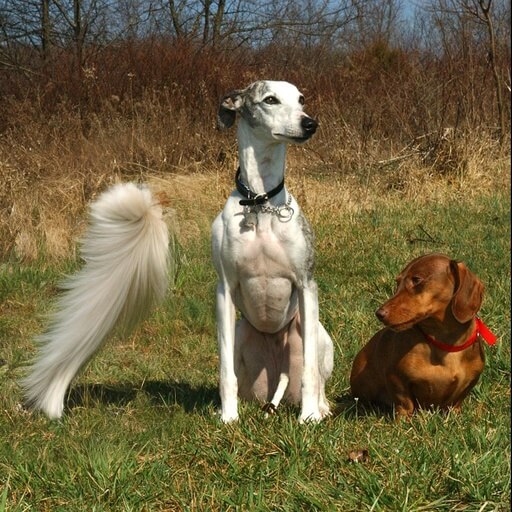}}                   &
        \raisebox{-.5\height}{\includegraphics[width=\ww]{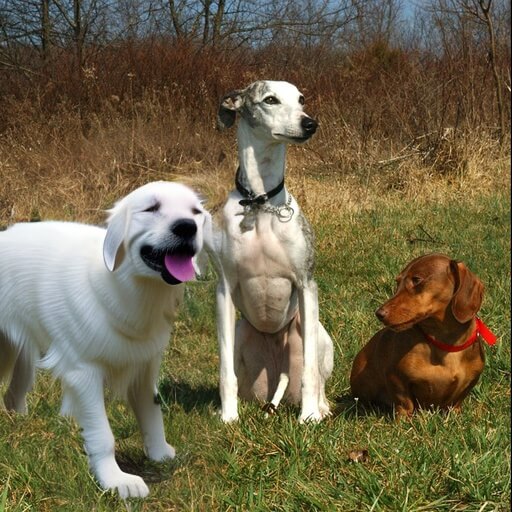}}                   &
        \raisebox{-.5\height}{\includegraphics[width=\ww]{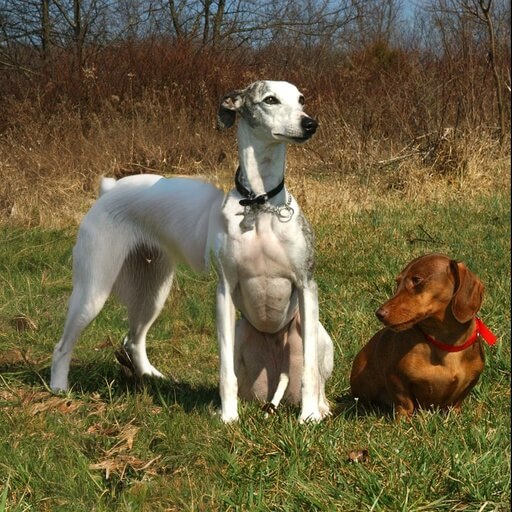}}
        \\
        [\rss]
        \sizedtext{\methsize}{Input}
        &
        \prunmerge{\prsize}{4}{A white dog}
        
    \end{tabular}
    
    \caption{\textbf{Limitations.} Top: the evolving mask struggles to converge to a small, fine-detailed shape like a golden necklace. Bottom: when prompt content (\ie\ white dog) already exists near the generated mask, our Stable Diffusion and BLD backbones, which guide the image globally to the prompt, may fail to confine guidance to the masked region.}
    \label{fig:limitations}
    \vspace{\capspace}
\end{figure}

During the evolution process, our model encounters difficulty converging to small, finely detailed mask shapes (e.g., a dog collar). This stems from hyperparameter choices balancing an initial large mask to capture the object, and a non-aggressive shrinkage rate to avoid boundary cropping. Alternative configurations might achieve smaller masks.

Additionally, since text guidance in Stable Diffusion is not spatially driven, BLD sometimes has difficulty adding the desired object to the masked area when a similar object is nearby in the unmasked area (e.g., adding a Bigfoot next to a person). Since we use BLD as our backbone, we sometimes encounter this problem. However, we have considerably improved it in comparison to BLD by optimizing the progressive mask shrinking process, and applying it across all objects, not just thin objects, as part of our mask evolution process. Moreover, in comparison to other SOTA methods, they often fail to add the desired object even if a similar one is not present, and our method outperforms them in both cases. See \Cref{fig:limitations} for examples of these cases.

\section{Conclusion} 
\label{sec:conclusion}
\ctm\ presents a novel approach for local image generation, freeing users from having to specify a mask, or describing the input or target images, and without being constrained to existing objects. We look forward to users applying our method with the source code that is available in the project page\projecttext{ (see Footnote in \Cpageref{projectpage})}, either to edit images or to embed the method for generating or fine-tuning masks.

\FloatBarrier

\bigskip
\bibliography{main.bib}
\clearpage
\newpage

\appendix

\section*{\huge{Click2Mask: Local Editing with Dynamic Mask Generation} \\ \textit{Appendix}}

\bigskip

\section{Additional Experiments}
In \Cref{additional_results} we start by providing additional \ctm\ generated masks examples, as well as further comparisons to the baselines \emu, \mb, and \ipp. In \Cref{additional_ablation} additional ablation tests are provided. \Cref{statistical_analysis} shows a statistical analysis held on the results from the user case study in \Cref{sec:results}.

\begin{figure*}[t]
    \centering
    \setlength{\tabcolsep}{0.5pt}
    \renewcommand{\arraystretch}{1.2}
    \setlength{\ww}{0.128\textwidth}
    \renewcommand{\methsize}{\defbigmethsize}
    \renewcommand{\prsize}{\defbigprsize}
    \setlength{\rsm}{32px}
    
    \begin{tabular}{@{\hskip -0.25\ww}c ccc @{\hskip 0.4\ww}c ccc}
        \sizedtext{\methsize}Prompt&
        \sizedtext{\methsize}Input&
        \sizedtext{\methsize}Generated Mask&
        \sizedtext{\methsize}\ctm &
        \sizedtext{\methsize}Prompt&
        \sizedtext{\methsize}Input&
        \sizedtext{\methsize}Generated Mask&
        \sizedtext{\methsize}\ctm 
        \\
        
        \prs{\prsize}{A green bowl} &
        \raisebox{-.5\height}{\includegraphics[width=\ww]{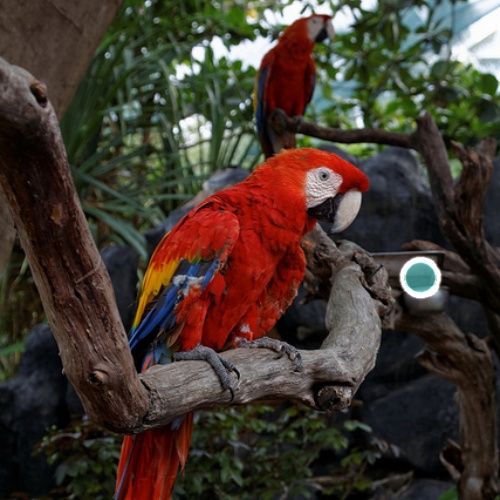}} &
        \raisebox{-.5\height}{\includegraphics[width=\ww]{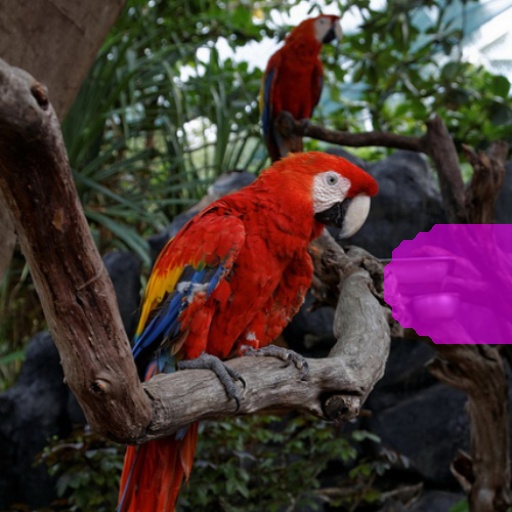}} &
        \raisebox{-.5\height}{\includegraphics[width=\ww]{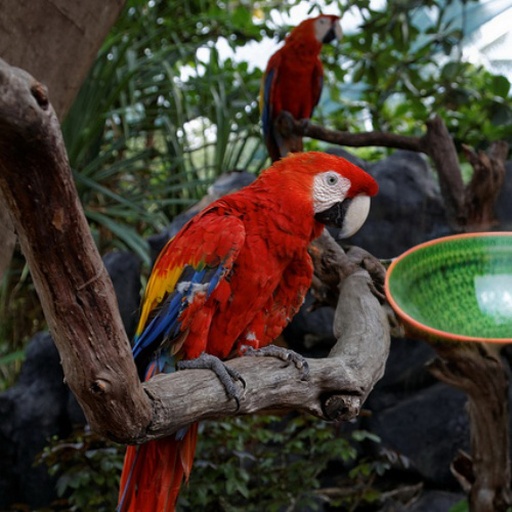}} 
        &
        
        \prs{\prsize}{A butterfly} &
        \raisebox{-.5\height}{\includegraphics[width=\ww]{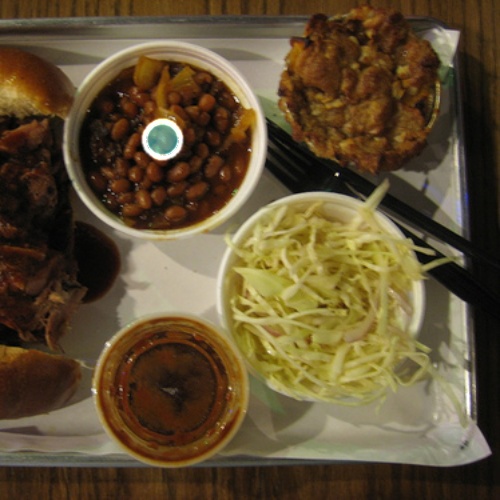}}   &
        \raisebox{-.5\height}{\includegraphics[width=\ww]{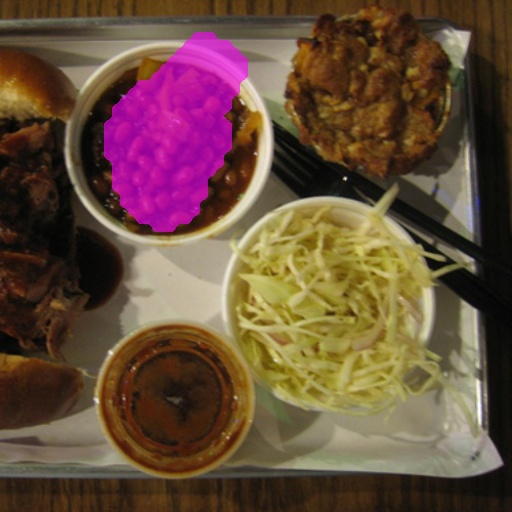}} &
        \raisebox{-.5\height}{\includegraphics[width=\ww]{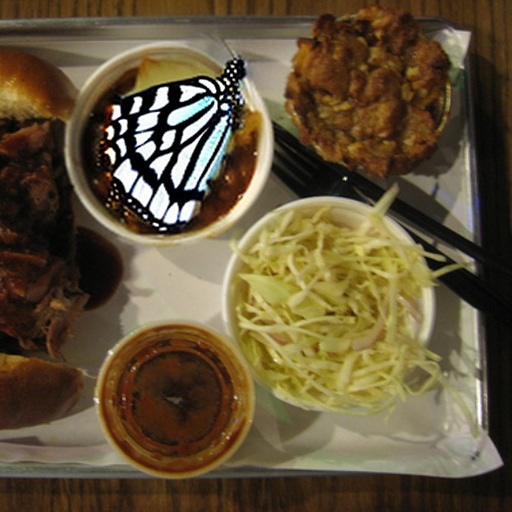}} 
        \\
        [\rsm]

        \prs{\prsize}{A dog sitting} &
        \raisebox{-.5\height}{\includegraphics[width=\ww]{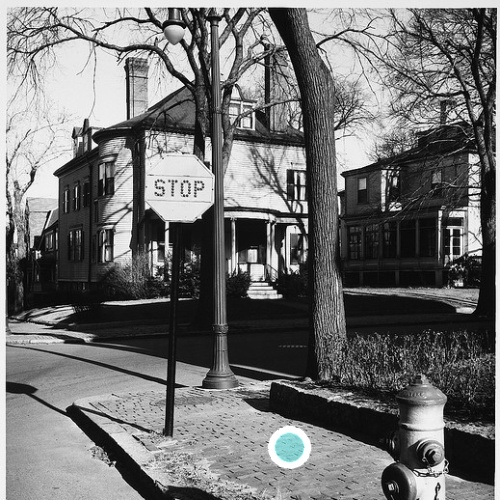}}   &
        \raisebox{-.5\height}{\includegraphics[width=\ww]{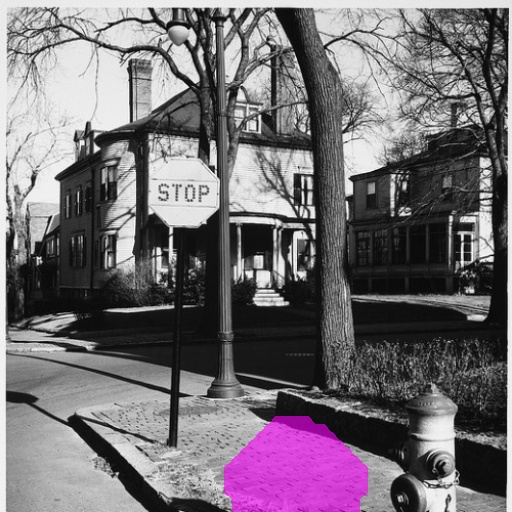}} &
        \raisebox{-.5\height}{\includegraphics[width=\ww]{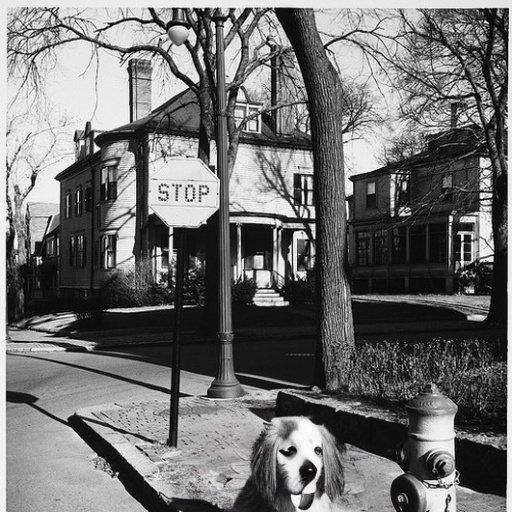}} 
        &

        \prs{\prsize}{A puppy} &
        \raisebox{-.5\height}{\includegraphics[width=\ww]{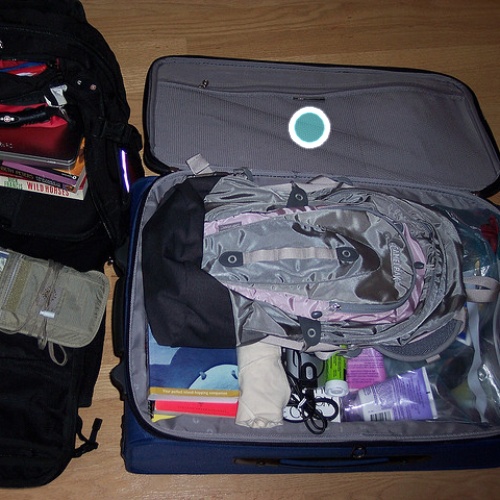}} &
        \raisebox{-.5\height}{\includegraphics[width=\ww]{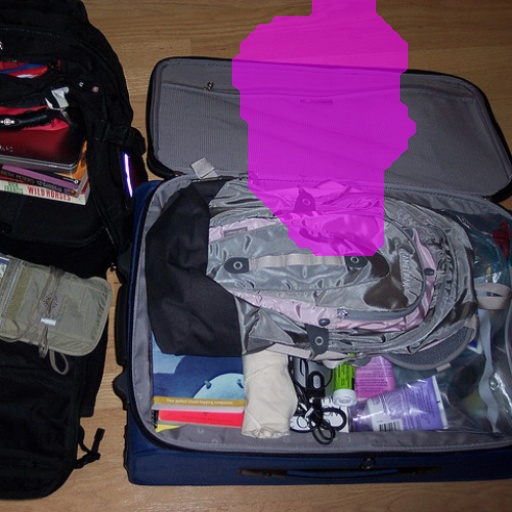}} &
        \raisebox{-.5\height}{\includegraphics[width=\ww]{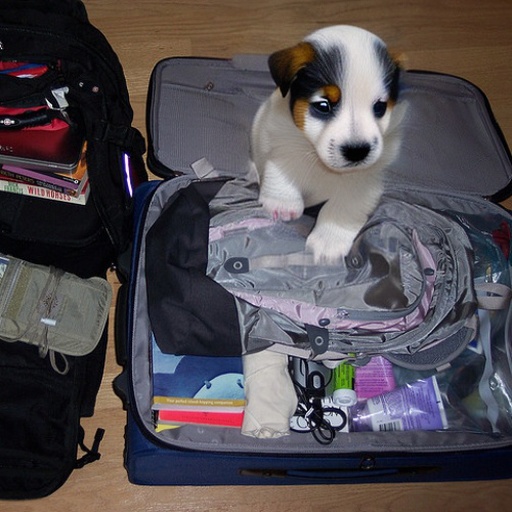}} 
        \\
        [\rsm]
        
        \prs{\prsize}{A baby cow} &
        \raisebox{-.5\height}{\includegraphics[width=\ww]{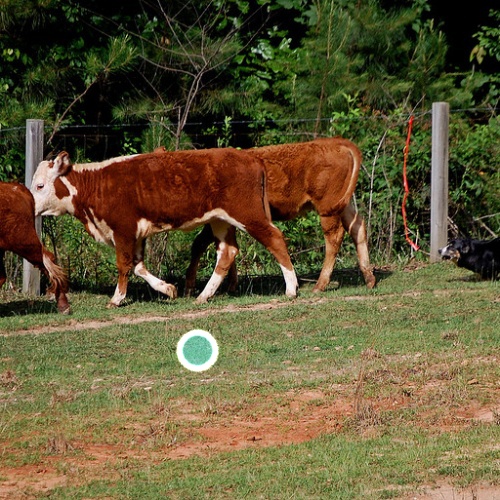}}   &
        \raisebox{-.5\height}{\includegraphics[width=\ww]{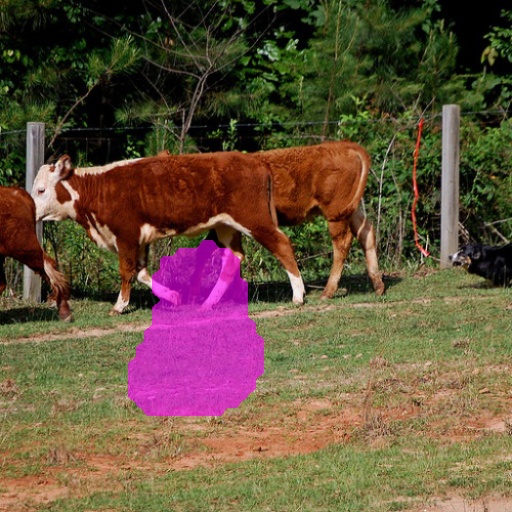}} &
        \raisebox{-.5\height}{\includegraphics[width=\ww]{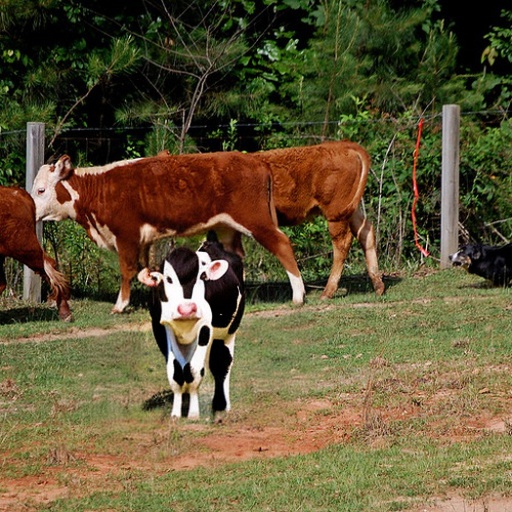}} 
        &

        \prs{\prsize}{Olives} &
        \raisebox{-.5\height}{\includegraphics[width=\ww]{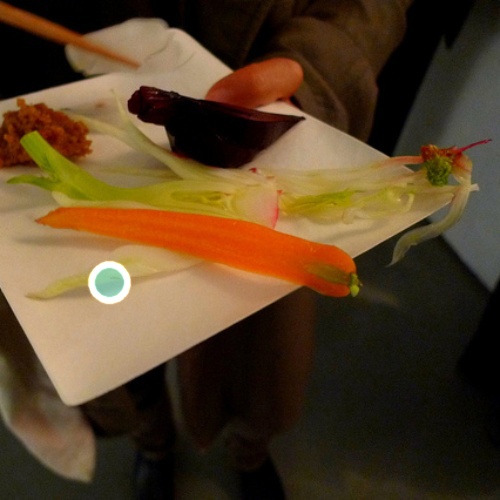}}   &
        \raisebox{-.5\height}{\includegraphics[width=\ww]{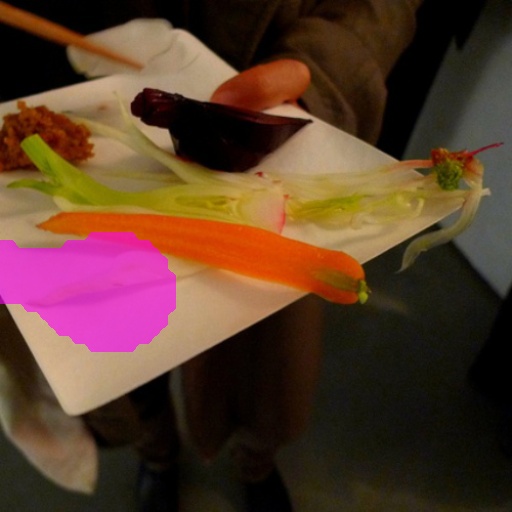}} &
        \raisebox{-.5\height}{\includegraphics[width=\ww]{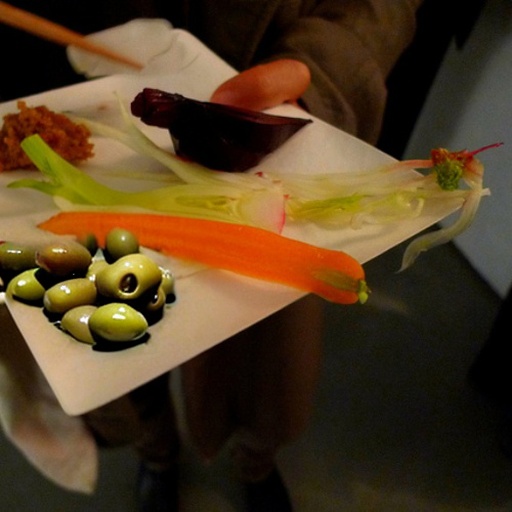}} 
        \\
        [0.95\rsm]
        
        \prs{\prsize}{A cat behind the glass window looking at the food} &
        \raisebox{-.5\height}{\includegraphics[width=\ww]{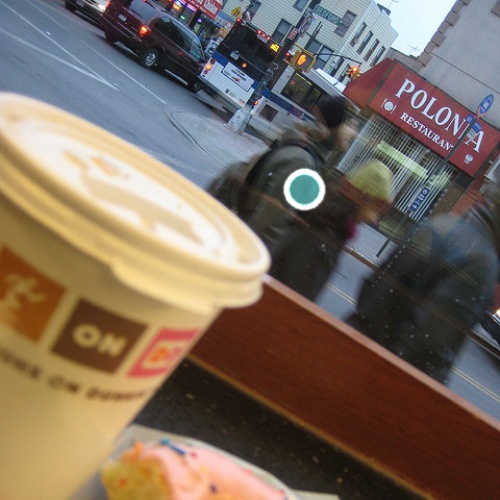}} &
        \raisebox{-.5\height}{\includegraphics[width=\ww]{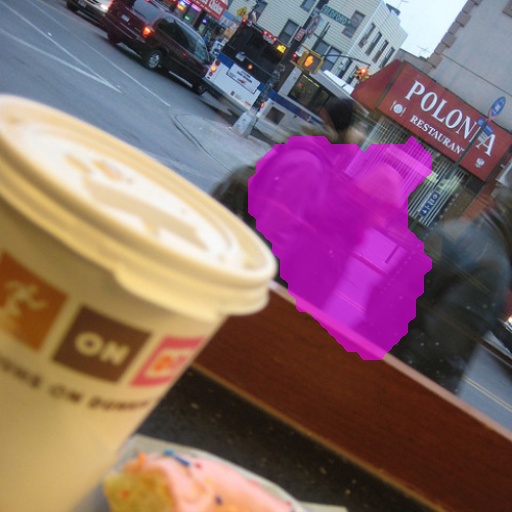}} &
        \raisebox{-.5\height}{\includegraphics[width=\ww]{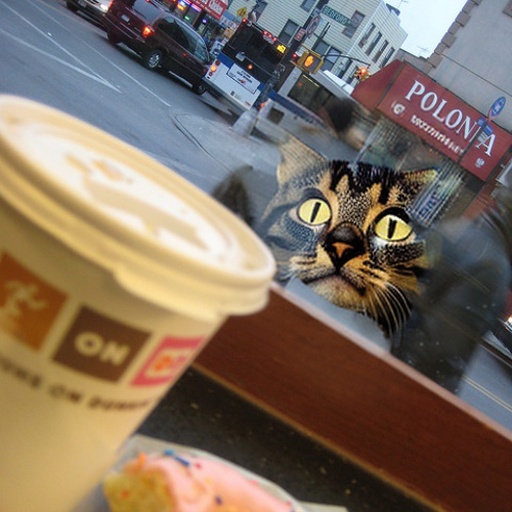}} 
        &
        
        \prs{\prsize}{A row of dolphins} &
        \raisebox{-.5\height}{\includegraphics[width=\ww]{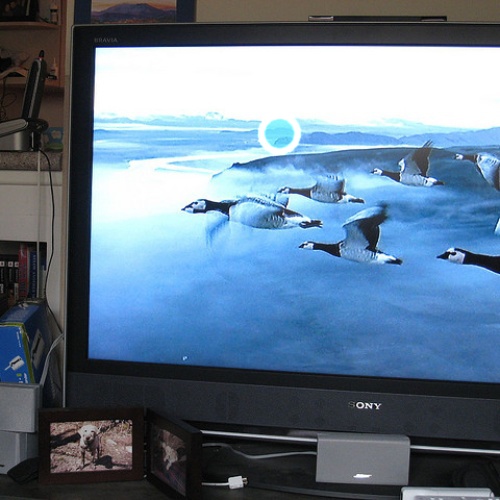}} &
        \raisebox{-.5\height}{\includegraphics[width=\ww]{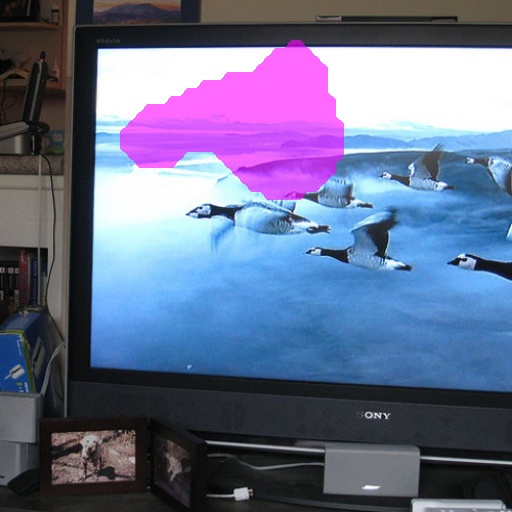}} &
        \raisebox{-.5\height}{\includegraphics[width=\ww]{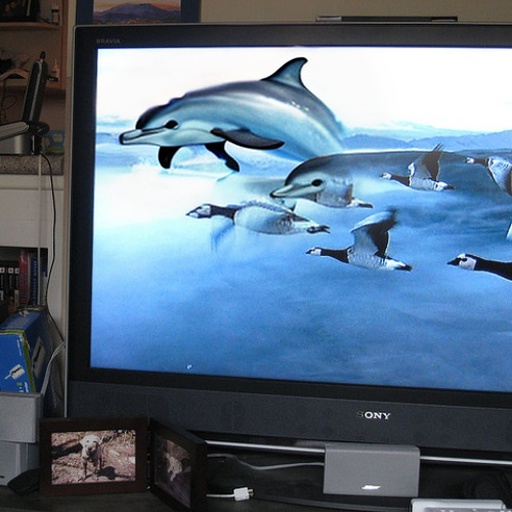}}
        \\
        [\rsm]
        
        \prs{\prsize}{A life jacket hanging} &
        \raisebox{-.5\height}{\includegraphics[width=\ww]{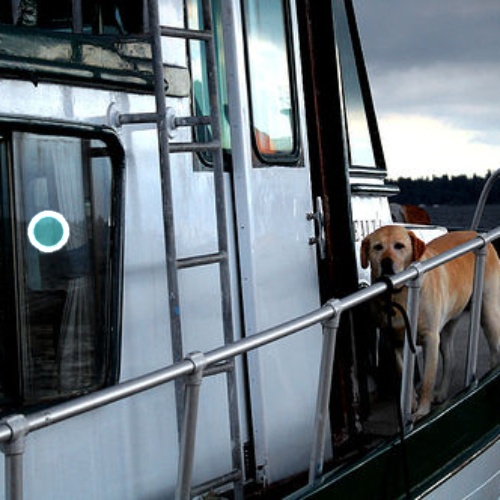}} &
        \raisebox{-.5\height}{\includegraphics[width=\ww]{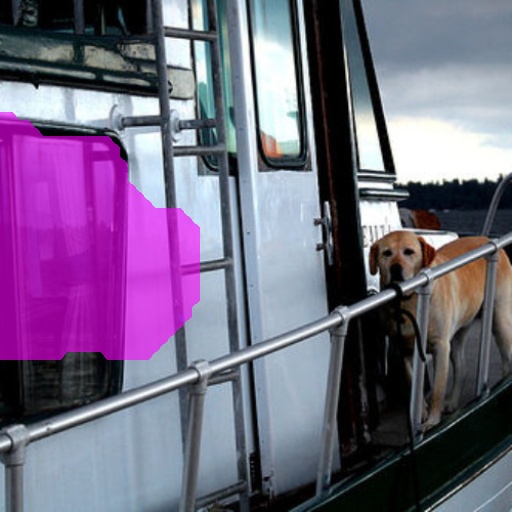}} &
        \raisebox{-.5\height}{\includegraphics[width=\ww]{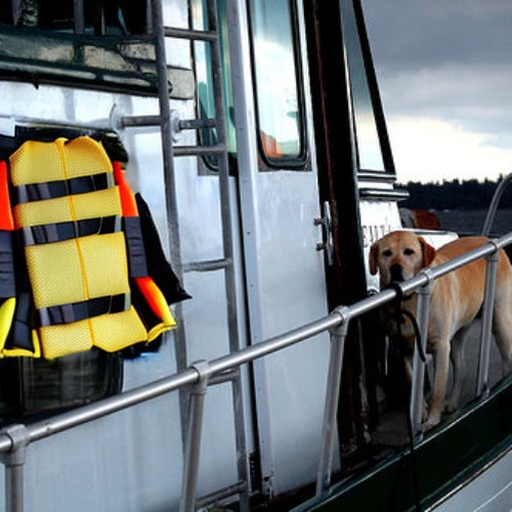}} 
        &

        \prs{\prsize}{A helmet} &
        \raisebox{-.5\height}{\includegraphics[width=\ww]{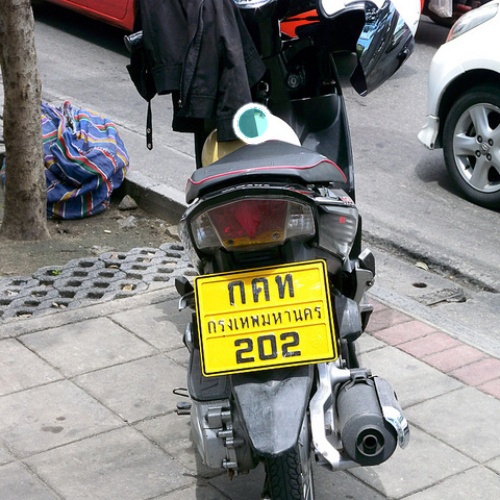}} &
        \raisebox{-.5\height}{\includegraphics[width=\ww]{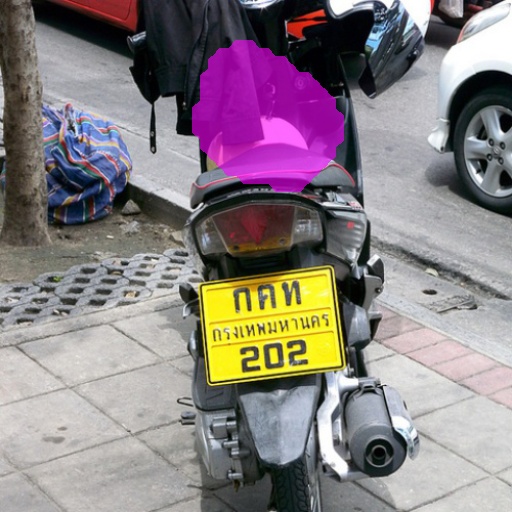}} &
        \raisebox{-.5\height}{\includegraphics[width=\ww]{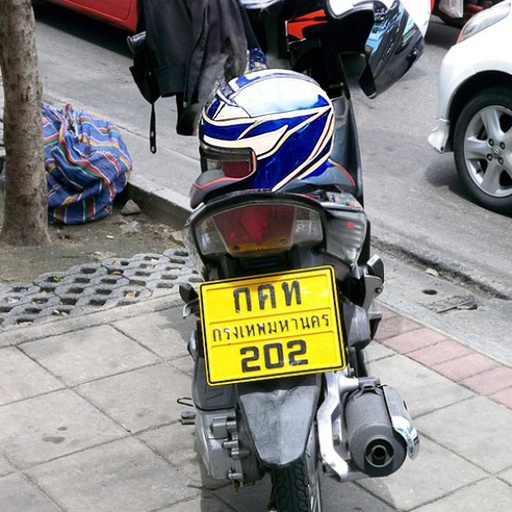}} 
        \\
        [\rsm]

        \prs{\prsize}{People swimming}&
        \raisebox{-.5\height}{\includegraphics[width=\ww]{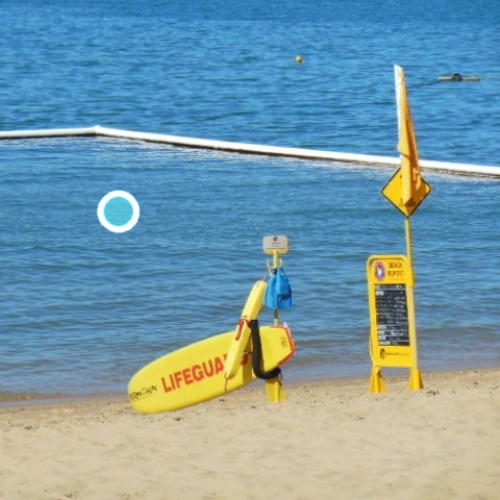}}   &
        \raisebox{-.5\height}{\includegraphics[width=\ww]{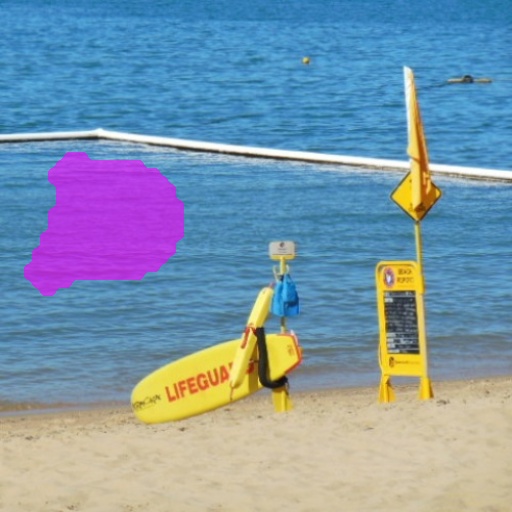}} &
        \raisebox{-.5\height}{\includegraphics[width=\ww]{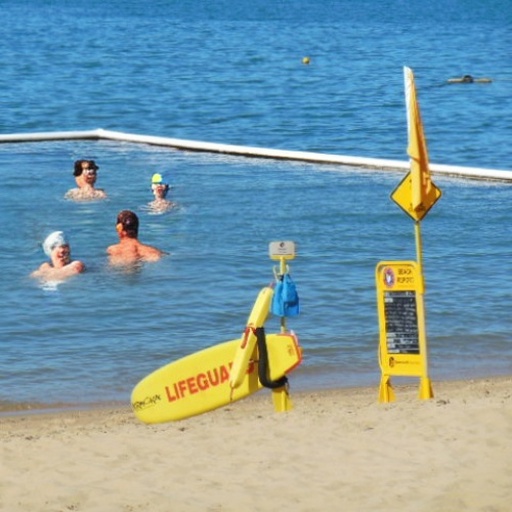}}
        &

        \prs{\prsize}{An otter swimming}&
        \raisebox{-.5\height}{\includegraphics[width=\ww]{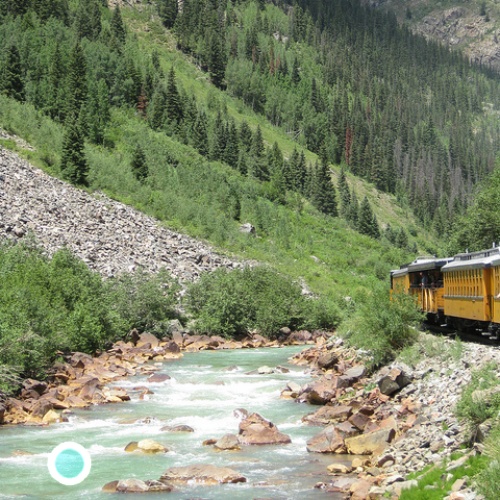}}   &
        \raisebox{-.5\height}{\includegraphics[width=\ww]{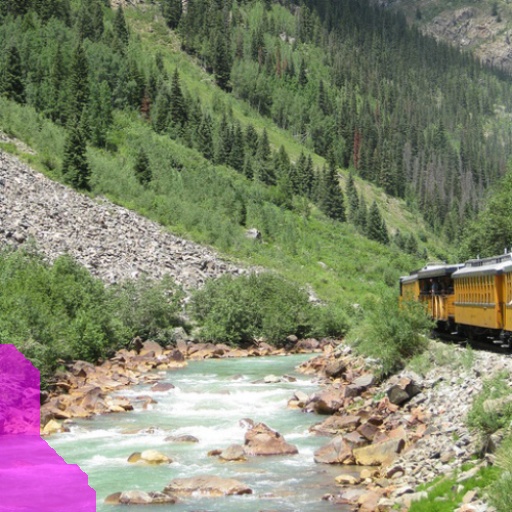}} &
        \raisebox{-.5\height}{\includegraphics[width=\ww]{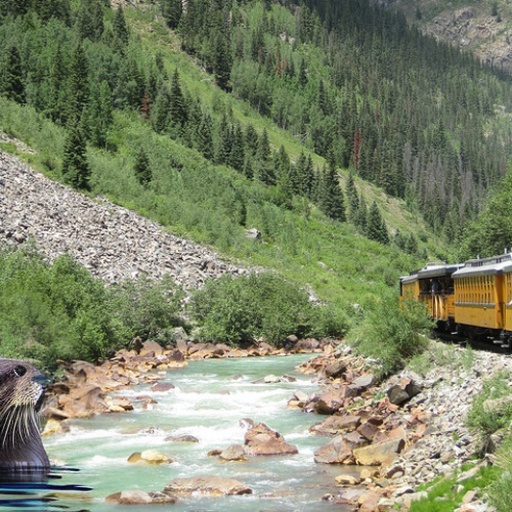}}       
        \\
        [\rsm]

        \prs{\prsize}{Sliced apples} &
        \raisebox{-.5\height}{\includegraphics[width=\ww]{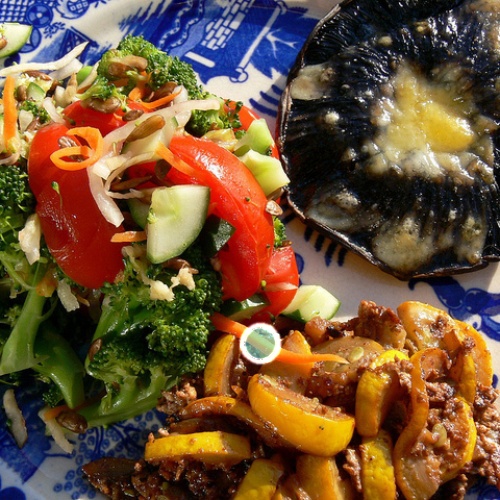}} &
        \raisebox{-.5\height}{\includegraphics[width=\ww]{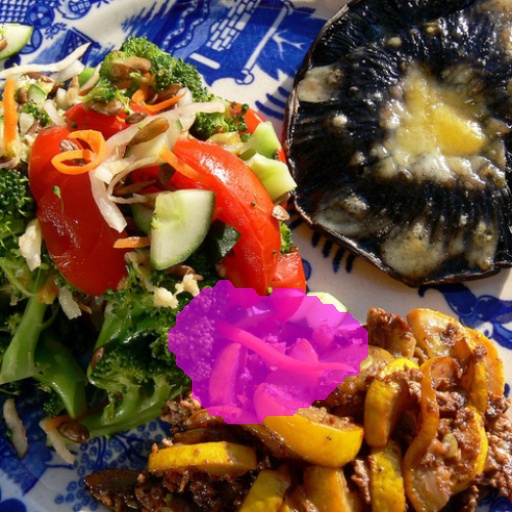}} &
        \raisebox{-.5\height}{\includegraphics[width=\ww]{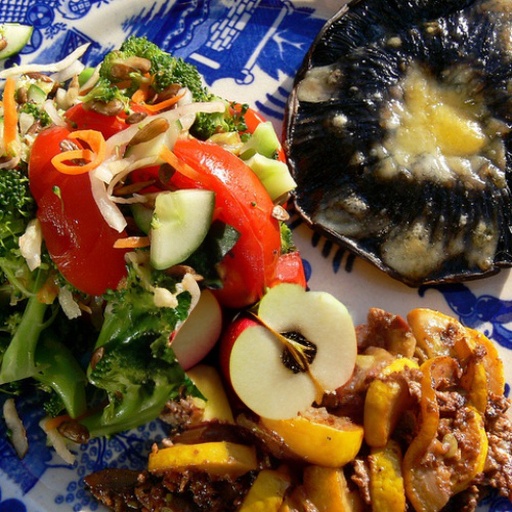}} 
        &

        \prs{\prsize}{A pizza sign} &
        \raisebox{-.5\height}{\includegraphics[width=\ww]{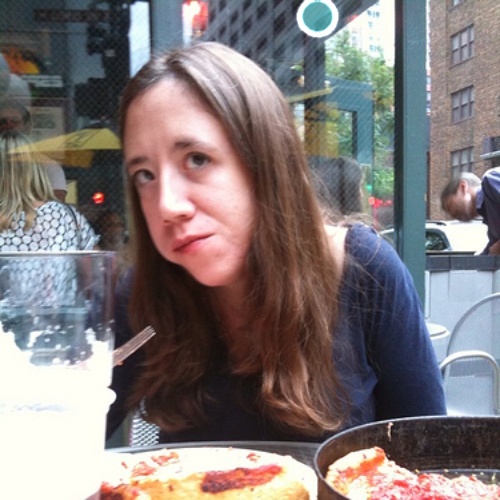}} &
        \raisebox{-.5\height}{\includegraphics[width=\ww]{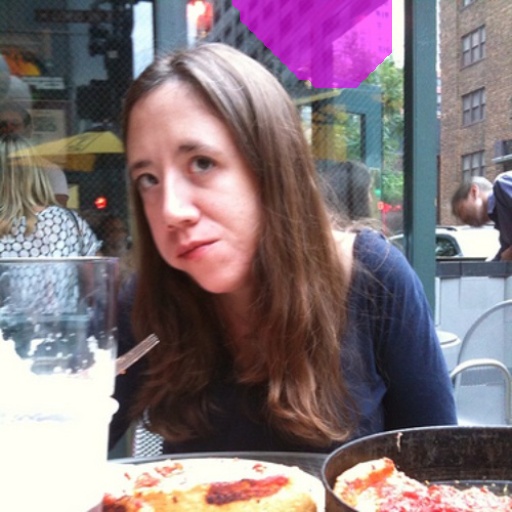}} &
        \raisebox{-.5\height}{\includegraphics[width=\ww]{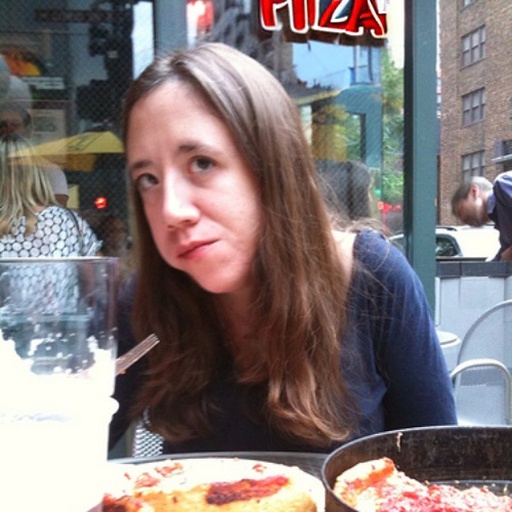}} 
        \\
        [\rsm]

        \prs{\prsize}{A soda bottle} &
        \raisebox{-.5\height}{\includegraphics[width=\ww]{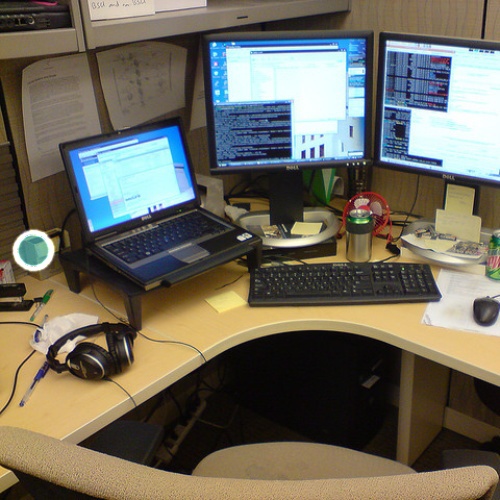}} &
        \raisebox{-.5\height}{\includegraphics[width=\ww]{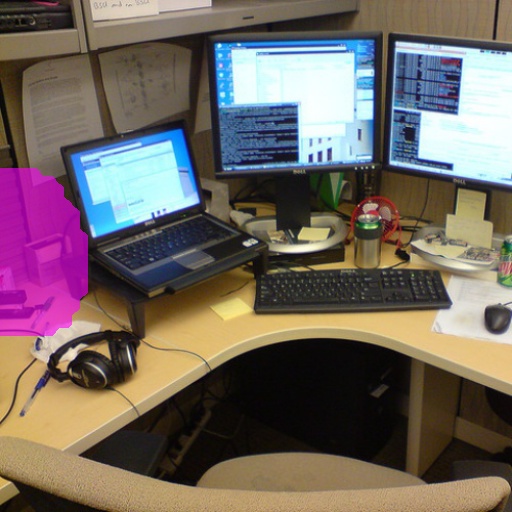}} &
        \raisebox{-.5\height}{\includegraphics[width=\ww]{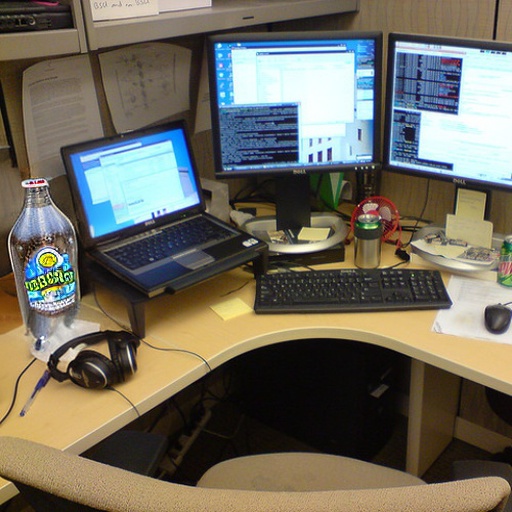}} 
        &

        \prs{\prsize}{A whale on the mural painting} &
        \raisebox{-.5\height}{\includegraphics[width=\ww]{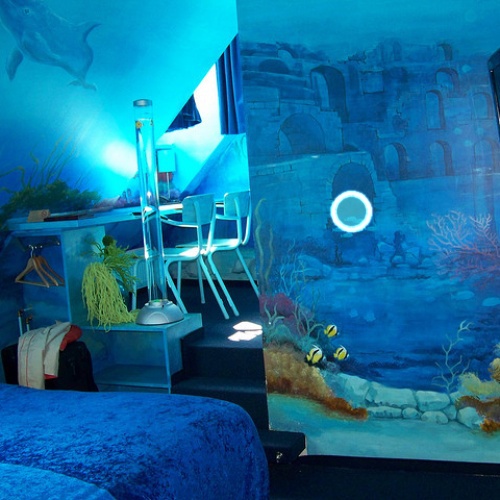}} &
        \raisebox{-.5\height}{\includegraphics[width=\ww]{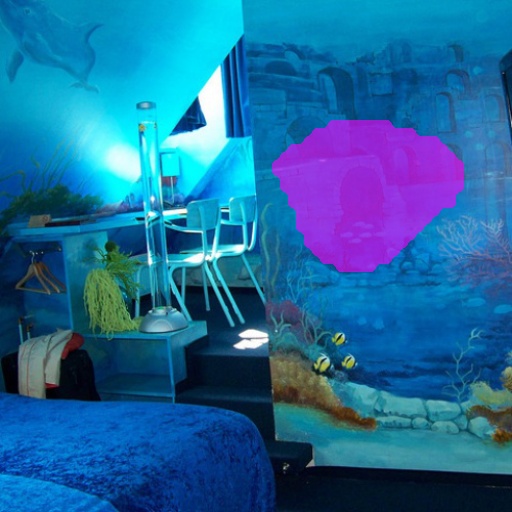}} &
        \raisebox{-.5\height}{\includegraphics[width=\ww]{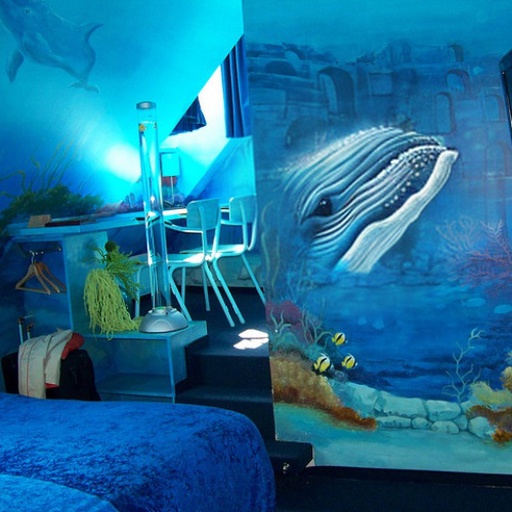}}

    \end{tabular}
    
     \caption{\textbf{Additional examples of generated masks.} In each image triplet, the leftmost image is the input with clicked point, accompanied by the given prompt on its left. The generated mask is demonstrated by a purple overlay on the input image (center image) and the rightmost image is the output of \ctm.} 
    \label{fig:generated_masks_appendix}
\end{figure*}

\subsection{Additional Results}
\label{additional_results}

\begin{figure}[h]
    \setlength{\ww}{0.99\columnwidth}

    \includegraphics[width=\ww]{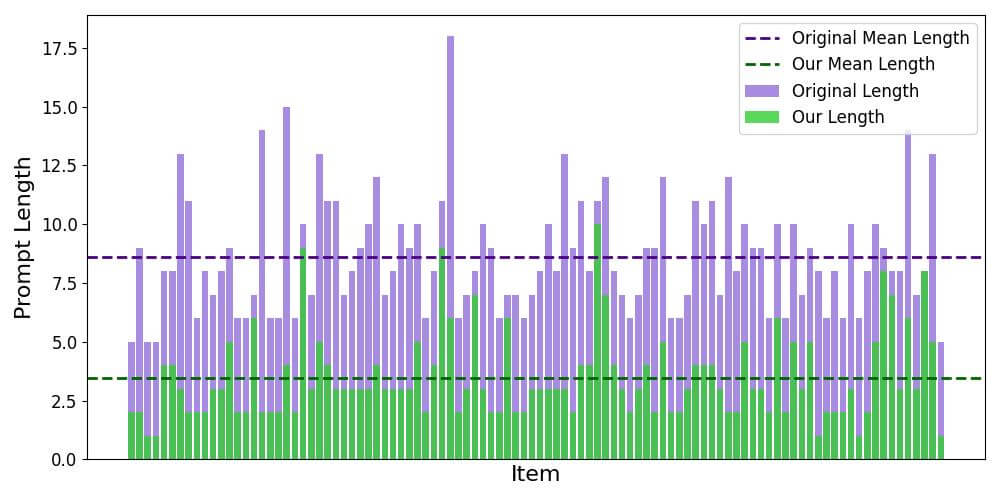}
        
    \caption{\textbf{Prompts word count.} An additional advantage of our method is that users can provide shorter prompts, which require less effort on their part. The bar plot above shows prompts word lengths of \emu\ benchmark in comparison to \ctm. Each purple high bar represents the number of words in an item in \emu\ benchmark, and the overlaid green low bar represents the corresponding prompt given to \ctm\ after removing the word that describes addition (\eg\ ``Add", ``Insert", etc.) and the words describing the desired edit location (\eg\ ``on the table next to the fridge"), as explained in the fixed routine in \Cref{sec:results}. The 100 bars correspond to the 100 samples we compared with \emu\ and \mb, as described also in \Cref{sec:results}. The purple higher horizontal line is the mean prompt length in \emu\ benchmark's samples, while the green lower one is the mean length of \ctm's shorter prompts.}
    \label{fig:word_count}
\end{figure}

\begin{figure*}[hbt!]
    \centering
    \setlength{\tabcolsep}{0.5pt}
    \renewcommand{\arraystretch}{0.5}
    \setlength{\ww}{0.172\linewidth}
    \renewcommand{\prsize}{\defbigprsize}
    \renewcommand{\methsize}{normalsize}
    \renewcommand{\pw}{60pt}
    \renewcommand{\rsm}{1.76cm}
    \renewcommand{\rsb}{3px}

    \begin{tabular}{cc cc cc}  
        \sizedtext{\methsize}Prompt &
        \sizedtext{\methsize}Input &
        \sizedtext{\methsize}\emu &
        \sizedtext{\methsize}\mb &
        \sizedtext{\methsize}\ipp &
        \sizedtext{\methsize}\ctmb 
        \\
        [\rsb]
        
        \prprs[\pw]{\prsize}{Add a baseball glove beside the bat}{A baseball glove}
        {\includegraphics[valign=c, width=\ww]{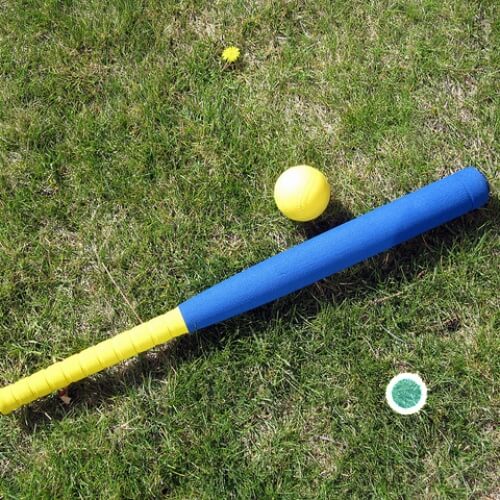}} &
        {\includegraphics[valign=c, width=\ww]{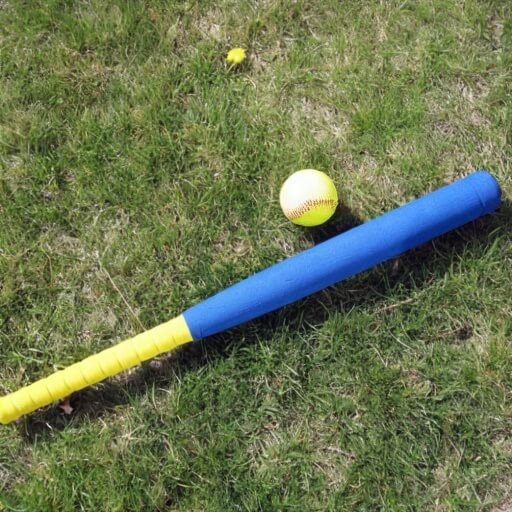}} &
        {\includegraphics[valign=c, width=\ww]{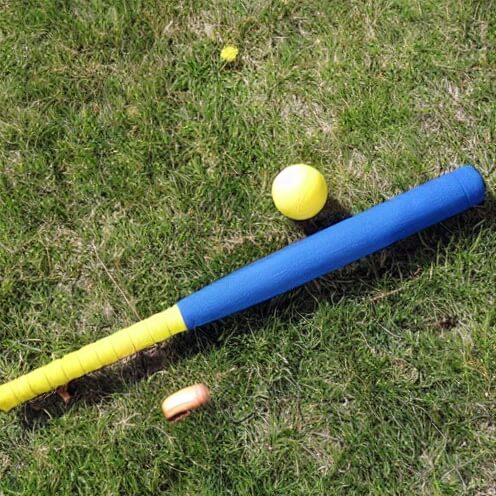}} &
        {\includegraphics[valign=c, width=\ww]{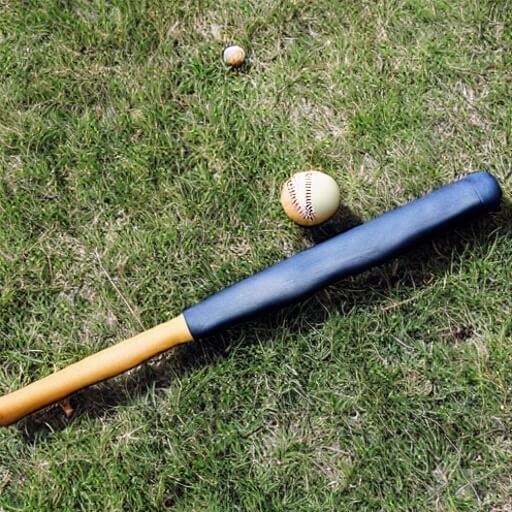}} &
        {\includegraphics[valign=c, width=\ww]{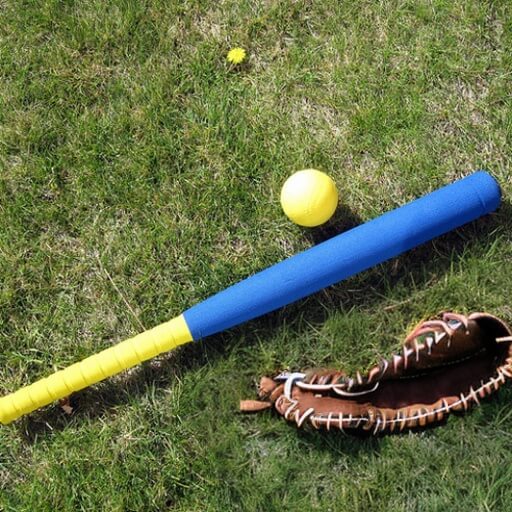}} 
        \\
        [\rsm]
        
        \prprs[\pw]{\prsize}{Add a butterfly on top of the beans}{A butterfly}
        {\includegraphics[valign=c, width=\ww] {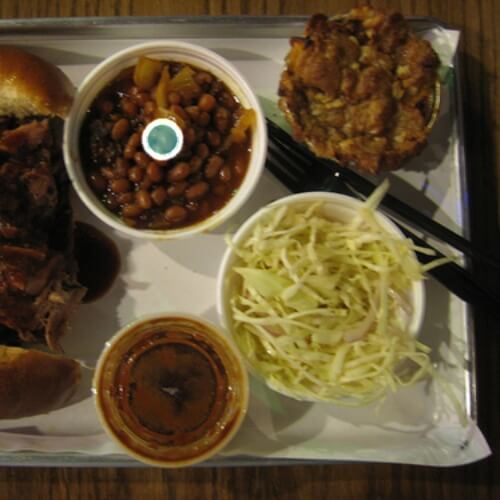}} &
        {\includegraphics[valign=c, width=\ww]{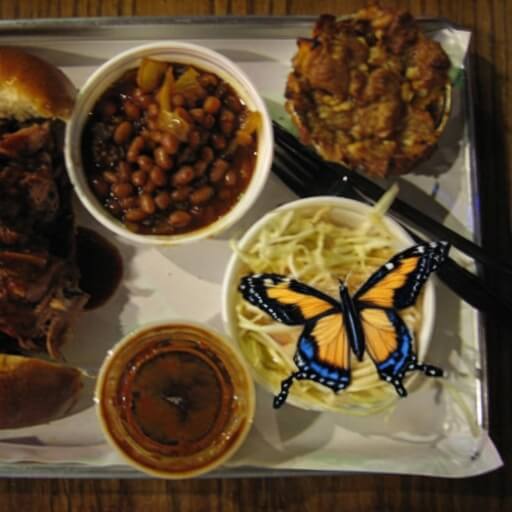}} &
        {\includegraphics[valign=c, width=\ww]{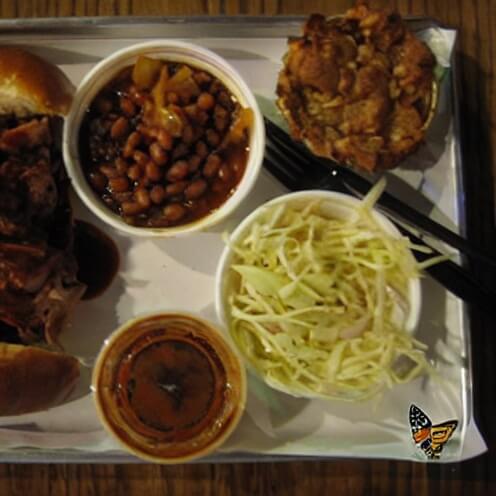}} &
        {\includegraphics[valign=c, width=\ww]{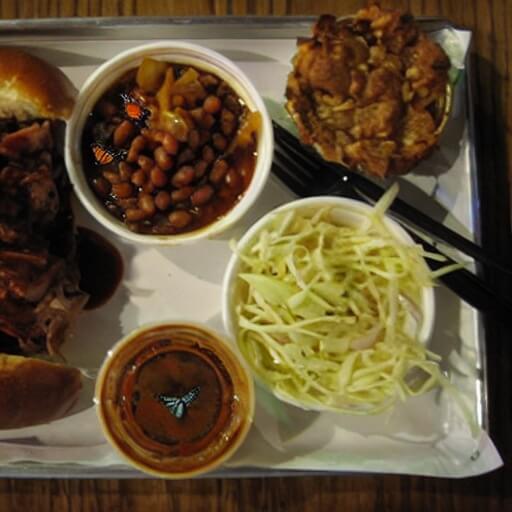}} &
        {\includegraphics[valign=c, width=\ww]{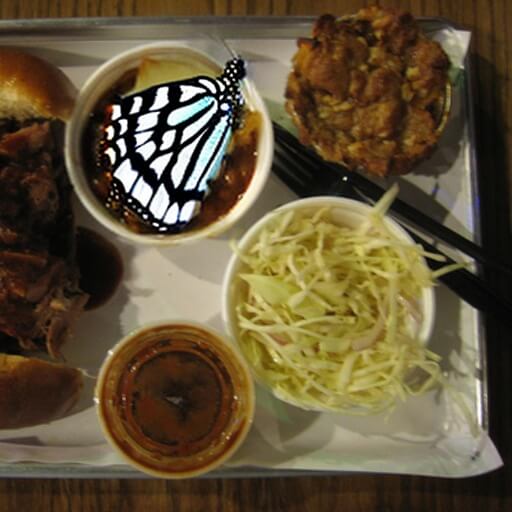}} 
        \\
        [\rsm]
   
        \prprs[\pw]{\prsize}{Add a monkey sitting on the fire hydrant}{Monkey sitting on the fire hydrant} 
        {\includegraphics[valign=c, width=\ww]{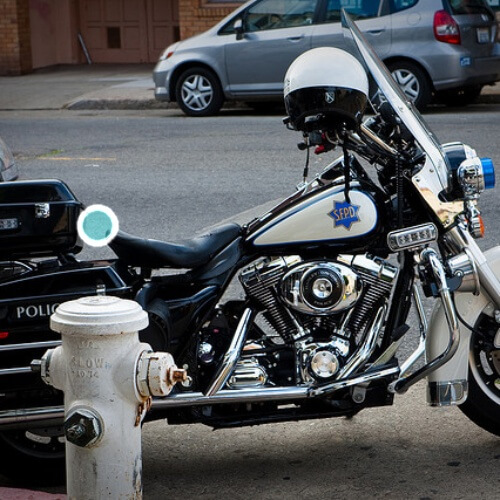}} &
        {\includegraphics[valign=c, width=\ww]{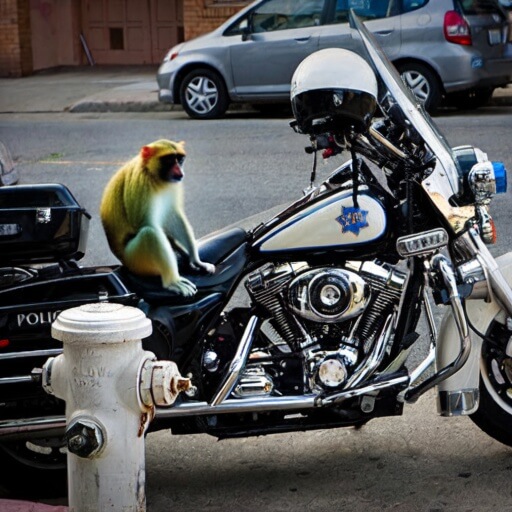}} &
        {\includegraphics[valign=c, width=\ww]{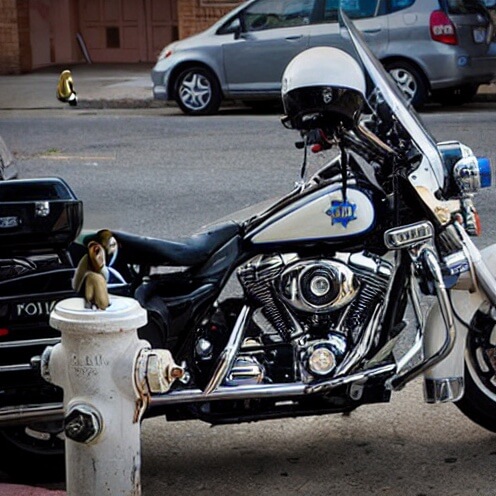}} &
        {\includegraphics[valign=c, width=\ww]{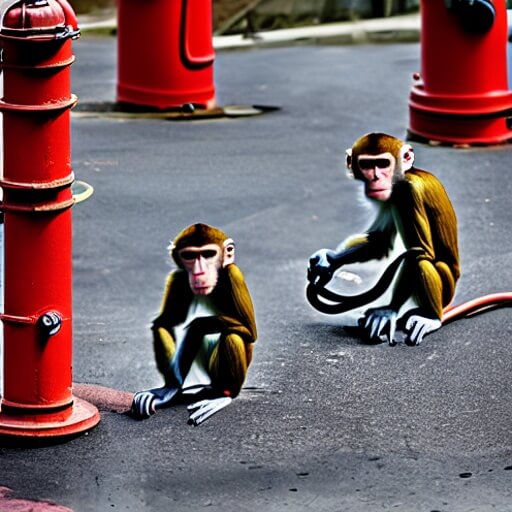}} &
        {\includegraphics[valign=c, width=\ww]{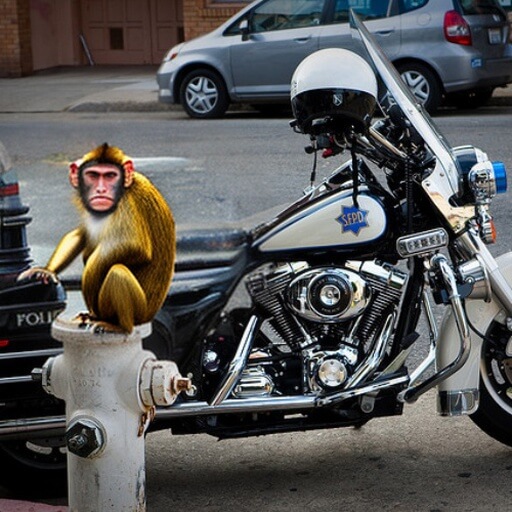}} 
        \\
        [\rsm]

        \prprs[\pw]{\prsize}{Add a splash of paint over the fridge}{A splash of paint} 
        {\includegraphics[valign=c, width=\ww]{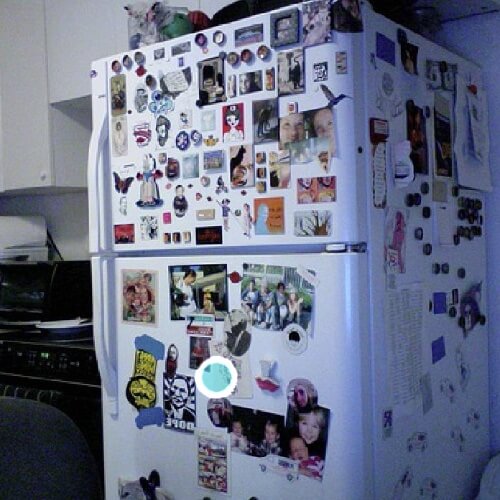}} &
        {\includegraphics[valign=c, width=\ww]{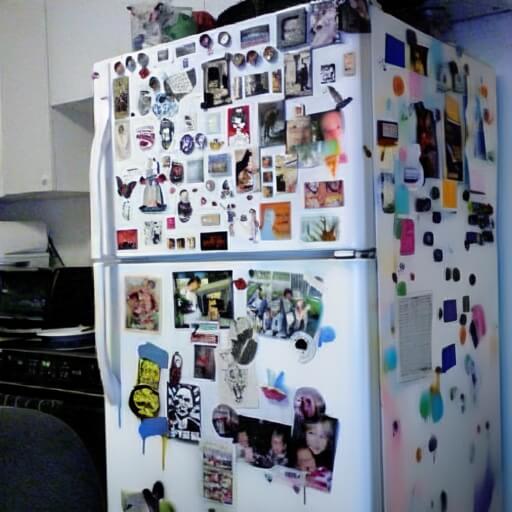}} &
        {\includegraphics[valign=c, width=\ww]{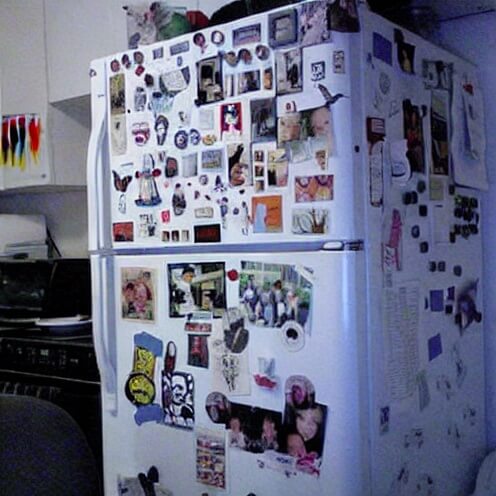}} &
        {\includegraphics[valign=c, width=\ww]{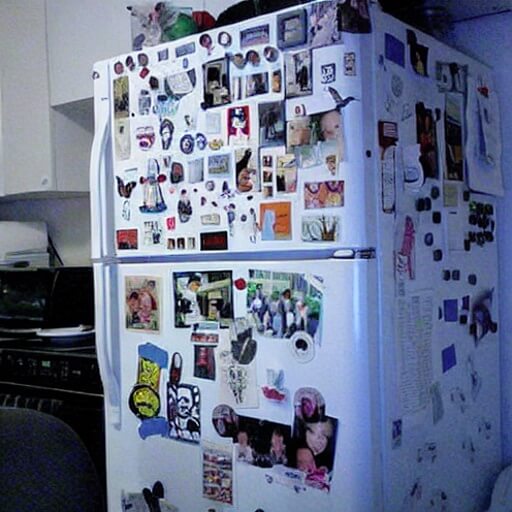}} &
        {\includegraphics[valign=c, width=\ww]{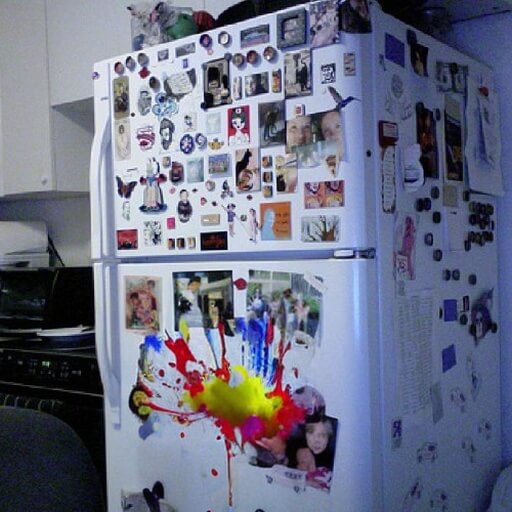}} 
        \\
        [\rsm]

        \prprs[\pw]{\prsize}{Add a small white bowl below the pile of pepper slice}{A small white bowl} 
        {\includegraphics[valign=c, width=\ww]{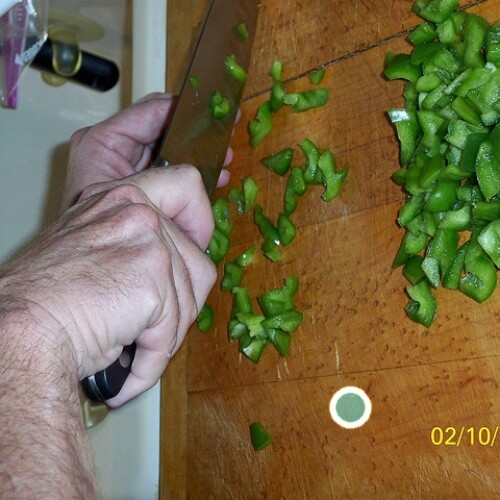}} &
        {\includegraphics[valign=c, width=\ww]{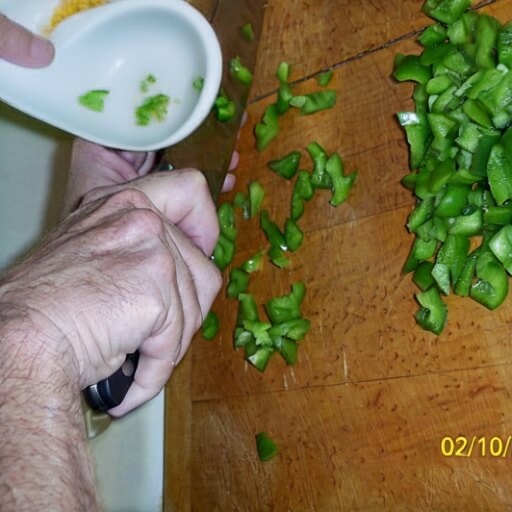}} &
        {\includegraphics[valign=c, width=\ww]{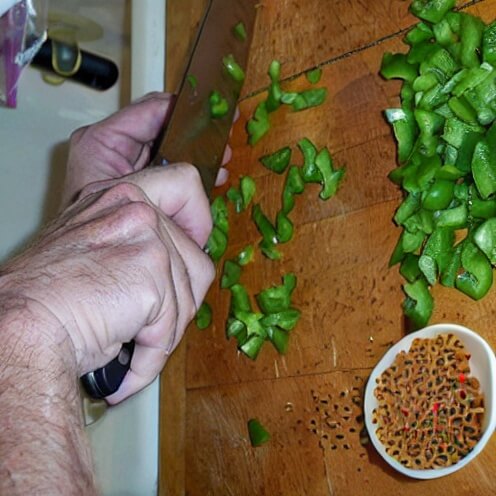}} &
        {\includegraphics[valign=c, width=\ww]{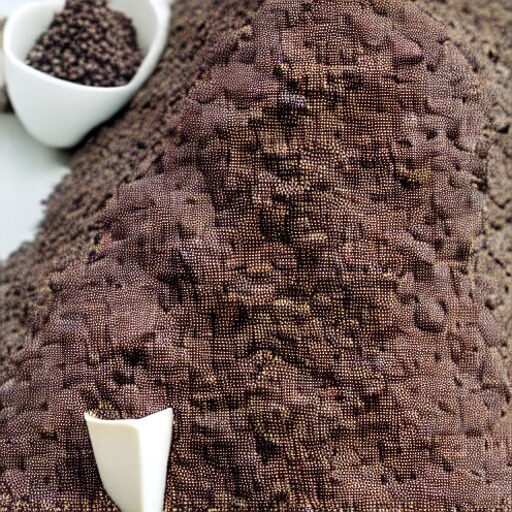}} &
        {\includegraphics[valign=c, width=\ww]{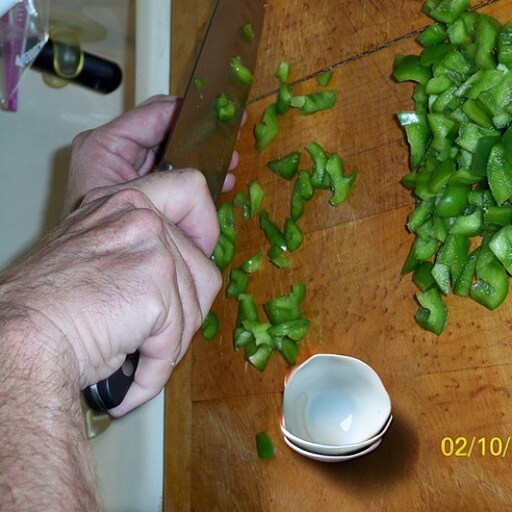}} 
        \\
        [\rsm]
     
        \prprs[\pw]{\prsize}{Add a sand castle to the image}{A sand castle}
        {\includegraphics[valign=c, width=\ww]{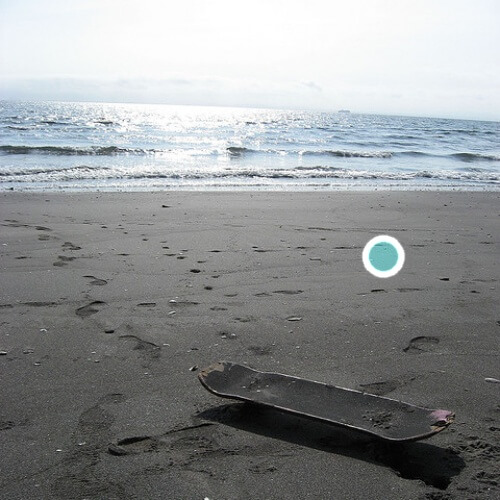}} &
        {\includegraphics[valign=c, width=\ww]{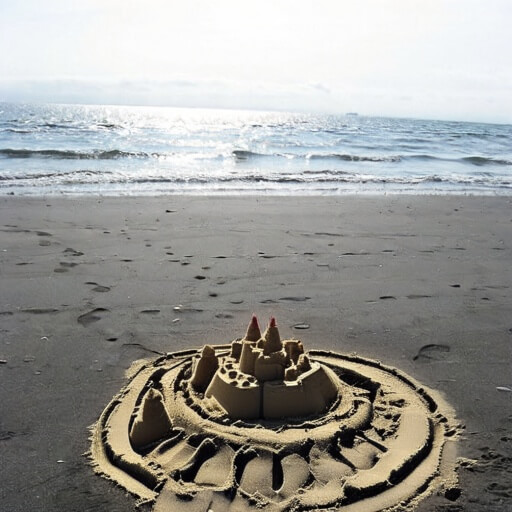}} &
        {\includegraphics[valign=c, width=\ww]{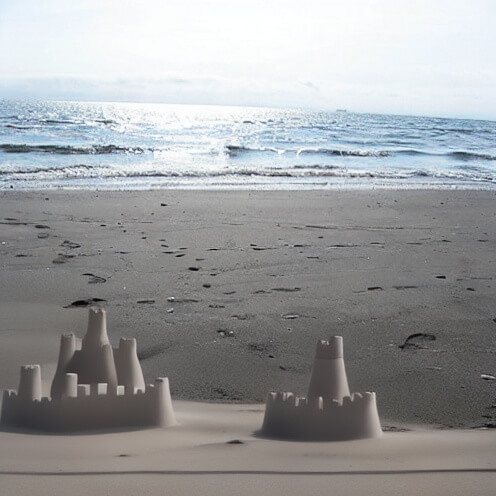}} &
        {\includegraphics[valign=c, width=\ww]{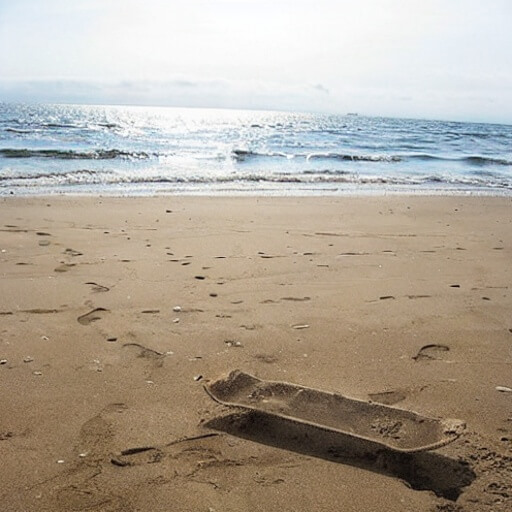}} &
        {\includegraphics[valign=c, width=\ww]{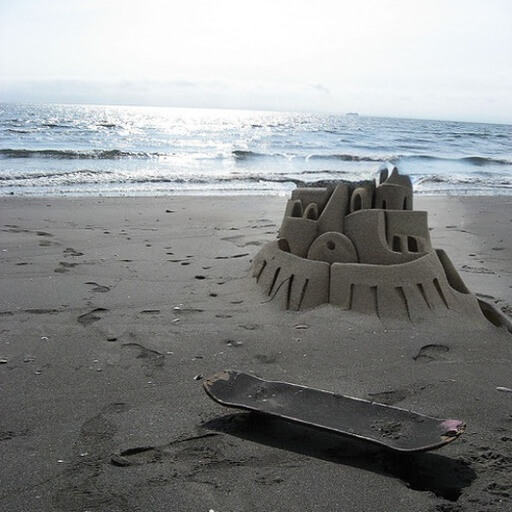}} 
        \\

    \end{tabular}

    \caption{\textbf{Additional comparisons with SoTA methods.} Additional comparisons of \emu\ \cite{sheynin2023emu}, \mb\ \cite{Zhang2023MagicBrush} and \ipp\ \cite{brooks2022instructpix2pix} with our model \ctmb. The upper prompts were given to baselines, and the lower ones to \ctm. Inputs contain the clicked point from \ctm.}
    
    \label{fig:comparison_2}
\end{figure*}

\begin{figure*}[h]
    \centering
    \setlength{\tabcolsep}{0.5pt}
    \renewcommand{\arraystretch}{0.5}
    \setlength{\ww}{0.172\linewidth}
    \renewcommand{\prsize}{\defbigprsize}
    \renewcommand{\methsize}{normalsize}
    \renewcommand{\pw}{60pt}
    \renewcommand{\rsm}{1.76cm}
    \renewcommand{\rsb}{3px}

    \begin{tabular}{cc cc cc}  
        \sizedtext{\methsize}Prompt &
        \sizedtext{\methsize}Input &
        \sizedtext{\methsize}\emu &
        \sizedtext{\methsize}\mb &
        \sizedtext{\methsize}\ipp &
        \sizedtext{\methsize}\ctmb 
        \\
        [\rsb]

        \prprs[\pw]{\prsize}{Add a small pond in the front}{a small pond}
        {\includegraphics[valign=c, width=\ww]{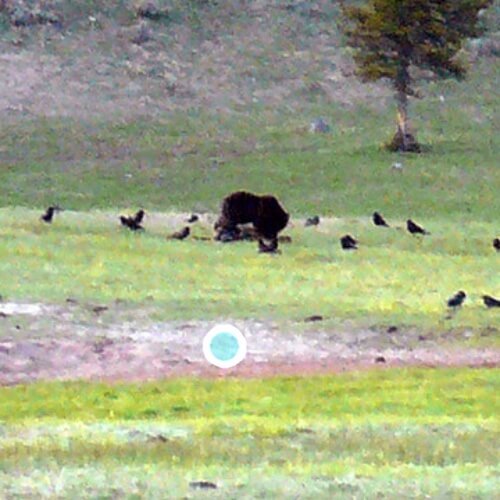}} &
        {\includegraphics[valign=c, width=\ww]{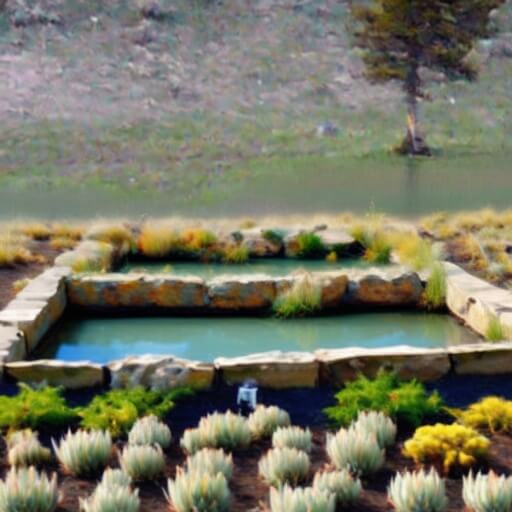}} &
        {\includegraphics[valign=c, width=\ww]{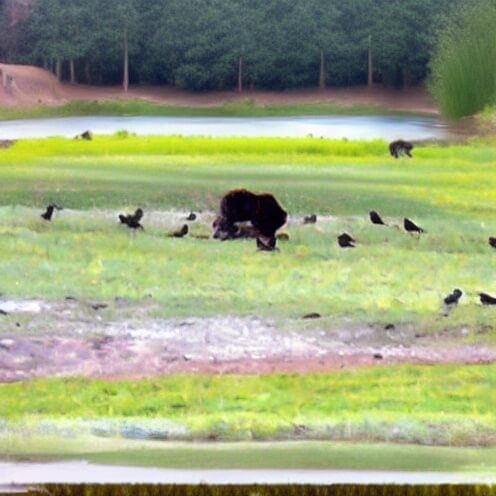}} &
        {\includegraphics[valign=c, width=\ww]{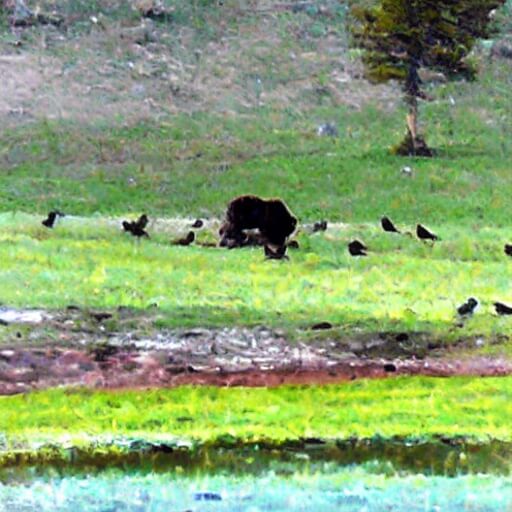}} &
        {\includegraphics[valign=c, width=\ww]{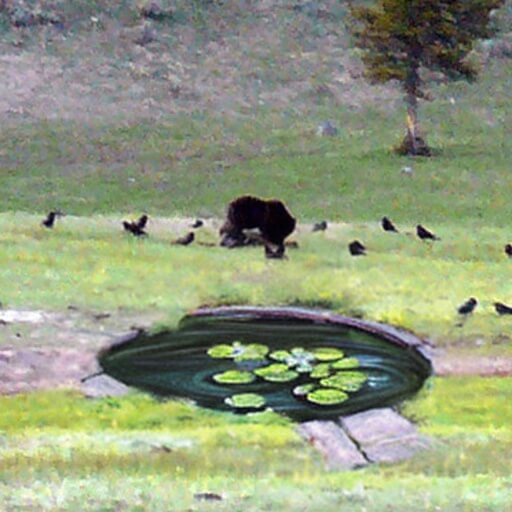}} 
        \\
        [\rsm]
        
        \prprs[\pw]{\prsize}{Add a smiley face on the wall between the cop and the stop sign}{A smiley face}
        {\includegraphics[valign=c, width=\ww]{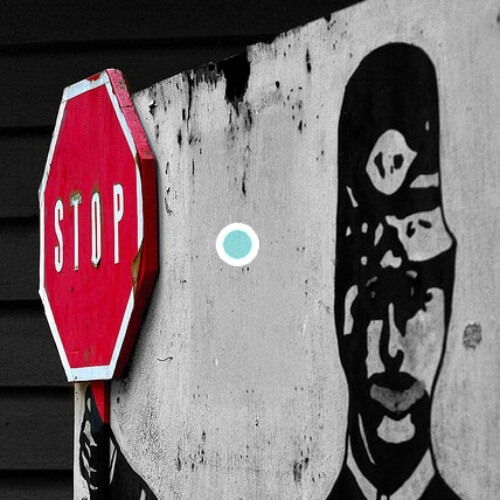}} &
        {\includegraphics[valign=c, width=\ww]{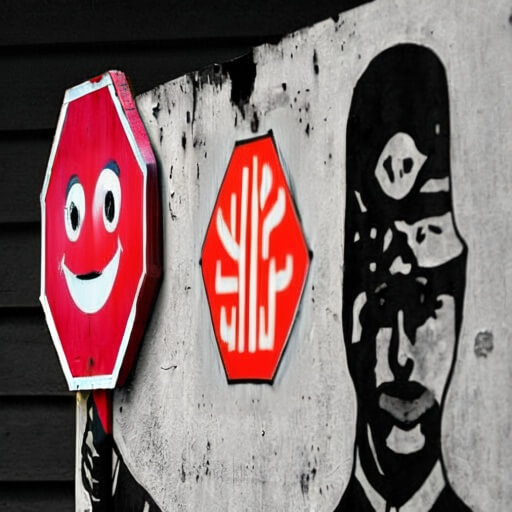}} &
        {\includegraphics[valign=c, width=\ww]{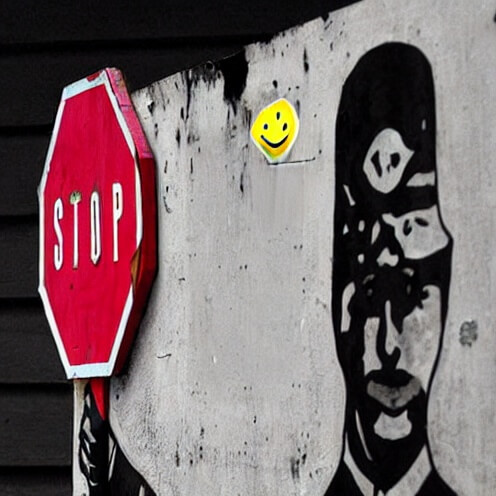}} &
        {\includegraphics[valign=c, width=\ww]{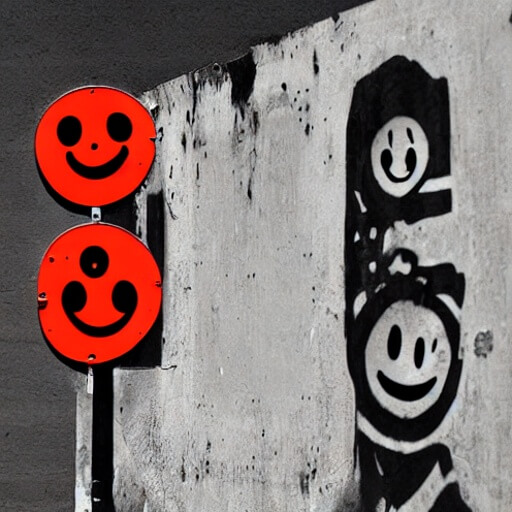}} &
        {\includegraphics[valign=c, width=\ww]{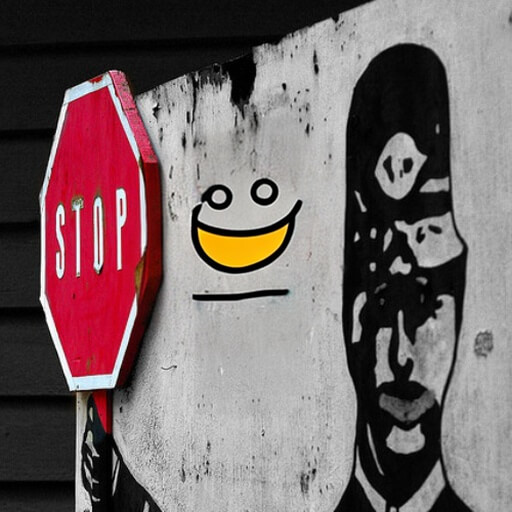}} 
        \\
        [\rsm]

        \prprs[\pw]{\prsize}{Add a tennis ball coming toward the man's racquet}{A tennis ball}
        {\includegraphics[valign=c, width=\ww]{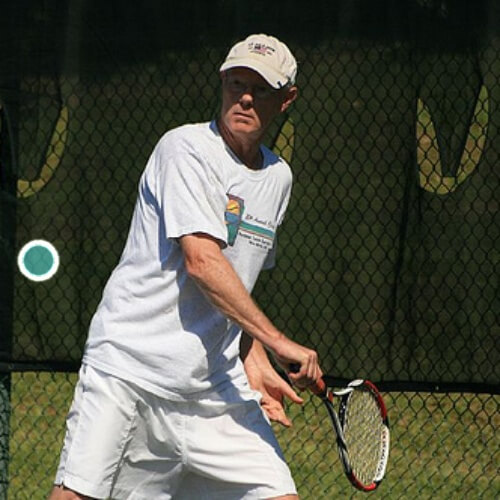}} &
        {\includegraphics[valign=c, width=\ww]{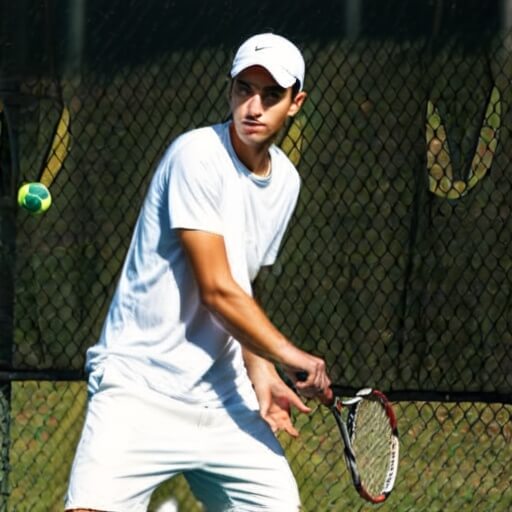}} &
        {\includegraphics[valign=c, width=\ww]{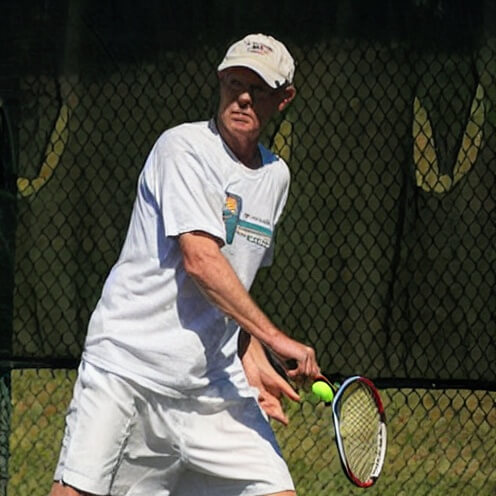}} &
        {\includegraphics[valign=c, width=\ww]{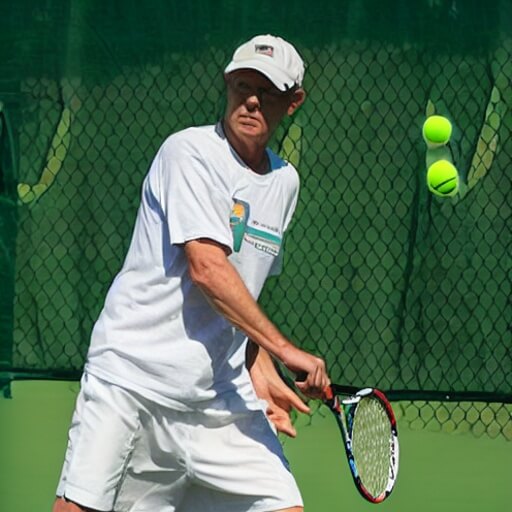}} &
        {\includegraphics[valign=c, width=\ww]{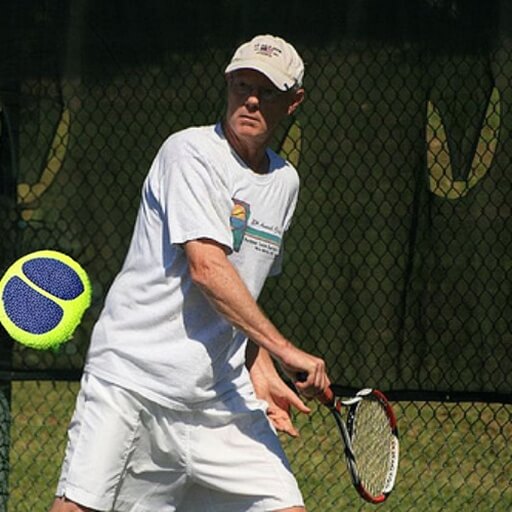}} 
        \\
        [\rsm]
        
        \prprs[\pw]{\prsize}{Add a small teddy bear in front of the book}{A small teddy bear}
        {\includegraphics[valign=c, width=\ww]{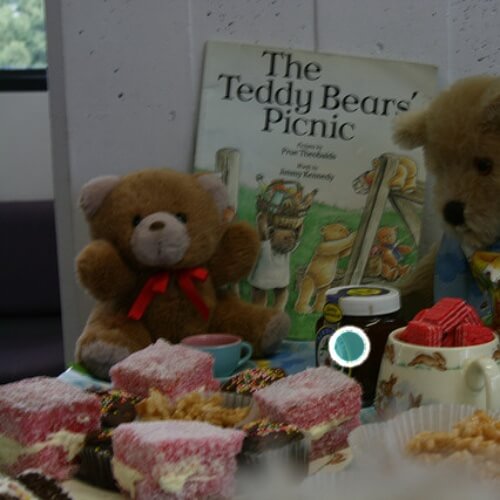}} &
        {\includegraphics[valign=c, width=\ww]{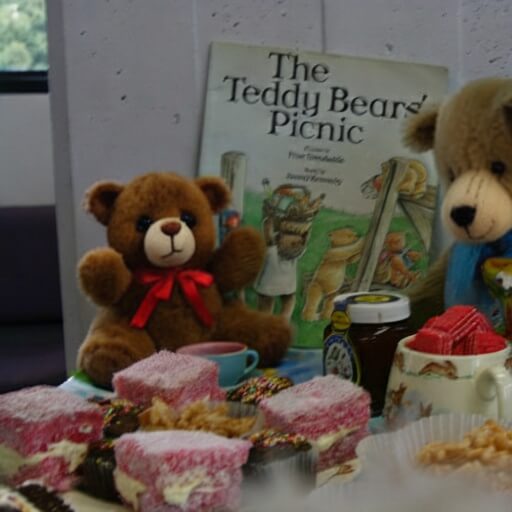}} &
        {\includegraphics[valign=c, width=\ww]{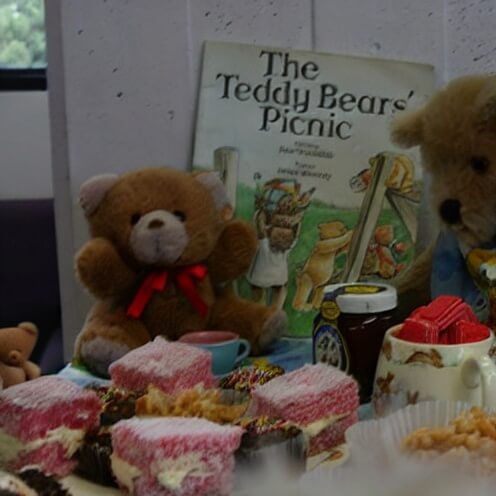}} &
        {\includegraphics[valign=c, width=\ww]{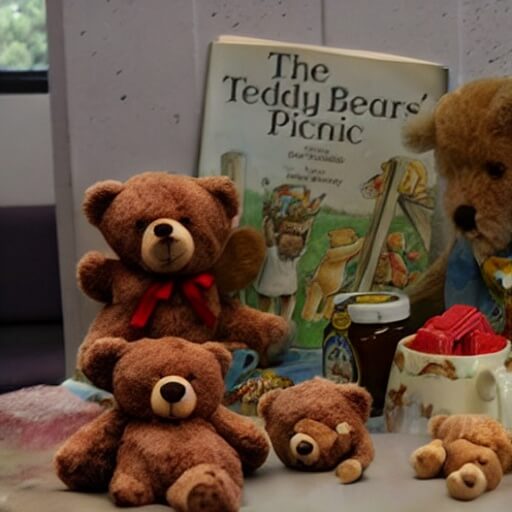}} &
        {\includegraphics[valign=c, width=\ww]{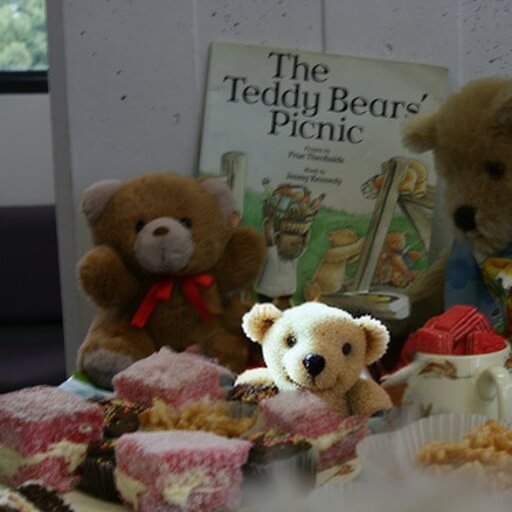}} 
        \\
        [\rsm]
    
        \prprs[\pw]{\prsize}{Add toys to the floor}{Toys}
        {\includegraphics[valign=c, width=\ww]{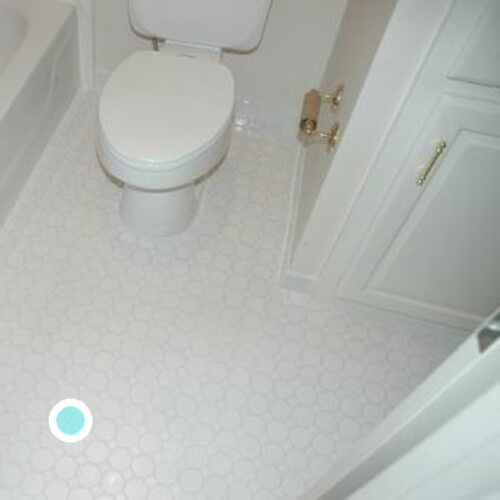}} &
        {\includegraphics[valign=c, width=\ww]{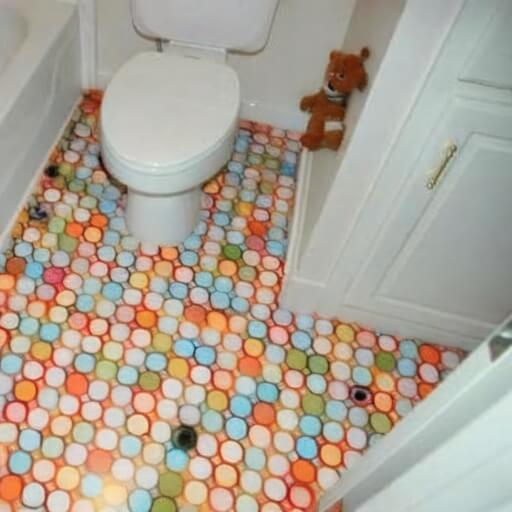}} &
        {\includegraphics[valign=c, width=\ww]{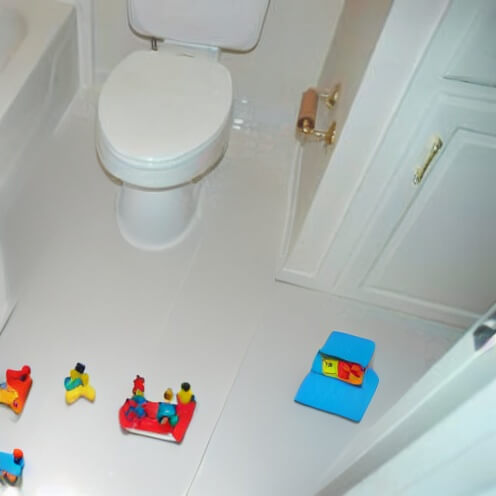}} &
        {\includegraphics[valign=c, width=\ww]{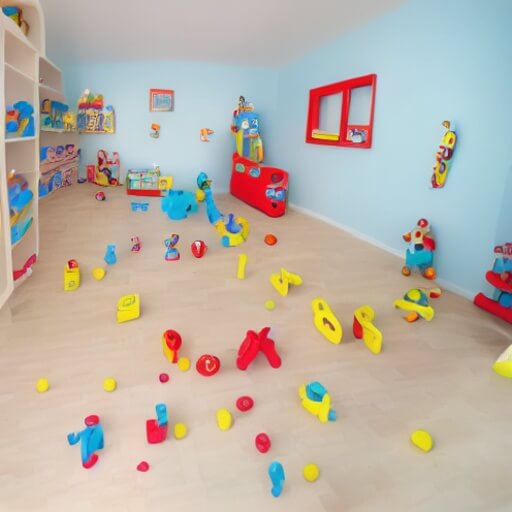}} &
        {\includegraphics[valign=c, width=\ww]{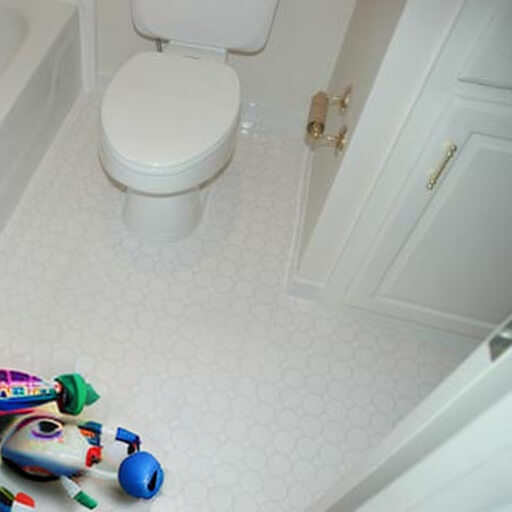}} 
        \\
        [\rsm]
    
        \prprs[\pw]{\prsize}{Add an additional Christmas tree behind the container}{A Christmas tree}
        {\includegraphics[valign=c, width=\ww]{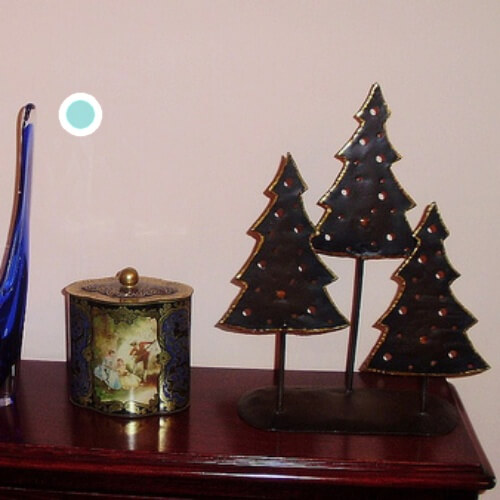}} &
        {\includegraphics[valign=c, width=\ww]{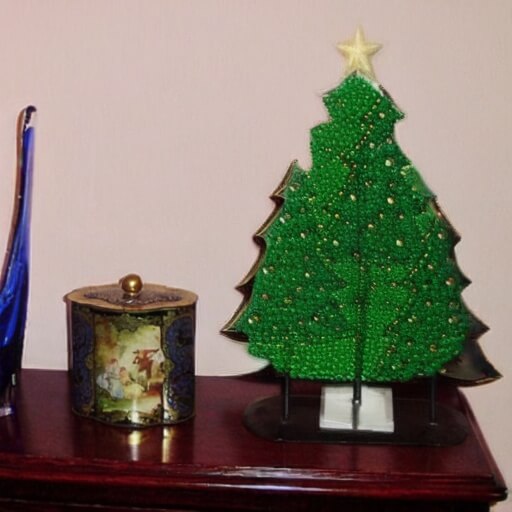}} &
        {\includegraphics[valign=c, width=\ww]{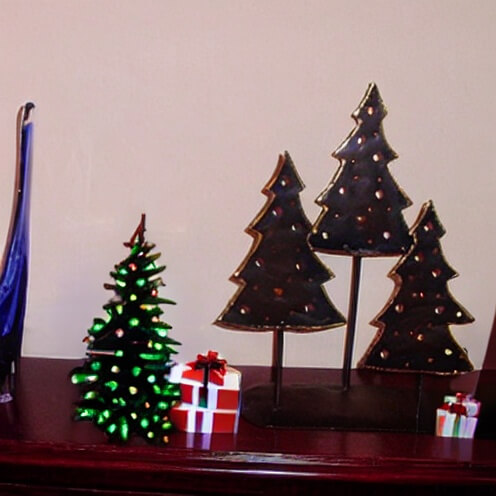}} &
        {\includegraphics[valign=c, width=\ww]{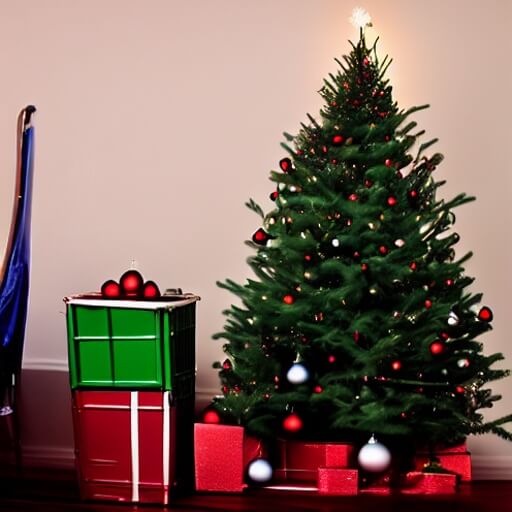}} &
        {\includegraphics[valign=c, width=\ww]{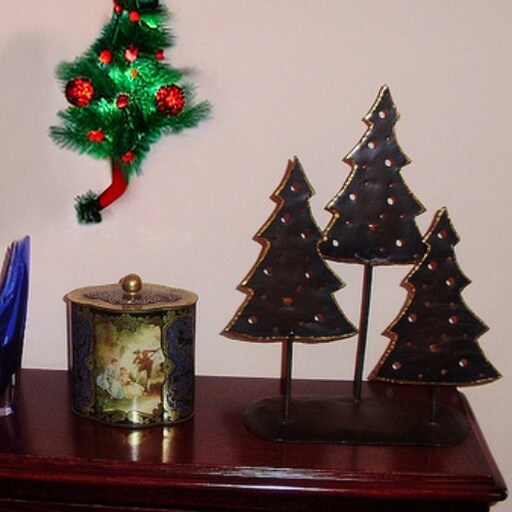}} 

    \end{tabular}
    
    \caption{\textbf{Additional comparisons with SoTA methods.}}
    \label{fig:comparison_3}
\end{figure*}

\begin{figure*}[h]
    \centering
    \setlength{\tabcolsep}{0.5pt}
    \renewcommand{\arraystretch}{0.5}
    \setlength{\ww}{0.172\linewidth}
    \renewcommand{\prsize}{\defbigprsize}
    \renewcommand{\methsize}{normalsize}
    \renewcommand{\pw}{60pt}
    \renewcommand{\rsm}{1.76cm}
    \renewcommand{\rsb}{3px}

    \begin{tabular}{cc cc cc}  
        \sizedtext{\methsize}Prompt &
        \sizedtext{\methsize}Input &
        \sizedtext{\methsize}\emu &
        \sizedtext{\methsize}\mb &
        \sizedtext{\methsize}\ipp &
        \sizedtext{\methsize}\ctmb 
        \\
        [\rsb]
        
        \prprs[\pw]{\prsize}{Add a baseball in front of the batter next to his face}{A baseball}
        \raisebox{-.5\height}{\includegraphics[width=\ww]{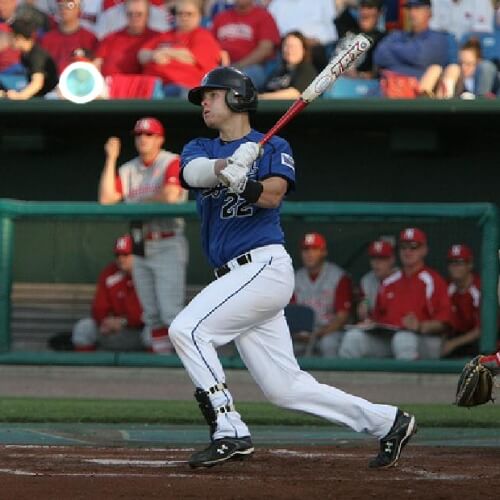}} &
        \raisebox{-.5\height}{\includegraphics[width=\ww]{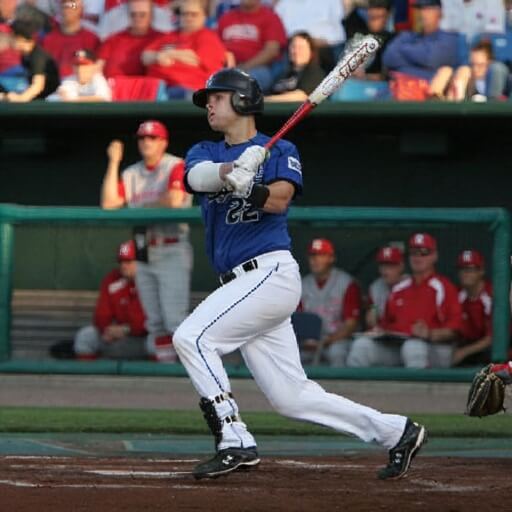}} &
        \raisebox{-.5\height}{\includegraphics[width=\ww]{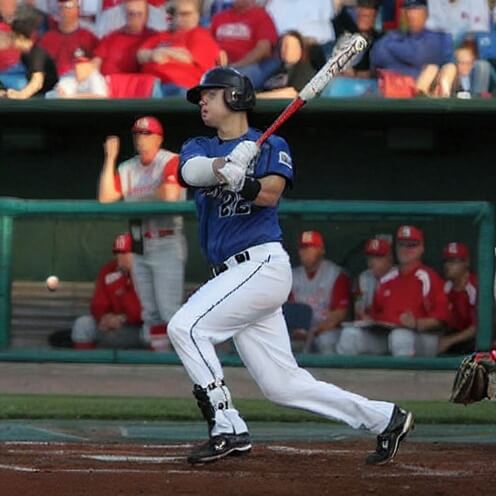}} &
        \raisebox{-.5\height}{\includegraphics[width=\ww]{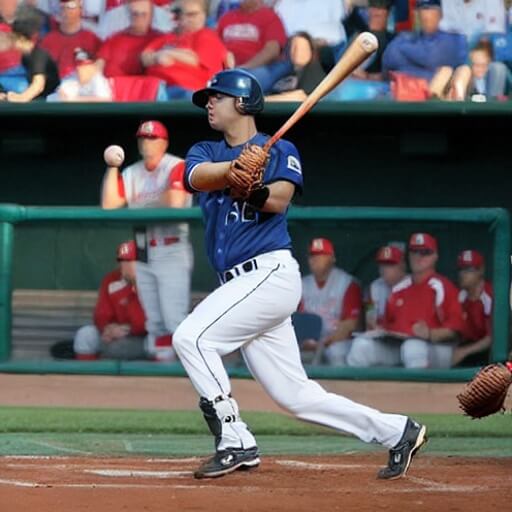}} &
        \raisebox{-.5\height}{\includegraphics[width=\ww]{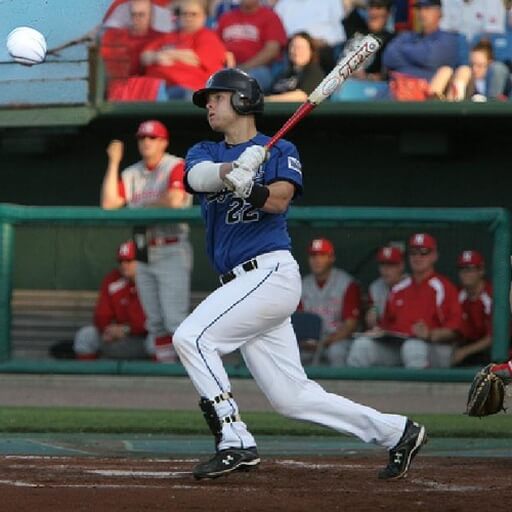}} 
        \\
        [\rsm]
      
        \prprs[\pw]{\prsize}{Insert a bag of chips on the left side on the hotdog}{A bag of chips}
        \raisebox{-.5\height}{\includegraphics[width=\ww]{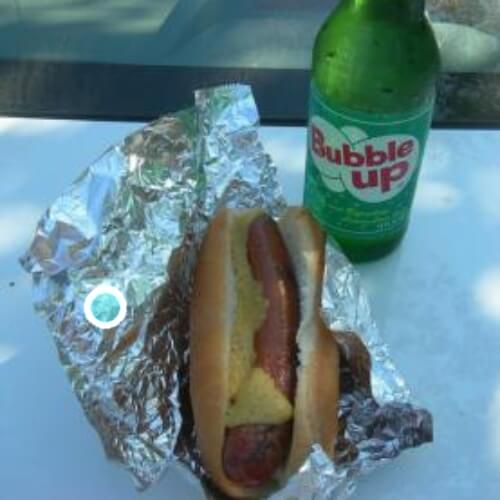}} &
        \raisebox{-.5\height}{\includegraphics[width=\ww]{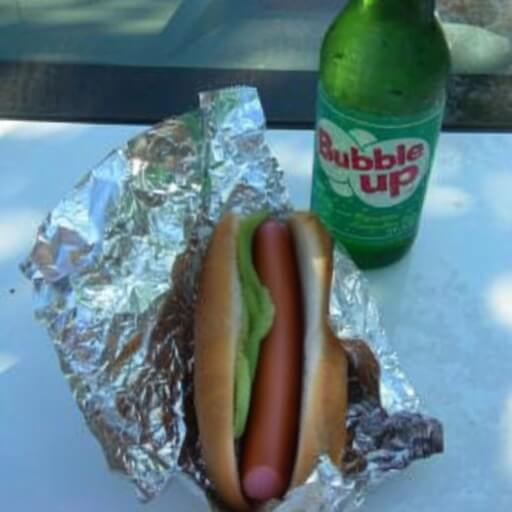}} &
        \raisebox{-.5\height}{\includegraphics[width=\ww]{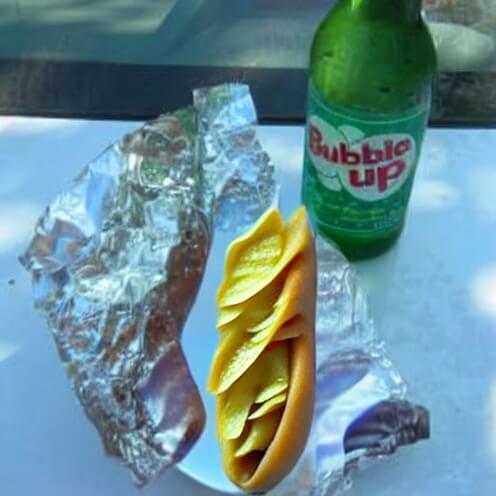}} &
        \raisebox{-.5\height}{\includegraphics[width=\ww]{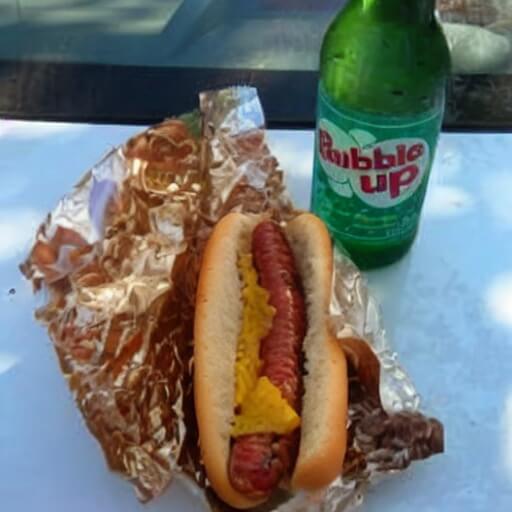}} &
        \raisebox{-.5\height}{\includegraphics[width=\ww]{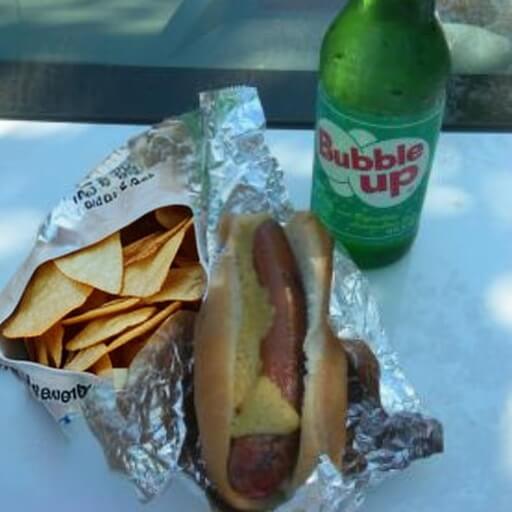}} 
        \\
        [\rsm]

        \prprs[\pw]{\prsize}{Add smoke to the planks}{Smoke}
        \raisebox{-.5\height}{\includegraphics[width=\ww]{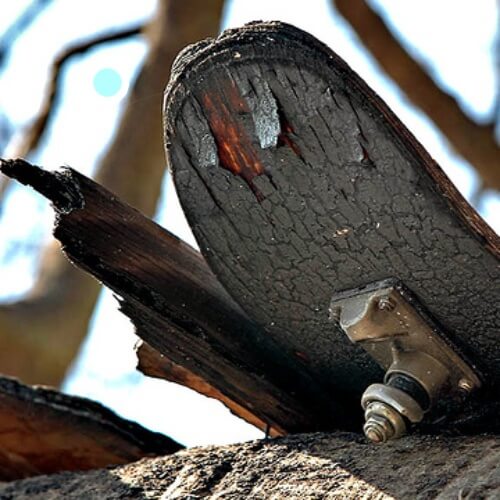}} &
        \raisebox{-.5\height}{\includegraphics[width=\ww]{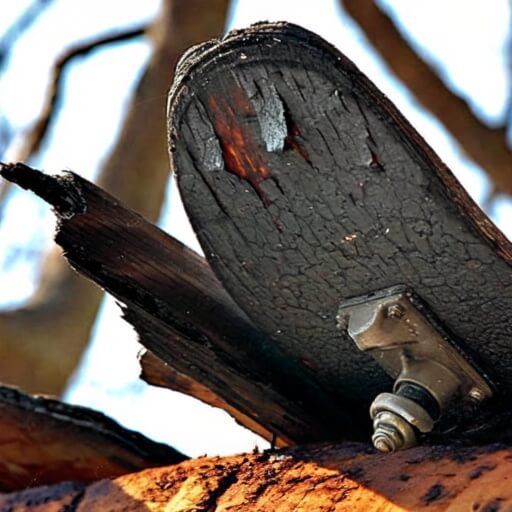}} &
        \raisebox{-.5\height}{\includegraphics[width=\ww]{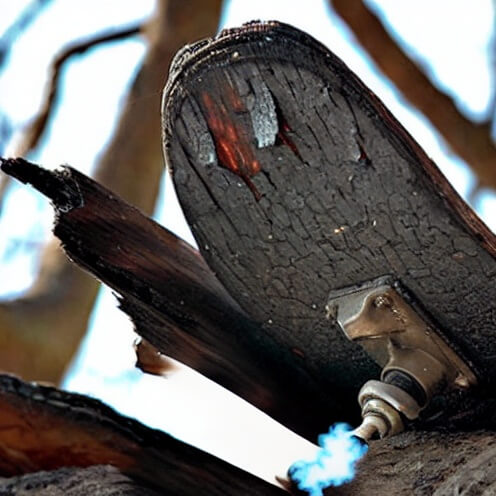}} &
        \raisebox{-.5\height}{\includegraphics[width=\ww]{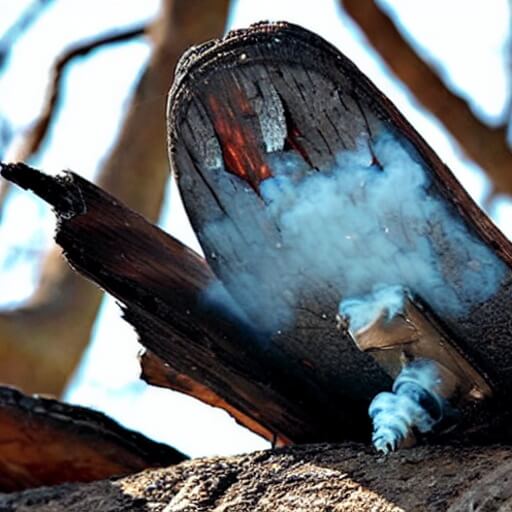}} &
        \raisebox{-.5\height}{\includegraphics[width=\ww]{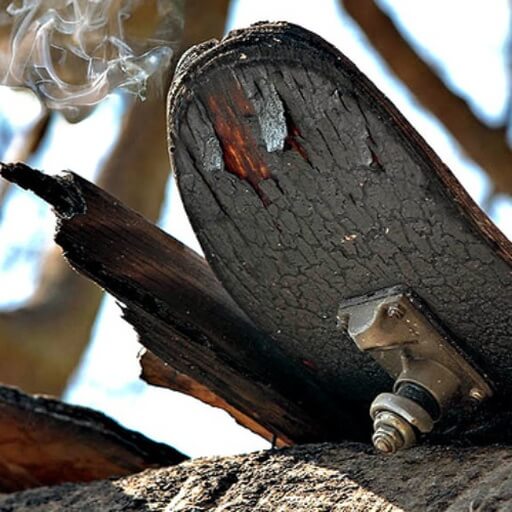}} 
        \\
        [\rsm]
     
        \prprs[\pw]{\prsize}{In the mirror show the reflection of a ghost}{A reflection of a ghost}
        \raisebox{-.5\height}{\includegraphics[width=\ww]{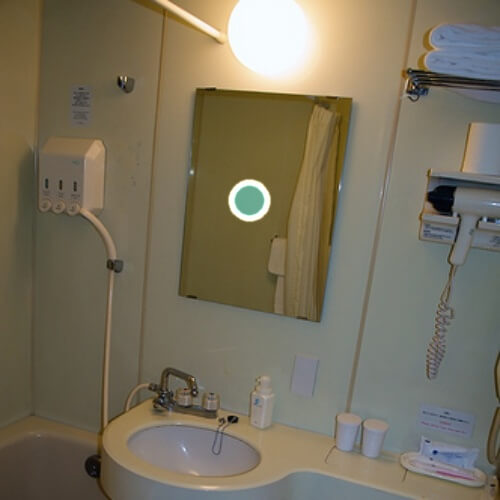}} &
        \raisebox{-.5\height}{\includegraphics[width=\ww]{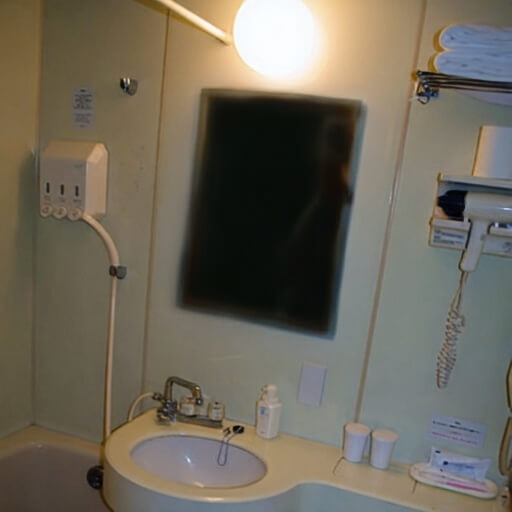}} &
        \raisebox{-.5\height}{\includegraphics[width=\ww]{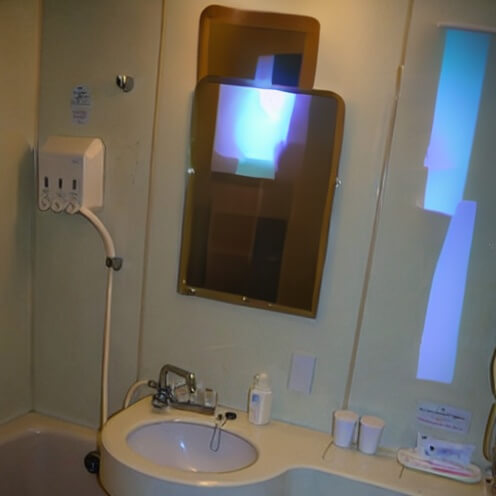}} &
        \raisebox{-.5\height}{\includegraphics[width=\ww]{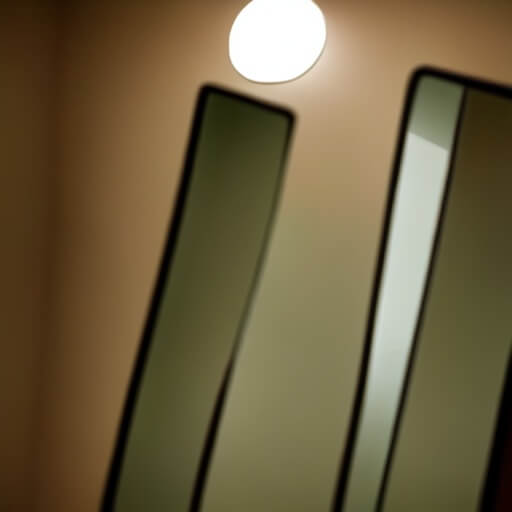}} &
        \raisebox{-.5\height}{\includegraphics[width=\ww]{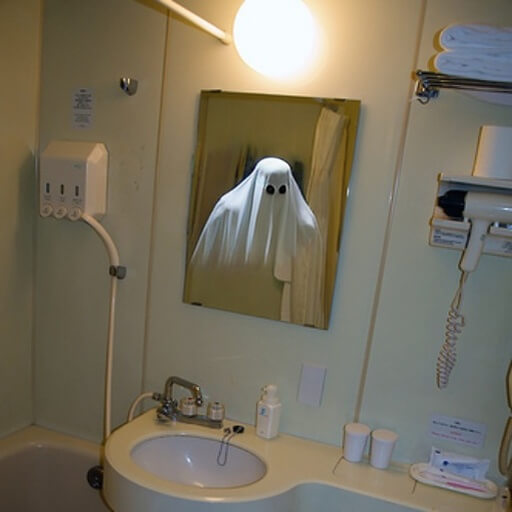}} 
        \\
        [\rsm]
      
        \prprs[\pw]{\prsize}{Add graffiti to the orange tiles.}{Graffiti}
        \raisebox{-.5\height}{\includegraphics[width=\ww]{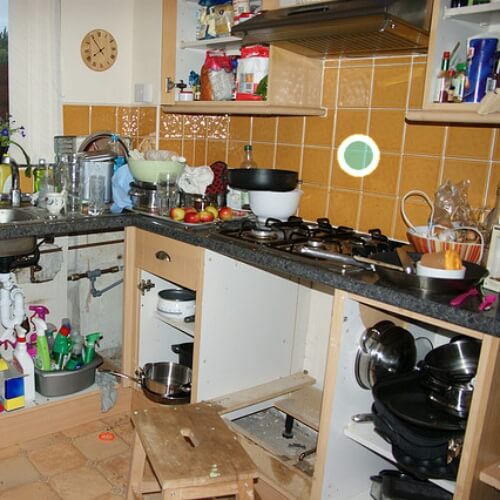}} &
        \raisebox{-.5\height}{\includegraphics[width=\ww]{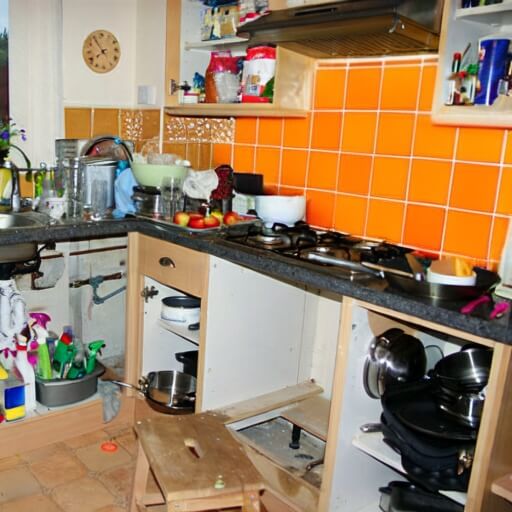}} &
        \raisebox{-.5\height}{\includegraphics[width=\ww]{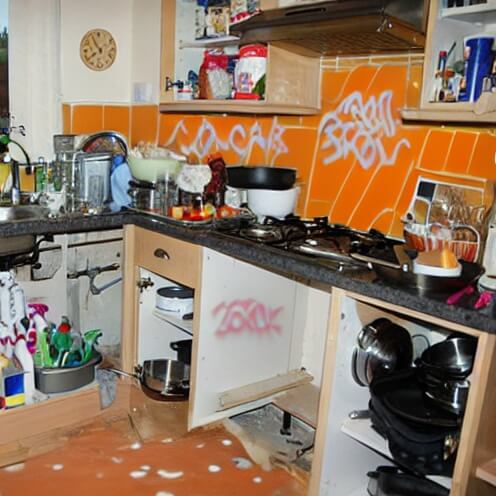}} &
        \raisebox{-.5\height}{\includegraphics[width=\ww]{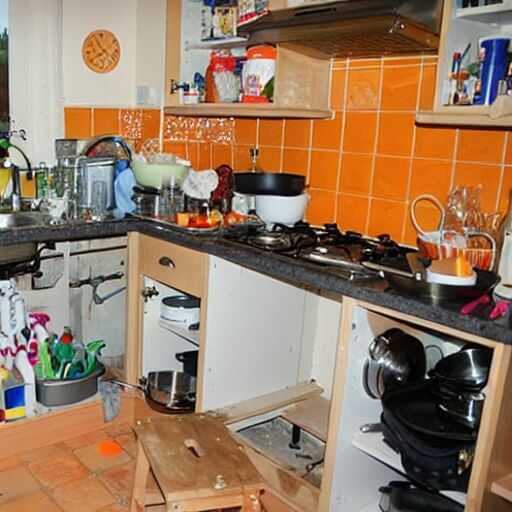}} &
        \raisebox{-.5\height}{\includegraphics[width=\ww]{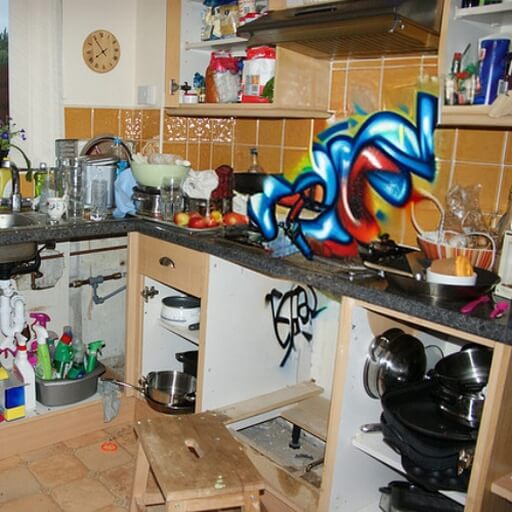}} 
        \\
        [\rsm]
        
        \prprs[\pw]{\prsize}{Add some toys in  the stand}{Some toys}
        \raisebox{-.5\height}{\includegraphics[width=\ww]{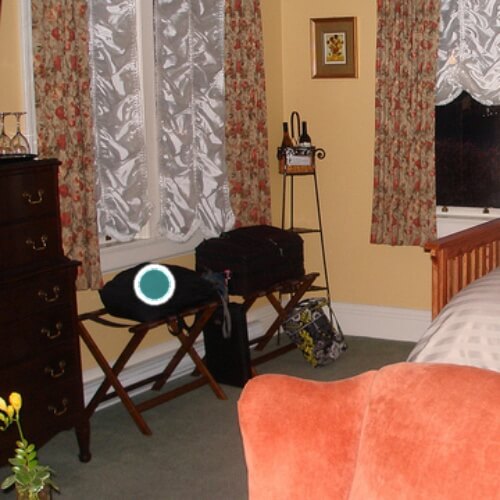}} &
        \raisebox{-.5\height}{\includegraphics[width=\ww]{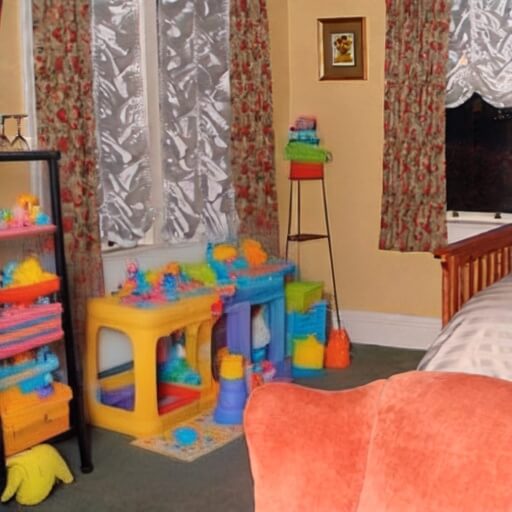}} &
        \raisebox{-.5\height}{\includegraphics[width=\ww]{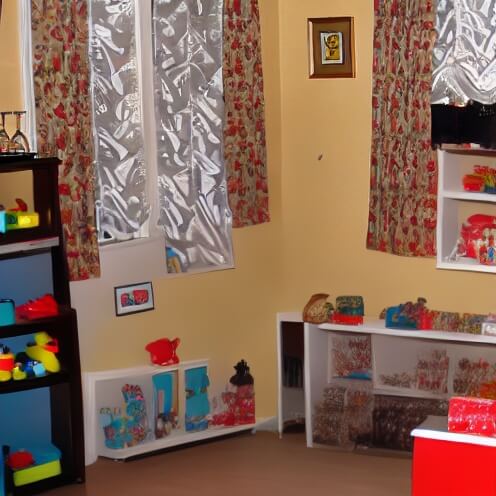}} &
        \raisebox{-.5\height}{\includegraphics[width=\ww]{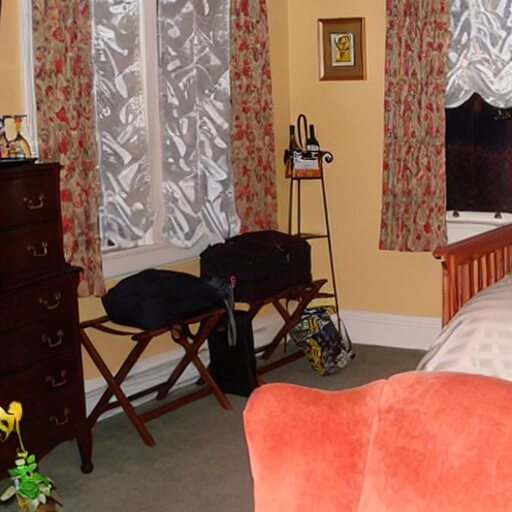}} &
        \raisebox{-.5\height}{\includegraphics[width=\ww]{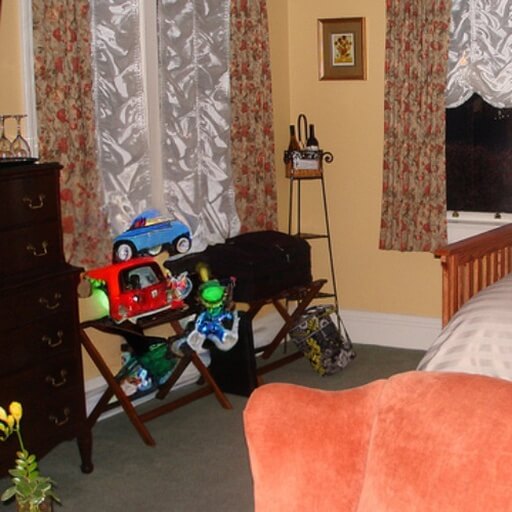}}

    \end{tabular}
    
    \caption{\textbf{Additional comparisons with SoTA methods.}}
    \label{fig:comparison_4}
\end{figure*}

\begin{figure*}[h]
    \centering
    \setlength{\tabcolsep}{0.5pt}
    \renewcommand{\arraystretch}{0.5}
    \setlength{\ww}{0.17\linewidth}
    \renewcommand{\prsize}{\defbigprsize}
    \renewcommand{\methsize}{normalsize}
    \renewcommand{\pw}{65pt}
    \renewcommand{\rsm}{1.76cm}
    \renewcommand{\rsb}{3px}

    \begin{tabular}{cc cc cc}  
        \sizedtext{\methsize}Prompt &
        \sizedtext{\methsize}Input &
        \sizedtext{\methsize}\emu &
        \sizedtext{\methsize}\mb &
        \sizedtext{\methsize}\ipp &
        \sizedtext{\methsize}\ctmb 
        \\
        [\rsb]
        
        \prprs[\pw]{\prsize}{Add a sandcastle to the right of the dog}{A sandcastle}
        \raisebox{-.5\height}{\includegraphics[width=\ww]{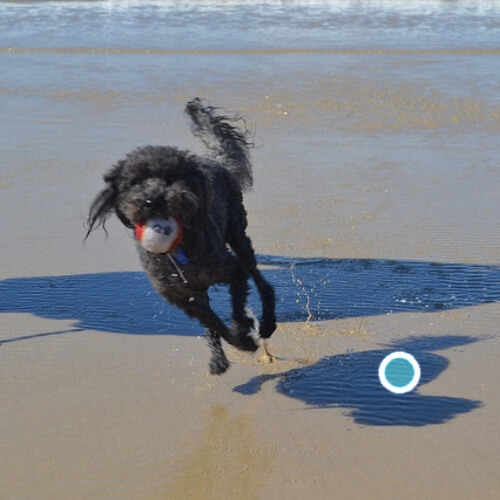}} &
        \raisebox{-.5\height}{\includegraphics[width=\ww]{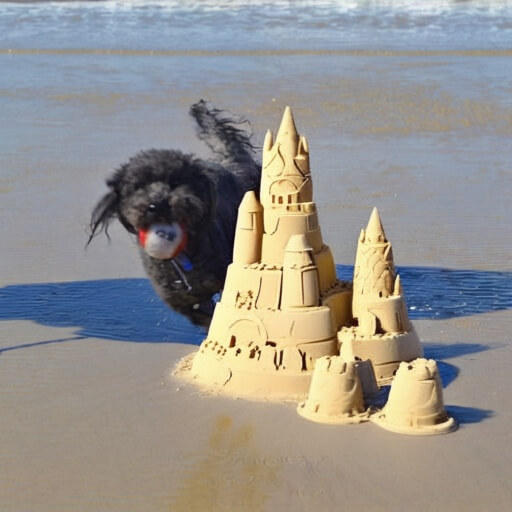}} &
        \raisebox{-.5\height}{\includegraphics[width=\ww]{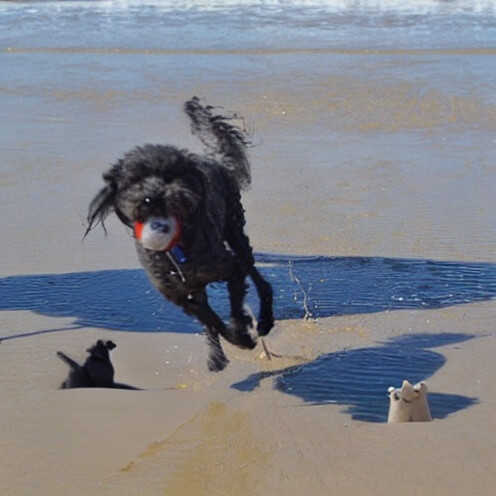}} &
        \raisebox{-.5\height}{\includegraphics[width=\ww]{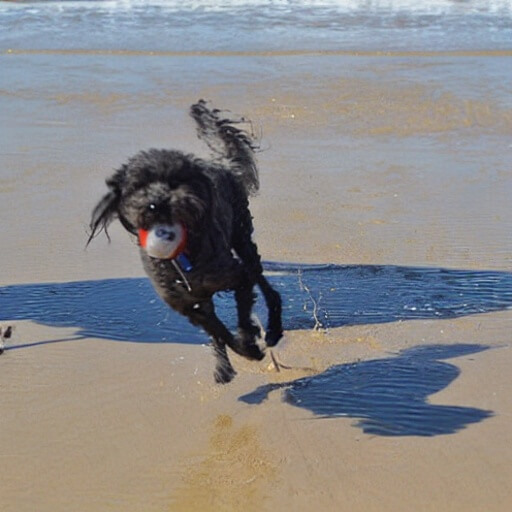}} &
        \raisebox{-.5\height}{\includegraphics[width=\ww]{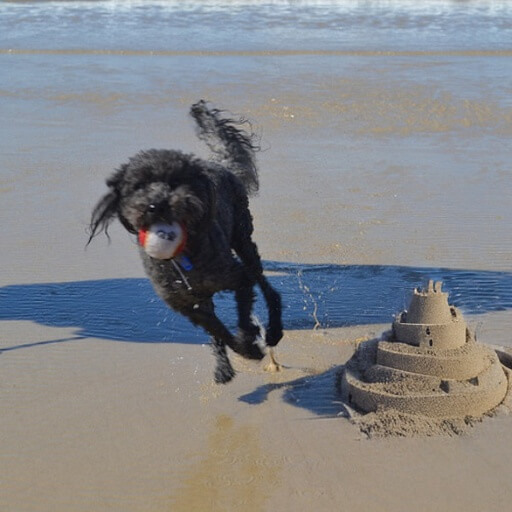}} 
        \\
        [\rsm]
       
        \prprs[\pw]{\prsize}{Add a grasshopper in the grass}{A grasshopper}
        \raisebox{-.5\height}{\includegraphics[width=\ww]{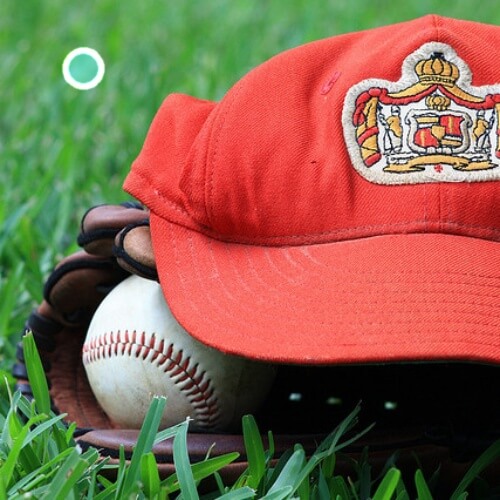}} &
        \raisebox{-.5\height}{\includegraphics[width=\ww]{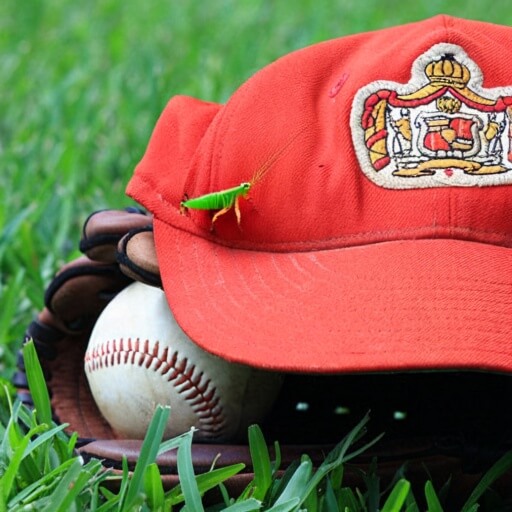}} &
        \raisebox{-.5\height}{\includegraphics[width=\ww]{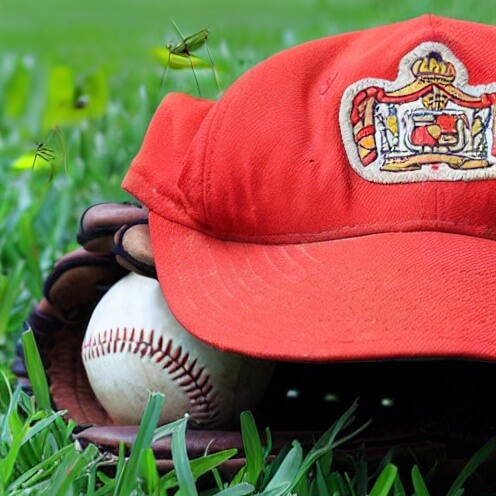}} &
        \raisebox{-.5\height}{\includegraphics[width=\ww]{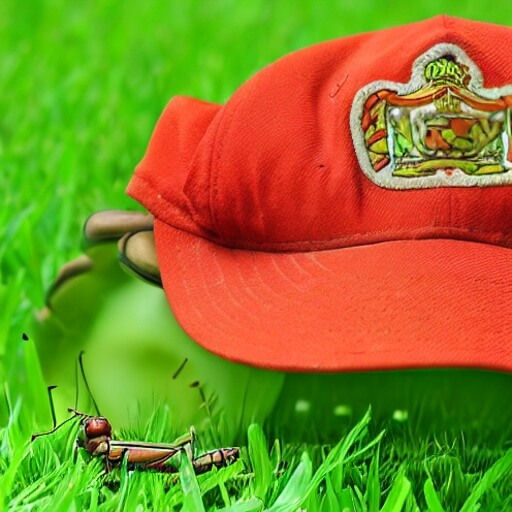}} &
        \raisebox{-.5\height}{\includegraphics[width=\ww]{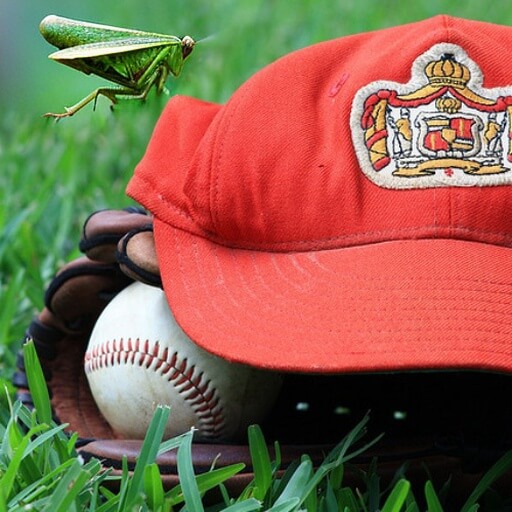}} 
        \\
        [\rsm]
        
        \prprs[\pw]{\prsize}{Add a hot air balloon to the background above the computer mouse}{A hot air balloon}
        \raisebox{-.5\height}{\includegraphics[width=\ww]{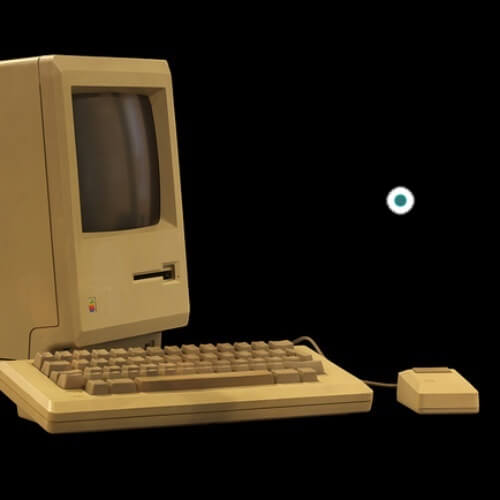}} &
        \raisebox{-.5\height}{\includegraphics[width=\ww]{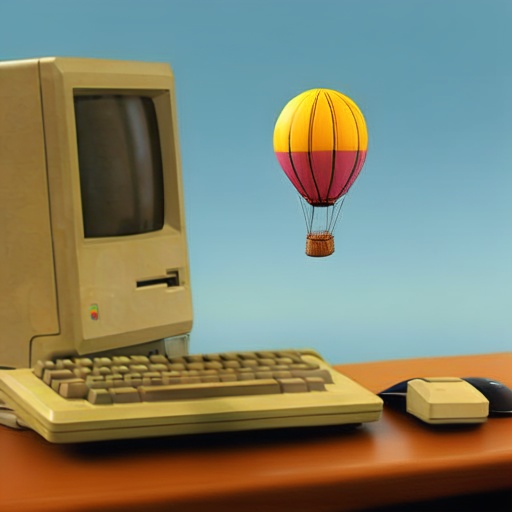}} &
        \raisebox{-.5\height}{\includegraphics[width=\ww]{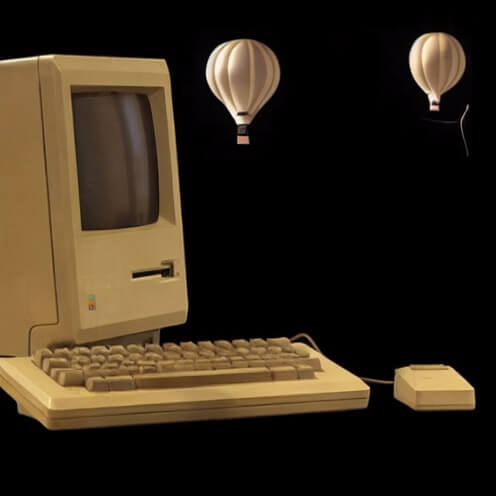}} &
        \raisebox{-.5\height}{\includegraphics[width=\ww]{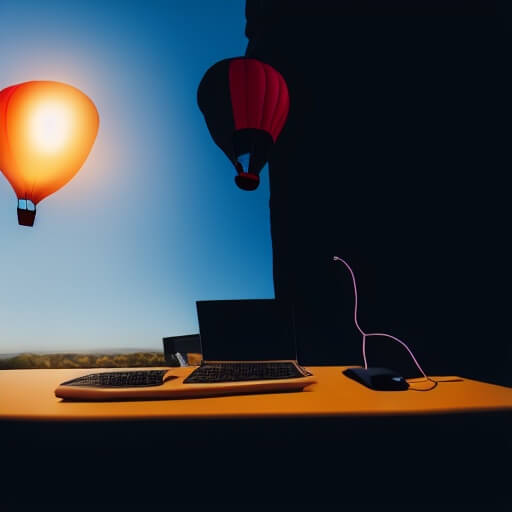}} &
        \raisebox{-.5\height}{\includegraphics[width=\ww]{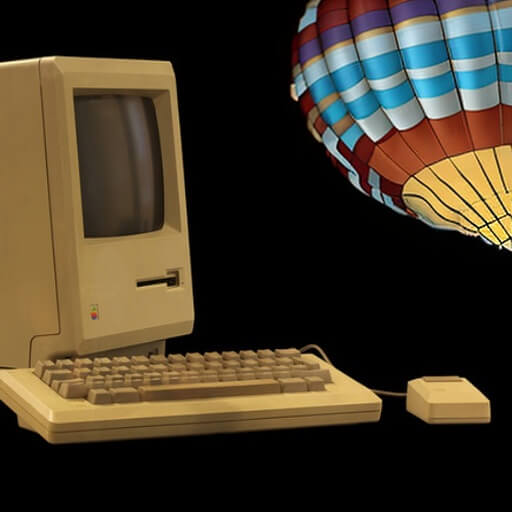}} 
        \\
         [\rsm]
        
        \prprs[\pw]{\prsize}{Add a pizza sign to the window}{A pizza sign}
        \raisebox{-.5\height}{\includegraphics[width=\ww]{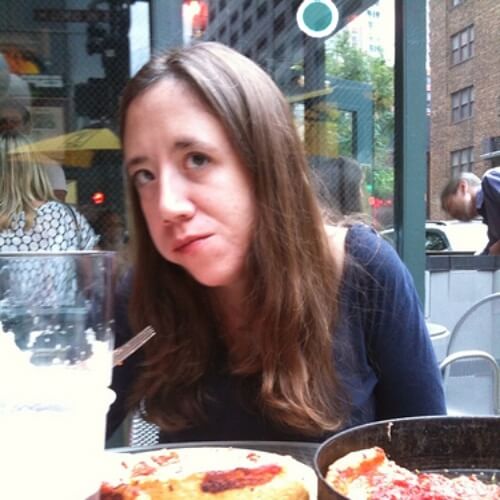}} &
        \raisebox{-.5\height}{\includegraphics[width=\ww]{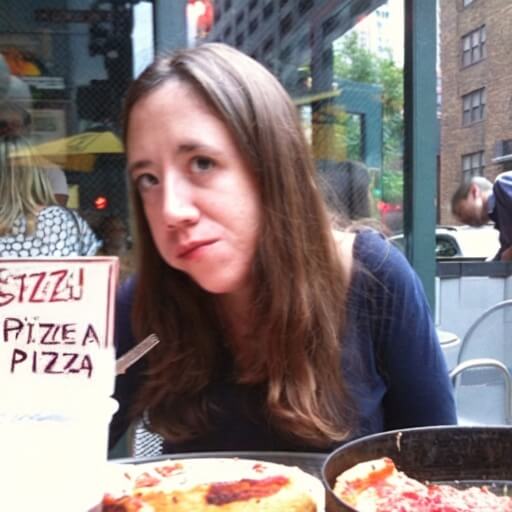}} &
        \raisebox{-.5\height}{\includegraphics[width=\ww]{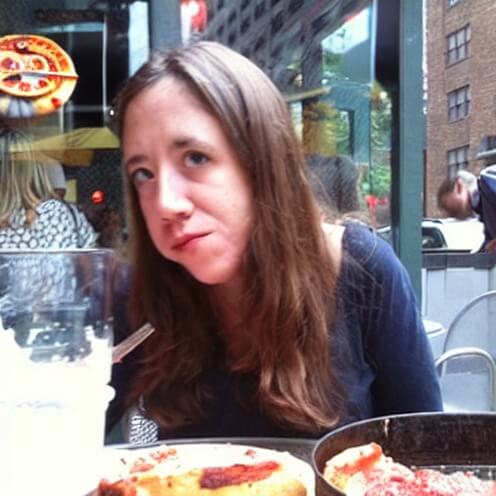}} &
        \raisebox{-.5\height}{\includegraphics[width=\ww]{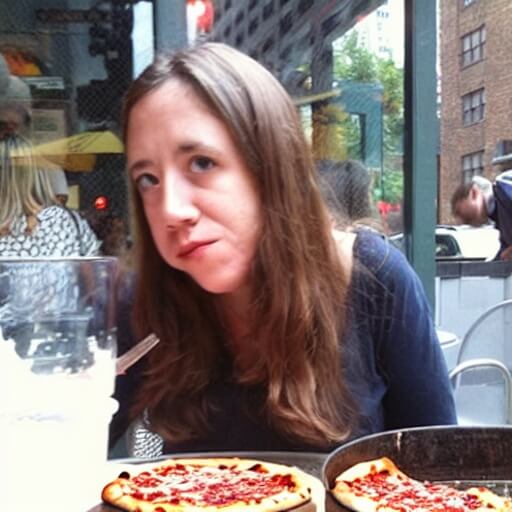}} &
        \raisebox{-.5\height}{\includegraphics[width=\ww]{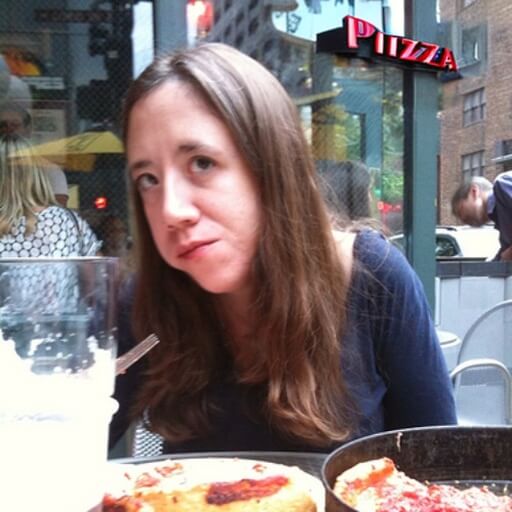}} 
        \\
        [\rsm]

       \prprs[\pw]{\prsize}{Add a soda bottle to the desk}{A soda bottle}
        \raisebox{-.5\height}{\includegraphics[width=\ww]{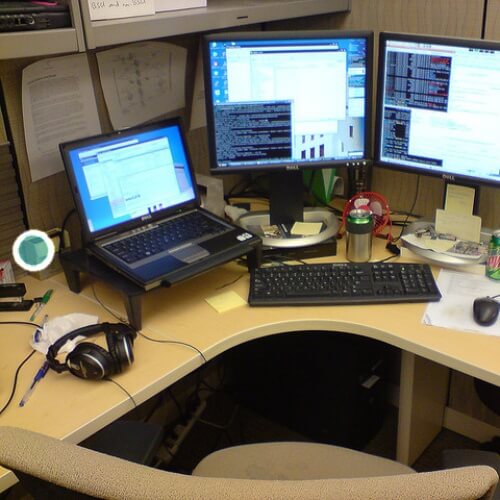}} &
        \raisebox{-.5\height}{\includegraphics[width=\ww]{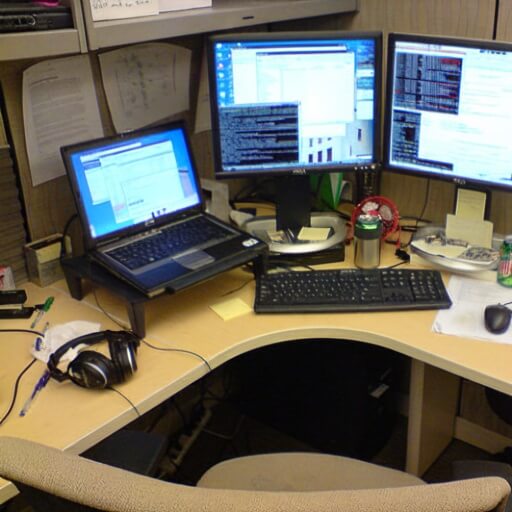}} &
        \raisebox{-.5\height}{\includegraphics[width=\ww]{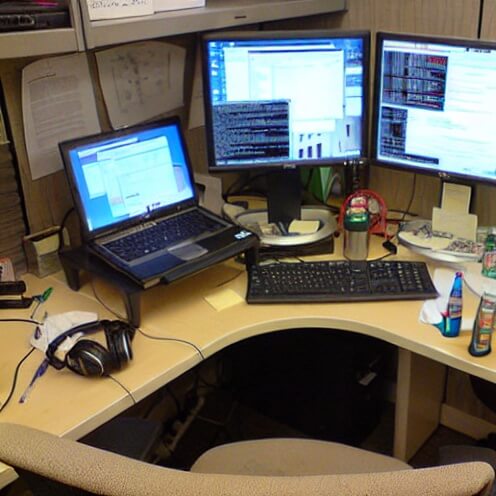}} &
        \raisebox{-.5\height}{\includegraphics[width=\ww]{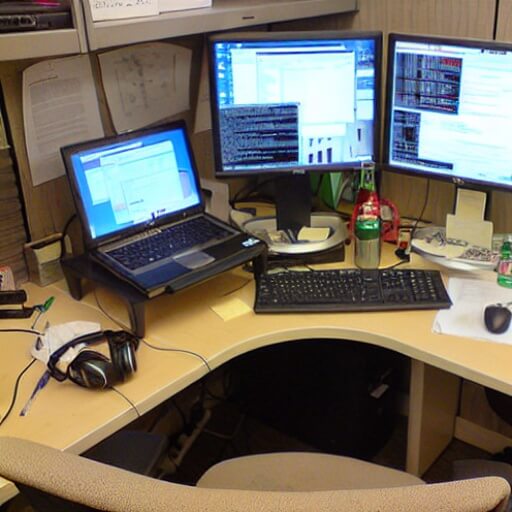}} &
        \raisebox{-.5\height}{\includegraphics[width=\ww]{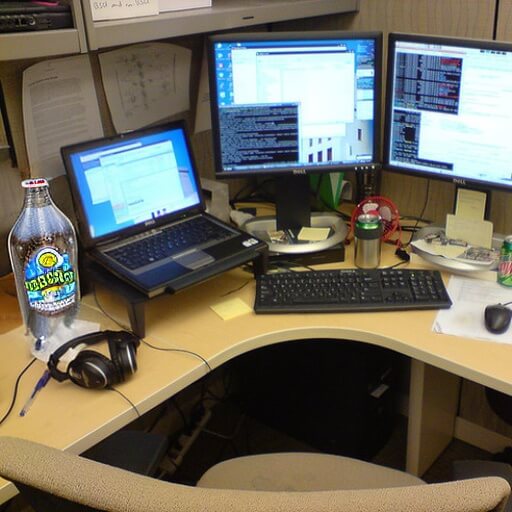}} 
        \\
        [\rsm]
        
        \prprs[\pw]{\prsize}{Add a saddle to the horse}{A saddle}
        \raisebox{-.5\height}{\includegraphics[width=\ww]{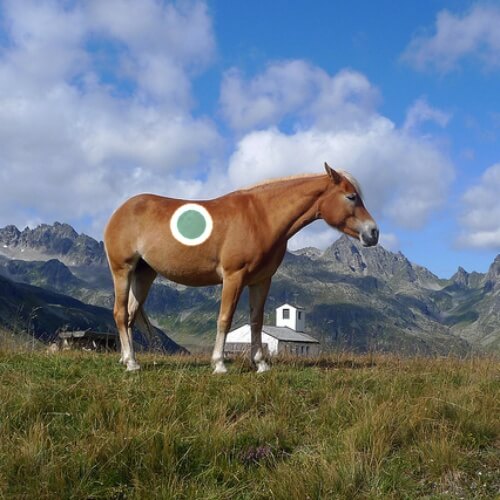}} &
        \raisebox{-.5\height}{\includegraphics[width=\ww]{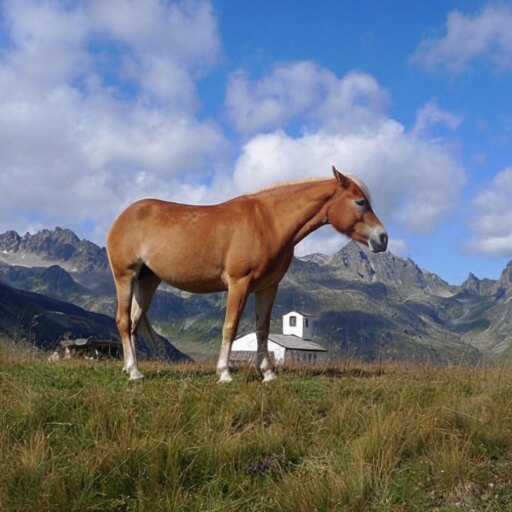}} &
        \raisebox{-.5\height}{\includegraphics[width=\ww]{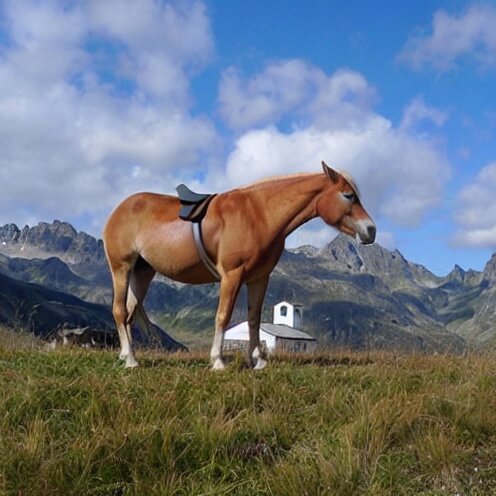}} &
        \raisebox{-.5\height}{\includegraphics[width=\ww]{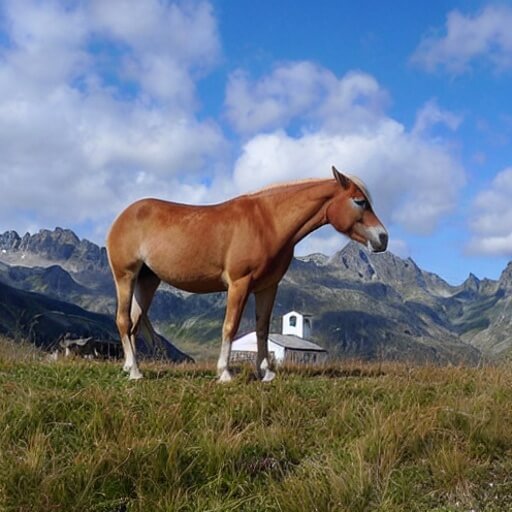}} &
        \raisebox{-.5\height}{\includegraphics[width=\ww]{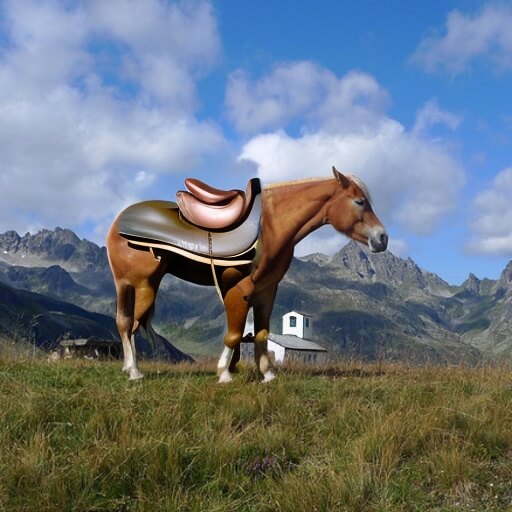}}

    \end{tabular}
    
    \caption{\textbf{Additional comparisons with SoTA methods.}}
    \label{fig:comparison_5}
\end{figure*}

Additional examples of \ctm generated masks can be found in \Cref{fig:generated_masks_appendix}. Further results comparing to baselines \emu, \mb, and \ipp\ are provided in \Cref{fig:comparison_2}, \Cref{fig:comparison_3}, \Cref{fig:comparison_4}, and \Cref{fig:comparison_5}. A comparison of prompt lengths with baselines is illustrated in \Cref{fig:word_count}.

\subsection{Additional Ablation Study}
\label{additional_ablation}
\begin{figure*}[h]
    \centering
    \setlength{\tabcolsep}{0.5pt}
    \renewcommand{\arraystretch}{0.5}
    \setlength{\ww}{0.17\linewidth}
    \renewcommand{\prsize}{footnotesize}
    \renewcommand{\methsize}{footnotesize}
    \renewcommand{\rsm}{1.6cm}
    \setlength{\rss}{4px}
    \setlength{\rsb}{15px}
    
    \begin{tabular}{cc cc cc}

        \sizedtext{\methsize}Prompt &
        \sizedtext{\methsize}Input &
        \sizedtext{\methsize}(a) \emu &
        \sizedtext{\methsize}(b) No blend &
        \sizedtext{\methsize}(c) Binary blend &
        \sizedtext{\methsize}(d) Gaussian blend 
        \\
        [0.3\rss]
        
        \prprs[60pt]{\prsize}{Add a massive sinkhole in front of the riders}{A massive sinkhole}
        {\includegraphics[valign=c, width=\ww]{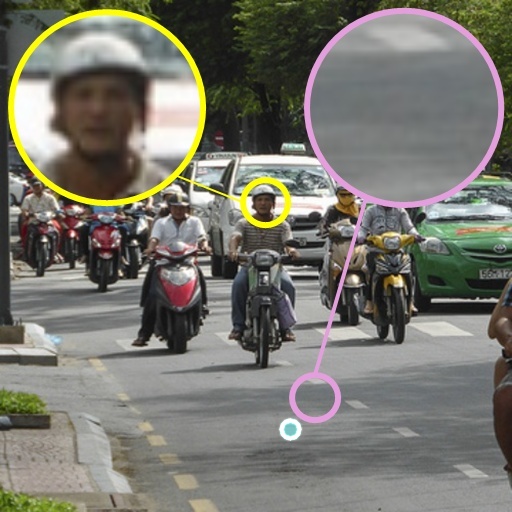}} &
        {\includegraphics[valign=c, width=\ww]{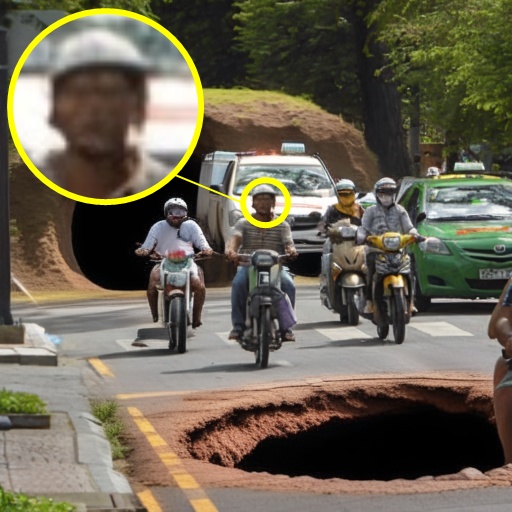}}  &
        {\includegraphics[valign=c, width=\ww]{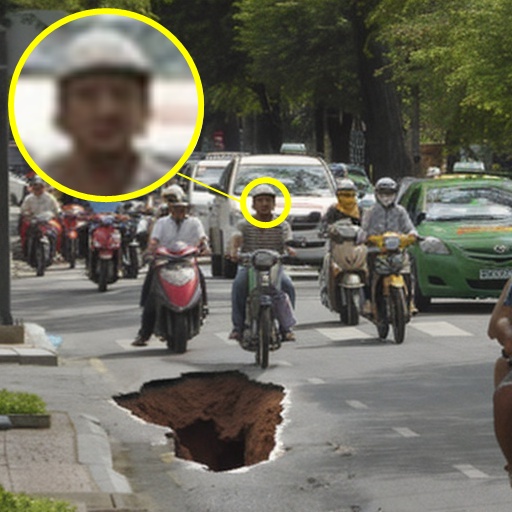}}  &
        {\includegraphics[valign=c, width=\ww]{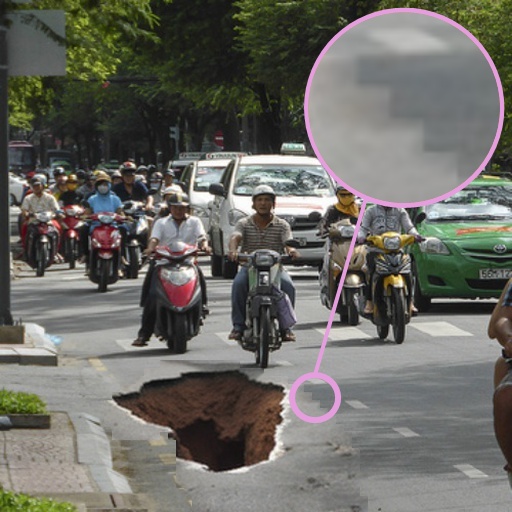}}  &
        {\includegraphics[valign=c, width=\ww]{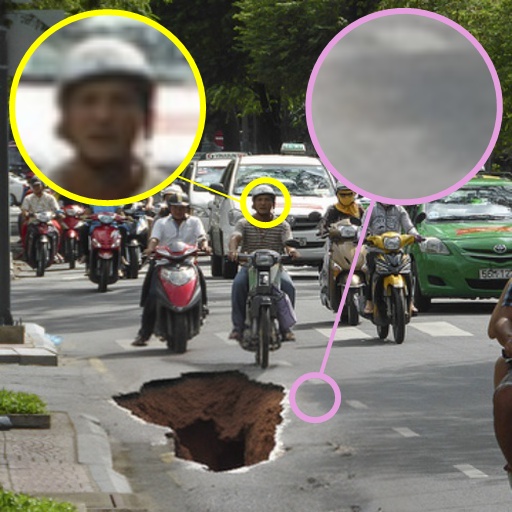}}
        \\
        [\rsm]
        
        \prprs[60pt]{\prsize}{Add a cat playing with the white mouse}{A cat playing}
        {\includegraphics[valign=c, width=\ww]{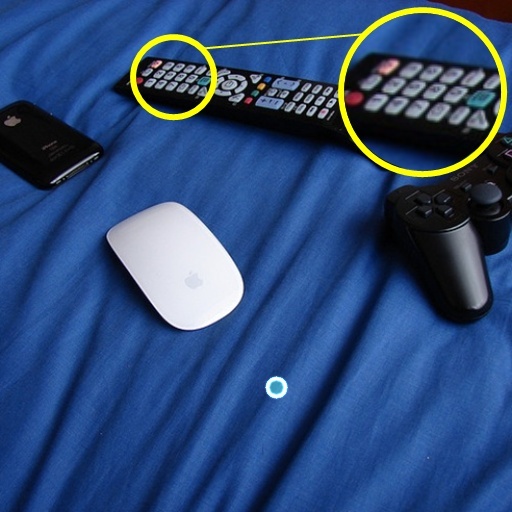}} &
        {\includegraphics[valign=c, width=\ww]{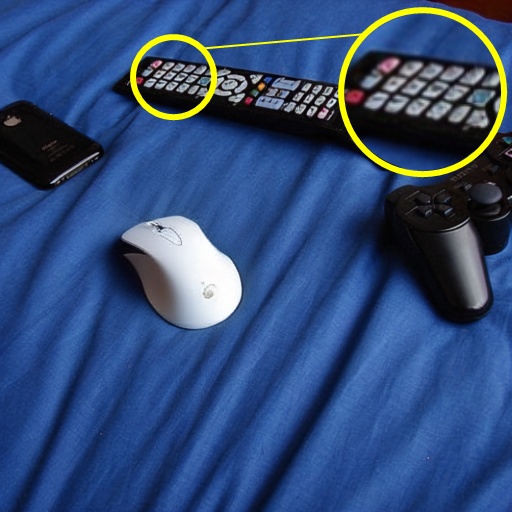}}  &
        {\includegraphics[valign=c, width=\ww]{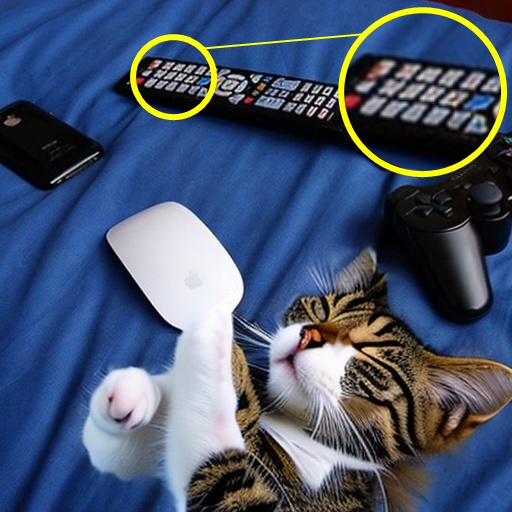}}  &
        {\includegraphics[valign=c, width=\ww]{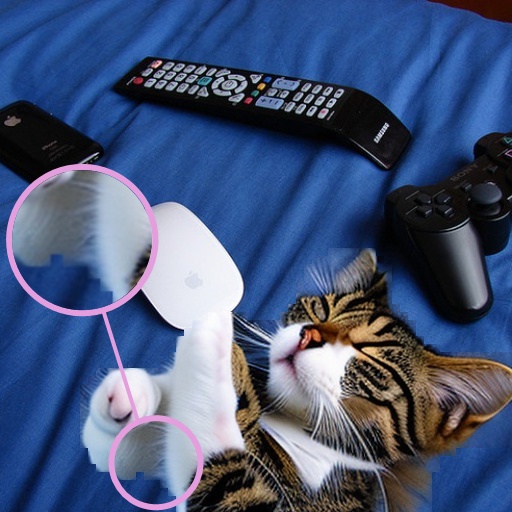}}  &
        {\includegraphics[valign=c, width=\ww]{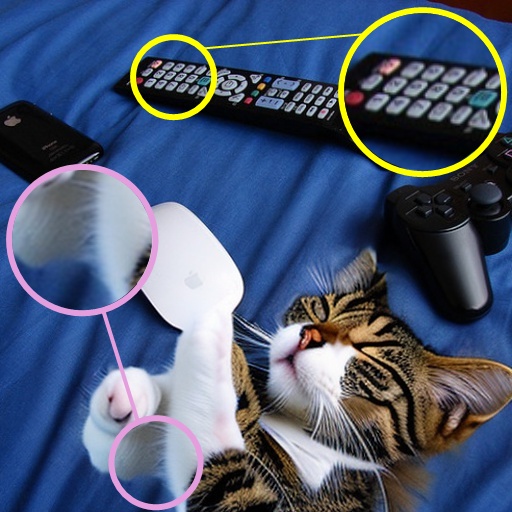}}

    \end{tabular}
    
    \caption{\textbf{Background preservation ablation study}. When decoding the final diffused latents, details are not fully preserved (b). A binary blending of the mask and the input image at pixel space will yield artifacts on pixels surrounding the mask's contour (c). \emu\ suffers as well from loss of details. As mentioned in \Cref{sec:method}, we suggest a Gaussian blend at pixel space (d), which preserves the background details, while creating a seamless blend. This also eliminates the need for a decoder weights optimization presented in BLD. Please zoom in for a vivid visual.}

    \label{fig:abl_background_preservation}
\end{figure*}

\begin{figure*}[h]
    \centering
    \setlength{\tabcolsep}{0pt}
    \renewcommand{\arraystretch}{0.5}
    \setlength{\ww}{0.121\linewidth}
    \renewcommand{\ablsize}{footnotesize}
    \renewcommand{\methsize}{footnotesize}
    \renewcommand{\rsm}{1.2cm}
    \renewcommand{\rsb}{1.1cm}

    \begin{tabular}{c @{\hspace{0.005\columnwidth}}c cc cc cc c @{\hspace{0.005\columnwidth}}c}  
        \abltitle{\ablsize}{Only inner}
        {\includegraphics[valign=c, width=\ww]{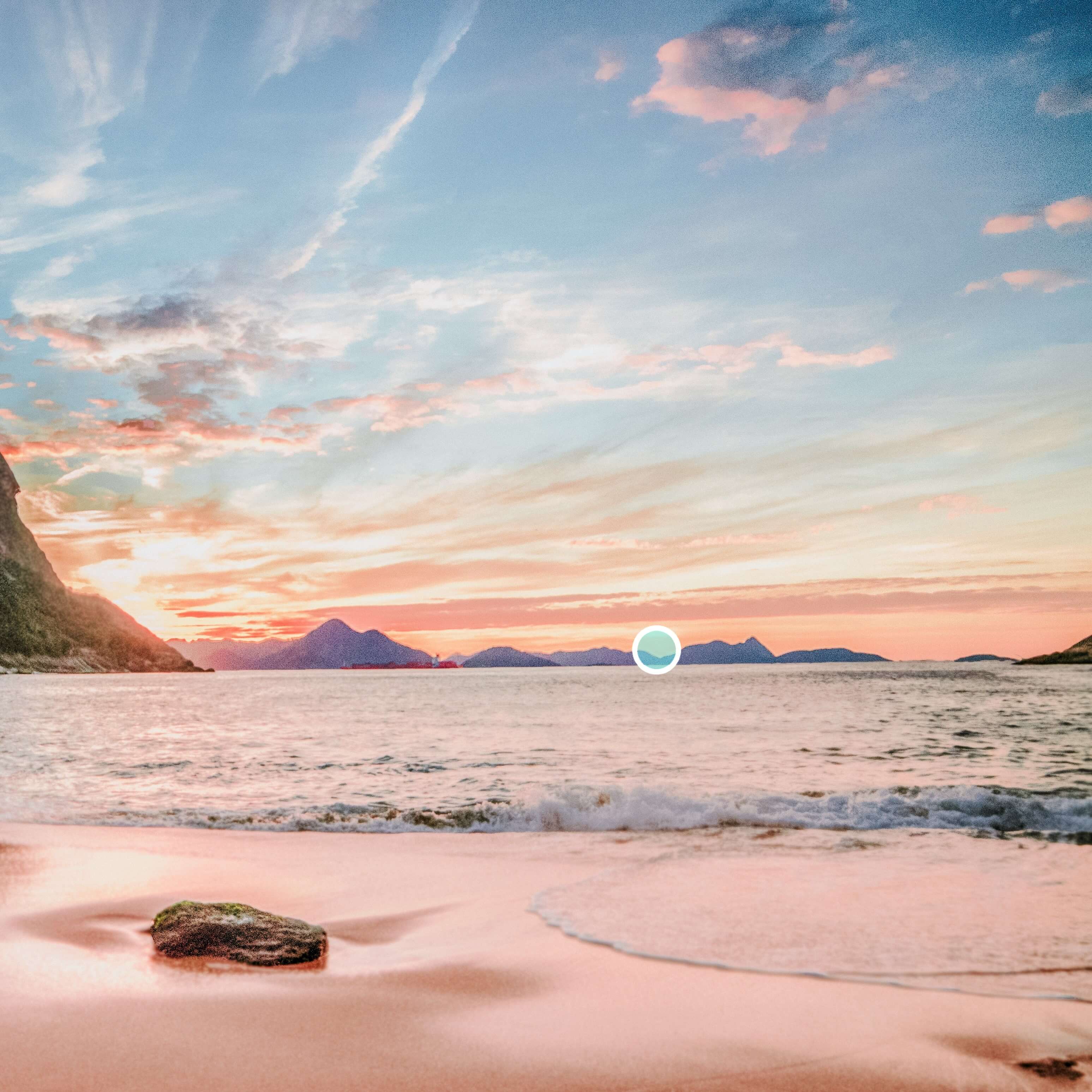}} &
        {\includegraphics[valign=c, width=\ww]{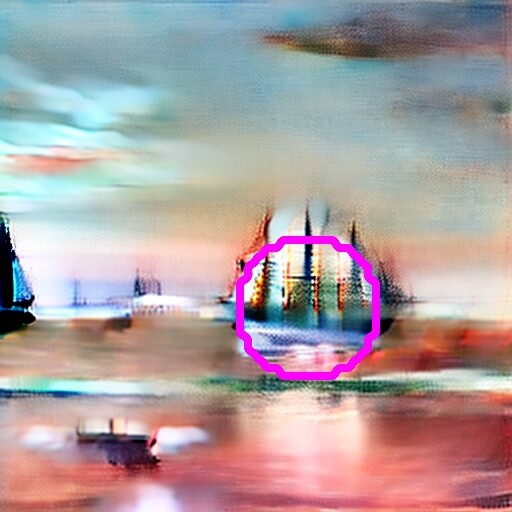}}  &
        {\includegraphics[valign=c, width=\ww]{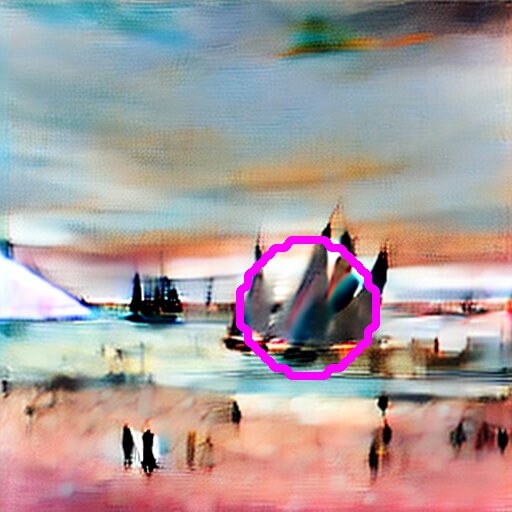}}  &
        {\includegraphics[valign=c, width=\ww]{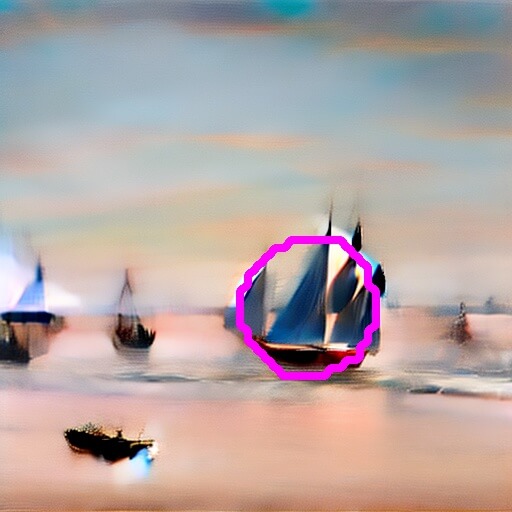}}  &
        {\includegraphics[valign=c, width=\ww]{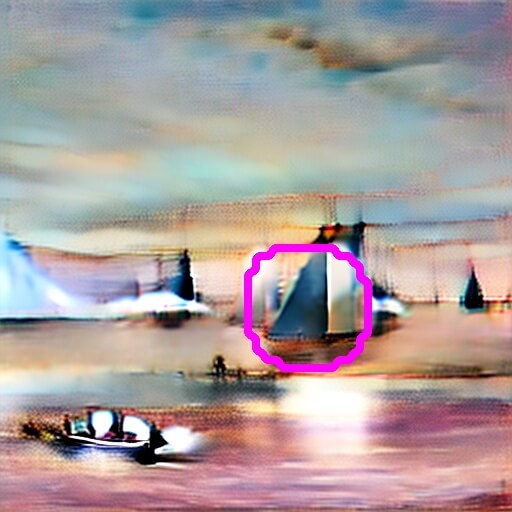}} &
        {\includegraphics[valign=c, width=\ww]{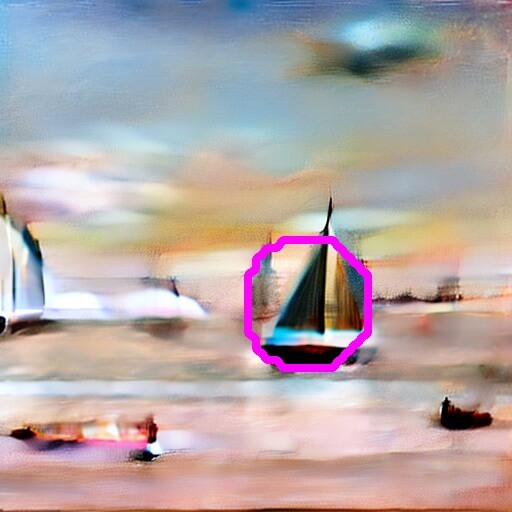}} &
        {\includegraphics[valign=c, width=\ww]{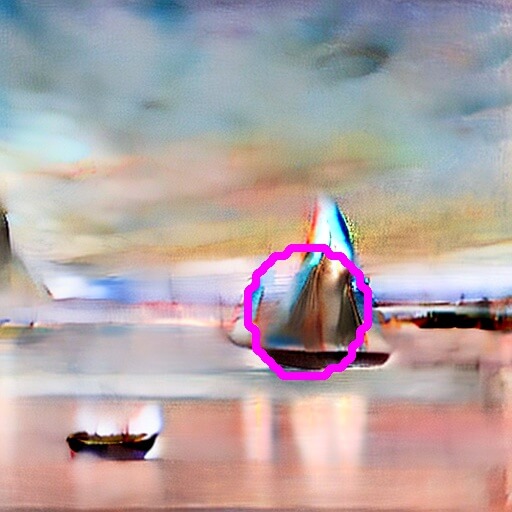}} &
        {\includegraphics[valign=c, width=\ww]{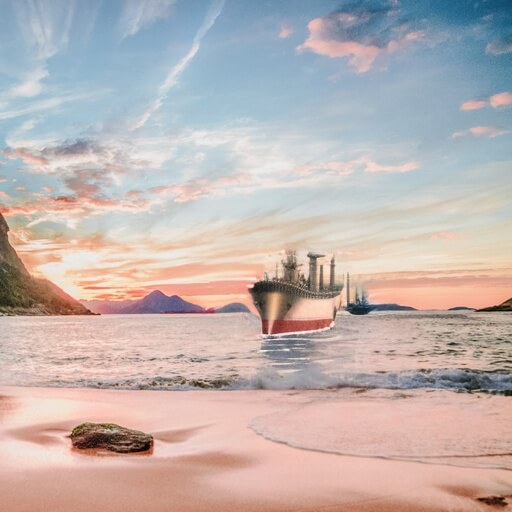}} 
        \\
        [\rsm]
        
        \abltitle{\ablsize}{Ours}
        {\includegraphics[valign=c, width=\ww]{figures/abl_no_outer_elev/no_outer_elev/fleet_in_click.jpg}} &
        {\includegraphics[valign=c, width=\ww]{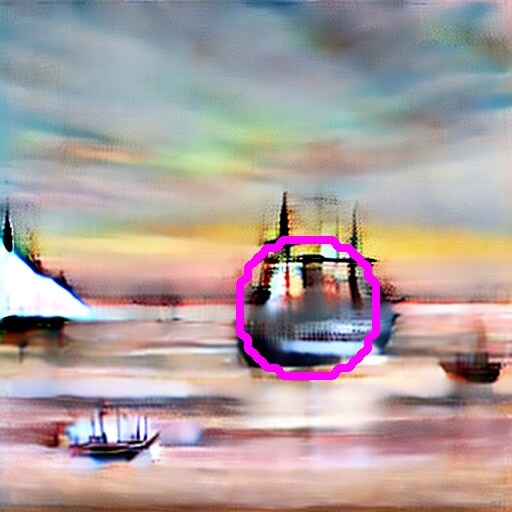}}  &
        {\includegraphics[valign=c, width=\ww]{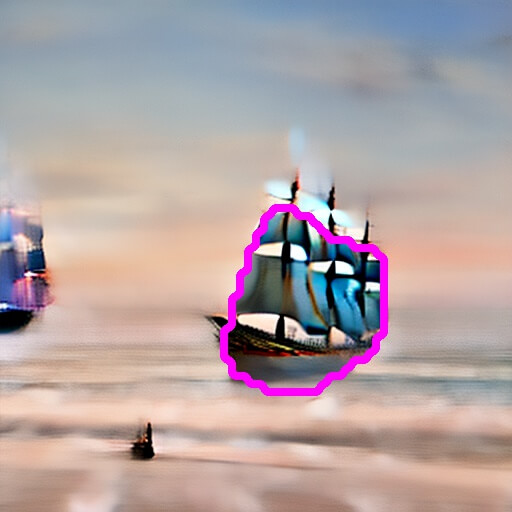}}  &
        {\includegraphics[valign=c, width=\ww]{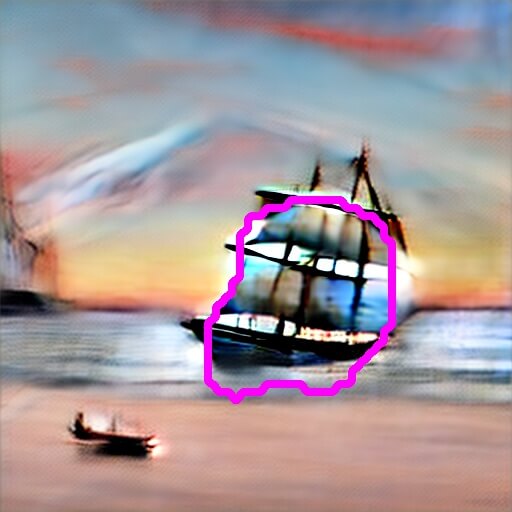}}  &
        {\includegraphics[valign=c, width=\ww]{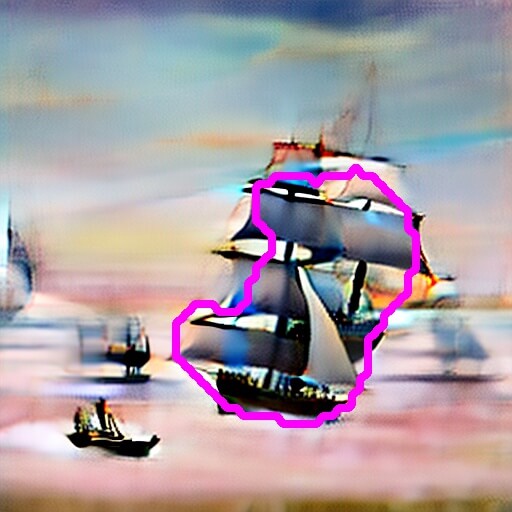}} &
        {\includegraphics[valign=c, width=\ww]{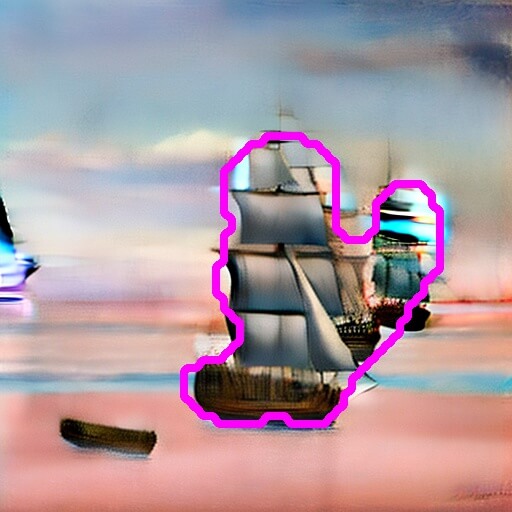}} &
        {\includegraphics[valign=c, width=\ww]{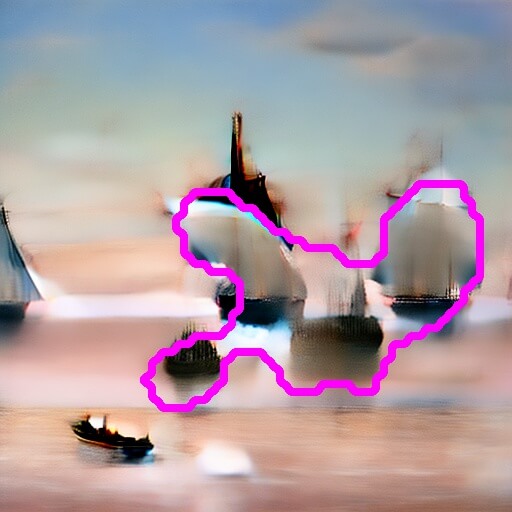}} &
        {\includegraphics[valign=c, width=\ww]{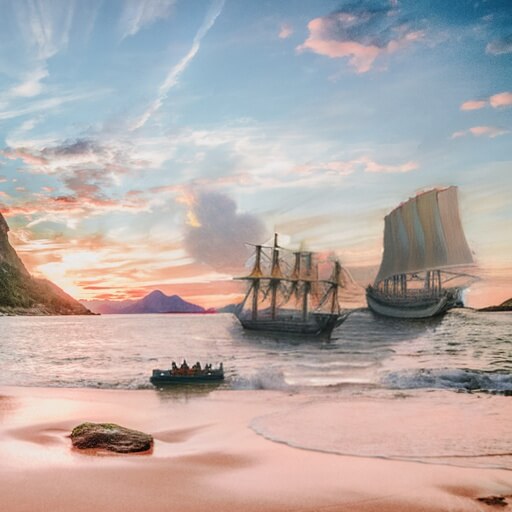}} 
        \\
        [\rsb]

        \sizedtext{\methsize}{Input}& \sizedtext{\methsize}{40\%}& \sizedtext{\methsize}{41\%}& \sizedtext{\methsize}{42\%}& \sizedtext{\methsize}{43\%}& \sizedtext{\methsize}{44\%} & \sizedtext{\methsize}{45\% (final $\mask$)}& \sizedtext{\methsize}{Output}

    \end{tabular}

    \caption{\textbf{Ablation study: elevating $\potential$ only inside $\boldsymbol{\mask}$}. Top row shows the evolution of $\mask$ (depicted by purple contours over $\est{z\un{fg}}$s during diffusion steps as indicated by percentages below) where potential $\potential$ elevation is contained within current $\mask$. Bottom row depicts $\mask$'s evolution in \ctm, where a surrounding ring of $\mask$ is also elevated in $\potential$. The prompt is \textit{``Fleet of ships"}. When elevating $\potential$ only within $\mask$, the mask shrinks continuously, unlike \ctm, where the outer ring elevation prevents the mask from shrinking in important areas, resulting in a mask shaped according to the generated objects.}
    \label{fig:ablation_no_outer_elevation}
\end{figure*}

\begin{figure*}[t]
    \centering
    \setlength{\tabcolsep}{0pt}
    \renewcommand{\arraystretch}{0.5}
    \setlength{\ww}{0.121\linewidth}
    \renewcommand{\ablsize}{footnotesize}
    \renewcommand{\methsize}{footnotesize}
    \renewcommand{\rsm}{1.2cm}
    \renewcommand{\rsb}{1.1cm}
    
    \begin{tabular}{c @{\hspace{0.005\columnwidth}}c cccccc @{\hspace{0.005\columnwidth}}c}    
        \abltitle{\ablsize}{All pixels}
        {\includegraphics[valign=c, width=\ww]{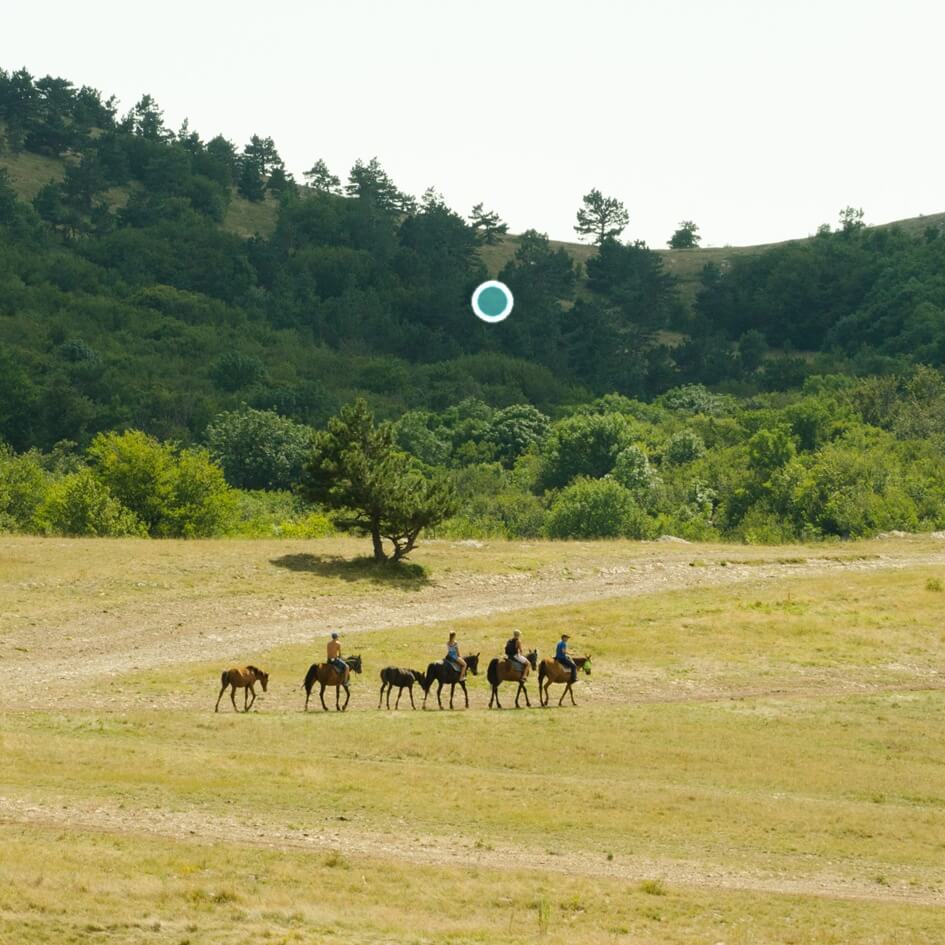}} &
        {\includegraphics[valign=c, width=\ww]{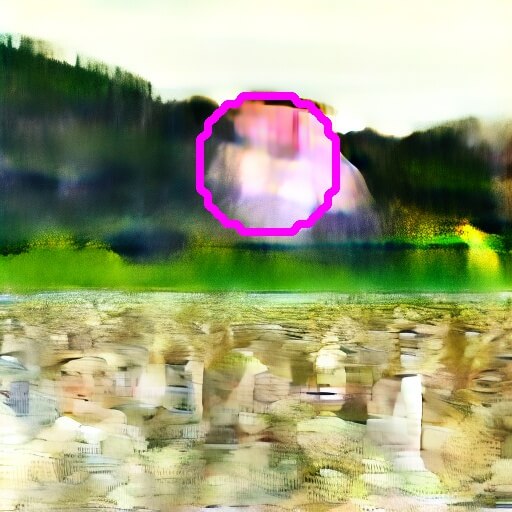}}  &
        {\includegraphics[valign=c, width=\ww]{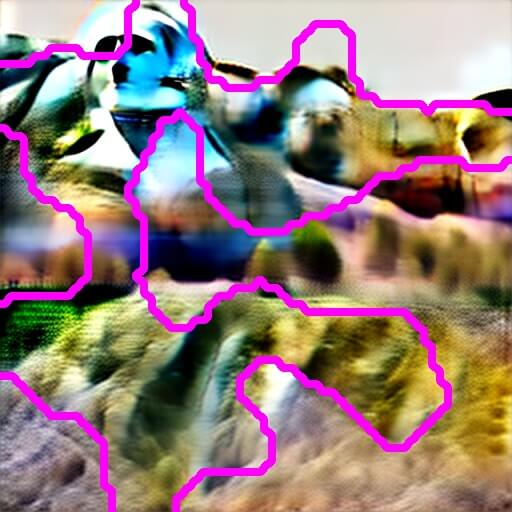}}  &
        {\includegraphics[valign=c, width=\ww]{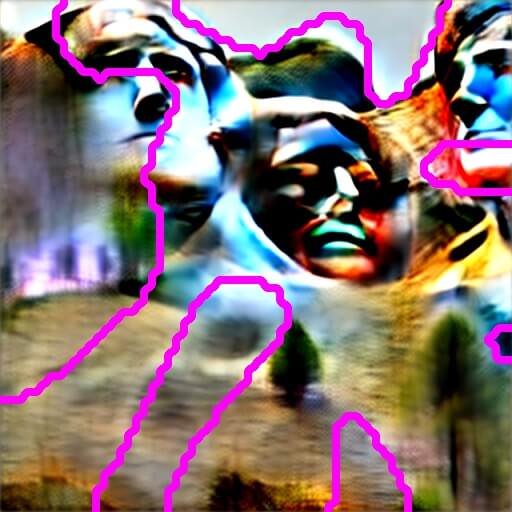}}  &
        {\includegraphics[valign=c, width=\ww]{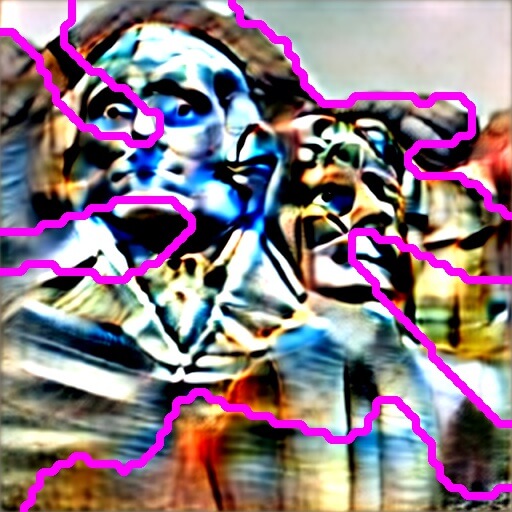}} &
        {\includegraphics[valign=c, width=\ww]{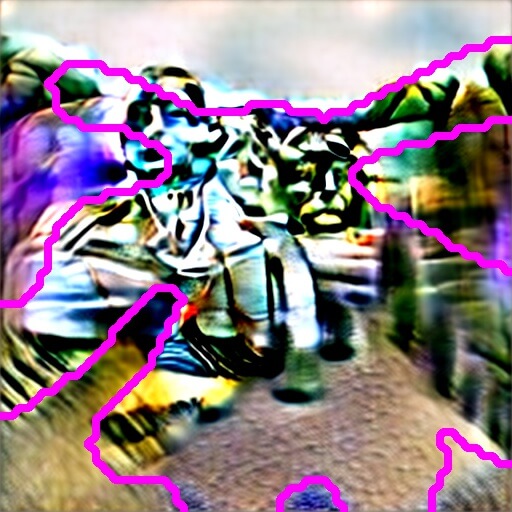}} &
        {\includegraphics[valign=c, width=\ww]{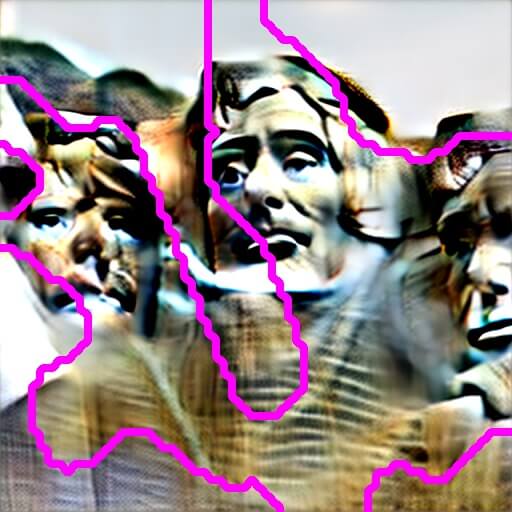}} &
        {\includegraphics[valign=c, width=\ww]{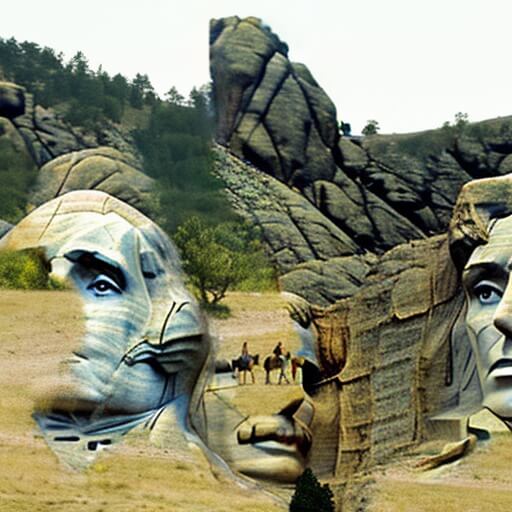}} 
        \\
        [\rsm]
        
        \abltitle{\ablsize}{Ours}
        {\includegraphics[valign=c, width=\ww]{figures/abl_elev_all/elev_all/rushmore_in_click.jpg}} &
        {\includegraphics[valign=c, width=\ww]{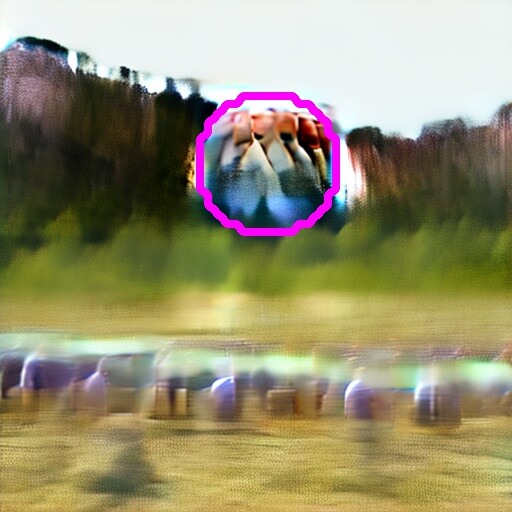}}  &
        {\includegraphics[valign=c, width=\ww]{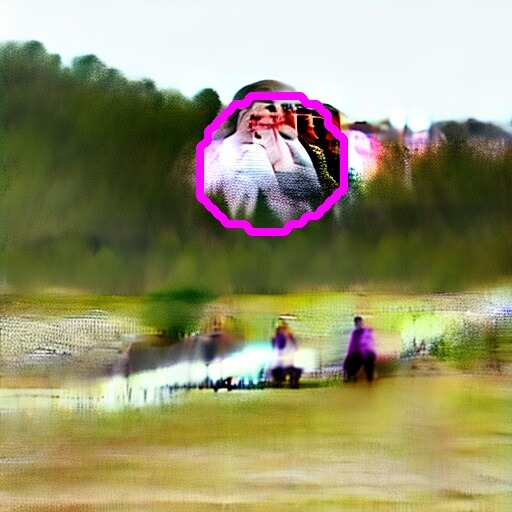}}  &
        {\includegraphics[valign=c, width=\ww]{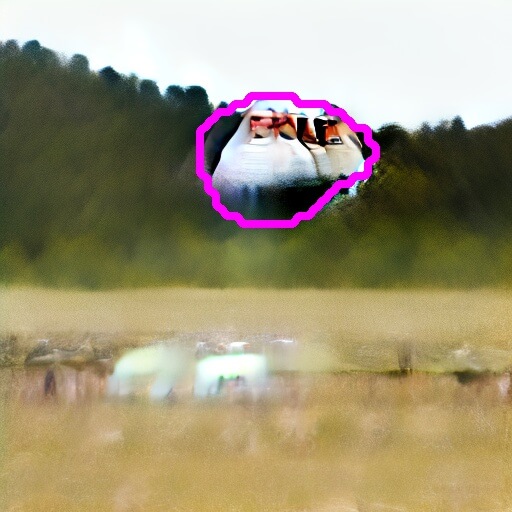}}  &
        {\includegraphics[valign=c, width=\ww]{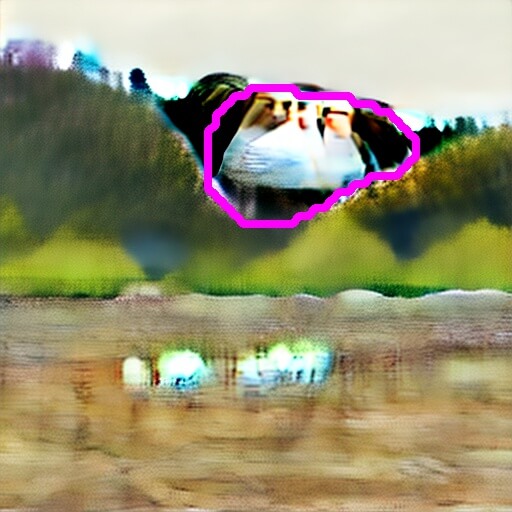}}  &
        {\includegraphics[valign=c, width=\ww]{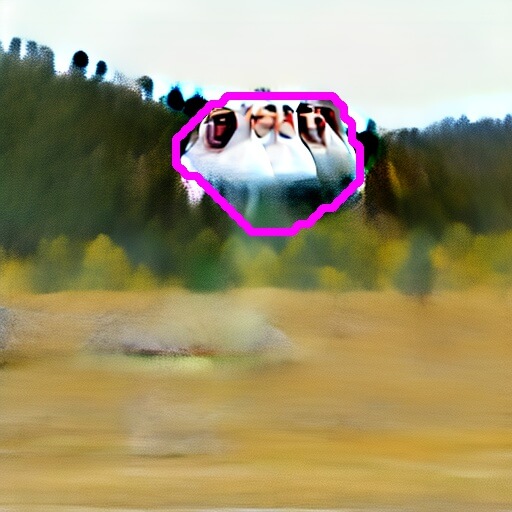}} &
        {\includegraphics[valign=c, width=\ww]{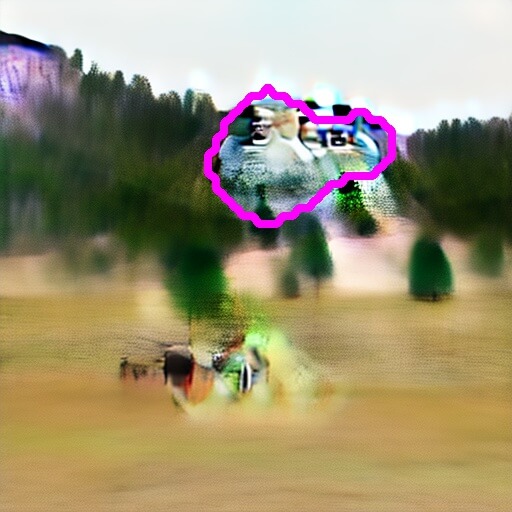}} &
        {\includegraphics[valign=c, width=\ww]{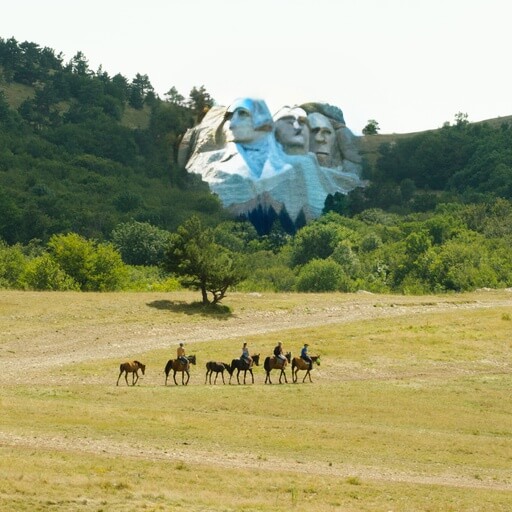}} 
        \\
        [\rsb]
        
        \sizedtext{\methsize}{Input}& \sizedtext{\methsize}{40\%}& \sizedtext{\methsize}{41\%}& \sizedtext{\methsize}{42\%}&  \sizedtext{\methsize}{44\%}& 
        \sizedtext{\methsize}{45\%}&
        \sizedtext{\methsize}{46\% (final $\mask$)}&
        \sizedtext{\methsize}{Output}
        \\
        
    \end{tabular}

    \caption{\textbf{Ablation study: elevating $\potential$ on all image. } 
    With the prompt \textit{``Figures from Mount Rushmore"}, the top row depicts mask $\mask$'s evolution (shown by a purple contour over $\est{z\un{fg}}$ throughout the diffusion steps indicated by percentages below) when elevating potential $\potential$ across the entire image. This results in an unstable and unsmooth mask progression, with an output disassociated from the input image. Bottom row depicts $\mask$'s evolution in \ctm, where only the surrounding area of $\mask$'s contour is elevated in $\potential$.}
    \label{fig:ablation_elevate_all}
\end{figure*}

\begin{figure*}[h]
    \centering
    \setlength{\tabcolsep}{0pt}
    \renewcommand{\arraystretch}{0.5}
    \setlength{\ww}{0.122\linewidth}
    \renewcommand{\ablsize}{footnotesize}
    \renewcommand{\rsm}{1.2cm}
    \renewcommand{\rsb}{1.1cm}
    \renewcommand{\methsize}{footnotesize}
    
    \begin{tabular}{c @{\hspace{0.005\columnwidth}}c ccccccc}  
        \abltitle{\ablsize}{No Rerun}
        {\includegraphics[valign=c, width=\ww]{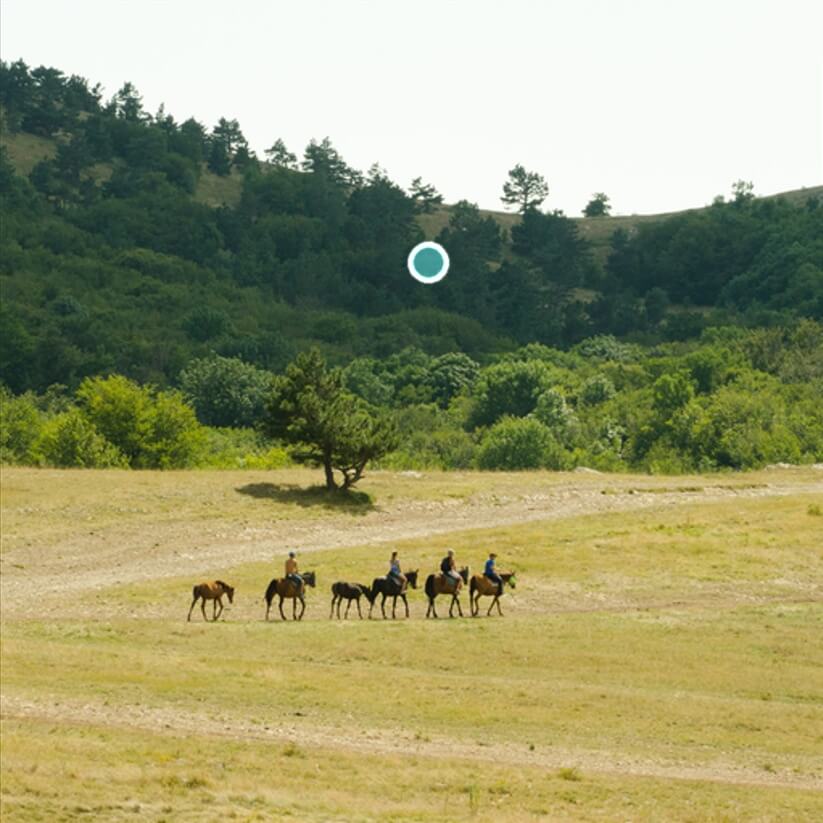}}&
        {\includegraphics[valign=c, width=\ww]{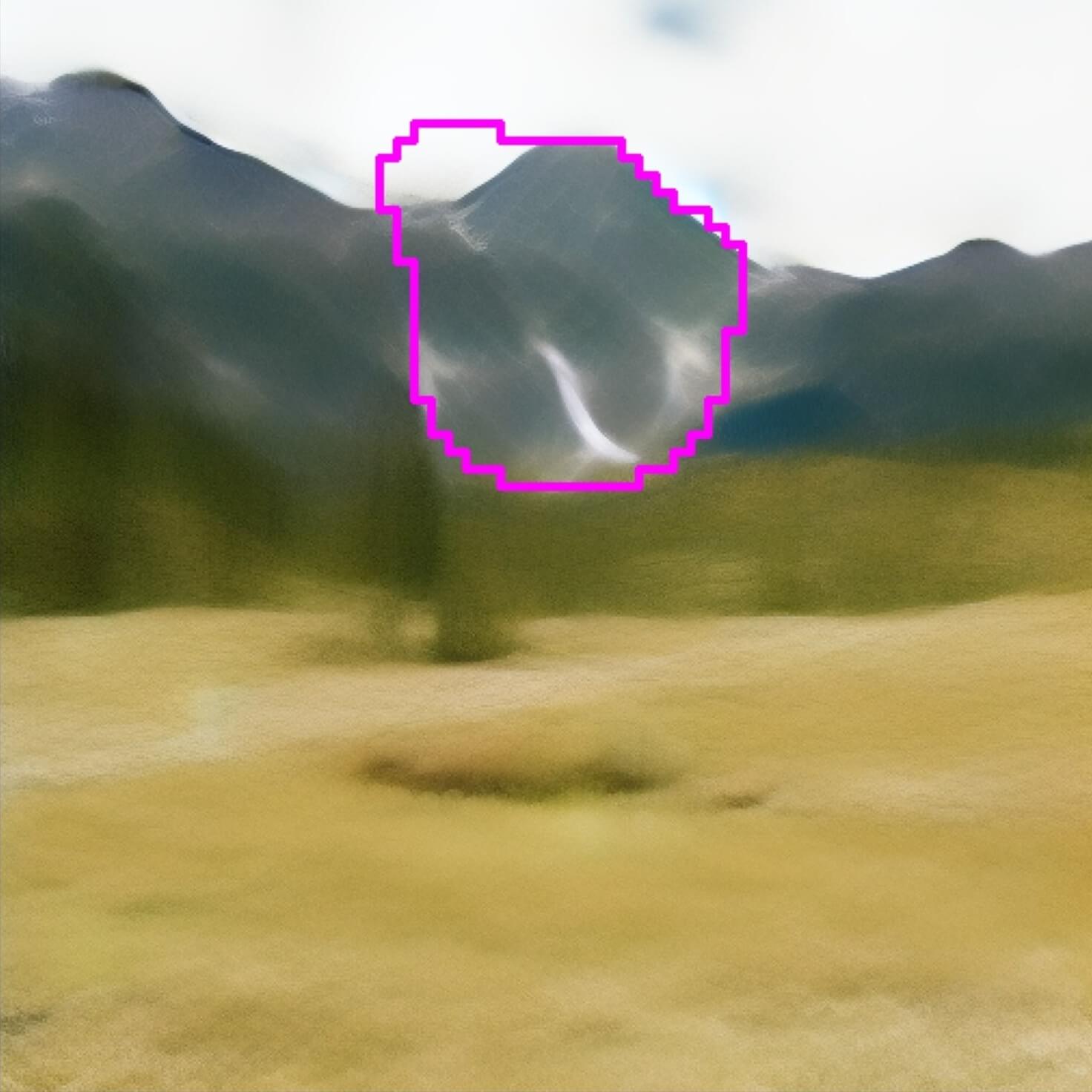}}&
        {\includegraphics[valign=c, width=\ww]{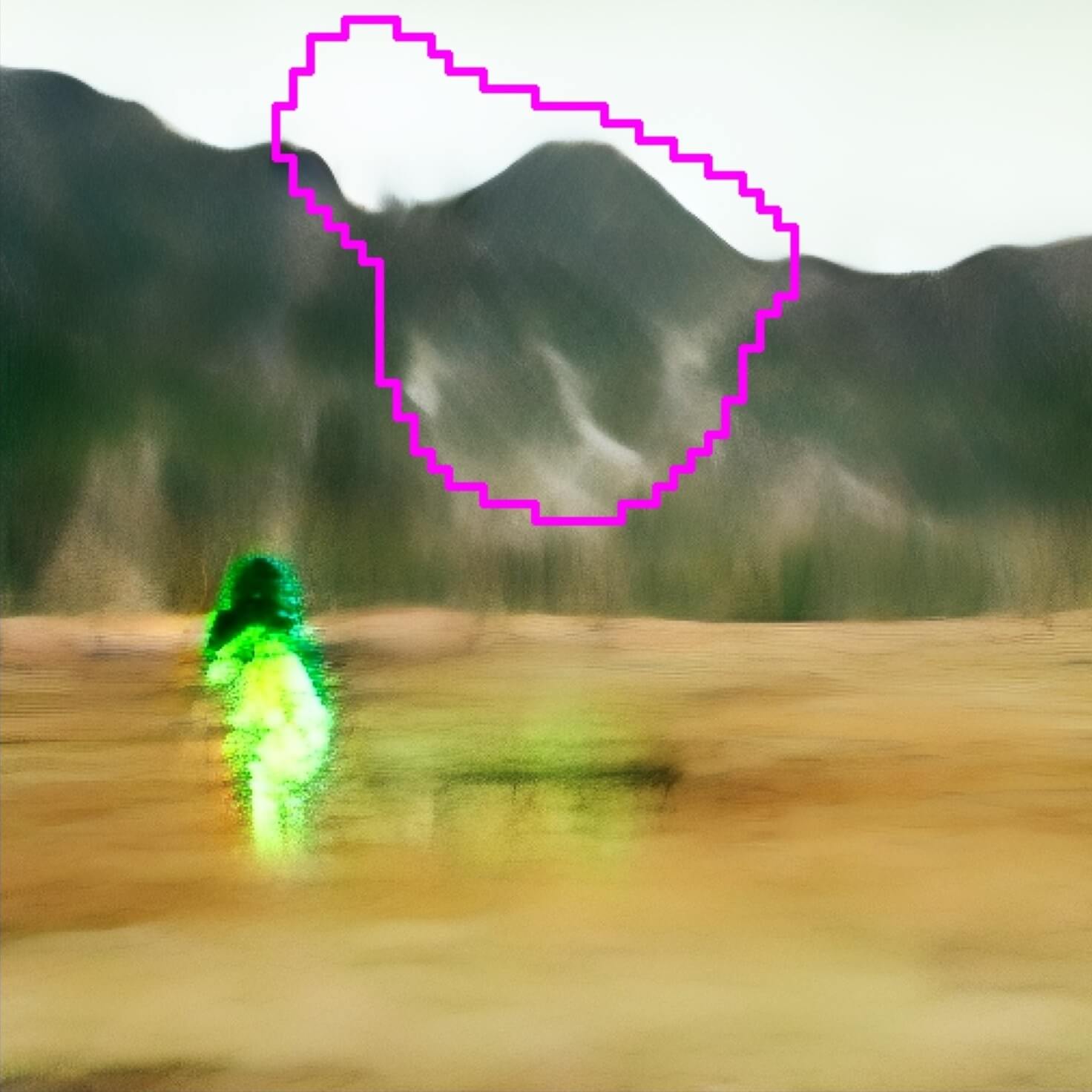}}&
        {\includegraphics[valign=c, width=\ww]{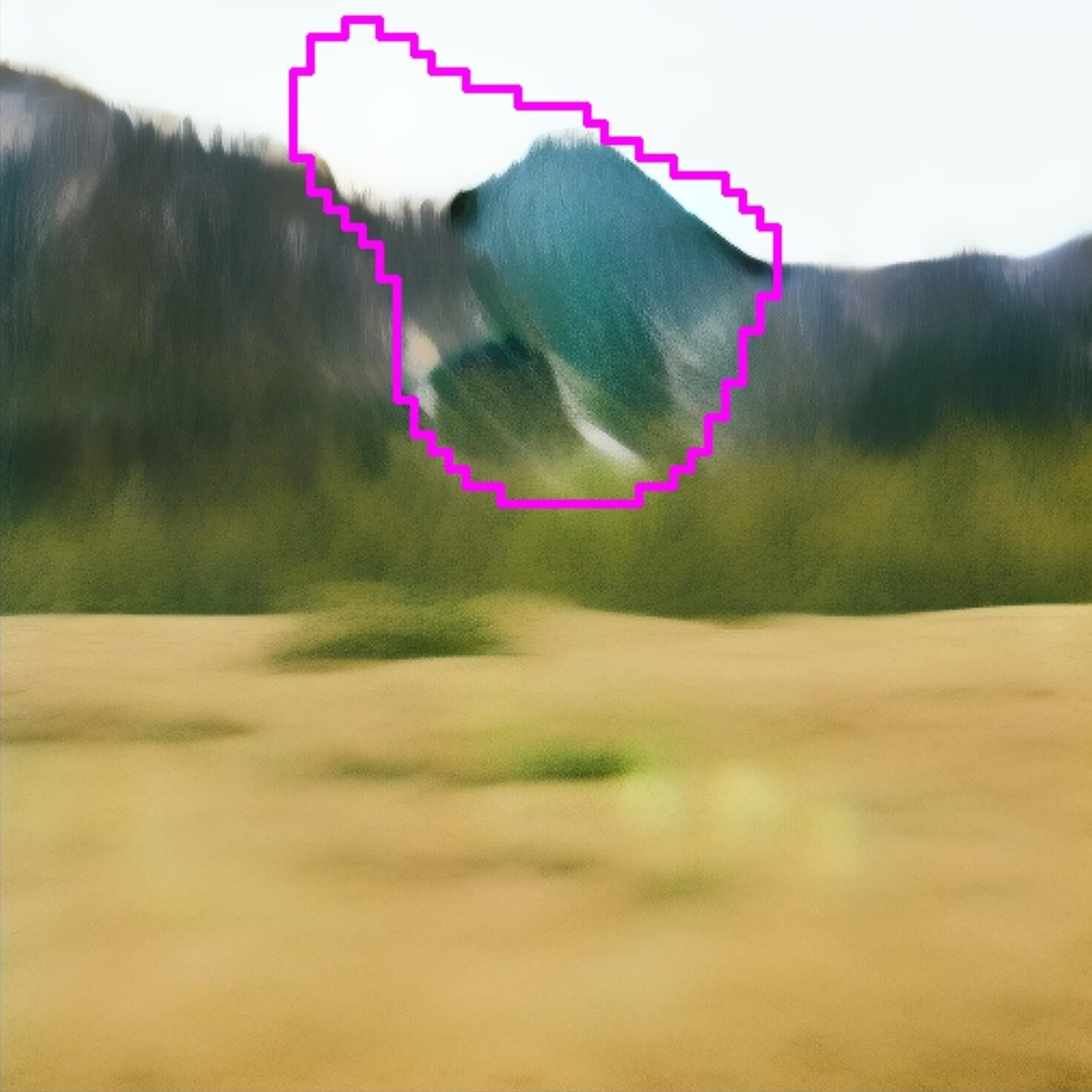}}&
        {\includegraphics[valign=c, width=\ww]{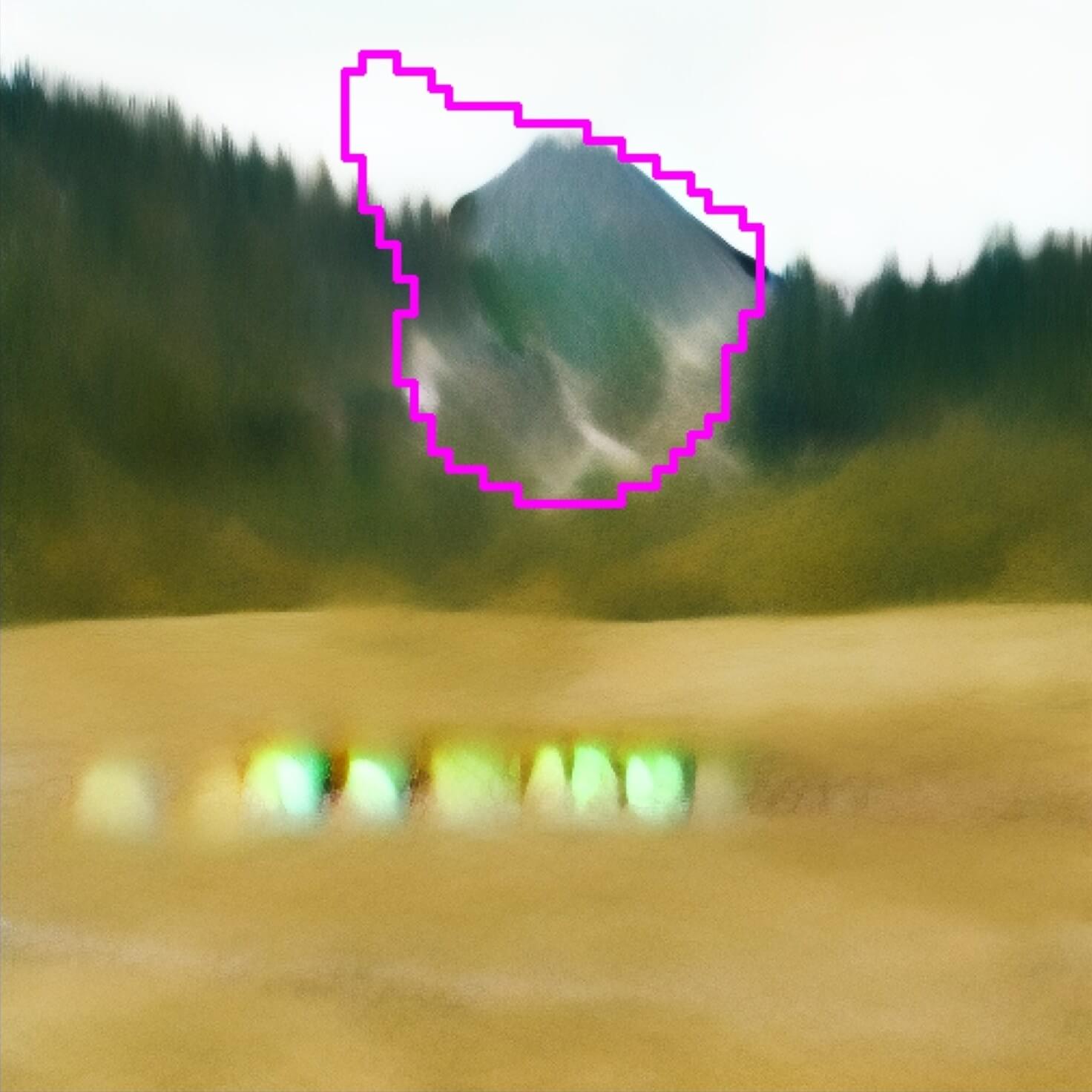}}&
        {\includegraphics[valign=c, width=\ww]{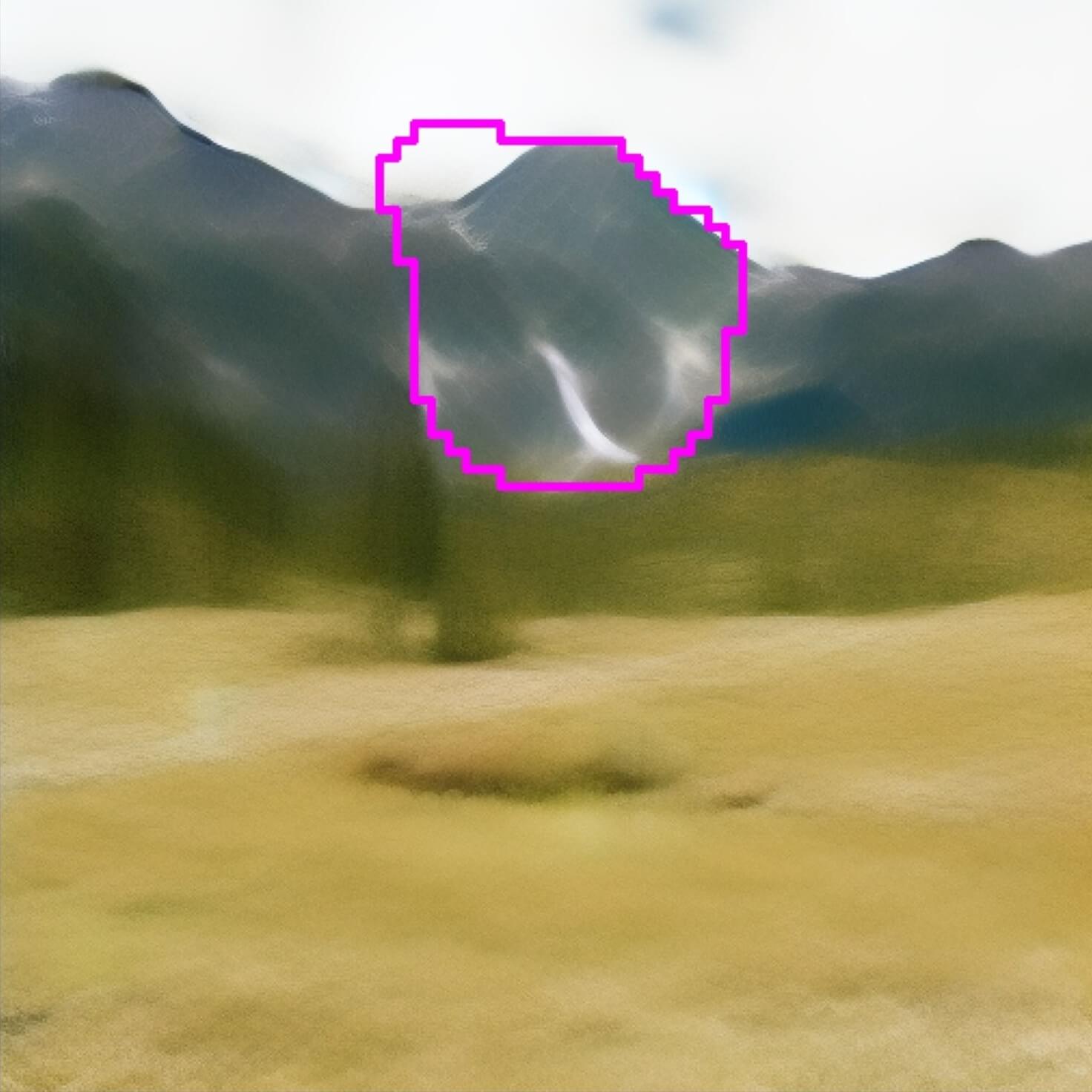}}&
        {\includegraphics[valign=c, width=\ww]{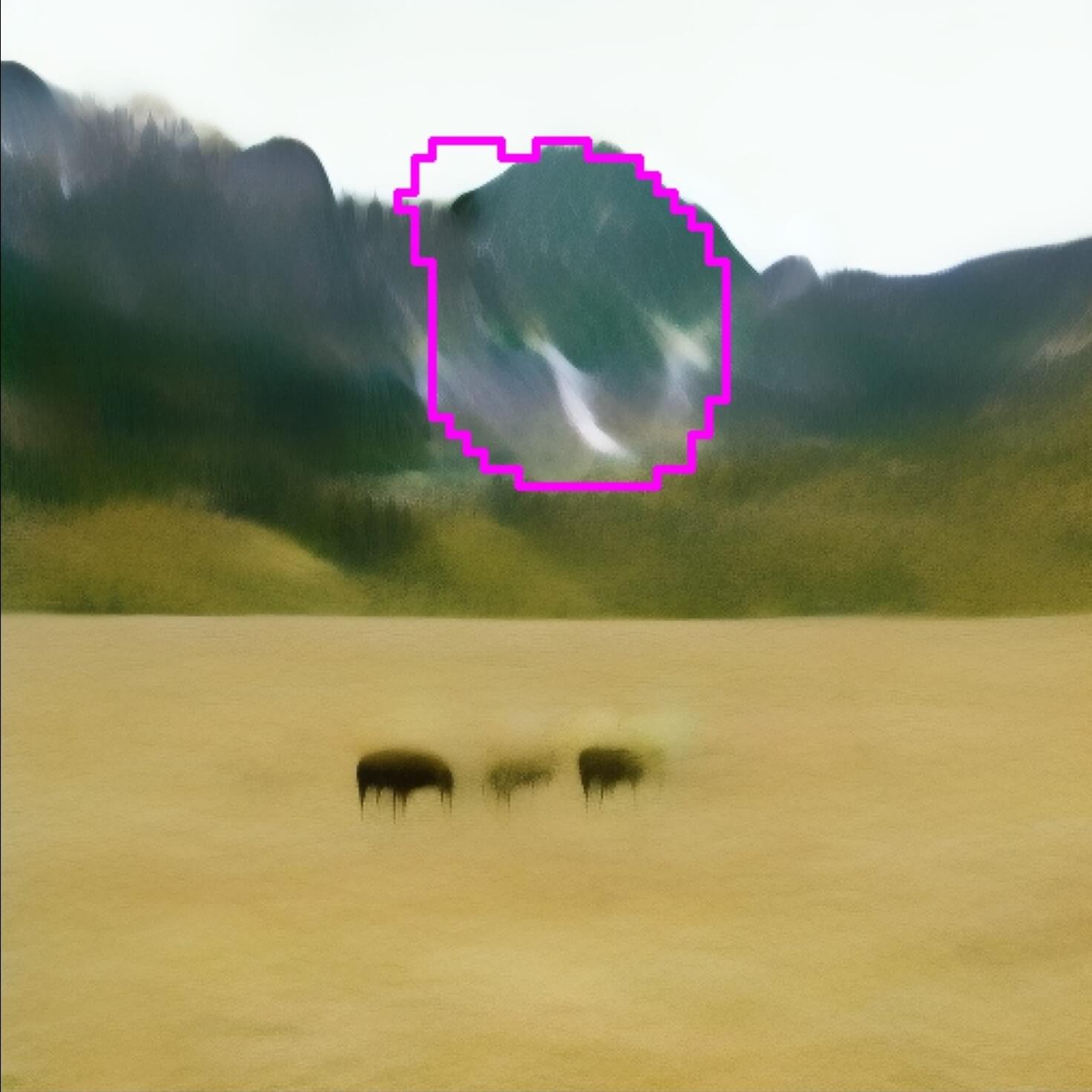}}&
        {\includegraphics[valign=c, width=\ww]{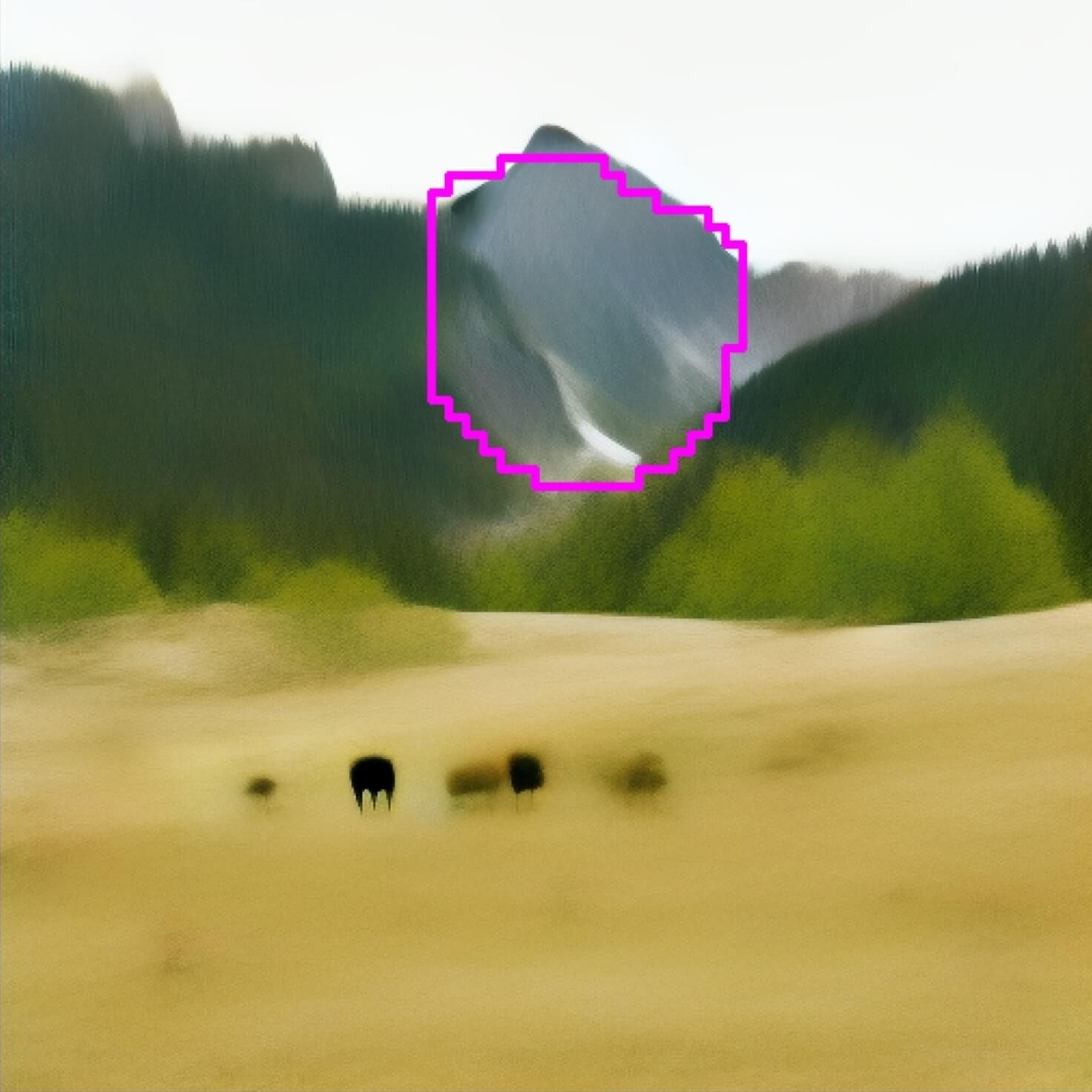}}
        \\
        [\rsm]

        \abltitle{\ablsize}{With Rerun}
        {\includegraphics[valign=c, width=\ww]{figures/abl_rerun/snow_in_click.jpg}}&
        {\includegraphics[valign=c, width=\ww]{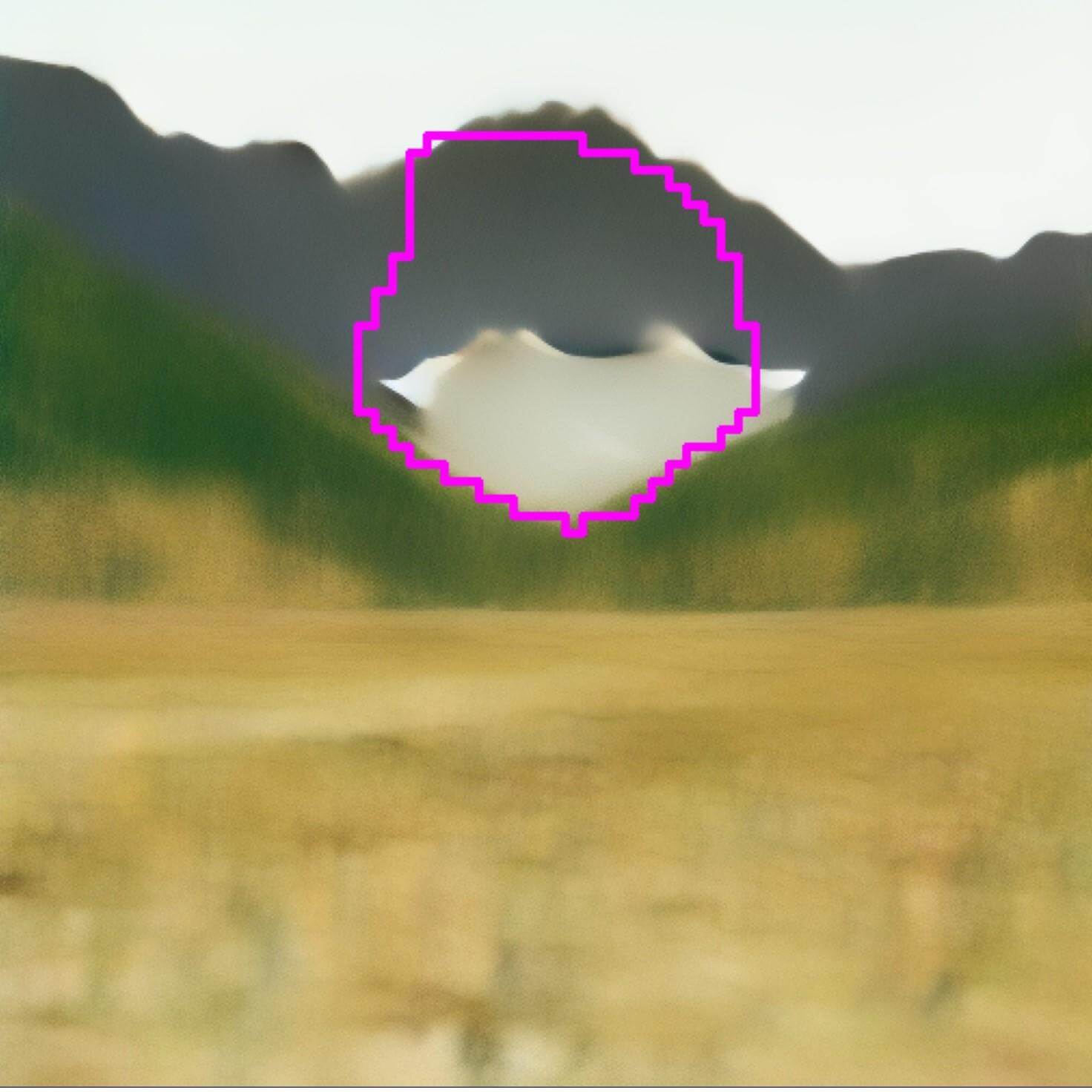}}&
        {\includegraphics[valign=c, width=\ww]{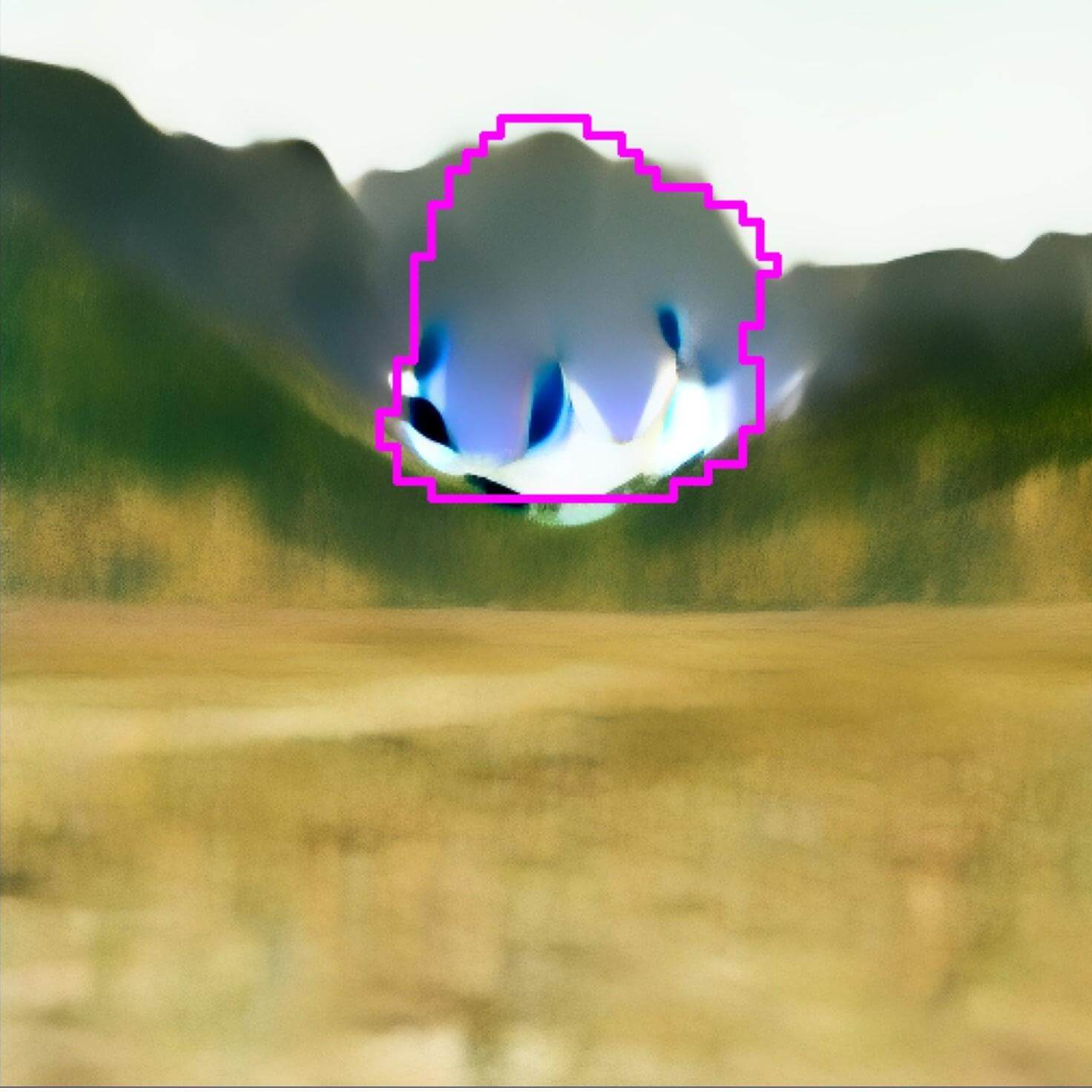}}&
        {\includegraphics[valign=c, width=\ww]{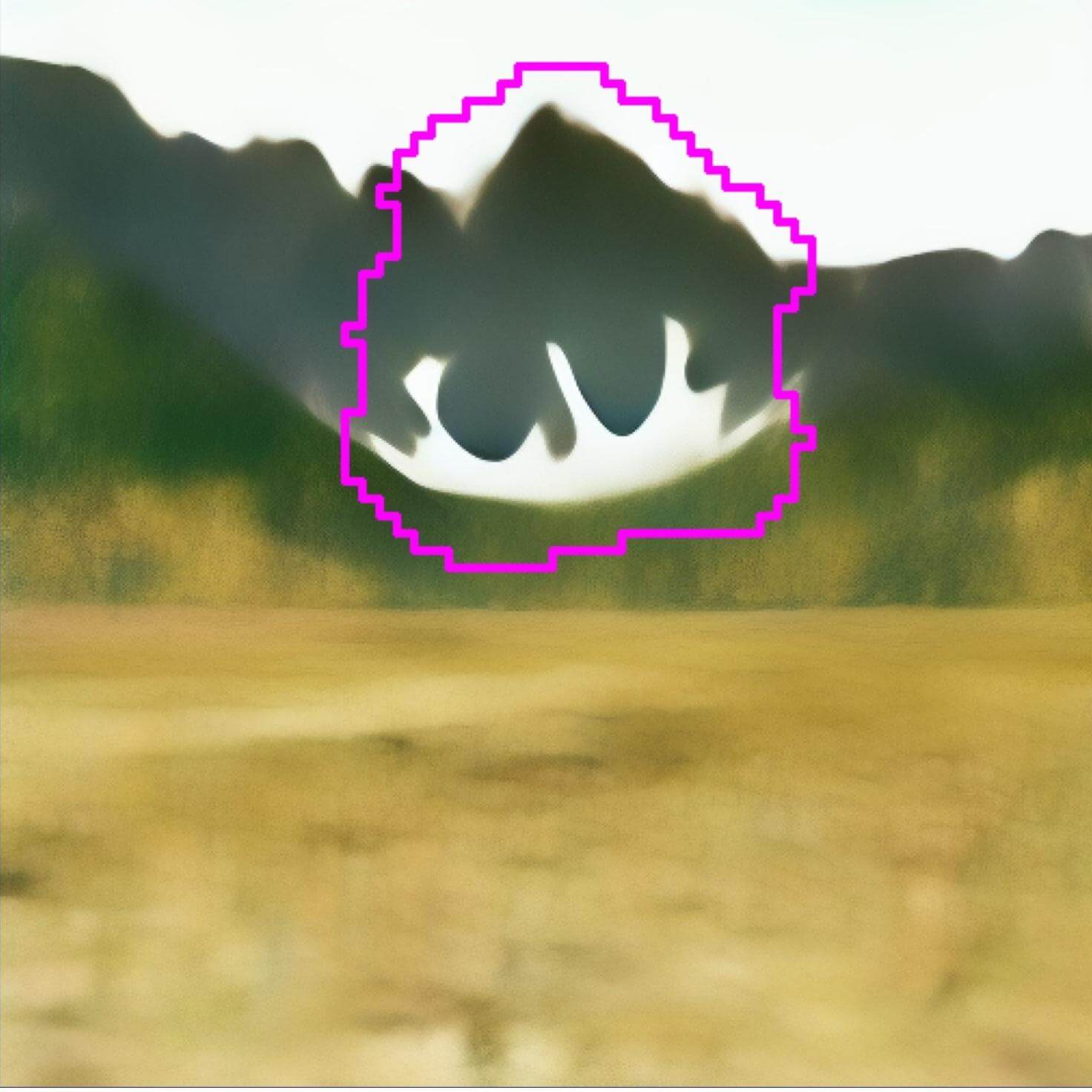}}&
        {\includegraphics[valign=c, width=\ww]{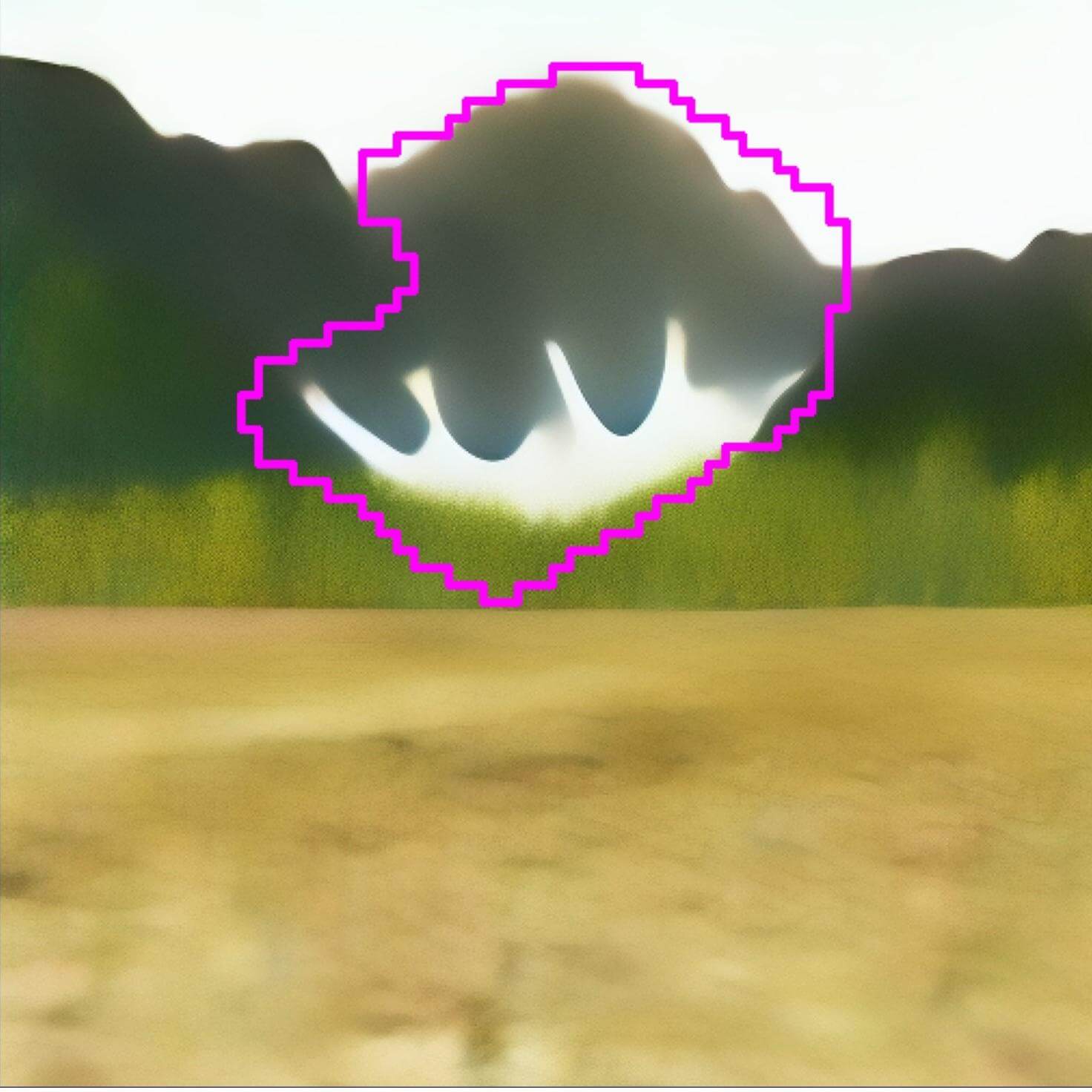}}&
        {\includegraphics[valign=c, width=\ww]{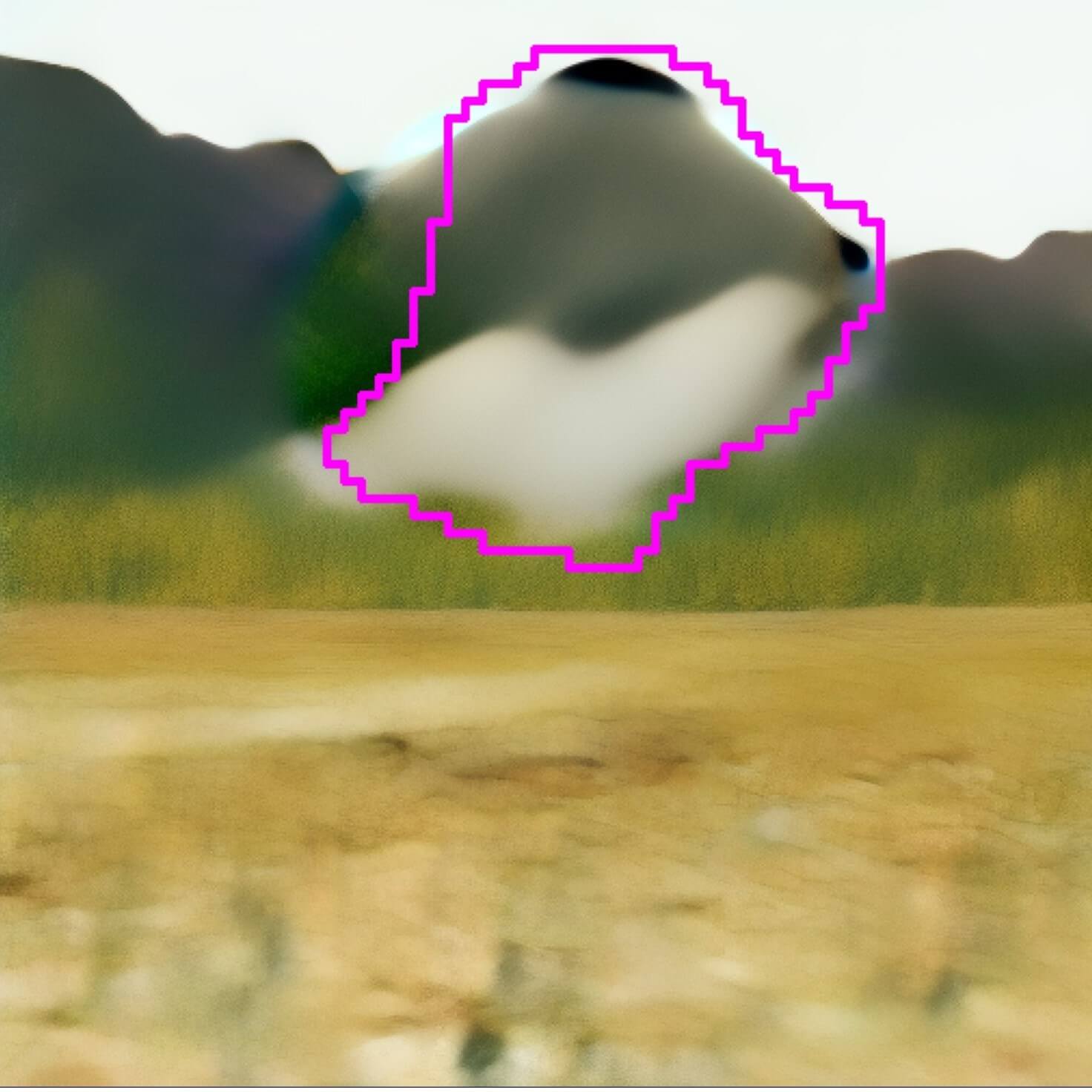}}&
        {\includegraphics[valign=c, width=\ww]{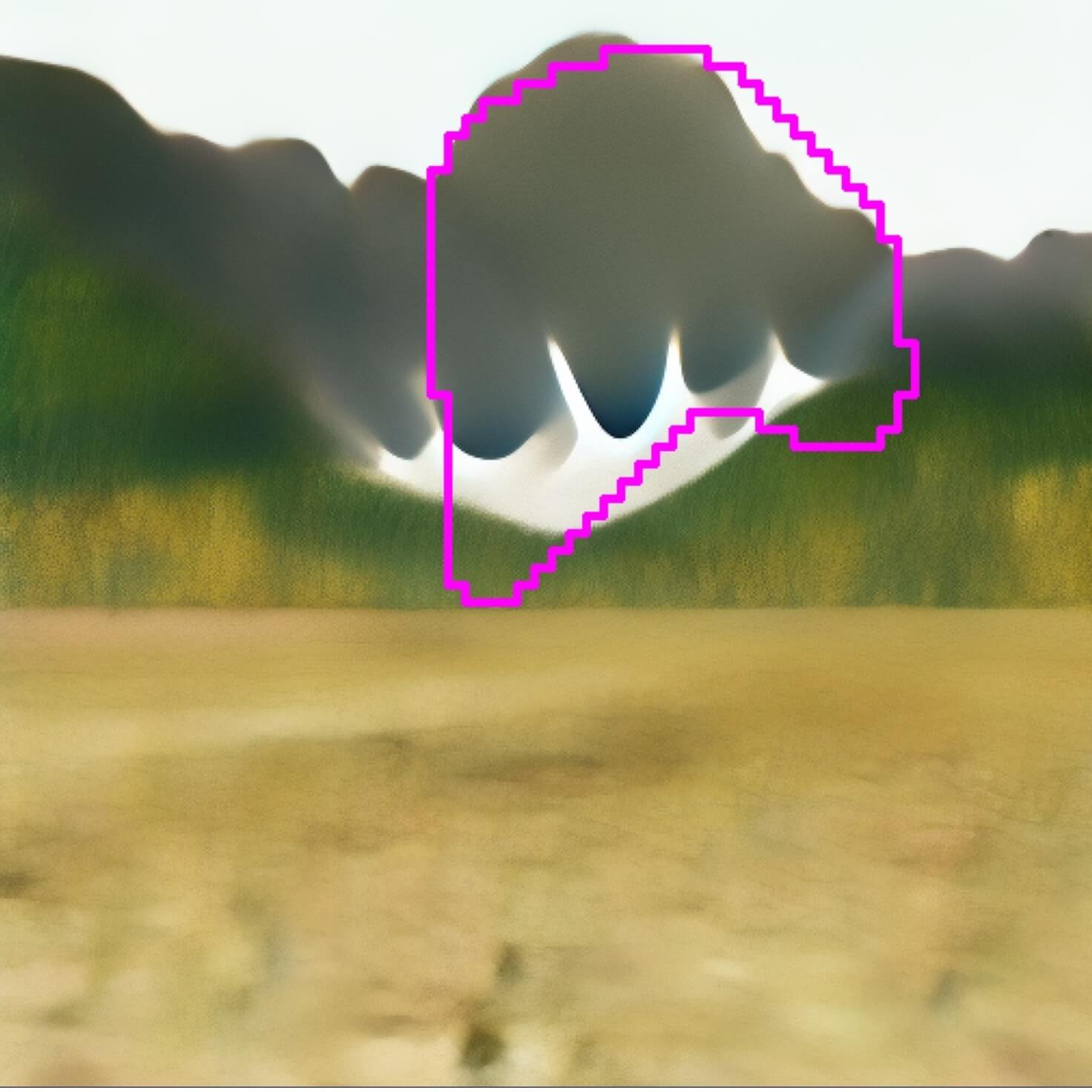}}&
        {\includegraphics[valign=c, width=\ww]{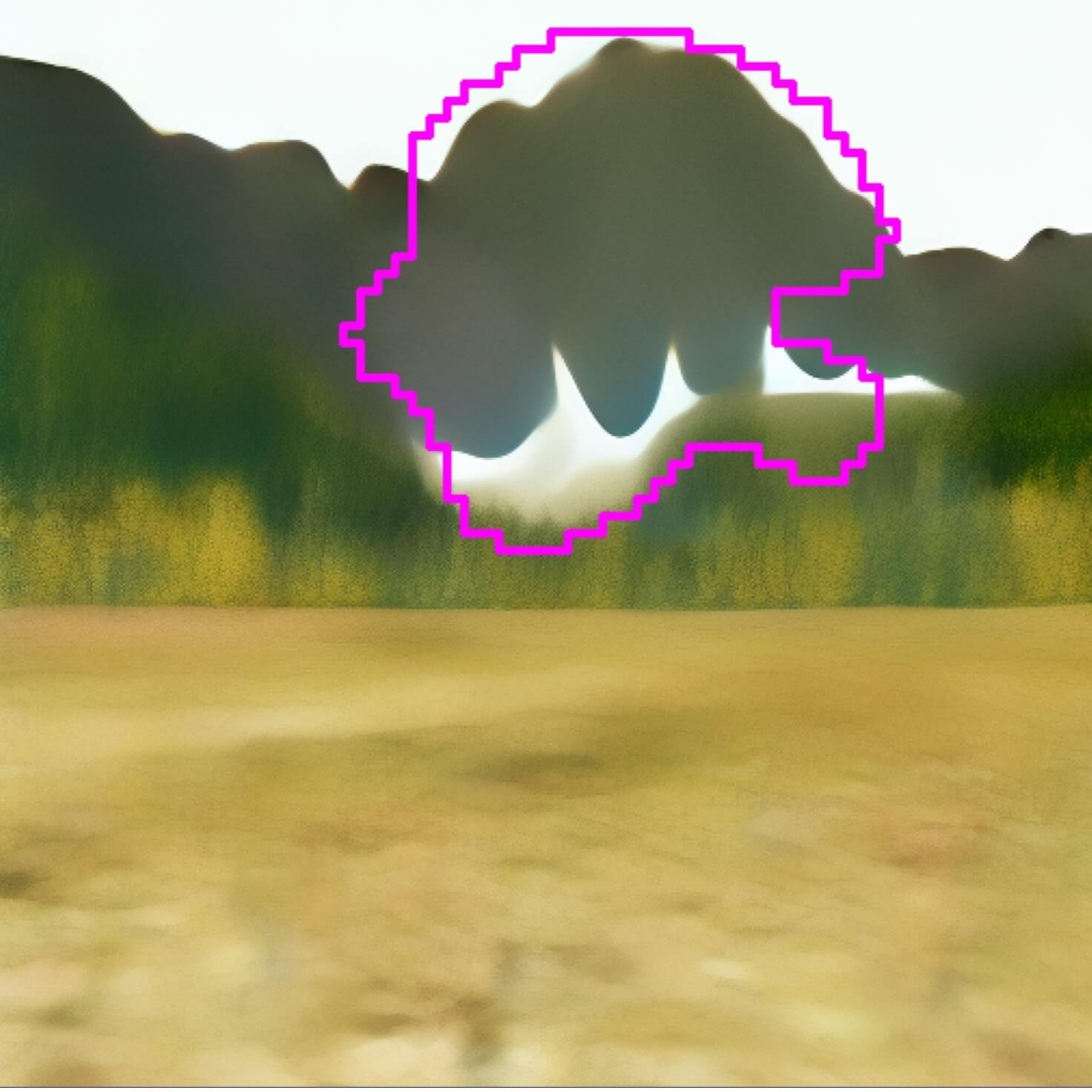}}
        \\
        [\rsb]

        \sizedtext{\methsize}{Input}& \sizedtext{\methsize}{42\%}& \sizedtext{\methsize}{43\%}&  \sizedtext{\methsize}{44\%}& \sizedtext{\methsize}{45\%}& \sizedtext{\methsize}{46\%}& 
        \sizedtext{\methsize}{47\%}& \sizedtext{\methsize}{48\%}

    \end{tabular}

    \caption{\textbf{Rerun ablation study.} The figure depicts the rerun procedure. The prompt is \textit{``Snowy mountains''}, and the upper row lacks rerun, while the lower low has. In the upper row (where the purple mask contours are marked over $\est{z\un{fg}}$s throughout diffusion steps, as percentages below indicate), pixels that are added to the mask $\mask$ at advanced stages, fail to comply to the guiding prompt, since the spacial information has already been determined. Rerun allows a ``refresh" of the information to be driven towards the guiding prompt.}
    \label{fig:ablation_rerun}
\end{figure*}

\begin{figure*}[h]
    \centering
    \setlength{\tabcolsep}{0pt}
    \renewcommand{\arraystretch}{0.5}
    \setlength{\ww}{0.121\linewidth}
    \renewcommand{\ablsize}{footnotesize}
    \renewcommand{\rsm}{1.2cm}
    \renewcommand{\rsb}{1.1cm}
    \renewcommand{\methsize}{footnotesize}
    
    \begin{tabular}{c @{\hspace{0.005\columnwidth}}c ccccccc}  
        \abltitle{\ablsize}{Predicted $\est{z\un{fg}}$s}
        {\includegraphics[valign=c, width=\ww]{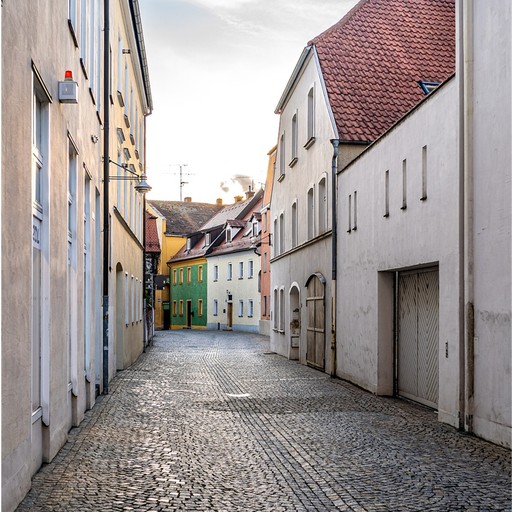}}&
        {\includegraphics[valign=c, width=\ww]{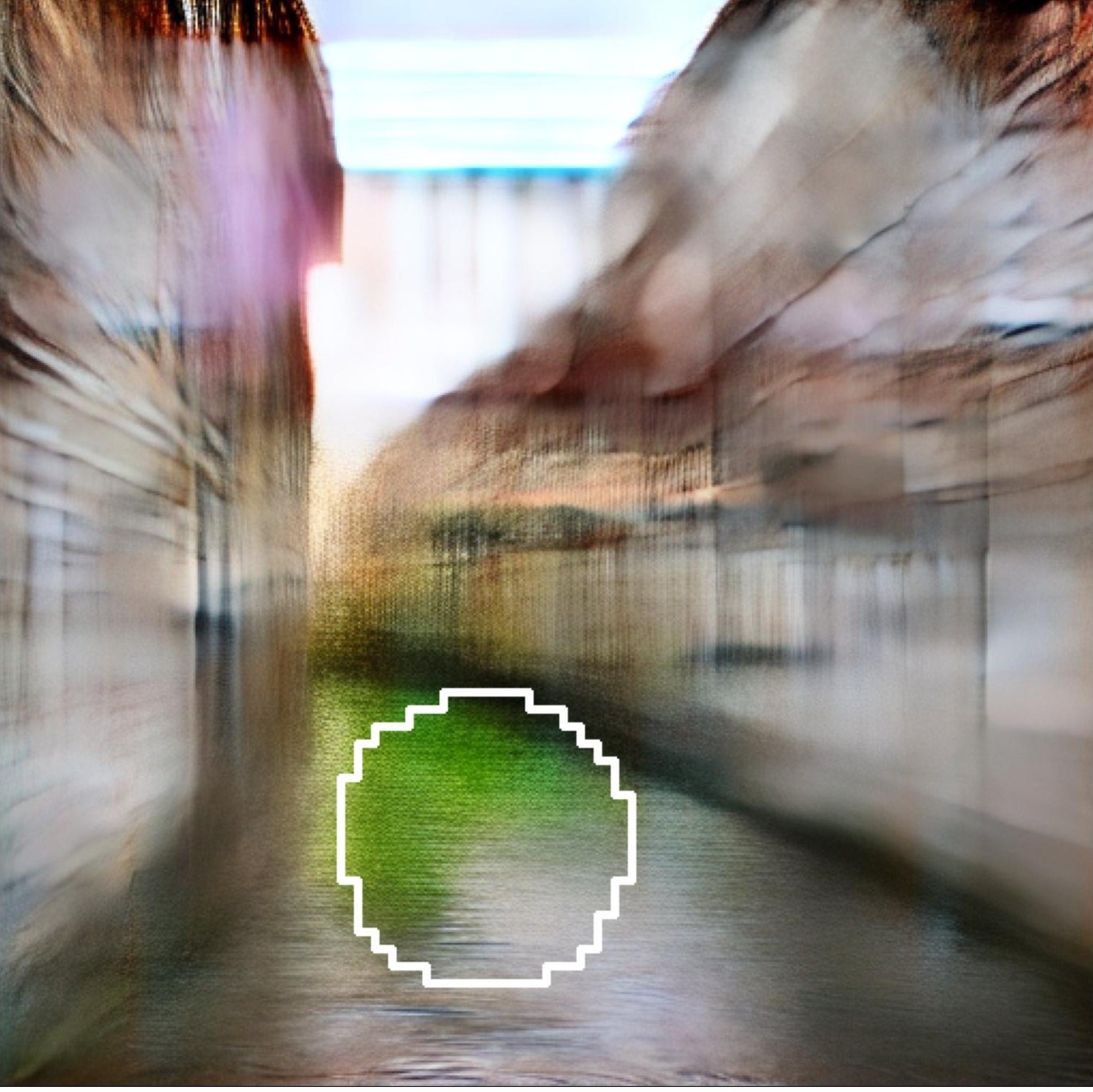}}&
        {\includegraphics[valign=c, width=\ww]{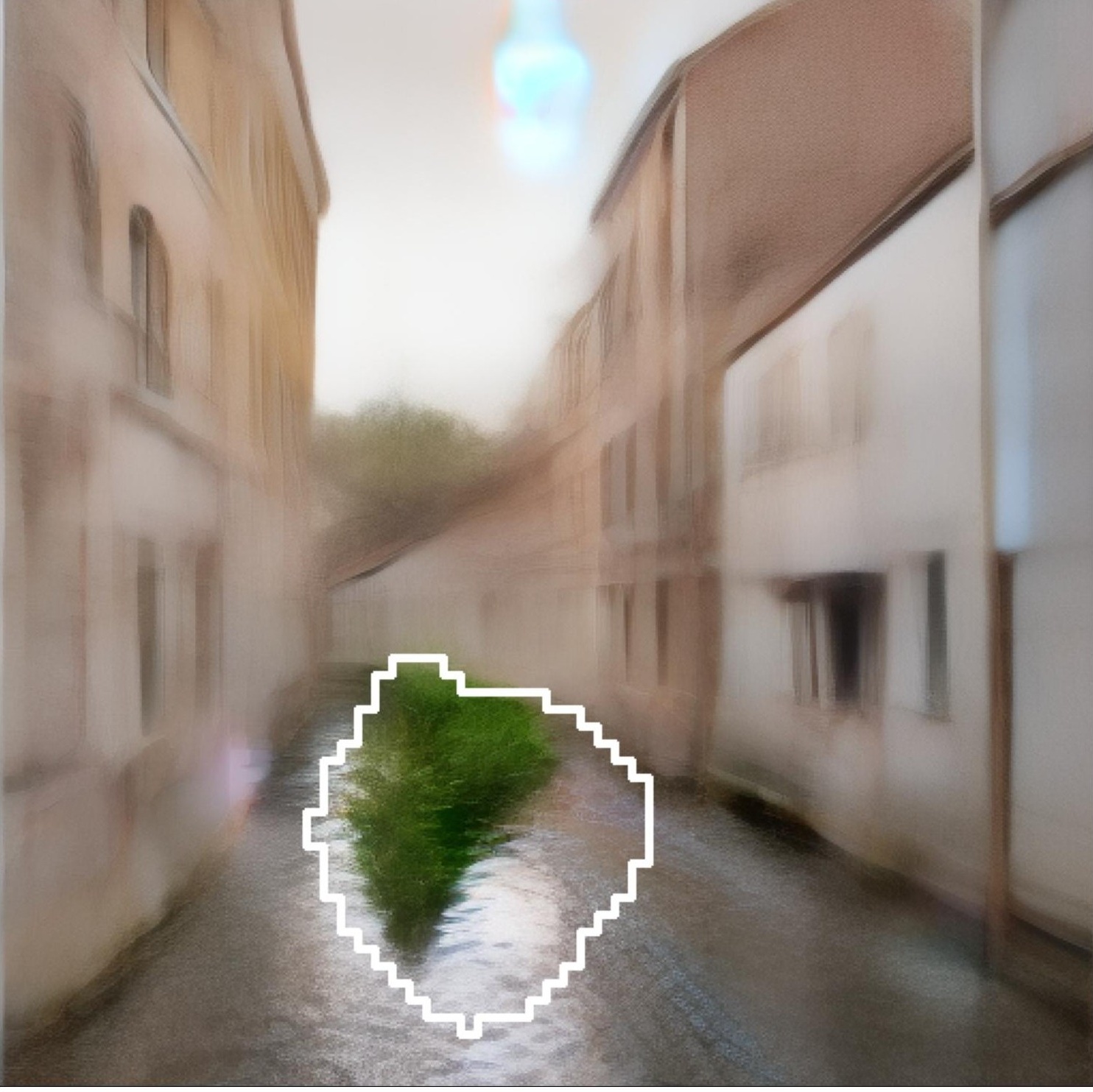}}&
        {\includegraphics[valign=c, width=\ww]{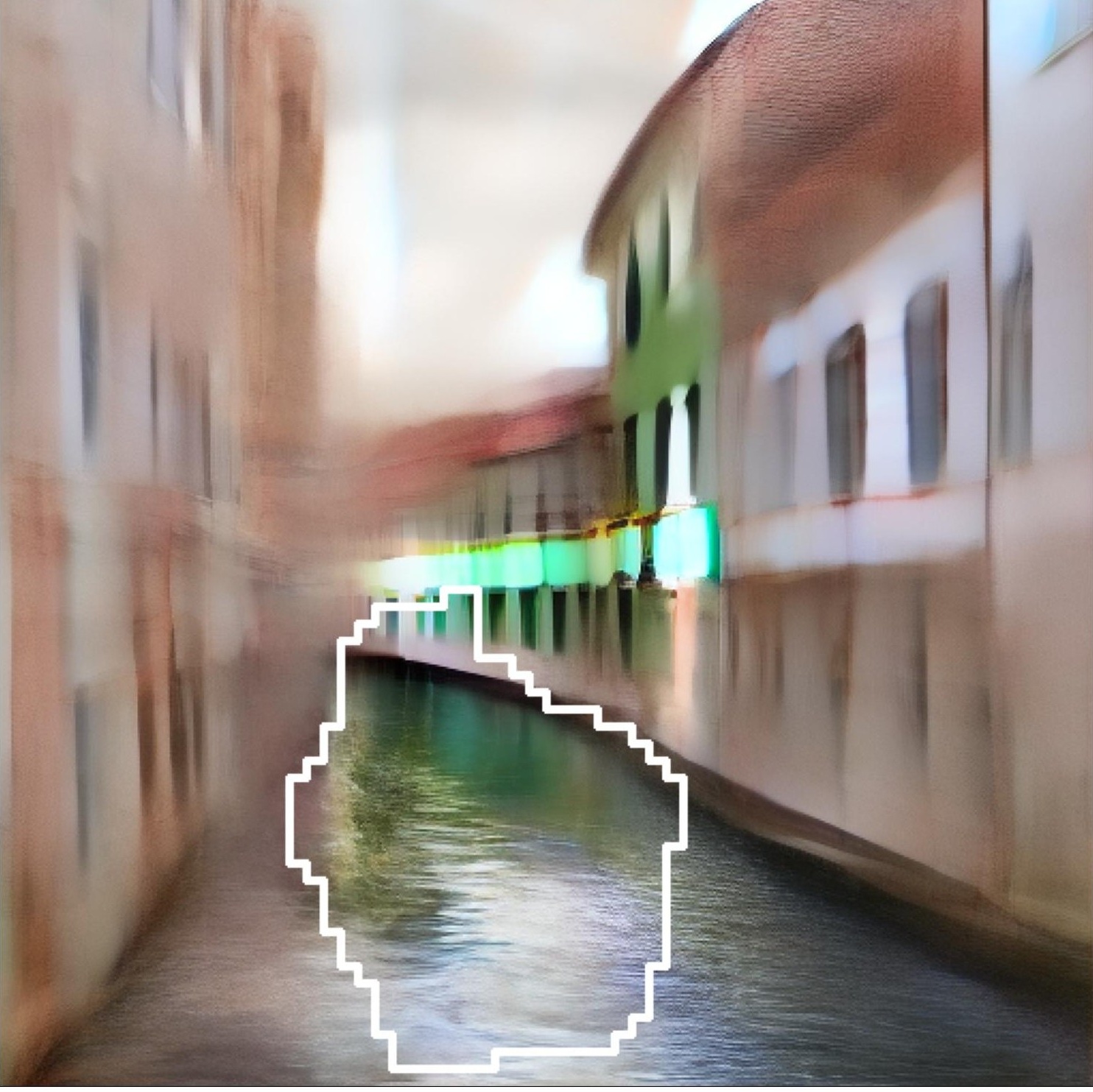}}&
        {\includegraphics[valign=c, width=\ww]{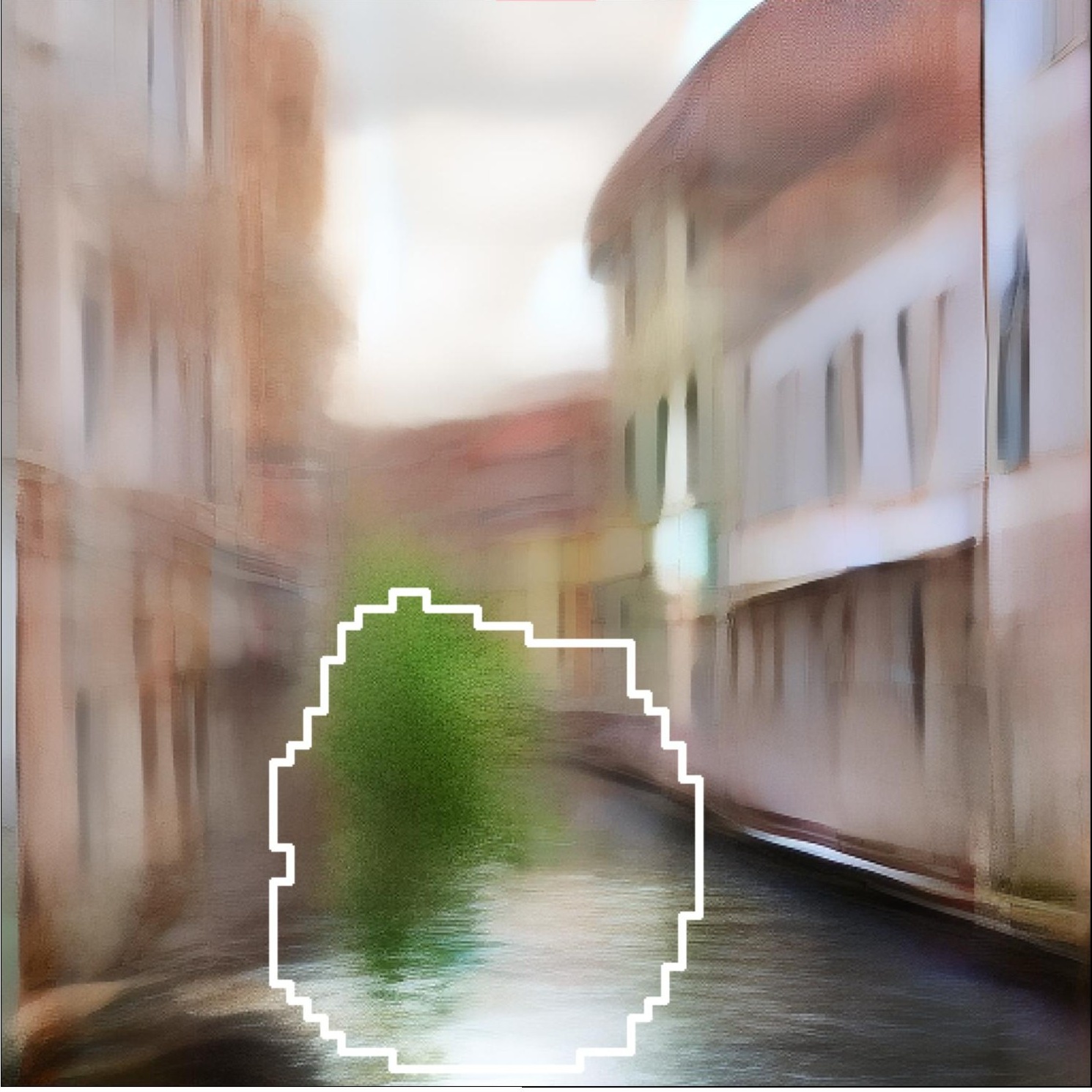}}&
        {\includegraphics[valign=c, width=\ww]{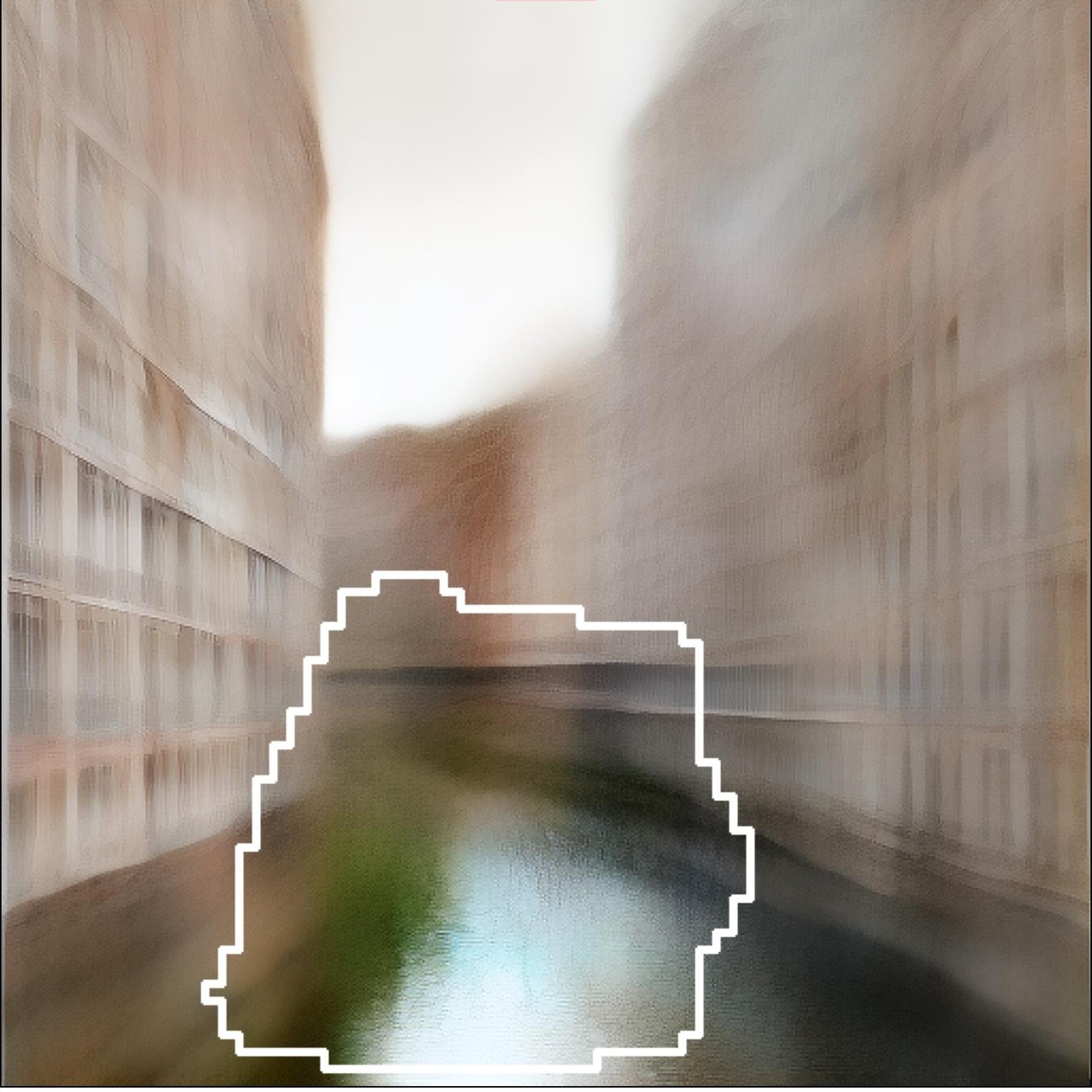}}&
        {\includegraphics[valign=c, width=\ww]{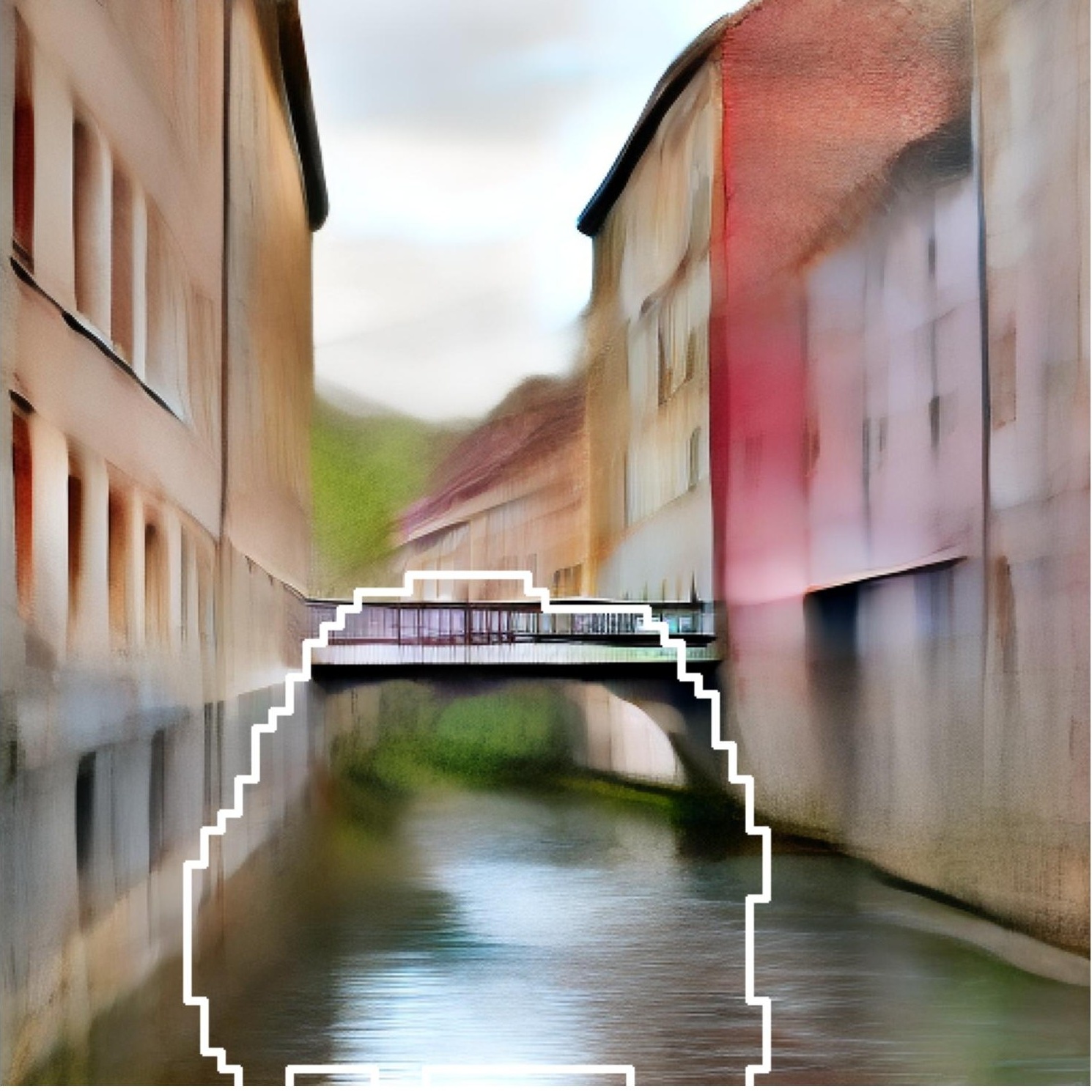}}&
        {\includegraphics[valign=c, width=\ww]{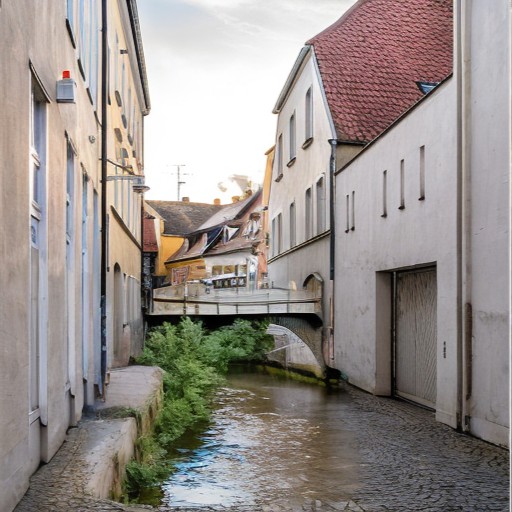}}
        \\
        [\rsm]
        \sizedtext{\methsize}{Input}& & & & & & & \sizedtext{\methsize}{Output} \\

        \abltitle{\ablsize}{Continuous Masks}
        {\includegraphics[valign=c, width=\ww]{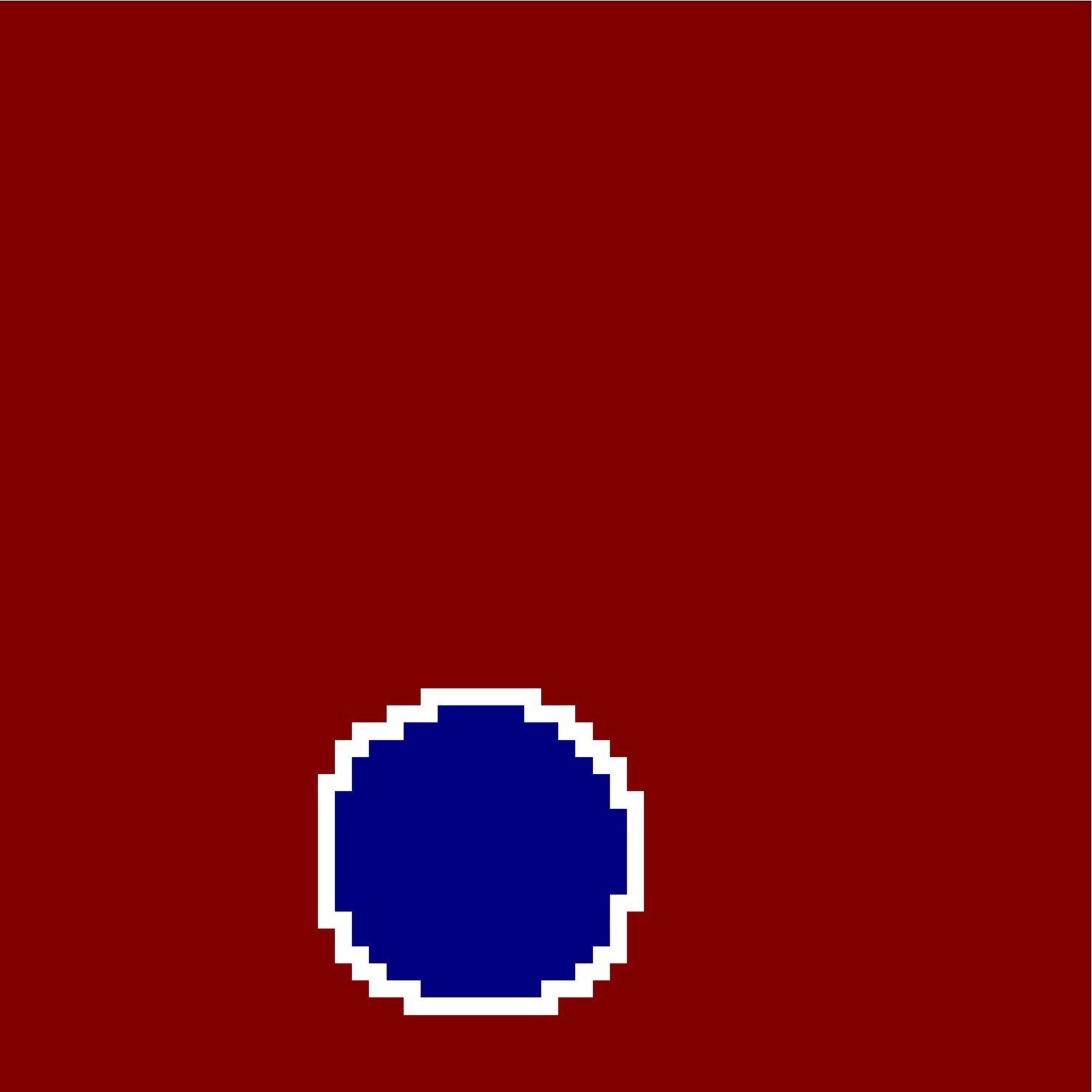}}&
        {\includegraphics[valign=c, width=\ww]{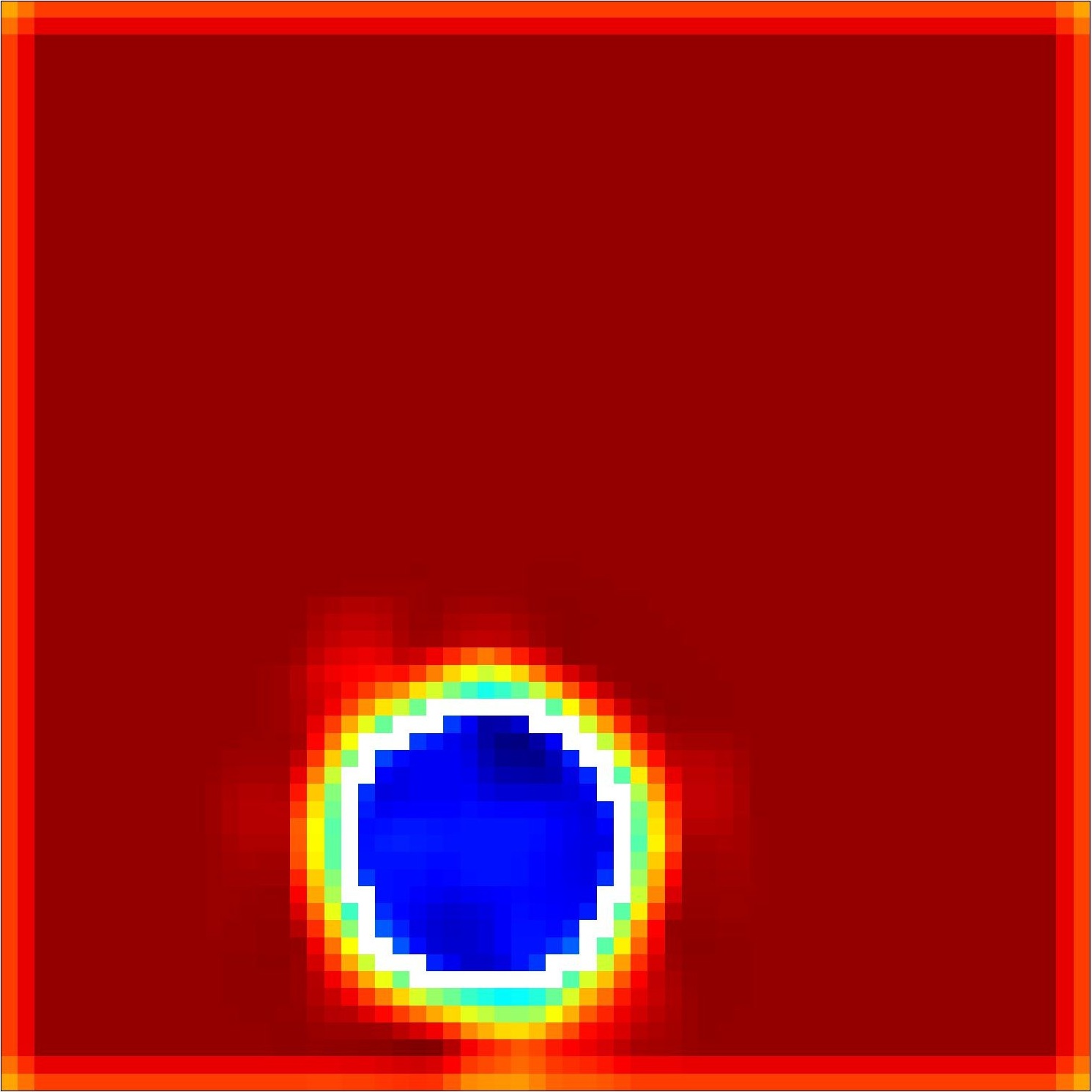}}&
        {\includegraphics[valign=c, width=\ww]{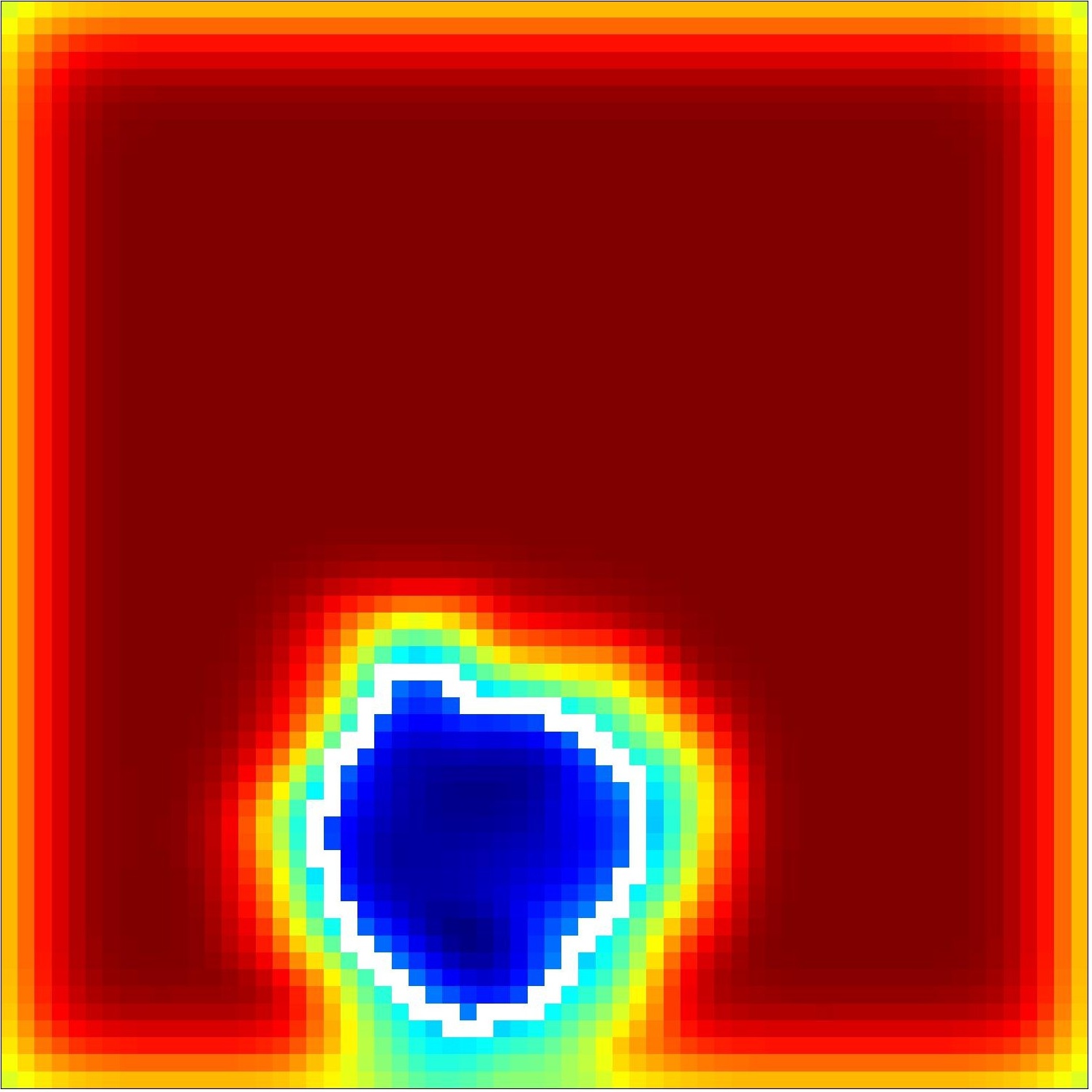}}&
        {\includegraphics[valign=c, width=\ww]{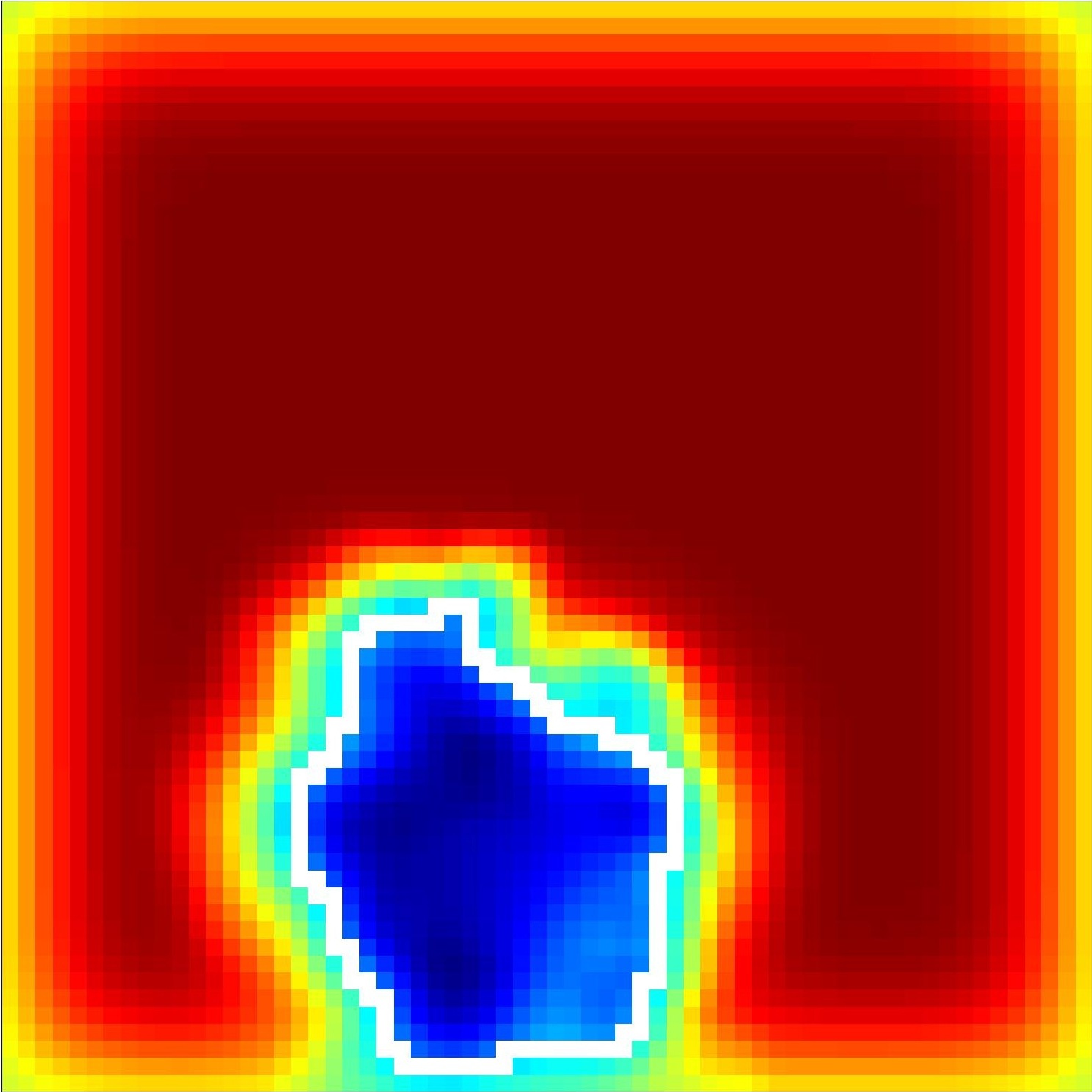}}&
        {\includegraphics[valign=c, width=\ww]{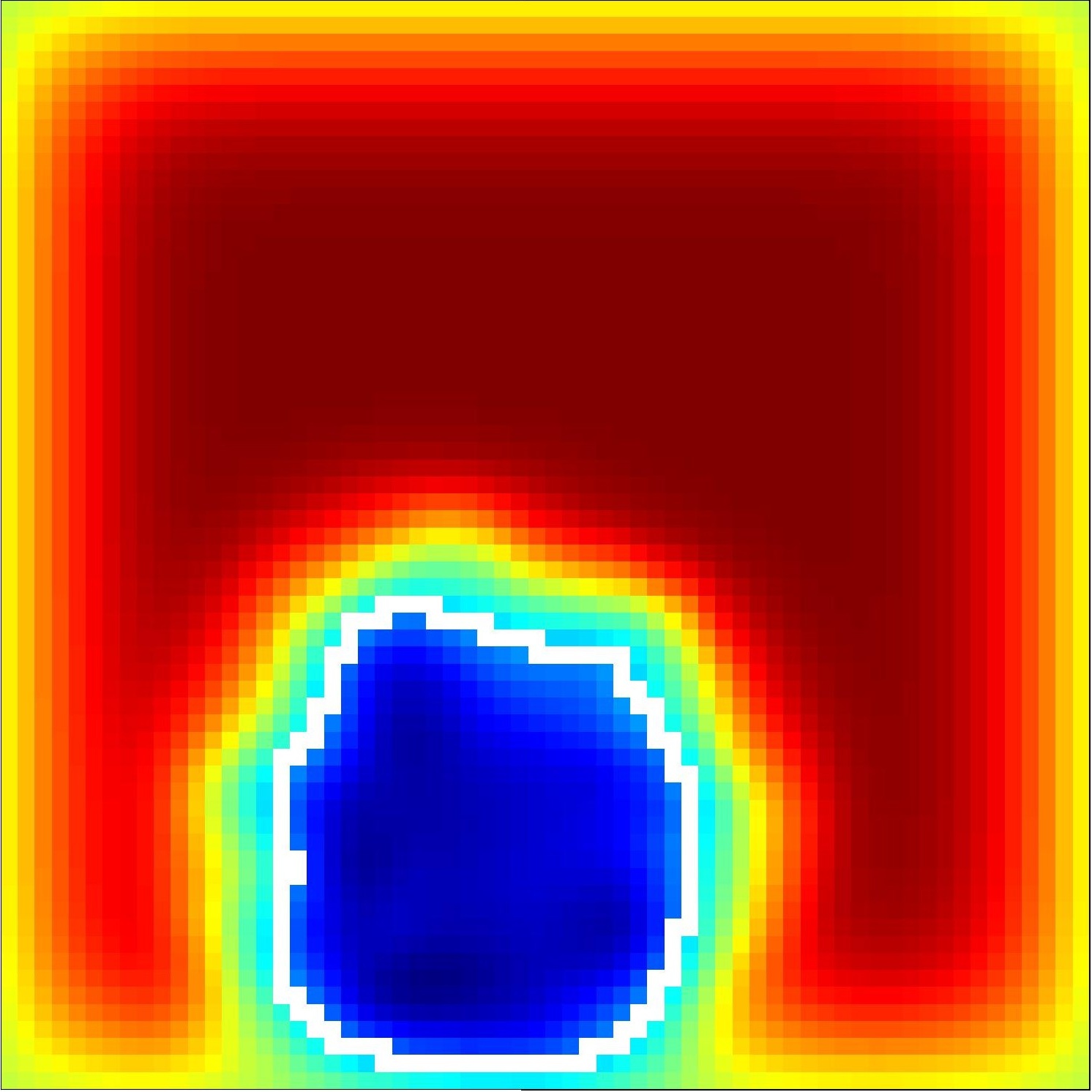}}&
        {\includegraphics[valign=c, width=\ww]{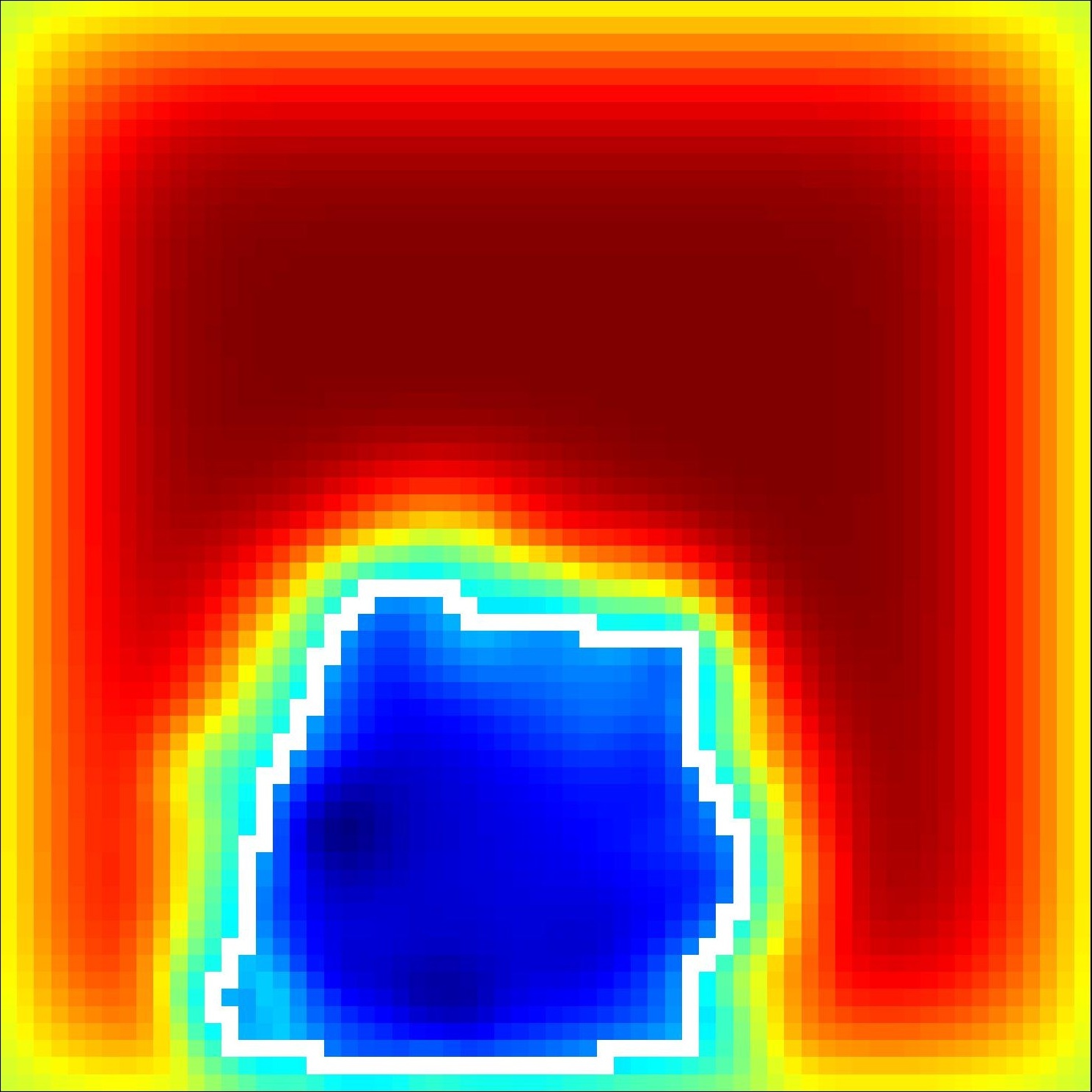}}&
        {\includegraphics[valign=c, width=\ww]{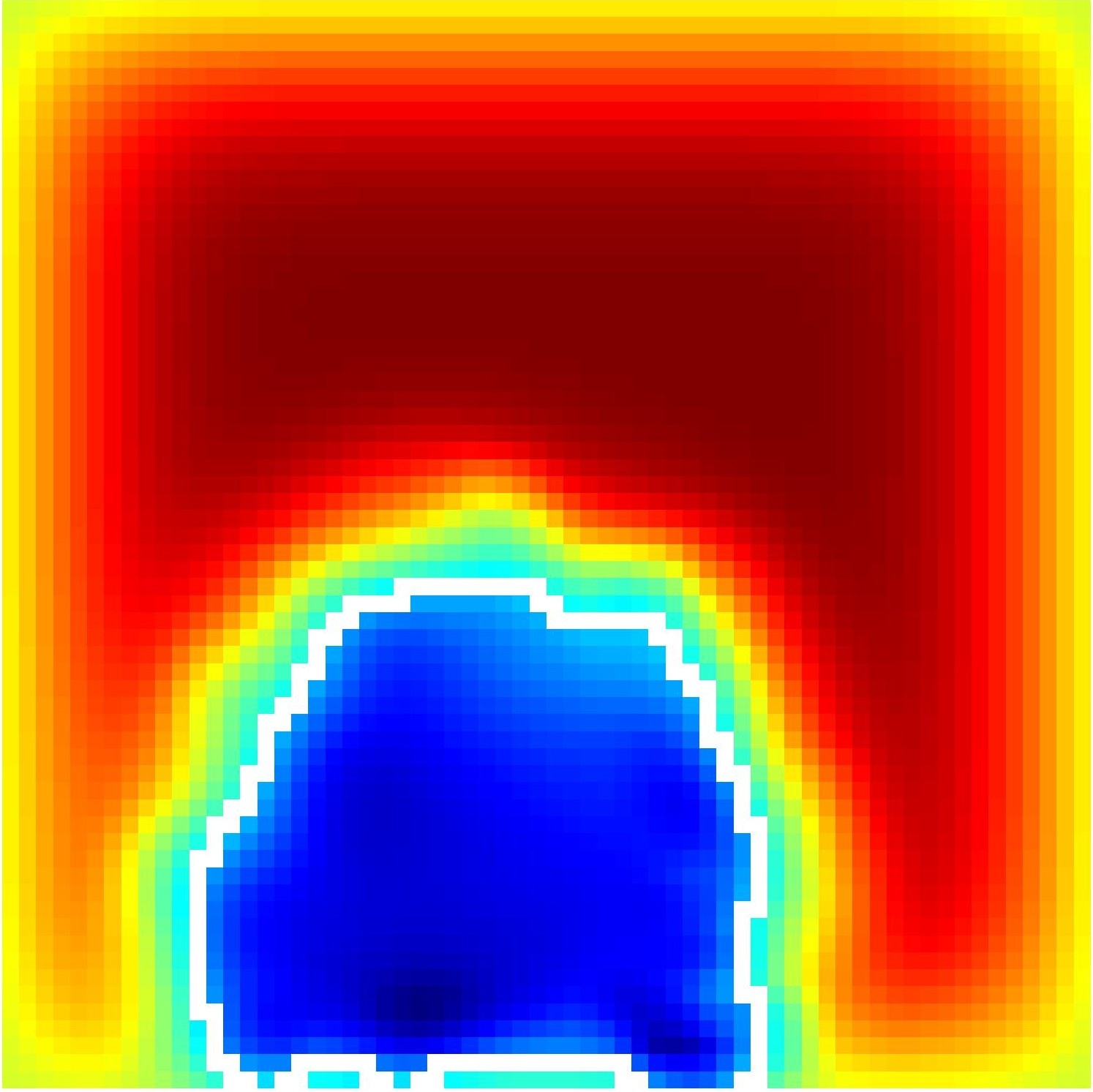}}&
        {\includegraphics[valign=c, width=\ww]{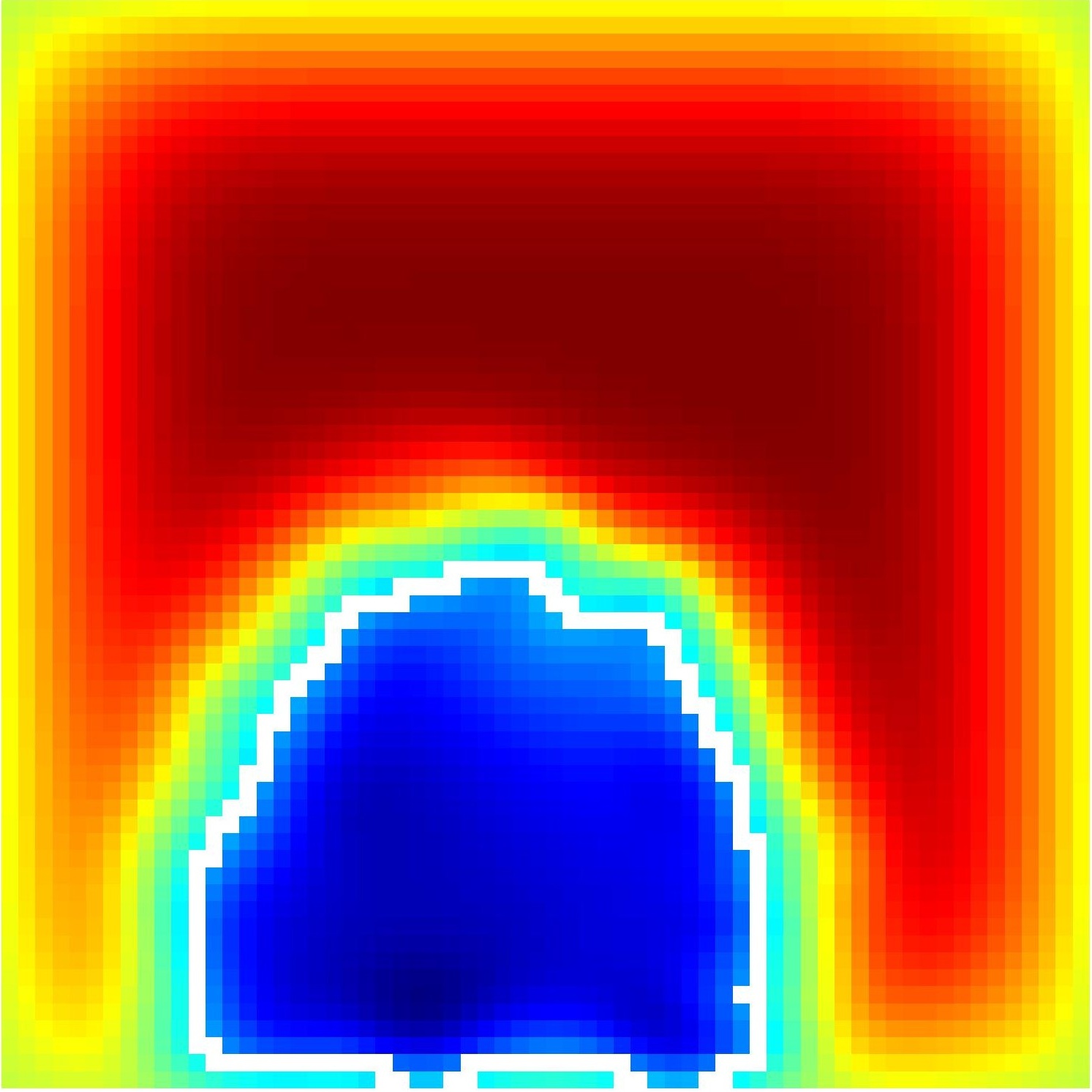}}
        \\
        [\rsb]
        \sizedtext{\methsize}{Initial}& & & & & & & \sizedtext{\methsize}{Final}

    \end{tabular}

    \caption{\textbf{Continuous mask study.} We investigated the use of a continuous mask, where instead of employing a potential height field ($\potential$) with binary thresholding ($\tau$), we utilized a continuous mask with a $\tanh$ normalization layer, followed by a shift to the (0, 1) range. Bottom row: The evolution of the continuous mask throughout the diffusion process, where blue represents high values and red represents low values. The white contour corresponds to a value of 0.5. Top row: The $\est{z\un{fg}}$ values throughout the diffusion process, with the white contour (indicating 0.5 values in the  continuous mask). The two approaches share similarities: both involve a continuous field (the continuous mask or $\potential$) and a transition to (almost) discrete values (via shifted $\tanh$ or binary thresholding with $\tau$). The continuous mask approach produced promising results, as illustrated in the figure, and represents a feasible alternative. However, our experiments indicated that the method based on binary thresholding ($\tau$) applied to a potential height field ($\potential$) ultimately performed better overall.}
    \label{fig:ablation_continuous}
\end{figure*}

\begin{figure*}[t!]
    \centering
    \setlength{\tabcolsep}{0pt}
    \renewcommand{\arraystretch}{0.5}
    \setlength{\ww}{0.12\linewidth}
    \renewcommand{\prsize}{footnotesize}
    \renewcommand{\methsize}{footnotesize}
    \setlength{\rsm}{1.28cm}
    \setlength{\rss}{4px}
    \setlength{\rsb}{15px}

    \begin{tabular}{c @{\hspace{2pt}}c @{\hspace{2pt}}cc @{\hspace{8pt}}c @{\hspace{2pt}}c @{\hspace{2pt}}cc}  
               
        \sizedtext{\methsize}{Input}&
        \sizedtext{\methsize}{No Evolution}&
        \multicolumn{2}{c}{\sizedtext{\methsize}{\ctm}} &
        \sizedtext{\methsize}{Input}&
        \sizedtext{\methsize}{No Evolution}&
        \multicolumn{2}{c}{\sizedtext{\methsize}{\ctm}} 
        \\
        [0.2\rss]
        
        \raisebox{-.5\height}{\includegraphics[width=\ww]{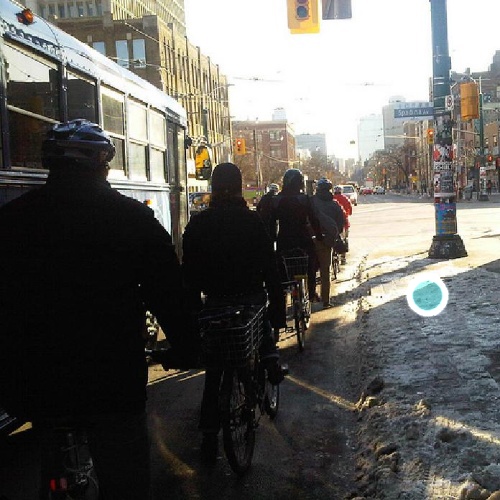}} &
        \raisebox{-.5\height}{\includegraphics[width=\ww]{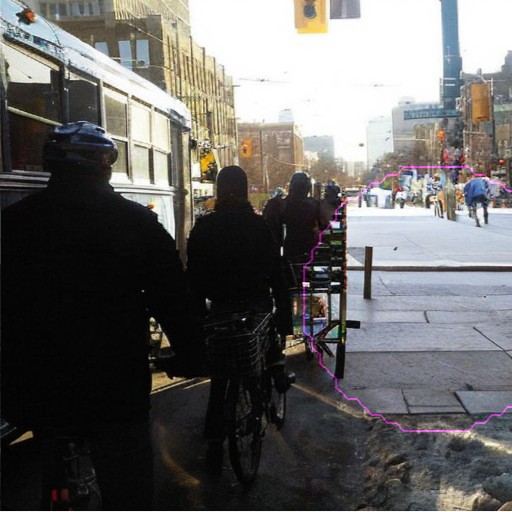}}  &
        \raisebox{-.5\height}{\includegraphics[width=\ww]{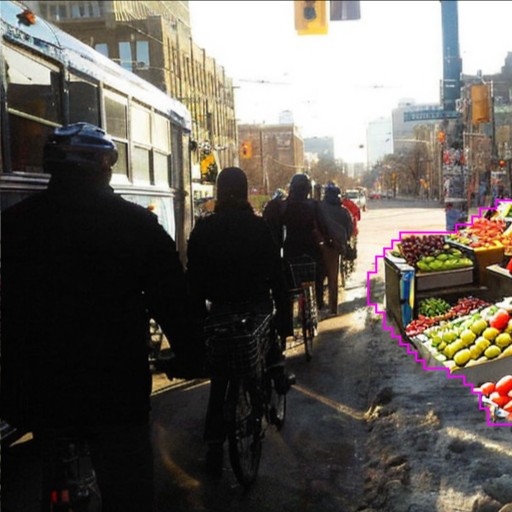}}  &
        \raisebox{-.5\height}{\includegraphics[width=\ww]{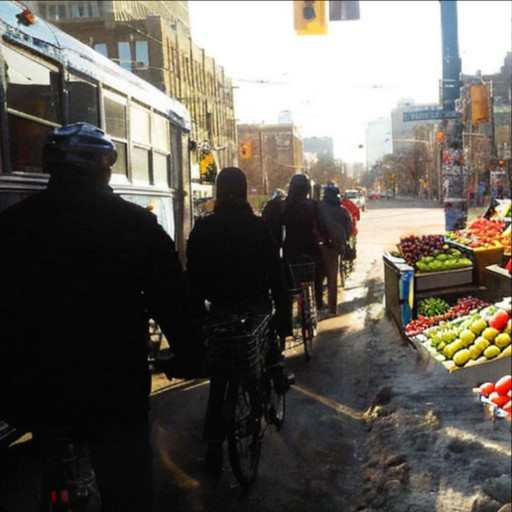}}  &
        \raisebox{-.5\height}{\includegraphics[width=\ww]{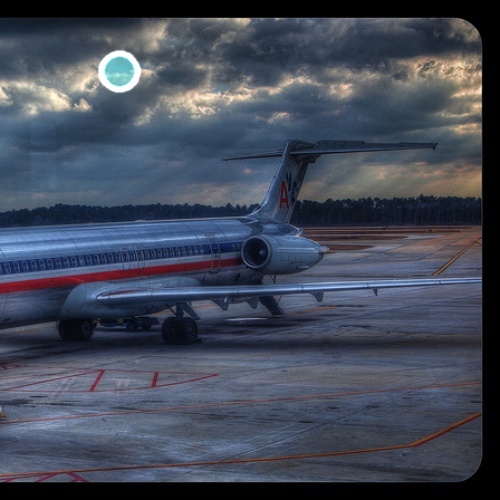}} &
        \raisebox{-.5\height}{\includegraphics[width=\ww]{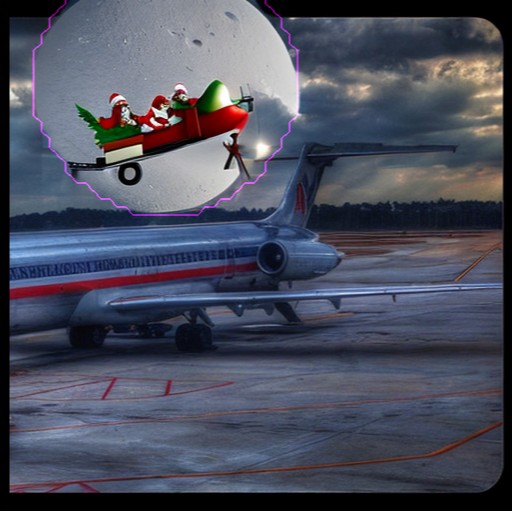}} &
        \raisebox{-.5\height}{\includegraphics[width=\ww]{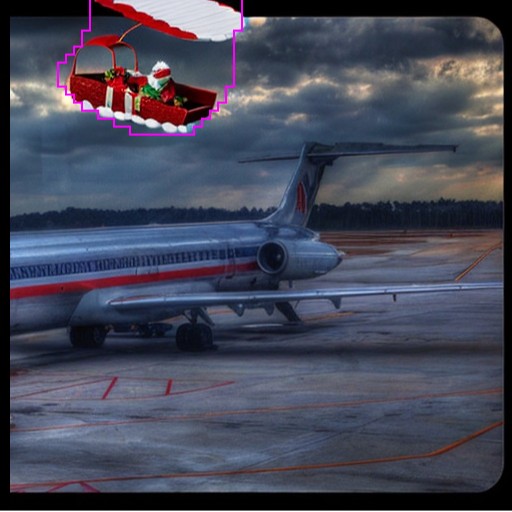}} & 
        \raisebox{-.5\height}{\includegraphics[width=\ww]{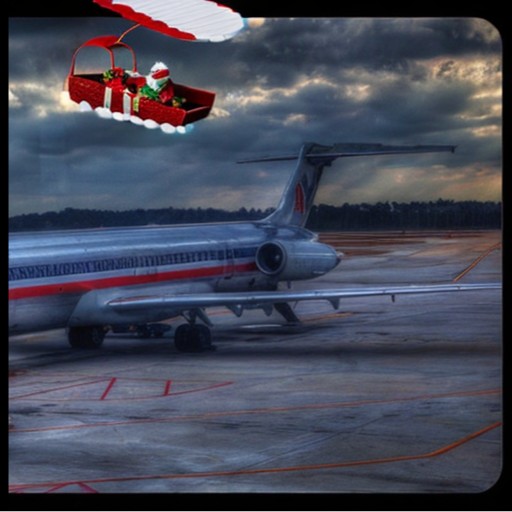}}
        \\
        [\rsm]
        \prunmerge{\prsize}{4}{A fruit stand} &
         \prunmerge{\prsize}{4}{Santa Clause in his sled flying}
        \\
        [\rsb]

    \end{tabular}

    \caption{\textbf{Ablation study: no mask evolution.} This figure presents an ablation study in which we employed a naive mask that remained fixed throughout the process. Specifically, we used the initial thresholded $\mask$ as a static mask, without any evolution, and applied BLD with it. Each quadruplet in the figure consists of (from left to right): the clicked input, the output generated using the fixed naive mask (indicated by a purple outline), the \ctm\ output with the final evolved mask (also depicted with a purple outline), and the \ctm\ output without a mask outline. As demonstrated, the naive approach results in significantly poorer performance.}
    \label{fig:abl_naive}
\end{figure*}

\Cref{fig:ablation_elevate_all} illustrates the importance of elevating potential $\potential$ only around the area of $\mask$’s contour, and not across the entire image. \Cref{fig:ablation_rerun} demonstrates an ablation study for the rerun component. \Cref{fig:abl_background_preservation} shows the importance of Gaussian mask feathering after the final diffusion step. \Cref{fig:ablation_no_outer_elevation} depicts the importance of adding a surrounding receptive around $\mask$'s area for gradient addition. 
\Cref{fig:abl_naive} illustrates a baseline using a fixed, non-evolving mask, while \Cref{fig:ablation_continuous} presents an alternative mask evolution approach we explored, based on a continuous mask.
An additional ablation study can be found in \Cref{sec:ablation_study}.

\subsection{Statistical Analysis}
\label{statistical_analysis}
As mentioned in \Cref{sec:results}, we conducted a user case study between \ctm\ with both \emu\ and \mb. To determine whether our comparisons are statistically significant, we use Pearson's Chi-squared test 
\cite{Pearson1900} with Yates's continuity correction \cite{c300c620-8718-3583-9438-7cb84c5d5aa0}. The tests show that the results are statistically significant, as can be seen in \Cref{tab:statistical_analysis}.

\subsection{Edited Alpha-CLIP Mask Extraction}
\label{edited_alphaclip_mask_extraction}
\begin{figure*}[h]
    \centering
    \setlength{\tabcolsep}{0pt}
    \renewcommand{\arraystretch}{0.5}
    \setlength{\ww}{0.135\linewidth}
    \renewcommand{\prsize}{footnotesize}
    \renewcommand{\methsize}{footnotesize}
    \setlength{\rsm}{1.25cm}
    \setlength{\rss}{4px}
    \setlength{\rsb}{15px}

    \begin{tabular}{c @{\hspace{5pt}}cc @{\hspace{5pt}}cc @{\hspace{5pt}}cc}  
               
        \sizedtext{\methsize}{Input}&
        \multicolumn{2}{c}{\sizedtext{\methsize}{\emu}} &
        \multicolumn{2}{c}{\sizedtext{\methsize}{\mb}} &
        \multicolumn{2}{c}{\sizedtext{\methsize}{\ctm}} 
        \\
        [0.2\rss]
        
        \raisebox{-.5\height}{\includegraphics[width=\ww]{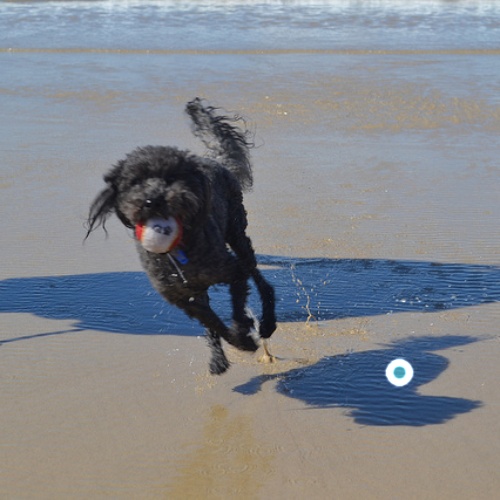}} &
        \raisebox{-.5\height}{\includegraphics[width=\ww]{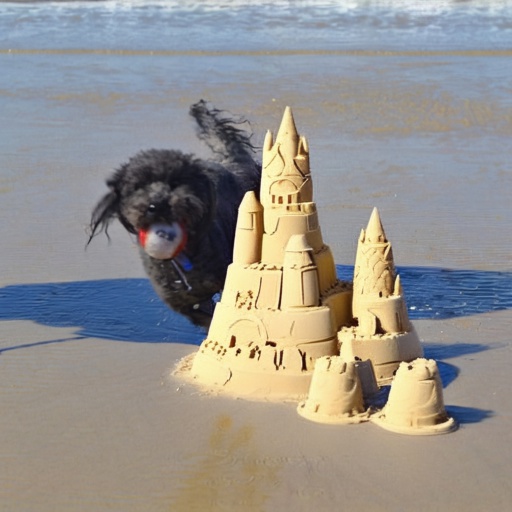}}  &
        \raisebox{-.5\height}{\includegraphics[width=\ww]{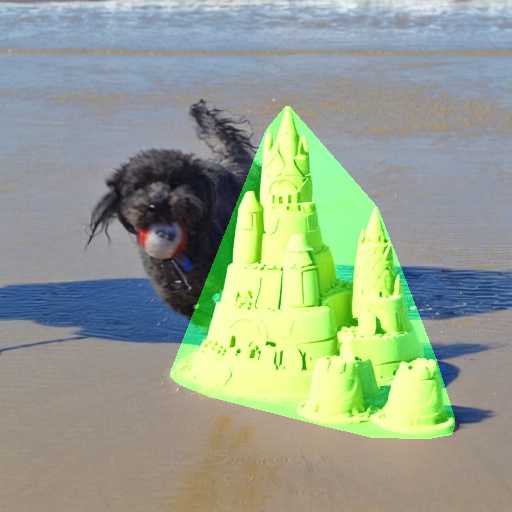}}  &
        \raisebox{-.5\height}{\includegraphics[width=\ww]{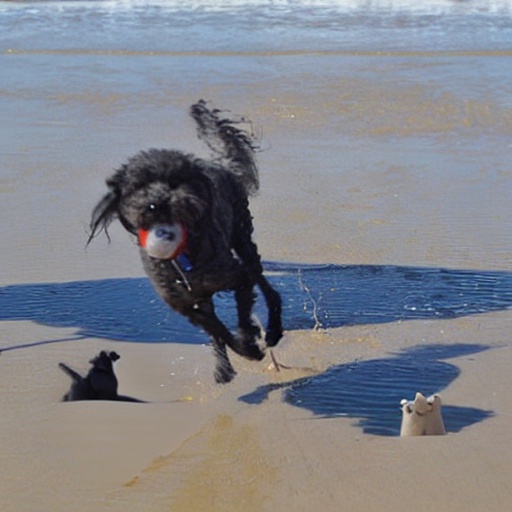}}  &
        \raisebox{-.5\height}{\includegraphics[width=\ww]{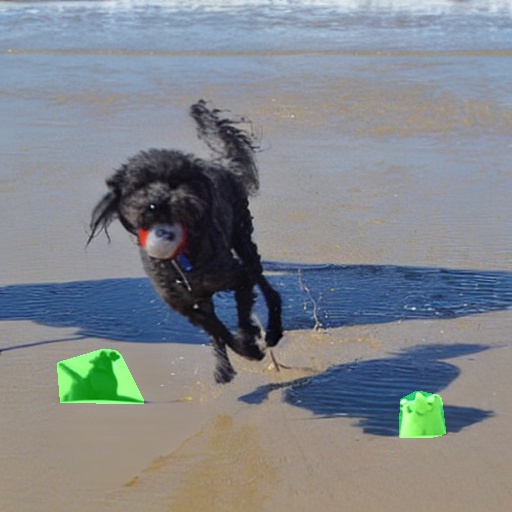}} &
        \raisebox{-.5\height}{\includegraphics[width=\ww]{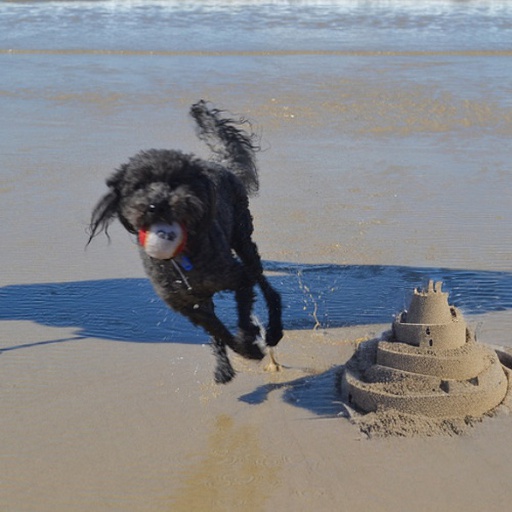}} &
        \raisebox{-.5\height}{\includegraphics[width=\ww]{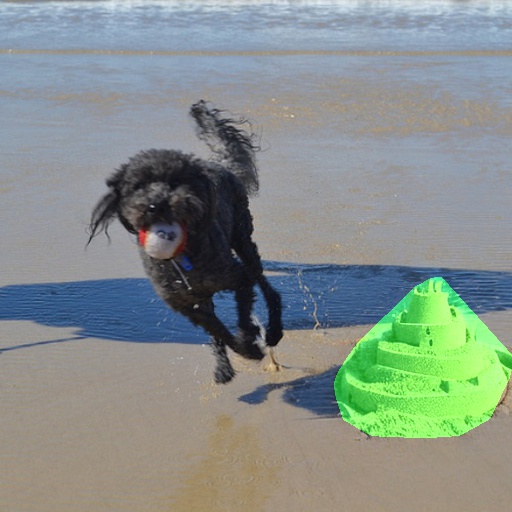}} 
        \\
        [\rsm]
        \prunmerge{\prsize}{7}{Add a sandcastle to the right of the dog}
        \\
        [\rss]
         \prunmerge{\prsize}{7}{A sandcastle}
        \\
        [\rsb]

        \raisebox{-.5\height}{\includegraphics[width=\ww]{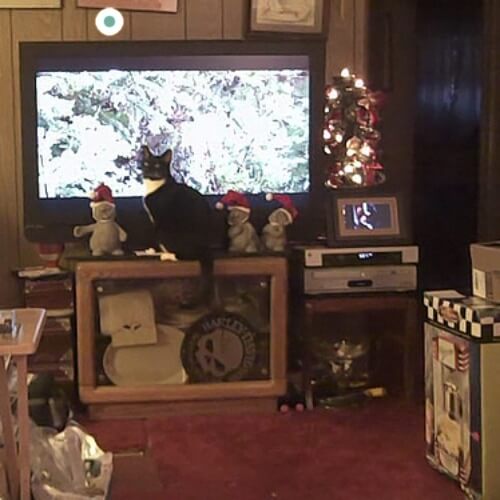}} &
        \raisebox{-.5\height}{\includegraphics[width=\ww]{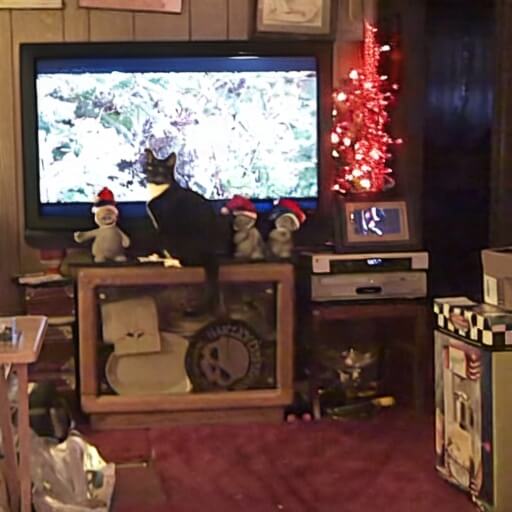}}  &
        \raisebox{-.5\height}{\includegraphics[width=\ww]{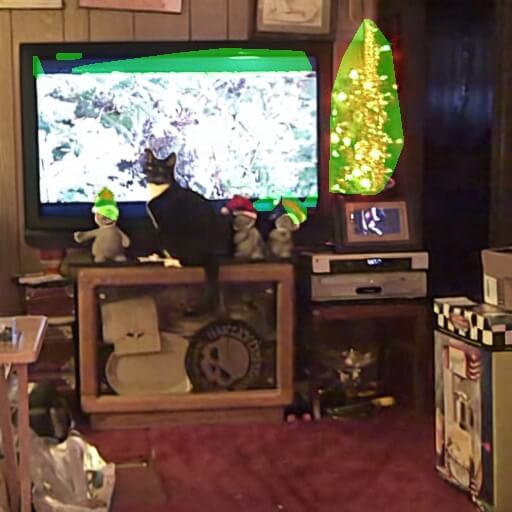}}  &
        \raisebox{-.5\height}{\includegraphics[width=\ww]{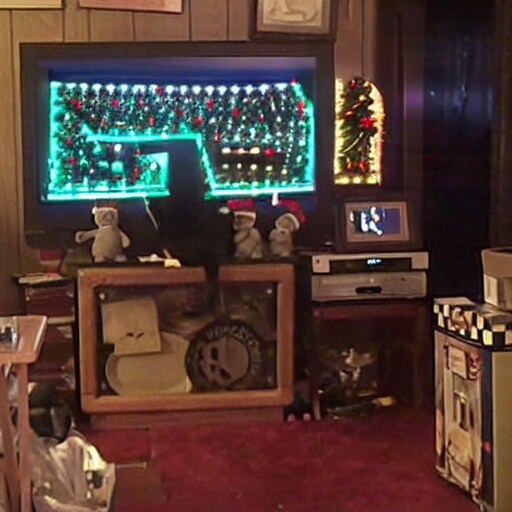}}  &
        \raisebox{-.5\height}{\includegraphics[width=\ww]{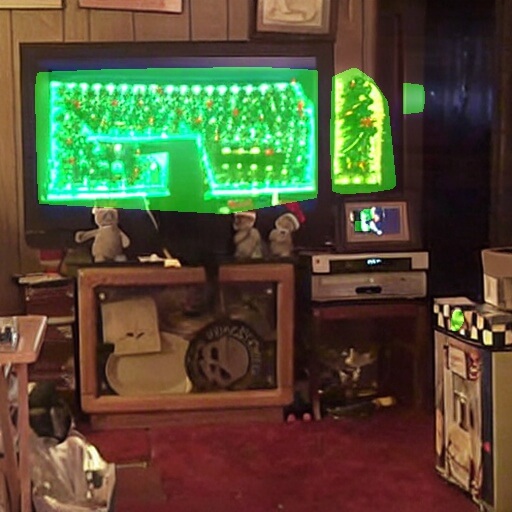}} &
        \raisebox{-.5\height}{\includegraphics[width=\ww]{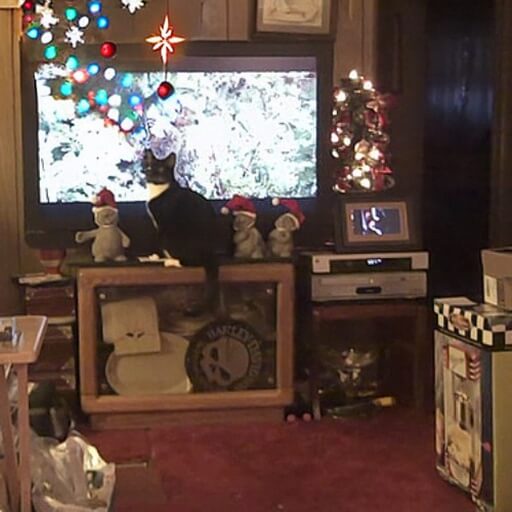}} &
        \raisebox{-.5\height}{\includegraphics[width=\ww]{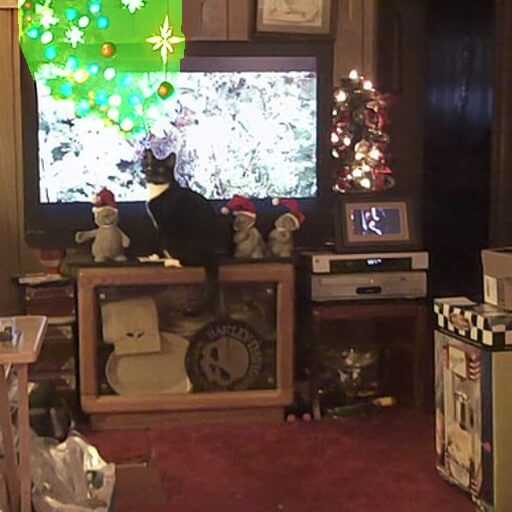}} 
        \\
        [\rsm]
        \prunmerge{\prsize}{7}{Add Christmas lights to the top of the television}
        \\
        [\rss]
        \prunmerge{\prsize}{7}{Christmas lights}

    \end{tabular}

    \caption{\textbf{Edited Alpha-CLIP.} A depiction of extracted masks as part of our Edited Alpha-CLIP metric presented in \Cref{sec:results}. Left column is the input with clicked point, where the text bellow each image row is the instruction given to \emu\ and \mb\ (higher text) and prompt given to \ctm\ (lower text). In each method's pair, the left image is the output, and the right image is the mask extracted by the Edited Alpha-CLIP metric.}
    \label{fig:edited_alpha_clip}
\end{figure*}

As mentioned in \Cref{fig:edited_alpha_clip}, in order to evaluate the edited region in methods that do not have input or output masks (as \emu\ and \mb), we extract a mask which specifies this region. The mask is extracted by first calculating the L1 distance between the input image and the generated image. We then take the mean value over the RGB channels for each pixel, and further clean noise by thersholding, Min-Pooling and Max-Pooling, and creating convex hulls. This provides us with an almost exact mask of the edited region, as demonstrated in \Cref{fig:edited_alpha_clip}.

\section{Implementation Details}
\label{sec:implementation_details}

\subsection{Pretrained Models}
\label{pretrained_models}
The pretrained models that we have used in all the experiments described in this paper are as follows:
\begin{itemize}[leftmargin=*]
    \item Blended Latent Diffusion model from Avrahami~\etal\ \shortcite{avrahami2023blendedlatent}.
    
    \item Text-to-image Latent Diffusion model from Rombach~\etal\ \shortcite{rombach2022highresolution} with checkpoint \url{https://huggingface.co/stabilityai/stable-diffusion-2-1-base}.  
    
    \item \alphaclip\ with \texttt{ViT-L/14@336px} by Sun~\etal\ \shortcite{sun2023alphaclip}.
    
    \item \emu\ benchmark from \url{https://huggingface.co/datasets/facebook/emu_edit_test_set} and \emu\ generated images from \url{https://huggingface.co/datasets/facebook/emu_edit_test_set_generations} by Sheynin \etal\ \shortcite{sheynin2023emu}.
    
    \item \mb\ by Zhang \etal\ \shortcite{Zhang2023MagicBrush} results were generated with latest checkpoint \newline\texttt{MagicBrush-epoch-52-step-4999.ckpt}.

    \item \ipp\ results generated from \url{https://huggingface.co/spaces/timbrooks/instruct-pix2pix} by Brooks \etal\ \shortcite{brooks2022instructpix2pix}.
   
\end{itemize}
All the above were implemented in PyTorch \cite{paszke2019pytorchimperativestylehighperformance}.

For \DALLE{3} \cite{BetkerImprovingIG}, we used OpenAI's ChatGPT-4o interface \url{https://chatgpt.com}. 

All input images are real and under free public domain or Creative Commons license (including Jeremy Bishop, Isaac Maffeis, Odysseas Chloridis and Cerqueira under Unsplash license; jenyalucy and Icecube11 under Pixabay license).

\subsection{Our Model}

When calculating the \alphaclip\ loss to derive gradients and to pick automatically the best output out of different random seeds (as discussed in \Cref{sec:results}), we augmented the image to mitigate adversarial results, as discussed in~\cite{Avrahami_2022_CVPR}, and dilate the mask region to add context.

The diffusion steps consisted of 100 steps. The total runtime for an edit on an Nvidia T4 Medium GPU is approximately 70 seconds.

To achieve unity over different samples in terms of learning rate $lr$ and potential $\potential$, a normalization is performed on the saliency map (i.e., absolute gradients backpropagated from \alphaclip\ loss function).

To reduce noise, maintain stability, and ensure a smooth mask, we perform Gaussian filtering to $\mask$ on a number of occasions, and post-process it each update step to account for gaps that can occur due to the landscape thresholding, such as filling holes, connecting disjointed mask parts, removing noise, etc. Additionally, we reset to the initial random seed on each BLD rerun for consistency of mask evolution. 

Source code of our model, which is implemented in PyTorch and runs on a GPU, is publicly available in the project page (see Footnote in \Cpageref{projectpage}).

\FloatBarrier

\begin{table}[!htb]
    \centering
    \small
    \begin{adjustbox}{width=1\columnwidth}
        \begin{tabular}{>{\columncolor[gray]{0.95}}c>{\columncolor[gray]{0.95}}c ccc}
            \toprule
            
            \textbf{Method 1} &
            \textbf{Method 2} &
            Majority (A)&
            Majority (B)&
            Total votes
            \\

            &
            &
            p-value &
            p-value &
            p-value
            \\
            \midrule

            \emu &
            Ours &
            $p < 10^{-21}$ &
            $p < 10^{-14}$ &
            $p < 10^{-192}$ 
            \\

            \mb &
            Ours &
            $p < 10^{-19}$ &
            $p < 10^{-15}$ &
            $p < 10^{-111}$
            \\
            
            \bottomrule
        \end{tabular}
    \end{adjustbox}
    \caption{\textbf{Statistical analysis.} We use Pearson's Chi-squared test with Yates's continuity correction to determine whether our results are statistically significant. Majority (A) refers to the comparison of majority votes for each item disregarding ties, and majority (B) refers to the comparison of votes disregarding items that most users rated as ties. Total votes are the total ratings for each method. See \Cref{sec:results} for further details.}
    \label{tab:statistical_analysis}
\end{table}

\FloatBarrier

\end{document}